\documentclass{article}
\usepackage{algorithm}
\usepackage{algorithmic}
\PassOptionsToPackage{numbers, compress}{natbib}
\usepackage{amsthm}
\usepackage{subcaption}
\usepackage{titletoc}
\newtheorem{lemma}{Lemma}
\newtheorem{proposition}{Proposition}

\newtheorem{theorem}{Theorem}
\newtheorem{remark}{Remark}
\newtheorem{assumption}{Assumption}
\usepackage{booktabs}
\usepackage{multirow}
\usepackage{adjustbox}
\usepackage{wrapfig}
\usepackage{graphicx}
\usepackage{amsmath, amssymb, amsfonts}
\usepackage{bm}
\usepackage[preprint]{neurips_2026}

\usepackage[utf8]{inputenc} 
\usepackage[T1]{fontenc}    
\usepackage{hyperref}       
\usepackage{url}            
\usepackage{booktabs}       
\usepackage{amsfonts}       
\usepackage{nicefrac}       
\usepackage{microtype}      
\usepackage{xcolor}         

\usepackage{xspace}
\providecommand{\ie}{i.e.\xspace}
\providecommand{\eg}{e.g.\xspace}
\title{When and Why Adversarial Training Improves PINNs: A Neural Tangent Kernel Perspective}

\author{%
   Yuandong Cao$^{1,3}$\thanks{work done during visiting University College London} \quad Chi Chiu So$^{2}$ \quad Jun-Min Wang$^{1}$ \quad He Wang$^{3}$\thanks{corresponding author, he\_wang@ucl.ac.uk}\\
   \\
   $^{1}$School of Mathematics and Statistics, 
   Beijing Institute of Technology, China\\
   $^{2}$School of Professional Education and Executive Development\\ The Hong Kong Polytechnic University, China \\
   $^{3}$Department of Computer Science \& UCL AI Centre, 
   University College London, UK \\
}
\input{preambles}

\begin{document}
	\maketitle
	\begin{abstract}

		Physics-informed neural networks (PINNs) are powerful surrogates for differential equations but are notoriously difficult to train due to spectral bias, stiffness, and poor accuracy on high-frequency or multiscale solutions. Adversarial training based on generative adversarial networks (GANs) has recently gained surprisingly strong empirical results in improving the training, but the underlying mechanisms remain elusive. To this end, we propose a new analysis framework for adversarially trained PINNs, based on the key observation of how the discriminator in GANs can influence the training dynamics of PINNs. The framework first provides a much needed theoretical grounding to why and when adversarial training is effective in PINNs, then presents a unified analysis of GANs variants in such training, and finally leads to a new, practical, efficient training algorithm for PINNs. Empirical results demonstrate that our method can significantly reduce the pathology of PINNs training, thereby providing better models with superior performance, often several magnitudes more accurate than alternative methods.
	\end{abstract}
	
	\section{Introduction}
	Partial Differential Equations (PDEs) are foundational modeling tools in science and engineering, but designing stable and efficient solvers for them is challenging, particularly for complex systems characterized by strong nonlinearity, stiffness, or multiscale structures. Neural networks recently emerged as a promising alternative for solving PDEs, where physics-informed neural networks (PINNs) have become widely used, achieving remarkable success across a broad range of applications. However, PINNs are notoriously hard to train, especially when applied to PDEs with solutions exhibiting high-frequency or multiscale features \cite{fuks2020limitations,raissi2018deep,zhu2019physics}. Recent studies have systematically analyzed these failures and attributed the pathological behavior of PINNs to multiscale interactions and spectral bias among different loss components \cite{wang2021understanding, wang2022and}. In practice, this manifests itself as a stagnation of the training loss near stiff regions of the PDE where the network prediction deviates significantly from the ground truth. A recent attempt to mitigate this pathology is to leverage adversarial training via generative adversarial networks (GANs), \ie adversarial PINNs training \cite{bullwinkel2022deqgan,ciftci2024physics, yang2020physics,song2024loss,cao2025adversarial}. 
	
	GANs consist of a generator and a discriminator, trained via a minmax game \cite{goodfellow2020generative}. When applied to PINNs, the prediction errors, \eg PDE residual, boundary, or/and initial condition errors, are modeled as generated samples, while the ideal zero error is treated as the real target. In other words, training amounts to driving the empirical distribution of the prediction error toward a Dirac measure centered at zero. A growing body of work has demonstrated the empirical effectiveness of this idea, including employing vanilla GANs with hard constraints \cite{bullwinkel2022deqgan}, Wasserstein GANs with penalty-based methods \cite{ciftci2024physics} for PDEs and Stochastic Differential Equations \cite{yang2020physics}. More recent research also investigates using separate discriminators for different loss terms \cite{song2024loss} and balancing physics constraints and control objectives in PDE-constrained optimal control \cite{cao2025adversarial}.

	Despite empirical successes, the theoretical grounding of adversarial PINNs training is missing, especially in why GANs succeed or fail in some settings. This has led to inconsistent or even contradictory observations and analysis. For example, Ciftci et al. \cite{ciftci2024physics} and Yang et al. \cite{yang2020physics} argued that vanilla GAN training is essentially Jensen-Shannon divergence minimization under suitable assumptions, and therefore should be avoided in PINN training due to its well-known instability, but Bullwinkel et al. \cite{bullwinkel2022deqgan} showed that vanilla GANs can in practice drive PINNs to extremely accurate solutions. More broadly, the generated distribution of adversarially trained PINNs is often a finite Dirac mixture supported on the PDE residual samples, whereas the target distribution is a Dirac measure with singular support at zero. Under this support mismatch, the classical divergence-based theory \cite{nowozin2016f, mao2017least} provides little useful guidance and may even suggest degenerate training behaviors, which is clearly inconsistent with empirical evidence. Overall, the above discrepancies show that existing analysis which is mainly based on \textit{static} optimal-discriminator behaviors is insufficient to explain the adversarial PINNs training. A \textit{dynamical} perspective is required.
	
	
	We draw inspiration from the gradient-flow analysis of neural networks, specifically through the lens of Neural Tangent Kernel (NTK) \cite{jacot2018neural}, to characterize the dynamics of adversarial PINNs training. NTKs have been employed successfully in GANs \cite{franceschi2022neural} to show that the study of the alternating updates between the generator and the discriminator is essential to understand the training dynamics. We generalize this analysis to adversarial PINNs training to reveal: (1) how discriminator training shapes the dynamics of PINNs, (2) how different GAN objectives induce different sample-wise weighting, and (3) how these effects interact with the generator training dynamics. These phenomena jointly explain the convergence (or the lack of) of adversarial PINNs training. 
	
	Overall, our main contributions include:
	\begin{itemize}
		\item A new unified NTK framework for analyzing adversarially trained PINNs. 
		\item A formal analysis of adversarial PINNs training under different GAN variants.
		\item New findings in how the discriminator shapes the spectral dynamics of PINNs training.
		\item A new training algorithm with improved optimization stability and convergence behavior. 
	\end{itemize}

	\section{Preliminaries}
	\paragraph{Generative Adversarial Networks}
	GAN consists of a generator and a discriminator. We denote the discriminator by
	\(
	f(\cdot;\phi):\mathcal X\to\mathbb R
	\),
	where \(\phi\) denotes the discriminator parameters. The generator $g(z;\theta)$ is parameterized by \(\theta\), where $z \sim p_z$ is a latent variable. Let \(\alpha_r\) and \(\alpha_g^{\theta}\) denote the real and generated data distributions on \(\mathcal X\), respectively. GAN training can be seen as a minmax game:
	\begin{align}
		\max_{\phi}\mathcal{L}(\phi)
		&=
		\mathbb{E}_{x\sim\alpha_r}\big[P(f(x;\phi))\big]
		+
		\mathbb{E}_{x\sim\alpha_g^{\theta}}\big[Q(f(x;\phi))\big],
		\label{eq:general_gan1_short}\\
		\min_{\theta}\mathcal{L}(\theta)
		&=
		\mathbb{E}_{x\sim\alpha_g^{\theta}}\big[R(f(x;\phi))\big],
		\label{eq:general_gan2_short}
	\end{align}
	where \(P,Q,R:\mathbb R\to\mathbb R\) are scalar functions. In particular, we consider three GAN variants:
	\begin{equation}
		\text{GAN \cite{goodfellow2020generative}:}\ \ 
		P(t)=\log \sigma(t),\ \ 
		Q(t)=R(t)=\log\!\bigl(1-\sigma(t)\bigr), \ \  \sigma(\cdot) \text{ is the sigmoid function,}
		\label{eq:gan_short}
	\end{equation}
	\begin{equation}
		\text{LSGAN \cite{mao2017least}: }
		P(t)=-\frac12(t-1)^2,\quad
		Q(t)=-\frac12 t^2,\quad
		R(t)=\frac12(t-1)^2,
		\label{eq:lsgan_short}
	\end{equation}
	and
	\begin{equation}
		\text{WGAN/IPM\cite{arjovsky2017wasserstein}: }
		P(t)=t,\quad Q(t)=R(t)=-t.
		\label{eq:wgan_short}
	\end{equation}
	
	\paragraph{Hard-constrained PINNs}
	Let \(\Omega\subset\mathbb R^d\) be the space--time domain and consider the PDE
	\begin{equation}
		\mathcal N(u)(x)=a(x),\quad x\in\Omega,
		\qquad
		\mathcal B(u)(x)=b(x),\quad x\in\partial\Omega,
	\end{equation}
	where \(\mathcal N\) is the differential operator, \(a(x)\) is the source term, and
	\(\mathcal B\) denotes the boundary and/or initial constraint operator. Through a proper
	hard-constraint parameterization \cite{lagaris1998artificial, flamant2020solving}, the neural solution \(u_\theta\) satisfies
	\(\mathcal B(u_\theta)(x)=b(x)\) exactly. Therefore, training reduces to minimizing the
	interior residual $\mathcal R(x;\theta):=\mathcal N[u_\theta](x)-a(x).$
	The hard-constrained PINN loss is then given by
	\begin{equation}\label{eq:hard_short}
		\mathcal L_{\mathrm{PINN}}(\theta)
		=
		\mathbb E_{x\sim\mu_\Omega}
		\big[
		\mathcal R(x;\theta)
		\big],
	\end{equation}
	where \(\mu_\Omega\) denotes the sampling distribution over the interior collocation domain.

    \paragraph{Adversarially trained PINNs}
	\label{sec:adversarial_pinn}
	Given collocation points \(\{x_i\}_{i=1}^{N_r}\subset\Omega\), define residual 
	\(r_i:=\mathcal R(x_i;\theta)\) and
	\(\mathbf r:= (r_1,\dots,r_{N_r})^\top\in\mathbb R^{N_r}\).
	Furthermore, we define
	\(\hat\mu_\Omega:=\frac{1}{N_r}\sum_{i=1}^{N_r}\delta_{x_i}\)
	to be the empirical collocation measure, where \(\delta_{x_i}\) is the Dirac measure at \(x_i\).
	Then the generated residual distribution is the pushforward empirical measure
	$
	\hat\mu_\theta^{\,r}
	:=
	(\mathcal R_\theta)_\#\hat\mu_\Omega
	=
	\frac{1}{N_r}\sum_{i=1}^{N_r}\delta_{r_i},
	$
	where \(\mathcal R_\theta:x\mapsto \mathcal R(x;\theta)\).
	The ideal target residual distribution is the Dirac measure at zero,
	\(\mu_{\mathrm{real}}=\delta_0\). Equivalently, we define the target residual map
	$
	\mathcal T(x_i):=0, i=1,\dots,N_r,
	$
	which represents the desired zero-residual state (real data).
	Therefore, the adversarial PINNs training objectives become:
	\begin{equation}
		\mathcal L(\theta)
		=
		\frac{1}{N_r}\sum_{i=1}^{N_r}
		R\Big(f(\mathcal R(x_i;\theta);\phi)\Big),
		\label{eq:g_short}
	\end{equation}
	and
	\begin{equation}
		\mathcal L(\phi)
		=
		\frac{1}{N_r}\sum_{i=1}^{N_r}
		\Big[
		P\Big(f(\mathcal T(x_i);\phi)\Big)
		+
		Q\Big(f(\mathcal R(x_i;\theta);\phi)\Big)
		\Big].
		\label{eq:d_short}
	\end{equation}
	Here \(P,Q,R\) are the same as in \cref{eq:general_gan1_short,eq:general_gan2_short}.
	The objectives in \cref{eq:g_short,eq:d_short} will be the starting point of our analysis. Hence, the generator $g$ is the PINN itself, while the discriminator operates on PDE residuals rather than collocation locations. Specifically, each collocation point passes through the following training pipeline:
    \begin{equation}
    \label{eq:adversarial_pinn_schematic}
    x_i
    \xrightarrow{\,g}
    u_\theta(x_i)
    \xrightarrow{\,\mathcal{R}\,}
    r_i
    \xrightarrow{\,f\,}
    f(r_i)
    \xrightarrow{\,\mathrm{backpropagation}\,}
    \nabla_\theta \mathcal{L}(\theta)
    \xrightarrow{\,\mathrm{parameter\ update}\,}
    g.
    \end{equation}
    The discriminator distinguishes these residuals from the target residual of zero, and its feedback is propagated through the residual operator to update
    the generator $g$.

	
	\paragraph{Neural tangent kernel regime}
	\label{sec:ntk_prelim}
	
	Our analysis is carried out in the neural tangent kernel (NTK) regime. 
	At a high level, this regime describes sufficiently wide neural networks trained in a lazy-training manner, so that the network output can be approximated by its first-order expansion around initialization, while the associated kernel remains nearly constant during training.
	
	We adopt the following standard assumptions:
	\begin{assumption}[\cite{jacot2018neural}]
		\label{assump1}
		(i) The network is a standard architecture, such as a fully connected, convolutional, or residual network, with standard nonlinearities. (ii) The activation function is sufficiently smooth, and the network output is differentiable with respect to the parameters. (iii) Linear layers contain nonzero bias terms. (iv) The network parameters are randomly initialized with a standard width-dependent scaling. (v) The hidden widths tend to infinity.
	\end{assumption}
	
	Under these assumptions, if \(g(x;\theta)\) denotes the network output, then around initialization \(\theta_0\),
	\begin{equation}
		g(x;\theta)
		\approx
		g(x;\theta_0)
		+
		\nabla_\theta g(x;\theta_0)^\top(\theta-\theta_0),
	\end{equation}
	and the NTK is defined by
	\begin{equation}
		k(x,x')
		=
		\big\langle
		\nabla_\theta g(x;\theta_0),
		\nabla_\theta g(x';\theta_0)
		\big\rangle.
	\end{equation}
	For a finite set of training points, the output dynamics under gradient flow are then approximated by
	\begin{equation}
		\dot{\mathbf g}(t)
		\approx
		-K\,\nabla_{\mathbf g}\mathcal L(\mathbf g(t)),
	\end{equation}
	where \(K\) is the NTK Gram matrix. Therefore, in the NTK regime, training can be analyzed in function space through an approximately fixed kernel rather than only through the original nonlinear parameter dynamics. A more detailed discussion of the related works and their connections to the present framework is provided in Appendix~\ref{appendixA}.
	
	\section{The Unreasonable Successes in Adversarial PINNs Training}
	\label{sec:unresasonable}
	In vanilla GAN, it is known that the generator's objective is, up to constants, minimizing the Jensen--Shannon divergence between the generated $\hat{\mu}_\theta^{\,r}$ and the target $\delta_0$, under proper assumptions. Similar conclusions also arise from other analysis based on \eg KL-type \cite{nowozin2016f} and $\chi^2$-type \cite{mao2017least} criteria. All analyses point to one phenomenon: \textit{when the supports of the generated and target distributions are nearly disjoint and the discriminator becomes nearly optimal, the generator receives a degenerate training signal}. This issue is particularly severe in adversarial PINNs. At the beginning of training, the residual samples \(\{r_i\}_{i=1}^{N_r}\) are typically far from zero, so the support of \(\hat{\mu}_\theta^{\,r}\) is almost disjoint from that of \(\delta_0\). Theoretically, one would expect the training signal for the generator to saturate or become uninformative. In reality, existing adversarial PINNs methods often perform well. This discrepancy indicates that the classical divergence-minimization framing does not directly apply to PINNs, as it assumes the generator is always trained against an optimal discriminator, whereas practical adversarial PINNs training uses a sub-optimal discriminator. As a result, the generator is driven not by the gradient of a fixed divergence alone, but also by the gradient field induced by a transient discriminator state.

\begin{wrapfigure}{r}{0.5\linewidth}
		\vspace{-1.5em}
		\centering
		\includegraphics[width=1.0\linewidth]{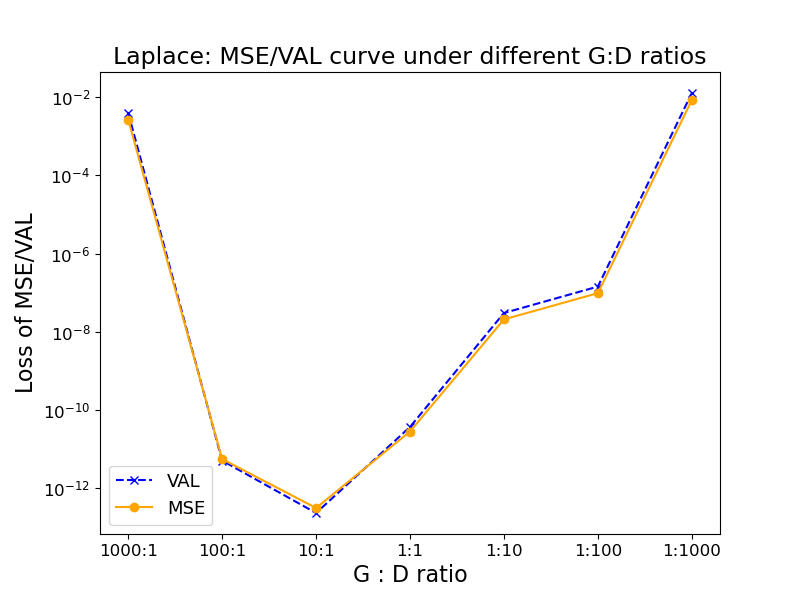}
		\caption{Laplace equation: final converged training and validation errors under different generator-to-discriminator update ratios (G:D).}
		\label{laplace1}
		\vspace{-1em}
	\end{wrapfigure}

	As a motivating example, we show how the generator--discriminator update scheme can significantly influence the training, using the two-dimensional Laplace equation:
	
	\begin{equation}
		\Delta u(x,y)=0,\qquad (x,y)\in[0,1]^2,
	\end{equation}
	with boundary conditions
	\begin{equation}
		u(x,0)=0,\ \  u(0,y)=0,\ \  u(1,y)=0,\ \ 
		u(x,1)=\frac{1}{2}\cosh(\pi)\sin(\pi x)\big(e^{\pi y}-e^{-\pi y}\big),
	\end{equation}
	and a hard-constrained PINNs. We vary the generator--discriminator update ratio by:
	\[
	G:D = 1000\!:\!1,\;100\!:\!1,\;\dots,\;1\!:\!1,\;\dots,\;1\!:\!100,\;1\!:\!1000,
	\]
	and show the training dynamics in Fig.~\ref{laplace1}.
	
    Fig.~\ref{laplace1} shows that both over-trained discriminators and over-trained generators can harm adversarial PINNs optimization. When \(D\) is updated too many times relative to \(G\), it may become nearly optimal for the current residual distribution, causing the learning signal passed to \(G\) to become saturated, weak, or poorly conditioned. Conversely, when \(G\) is updated too many times relative to \(D\), the discriminator remains under-trained and cannot provide a reliable gradient. In both cases, the imbalance leads to deteriorated training and validation errors. Therefore, the U-shaped curve in Fig.~\ref{laplace1} provides direct evidence that the success of adversarial PINNs is governed by finite-step alternating dynamics between \(G\) and \(D\), rather than by the static optimal-discriminator alone. The best performance appears in an intermediate regime where \(D\) is informative but not over-trained, and \(G\) is updated sufficiently but does not outrun the discriminator.
    
    These observations naturally raise two questions: how can we formally characterize such finite-step alternating dynamics, and how can we use this characterization to choose or control the generator--discriminator update schedule? Addressing these questions is the main goal of the following analysis.

	\section{NTK Analysis of Adversarially Trained PINNs}\label{NTK_analysis}
	
	
	We establish a dynamical analysis of the alternating optimization of adversarially trained PINNs. To this end, we first formulate the discriminator-induced residual-space gradient field, then show how it influences the training of the generator.
	
	\subsection{Discriminator-induced gradient field in the NTK regime}
	
	Given a fixed generator with $\theta$ and the objective in \cref{eq:d_short}, the discriminator gradient flow is:
	
	\begin{equation}\label{eq:disc_param_flow_final}
		\dot{\phi}(t)
		=
		\frac{1}{N_r}\sum_{i=1}^{N_r}
		\Bigg[
		P'\!\big(f(\mathcal T(x_i);\phi(t))\big)\,
		\frac{\partial f(\mathcal T(x_i);\phi(t))}{\partial \phi}
		+
		Q'\!\big(f(r_i;\phi(t))\big)\,
		\frac{\partial f(r_i;\phi(t))}{\partial \phi}
		\Bigg].
	\end{equation}
	If we define the discriminator NTK as:
	\begin{equation}
		k_t^D(x,y)
		:=
		\left\langle
		\frac{\partial f(x;\phi(t))}{\partial \phi(t)},
		\frac{\partial f(y;\phi(t))}{\partial \phi(t)}
		\right\rangle,
	\end{equation}
	then NTK theory implies that $k_t^D(x,y)\approx k^D(x,y)$ during training.
	Therefore, for any residual-space input $x$, the discriminator admits the function-space evolution:
	\begin{equation}\label{eq:disc_functional_flow_final}
		\frac{d}{dt}f(x;\phi(t))
		=
		\frac{1}{N_r}\sum_{i=1}^{N_r}
		\Big[
		P'\!\big(f(\mathcal T(x_i);\phi(t))\big)\,k^D(x,\mathcal T(x_i))
		+
		Q'\!\big(f(r_i;\phi(t))\big)\,k^D(x,r_i)
		\Big].
	\end{equation}
	NTK regularity implies that $k^D(\cdot,y)$ is differentiable with respect to its first argument, so $f(\cdot;\phi(t))$ is differentiable w.r.t. the input; see Appendix \ref{app:disc_gen_dynamics}.
    
	\begin{lemma}[Input-gradient representation of the discriminator \cite{franceschi2022neural}]
		\label{lem:disc_input_gradient_final}
		Under Assumptions \ref{assump1} the discriminator input gradient satisfies
		\begin{align}\label{eq:disc_input_grad_final}
			\partial_x f(x;\phi(t))
			=
			\partial_x f_0(x)
			+
			\int_0^t
			\frac{1}{N_r}\sum_{i=1}^{N_r}
			\Big[
			&P'\!\big(f(\mathcal T(x_i);\phi(s))\big)\,\partial_x k^D(x,\mathcal T(x_i))
			\nonumber \\+
			&Q'\!\big(f(r_i;\phi(s))\big)\,\partial_x k^D(x,r_i)
			\Big]\,ds,
		\end{align}
		where $f_0(x):=f(x;\phi(0))$.
	\end{lemma}
	\begin{remark}
        Lemma~\ref{lem:disc_input_gradient_final} describes the evolution of the discriminator gradient field w.r.t. the input during training. It shows that \(\partial_r f(r;\phi(t))\) is governed by the derivatives of the discriminator objective and the derivatives of the discriminator NTK kernel \(k^D\) w.r.t. the input. Therefore, the residual-space gradient field produced by the discriminator is not arbitrary, but is structured by the discriminator architecture, objective, and tangent kernel. This discriminator gradient field will later enter the generator dynamics through the sample-wise weighting (see Appendix~\ref{app:disc_gen_dynamics}). The corresponding NTK-based discriminator dynamics for vanilla GAN, LSGAN, and IPM-type GANs are given in Appendix~\ref{app:disc_dynamics}--\ref{app:disc_ntk_residual_space}.
    \end{remark}
	
	\subsection{Generator residual dynamics and objective-dependent weighting}
	
	For the generator, given the generator objective in \cref{eq:g_short}, 
	the generator gradient flow is:
	\begin{equation}\label{gen_gra_flow}
		\dot{\theta}(t)
		=
		-\frac{\partial \mathcal{L}(\theta(t))}{\partial \theta(t)}.
	\end{equation}
	
	We further define the generator residual NTK:
	\begin{equation}
		(K_{rr}^G)_{ij}(t)
		=
		\left\langle
		\frac{\partial \mathcal{R}(x_i;\theta(t))}{\partial \theta(t)},
		\frac{\partial \mathcal{R}(x_j;\theta(t))}{\partial \theta(t)}
		\right\rangle
	\end{equation}
	and a discriminator-induced weighting vector
	\begin{equation}\label{eq:gamma_vector_final}
		\boldsymbol{\gamma}(t)
		:=
		\big[\gamma_1(t),\dots,\gamma_{N_r}(t)\big]^\top,
	\end{equation}
	where
	\begin{equation}\label{eq:gamma_entry_final}
		\gamma_i(t)
		=
		-\frac{1}{N_r}
		R'\!\Big(f(r_i(t);\phi(t))\Big)\,
		\partial_r f(r_i(t);\phi(t)),
		\qquad i=1,\dots,N_r.
	\end{equation}
	
	\begin{proposition}[Residual dynamics under adversarial training]
		\label{prop:residual_dynamics_final}
		Under the generator gradient flow~\eqref{gen_gra_flow}, the residual vector evolves according to
		\begin{equation}\label{eq:residual_dyn_compact_final}
			\dot{\mathbf r}(t)
			=
			K_{rr}^G(t)\,\boldsymbol{\gamma}(t).
		\end{equation}
	\end{proposition}
	\begin{remark}
		Equation~\eqref{eq:residual_dyn_compact_final} shows that adversarial PINNs training takes the form of a \emph{preconditioned residual descent}. The generator residual NTK $K_{rr}^G(t)$ determines how a weighted residual signal is propagated across collocation points, whereas the discriminator enters through the sample-wise weighting vector $\boldsymbol{\gamma}(t)$. In other words, the discriminator reshapes the residual descent field through its input gradient rather than through its scalar output alone. Detailed derivations of the generator--discriminator update dynamics, together with the corresponding propositions and theorems, are provided in Appendix \ref{app:gen_dynamics}.
	\end{remark}
	\begin{remark}
		The NTK-based generator dynamics above are derived for the hard-constrained setting with the residual itself as the discriminator input. When the discriminator input is changed from the residual to squared residual, the dynamics become fundamentally different and take a closed-loop multiplicative form. The detailed derivation of the dynamics based on squared residual input and the corresponding residual-energy law under hard constraints is provided in Appendix \ref{app:x_vs_x2_dynamics}.
	\end{remark}
	\begin{remark}
		Beyond the hard-constrained setting, Appendix \ref{app:soft_adv_pinn} extends the analysis to soft-constrained adversarial PINNs, where we derive the coupled residual and energy dynamics for interior, boundary, and initial-condition violations, and further motivate our research.
	\end{remark}
	\begin{remark}
		More broadly, we emphasize that the present framework is not limited to standard GAN variants. Existing adversarially trained PINNs such as SA-PINN \cite{mcclenny2023self}, LA-PINN\cite{song2024loss}, and WAN\cite{zang2020weak} are all special cases of our analysis with the discriminator optimization restricted to structured function classes; the corresponding derivations are given in Appendix \ref{app:constrained_discriminator_unification}. This perspective also reveals a much broader and still underexplored design space of adversarial PINNs methods, which we summarize in Appendix \ref{app:design_space_adv_pinn}.
	\end{remark}

	\subsection{Residual-energy Law, Spectral Interpretation, and Practical Optimization}
	
	To characterize the optimization effect of the discriminator-induced weighting $\boldsymbol{\gamma}(t)$ (\cref{eq:gamma_vector_final}), we define a residual energy $E(t):=\frac12\|\mathbf r(t)\|^2$.
	\begin{theorem}[Residual-energy law under adversarial training]
		\label{thm:energy_law_bridge}
		Under the residual dynamics in \cref{eq:residual_dyn_compact_final}, the residual energy satisfies
		\begin{equation}\label{eq:energy_law_bridge}
			\frac{d}{dt}E(t)
			=
			\mathbf r(t)^\top K_{rr}^G(t)\boldsymbol{\gamma}(t) = S(t),
		\end{equation}
		where $S(t)$ is the instantaneous residual-energy variation rate induced by the current discriminator state. In particular,
		\begin{equation}
			S(t)<0 \Longrightarrow \frac{d}{dt}E(t)<0,
			\qquad
			S(t)>0 \Longrightarrow \frac{d}{dt}E(t)>0.
		\end{equation}
	\end{theorem}
	
	Theorem~\ref{thm:energy_law_bridge} shows that the first-order effect of adversarial interaction on the generator optimization is fully characterized by the scalar quantity $S(t)$. To interpret this quantity from a modal viewpoint, we consider the constant-NTK regime
	$K_{rr}^G(t)\approx K_{rr}^G$, decompose it as
	\begin{equation}
		K_{rr}^G = U\Lambda U^\top,
		\qquad
		\Lambda=\mathrm{Diag}(\lambda_1,\dots,\lambda_{N_r}),
		\qquad
		\lambda_j\ge 0,
	\end{equation}
	and define $\widetilde{\mathbf r}(t):=U^\top \mathbf r(t), \widetilde{\boldsymbol{\gamma}}(t):=U^\top \boldsymbol{\gamma}(t).$
	Then we obtain
	\begin{equation}\label{spectrum_format}
		S(t)
		=
		\mathbf r(t)^\top K_{rr}^G \boldsymbol{\gamma}(t)
		=
		\widetilde{\mathbf r}(t)^\top \Lambda \widetilde{\boldsymbol{\gamma}}(t)
		=
		\sum_{j=1}^{N_r}\lambda_j\,\widetilde r_j(t)\,\widetilde\gamma_j(t).
	\end{equation}

	\begin{remark}[Energy decay rate and spectral analysis]
		\label{rem:energy_rate_spectral_balance_short}
		By \cref{eq:energy_law_bridge,spectrum_format}, the decay of \(E(t)\) is jointly determined by the generator spectrum \(\{\lambda_j\}\), the residual amplitudes \(\widetilde r_j(t)\), and the discriminator-induced weights \(\widetilde\gamma_j(t)\). As training proceeds, modes with large $\lambda_j$s decay fast, then the energy decay increasingly depends on whether $\widetilde\gamma_j(t)$ remains aligned with the dominant modes. If it aligns with already decayed modes thereby underweighting the others, the energy decay slows down. 
	\end{remark}
	
	\begin{remark}[Representative failures]
		\label{rem:failure_modes_short}
		The analysis explains several representative failures in adversarial PINNs training: 
		(i) \emph{energy increase}, where \(S(t)>0\); 
		(ii) \emph{vanishing discriminator gradients}, where \(\boldsymbol{\gamma}(t)\approx 0\); and 
		(iii) \emph{noisy or misaligned weighting}, where $\widetilde\gamma_j(t)$ becomes unstable or ineffective. 
	\end{remark}
	More detailed analysis and justifications for Remark \ref{rem:energy_rate_spectral_balance_short} and \ref{rem:failure_modes_short} are in Appendix \ref{app:energy_decay}.
	
	To systematically address the above possible failures, one could explicitly control individual modes, where modes would need to be monitored during training and the discriminator feedback would need to be adjusted. However, a straightforward solution from \cref{eq:energy_law_bridge} is to ensure the continuous decrease of $E(t)$, \ie $S(t) < 0$. Therefore, we propose a practical approach below.
	
	We denote by $\theta^m$ the generator parameters at the beginning of the $m$-th iteration, and define residual 
	\begin{equation}
		\mathbf r^m:=[\mathcal R(x_1;\theta^m),\dots,\mathcal R(x_{N_r};\theta^m)]^\top.
	\end{equation}
	After a discriminator substep, we denote the frozen discriminator by $f^{m+\frac12}$, and define
	\begin{equation}\label{eq:gamma_m_final}
		\gamma_i^m
		=
		-\frac{1}{N_r}
		R'\!\big(f^{m+\frac12}(r_i^m)\big)\,
		\partial_r f^{m+\frac12}(r_i^m),
		\qquad
		\boldsymbol{\gamma}^m=[\gamma_1^m,\dots,\gamma_{N_r}^m]^\top.
	\end{equation}
	We also let $K_{rr}^{G,m}$ denote the NTK at $\theta^m$.
	
	\begin{proposition}[One-step discrete residual-energy variation]
		\label{prop:discrete_energy_bridge}
		A first-order generator update with frozen discriminator yields
		\[
		\mathbf r^{m+1}
		=
		\mathbf r^m+\eta_G K_{rr}^{G,m}\boldsymbol{\gamma}^m+\mathcal O(\eta_G^2).
		\]
		Consequently, if $E^m:=\frac12\|\mathbf r^m\|^2$, then
		\[
		E^{m+1}-E^m
		=
		\eta_G (\mathbf r^m)^\top K_{rr}^{G,m}\boldsymbol{\gamma}^m
		+\mathcal O(\eta_G^2).
		\]
	\end{proposition}
	\begin{algorithm}[H]
		\caption{Adaptive Alternating Adversarial PINNs training with Rollback}
		\label{alg:adaptive_adv_pinn_final}
		\begin{algorithmic}[1]
			\REQUIRE Initial parameters $\theta^0,\phi^0$, discriminator budget $T_D$, generator budget $T_G$
			\FOR{$m=0,1,2,\dots$}
			\STATE Build a fixed evaluation set and compute
			\[
			\mathbf r^m=
			[\mathcal R(x_1;\theta^m),\dots,\mathcal R(x_{N_r};\theta^m)]^\top,
			\qquad
			E^m=\frac12\|\mathbf r^m\|^2.
			\]
			
			\STATE \textbf{Discriminator:} with $\theta^m$ frozen, run up to $T_D$ discriminator updates; after each step compute
			\[
			S^{m,t}=(\mathbf r^m)^\top K_{rr}^{G,m}\boldsymbol{\gamma}^{m,t}.
			\]
			Keep the discriminator state attaining the smallest score, and set it as $\phi^{m+\frac12}$.
			
			\STATE \textbf{Generator:} with $\phi^{m+\frac12}$ frozen, run up to $T_G$ generator updates; after each step compute
			\[
			E^{m,t}=\frac12\|\mathbf r^{m,t}\|^2.
			\]
			Keep the generator state attaining the smallest residual energy, and set it as $\theta^{m+1}$.
			\ENDFOR
		\end{algorithmic}
	\end{algorithm}
	From Proposition \ref{prop:discrete_energy_bridge}, we derive Algorithm~\ref{alg:adaptive_adv_pinn_final}, which is essentially a dynamic update scheme for both the generator and the discriminator. Its key principle is that the discriminator should be retained only when it can induce a residual weighting that is able to improve the generator's ability to reduce the residual. At the same time, the generator update should be conducted only when the residual energy actually decreases. In this way, the discriminator acts not as an always accepted adversary but as an adaptive controller to balance the residual modes for continuous residual-energy decay.

	\section{Experiments}
	\label{sec:experiments}

	\subsection{Experimental setup}
	\label{sec:exp_setup}
	
	We compare DEQGAN \cite{bullwinkel2022deqgan}, GAN \cite{goodfellow2020generative}, LSGAN \cite{mao2017least}, and WGAN-GP \cite{arjovsky2017wasserstein} with their variants enhanced versions GAN-RB, LSGAN-RB, and WGAN-GP-RB, where ``RB'' denotes our rollback strategy. The benchmark suite includes the Poisson, Laplace, viscous Burgers, Reaction--Diffusion, and Klein--Gordon equations, covering linear and nonlinear, stationary and time-dependent PDEs.
	
	For fairness, all methods share the same generator architecture, while the discriminator architecture is chosen according to the adversarial objective. Except for DEQGAN, which follows its original tuning-dependent protocol, all methods are trained under the same default setting. We report training and validation errors as the main accuracy metrics, and additionally monitor the residual energy and the first-order quantity \(r^\top K_{rr}^{G}\gamma\). Full implementation details are deferred to Appendix~\ref{app:exp_setup_details}. 
	
	\subsection{Mechanistic validation: when and why adversarial training improves PINNs}
	\label{sec:exp_mechanistic}
	\begin{figure}[tb]
		\centering
		\begin{subfigure}[t]{0.49\linewidth}
			\centering
			\includegraphics[width=\linewidth]{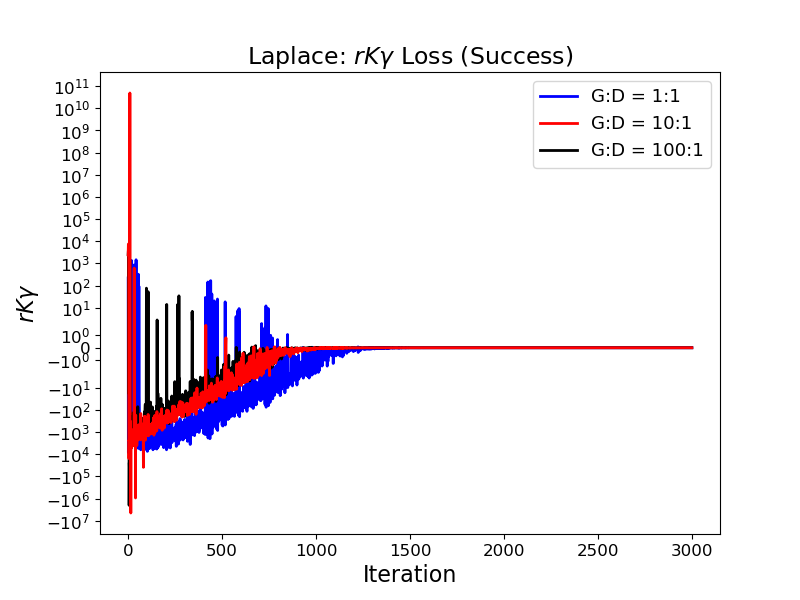}
		\end{subfigure}
		\hfill
		\begin{subfigure}[t]{0.49\linewidth}
			\centering
			\includegraphics[width=\linewidth]{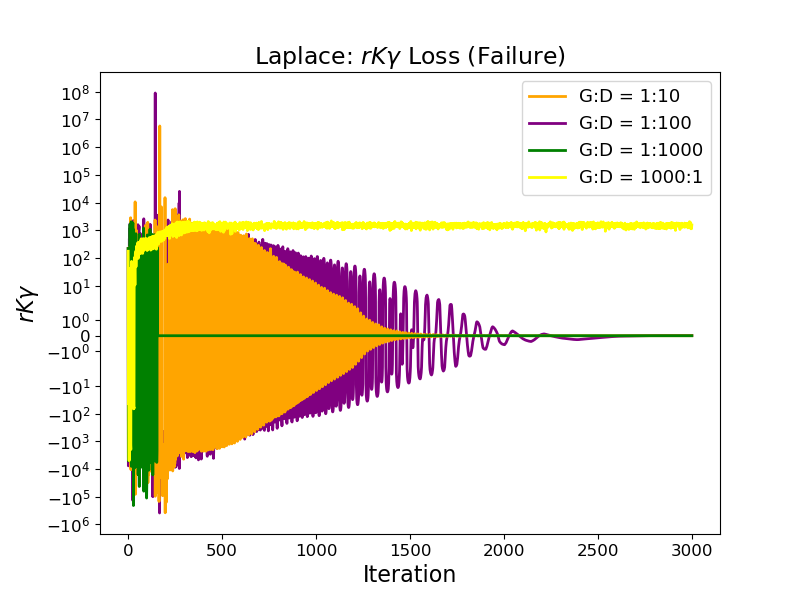}
		\end{subfigure}
		\caption{Left: successful training regimes under balanced or moderately imbalanced \(G:D\) ratios. Right: three representative failure modes caused by extreme update imbalance.}
		\label{fig:poigdloss}
	\end{figure}
	\begin{table}[tb]
		\centering
		\caption{Statistics of the positive part of \(r^\top K_{rr}^{G}\gamma\) for the three successful update ratios (\cref{fig:poigdloss} Left). Smaller values indicate fewer or weaker violations of the energy descent requirement.}
		\label{tab:rkgamma_stats}
		\begin{tabular}{lccccc}
			\toprule
			\(G:D\) & Positive ratio & Positive mean & Max value & \(L^1\) violation & \(L^2\) violation \\
			\midrule
			\(1:1\)   & 0.0220 & 18.1644 & 322.8066 & 1180.6849 & 190919.5174 \\
			\(10:1\)  & \textbf{0.0075} & \textbf{0.1429} & \textbf{1.8219} & \textbf{3.1449} & \textbf{4.0658} \\
			\(100:1\) & 0.0661 & 1.0747 & 75.9468 & 209.5676 & 10192.1785 \\
			\bottomrule
		\end{tabular}
	\end{table}
	We revisit DEQGAN under different generator--discriminator update ratios (Fig.~\ref{laplace1}) to clarify a central question of this paper: \emph{when} adversarial training improves PINNs, and \emph{why}. Rather than merely comparing schedules, we use the first-order quantity \(r^\top K_{rr}^{G}\gamma\) to distinguish successful alternating regimes from failed ones.
	
	A clear pattern emerges from Fig.~\ref{fig:poigdloss}. \cref{fig:poigdloss} Left shows three ratios \(G:D=100\!:\!1\), \(10\!:\!1\), and \(1\!:\!1\) which are successful regimes because their \(r^\top K_{rr}^{G}\gamma\) curves remain predominantly negative. This indicates that, on average, the discriminator-induced signal is aligned with residual descent after modulation by the generator kernel. However, the behaviors of the three are not identical. Table~\ref{tab:rkgamma_stats} reports aggregated statistics of the positive part of \(r^\top K_{rr}^{G}\gamma\), including the positive ratio, positive mean, maximum value, and \(L^1/L^2\) violation measures. \(G:D=10\!:\!1\) yields the smallest positive ratio and the weakest violation of the energy descent requirement, indicating the best overall performance. By contrast, although \(G:D=1\!:\!1\) has a smaller positive ratio than \(100\!:\!1\), its other indicators are much larger. So, successful adversarial training depends not only on whether \(r^\top K_{rr}^{G}\gamma\) is mostly negative, but also on how strongly it deviates from the energy descent regime when violations occur.
	
	\cref{fig:poigdloss} Right illustrates three failure modes described in Remark~\ref{rem:failure_modes_short}. \(G:D=1000\!:\!1\) corresponds to persistent misalignment: \(r^\top K_{rr}^{G}\gamma\) remains almost always positive between \(10^2\) and \(10^3\), indicating that the discriminator is updated too slowly to provide an effective gradient. Also, \(G:D=1\!:\!1000\) corresponds to discriminator overtraining: the curve rapidly collapses to zero after roughly 200 steps, showing that many residual samples receive nearly zero gradients, consistent with the classical analysis of vanishing generator gradients under an overly strong discriminator. Next, the intermediate discriminator-dominant ratios \(G:D=1\!:\!10\) and \(1\!:\!100\) correspond to unstable weighting: \(r^\top K_{rr}^{G}\gamma\) oscillates repeatedly across positive and negative regions, indicating that the effective alignment direction after modulation by \(K_{rr}^{G}\) is unstable.
	
	Overall, these results support our main claims. Adversarial training can improve PINNs but only when the alternating generator--discriminator dynamics induce a weighting field that is sufficiently aligned with the residual descent. \(r^\top K_{rr}^{G}\gamma\) is useful not only as a strict per-step certificate, but also as an aggregated indicator that separates successful regimes from qualitatively different failure modes.

	\subsection{Comparison on Diverse PDEs}
	\label{sec:exp_main_results}
	
	We next compare the proposed rollback strategy with existing adversarial PINNs baselines on the benchmark suite above. The main results are summarized in Table~\ref{tab:main_results}, while detailed per-equation curves are in Appendix \ref{app:comparision}.
	
	Four observations are most important. First, LSGAN is already a strong adversarial PINNs baseline, and even the classical GAN performs well under practical alternating optimization, which again suggests that the standard optimal-discriminator divergence picture is insufficient. Second, rollback is particularly effective for GAN and LSGAN: GAN-RB and LSGAN-RB consistently outperform their non-RB counterparts and achieve the best overall results on this benchmark suite. Third, the benefit is not uniform across objectives: rollback is weaker for WGAN-GP, and can even hurt performance on some problems, likely because the gradient penalty in WGAN-GP already regularizes the discriminator and the witness-style objective is less aligned with the proposed first-order criterion. Finally, LSGAN has rarely been explored in adversarial PINNs training so far but appears particularly promising, suggesting substantial room to revisit classical adversarial objectives from the perspective of practical alternating dynamics rather than nominal divergence form alone.
	
	\begin{table*}[tb]
		\centering
		\caption{Comparison of adversarial PINNs methods on diverse benchmark equations. Each entry is reported as ``Validation Error / Training Error''.}
		\label{tab:main_results}
		\resizebox{\textwidth}{!}{
			\begin{tabular}{lccccccc}
				\toprule
				Equation 
				& DEQGAN 
				& WGAN-GP 
				& WGAN-GP + RB 
				& LSGAN 
				& LSGAN + RB 
				& GAN
				& GAN + RB \\
				\midrule
				Poisson
				& 9.78e-13 / 9.75e-13
				& 1.33e-04 / 1.33e-04
				& 7.20e-05 / 7.18e-05
				& 1.02e-08 / 1.01e-08
				& \textbf{5.86e-13 / 5.84e-13} 
				& 4.80e-07 / 4.79e-07
				& 8.43e-12 / 8.47e-12\\
				
				Laplace
				& 3.65e-11 / 2.79e-11
				& 9.04e-06 / 6.30e-06
				& 6.51e-05 / 4.29e-05
				& 5.86e-10 / 4.25e-10
				& \textbf{3.49e-12 / 2.74e-12}
				& 8.12e-06 / 5.58e-06
				& 2.39e-11 / 1.69e-11 \\
				
				Viscous-Burgers
				& 6.49e-06 / \textbf{1.84e-04}
				& 7.36e-02 / 7.38e-02
				& 1.84e-01 / 1.84e-01
				& 2.35e-02 / 2.37e-02
				& 4.34e-05 / 3.49e-04
				& 6.48e-04 / 7.78e-04
				& \textbf{2.57e-06} / 3.87e-04\\
				
				Reaction-Diffusion 
				& 2.92e-09 / 2.45e-09
				& 2.13e-02 / 1.79e-02
				& 7.15e-02 / 5.96e-02
				& 8.16e-07 / 7.09e-07
				& 1.18e-09 / 9.24e-10
				& 9.81e-08 / 8.19e-08
				& \textbf{1.40e-13 / 1.08e-13} \\
				
				Klein-Gordon
				& 3.79e-06 / 3.16e-06
				& 1.62e-02 / 1.35e-02
				& 2.39e-04 / 1.78e-04
				& 3.91e-03 / 3.23e-03
				& 5.45e-05 / 4.36e-05
				& 1.65e-02 / 1.40e-02
				& \textbf{4.75e-10 / 4.02e-10} \\
				\bottomrule
		\end{tabular}}
	\end{table*}
	An important advantage of our method is that it avoids expensive hyperparameter search. Except for DEQGAN, all methods are trained with the same default learning rate \(10^{-3}\) and default optimizer settings. By contrast, DEQGAN relies on exhaustively tuning both generator/discriminator hyperparameters. Therefore, our method provides a more efficient route to strong PINNs performance. In addition, the rollback strategy can be used as a plugin in existing adversarial PINNs methods, also demonstrated by additional ablations in the appendix \ref{app:ablation}: (i) robustness under modified boundary conditions; (ii) compatibility with DEQGAN, and (iii) controlled comparisons under a fixed \(G:D=20\!:\!20\) update budget. These results indicate that the observed gains come from the rollback mechanism itself rather than merely from extra update steps.
	Our code is publicly available.\footnote{\url{https://github.com/realcrane/PINN-GAN}}
\section{Related work}
\label{sec:related_work}
\citet{wang2022and} studied PINNs training from the NTK perspective and explained optimization difficulties such as spectral bias and unbalanced convergence. However, their analysis is restricted to gradient descent on a fixed PINNs loss. By contrast, we study adversarial PINNs training, where the generator residual dynamics are continuously modulated by discriminator feedback. Classical GAN theory explains training through min--max optimization and divergence minimization \cite{mao2017least, nowozin2016f}, but this static picture does not explain why adversarial PINNs can still train successfully when the generated residual distribution and the zero target are initially far apart. \citet{franceschi2022neural} later analyzed GANs from an NTK viewpoint, but their setting is standard generative modeling rather than PDE residual learning. Our work instead studies adversarial PINNs from a residual-dynamics viewpoint. \citet{ciftci2024physics} proposed a physics-informed GAN framework and reported strong empirical performance, while \citet{bullwinkel2022deqgan} showed that adversarial objectives can improve differential equation solvers. However, these works mainly demonstrate that adversarial training can work well in practice, rather than explaining when and why it works. In contrast, we develop a residual-dynamics framework for adversarial PINNs and, based on it, derive a practical first-order rollback strategy. To the best of our knowledge, this is the first work to provide a unified theoretical framework for analyzing when and why adversarial training improves PINNs.
	
	\section{Limitations, Conclusions, and Future Work}
	\label{sec:limitation}
    
	
	There are limitations to our analysis. The spectral results rely on a fixed residual NTK $K_{rr}$, which provides a local approximation to the training dynamics. For finite-width networks, particularly when solving nonlinear PDEs, the residual NTK $K_{rr}^{(t)}$ may evolve as the network parameters change. The fixed-NTK assumption is used only for theoretical analysis, whereas the practical rollback algorithm recomputes quantities involving the current residual NTK at each iteration. Consequently, the rollback criterion does not require the kernel to remain constant throughout training. Nevertheless, the fixed-NTK analysis may not fully explain the behavior observed when the kernel changes substantially. Training can also oscillate under some PDEs and training settings despite rollback; identifying the mechanisms behind these oscillations requires further investigation.

    In summary, we have proposed a new analysis framework to explain the success of adversarially trained PINNs. The framework is rigorous in providing the much needed theoretical grounding, general in unifying the analysis of existing PINNs training, and practical by presenting effective and efficient training guidance. It reconciles the conflicts between the theories in GANs analysis and their empirically successful applications in training PINNs. 
	
    In future work, we will move away from the constant-NTK approximation and derive a more refined update strategy based on an evolving NTK spectrum. We include detailed discussions and limitations in Appendix \ref{limitation}.

    \section*{Acknowledgments}
    This work was supported in part by the UK Research and Innovation AIRR Innovator Award (0261-5654-9320-1) and the National Natural Science Foundation of China under Grants 92471108 and 12131008.
    
    \newpage
    \bibliographystyle{unsrtnat}
	\bibliography{document}

    \newpage
    \appendix
    
    \section*{Appendix Contents}
    \addcontentsline{toc}{section}{Appendix Contents}
    \startcontents[appendices]
    \printcontents[appendices]{}{1}{}
    \newpage
	\section{Preliminary Knowledge}\label{appendixA}
	\subsection{Generative adversarial networks}
	Generative Adversarial Networks (GANs) provide a powerful framework for learning probability distributions through an adversarial game between a generator and a discriminator. In this work, we adopt a unified functional perspective that allows several popular GAN variants to be represented under a common formulation.
	
	Let $\alpha_r$ denote the real data distribution and $\alpha_g^{\theta}$ denote the generated distribution induced by the generator $g(z;\theta)$, where $z \sim p_z$ is a latent variable. The discriminator is represented by a function $f(\cdot\ ; \phi): \mathcal{X} \rightarrow \mathbb{R}$. Here, $\theta$ and $\phi$ denote the trainable parameters of the generator and discriminator networks, respectively.
	
	We consider the following general adversarial objective:
	\begin{align}
		\max_{\phi}\mathcal{L}(\phi)
		=& \mathbb{E}_{x \sim \alpha_r} \big[ P(f(x;\phi)) \big]+\mathbb{E}_{x \sim \alpha_g^{\theta}} \big[ Q(f(x;\phi)) \big]\label{eq:general_gan1} \\	
		\min_{\theta}\mathcal{L}(\theta)=& \mathbb{E}_{x \sim \alpha_g^{\theta}} \big[ R(f(x; \phi)) \big]	\label{eq:general_gan2}
	\end{align}
	where $P(\cdot)$, $Q(\cdot)$ and $R(\cdot)$ are scalar-valued functions that determine the specific adversarial divergence or discrepancy being minimized between $\alpha_r$ and $\alpha_g^{\theta}$. This formulation encompasses a broad family of GAN objectives depending on the choice of $P$, $Q$ and $R$.
	
	\paragraph{Vanilla GAN.}
	The vanilla GAN objective corresponds to the minimization of the Jensenv--Shannon divergence between the real and generated distributions. This is recovered by choosing
	\begin{equation}\label{gan}
		P(t) =  \log \sigma(t), \quad
		Q(t) = R(t) = \log \big(1 - \sigma(t)\big),
	\end{equation}
	where $\sigma(\cdot)$ denotes the sigmoid function. 
	\paragraph{Least-Squares GAN (LSGAN).}
	The least-squares GAN replaces the logistic loss with a quadratic penalty, which improves gradient stability and alleviates saturation. This corresponds to choosing
	\begin{equation}\label{lsgan}
		P(t) =  -\frac{1}{2}(t-1)^2, \quad
		Q(t) = -\frac{1}{2}t^2, \quad
		R(t) = \frac{1}{2}(t-1)^2.
	\end{equation}
	
	\paragraph{Integral Probability Metric (IPM) GANs.}
	IPM-based GANs measure distribution discrepancy via a supremum over a constrained function class $\mathcal{F}$. In this setting, the adversarial objective becomes
	\begin{equation}
		\min_{\theta} \max_{f(x; \phi) \in \mathcal{F}}
		\mathbb{E}_{x \sim \alpha_r} f(x; \phi)
		-
		\mathbb{E}_{x \sim \alpha_g^{\theta}} f(x; \phi).
	\end{equation}
	This formulation is recovered from \ref{eq:general_gan1}-\ref{eq:general_gan2} by choosing
	\begin{equation}\label{wgan}
		P(t) = t, \quad Q(t) = R(t) = -t.
	\end{equation}
	together with structural constraints on $f$, such as Lipschitz continuity. Prominent examples include Wasserstein GANs (WGAN) and Wasserstein GANs with gradient penalty (WGAN-GP).  At this point, $f$ is often referred to as a critic in the literature. For consistency and ease of understanding, we will uniformly refer to it as the discriminator throughout this paper.
	

	\subsection{Physics-informed neural network}
	Physics-Informed Neural Networks (PINNs) provide a framework for solving partial differential equations (PDEs) by embedding governing physical laws into neural network training. Instead of relying solely on labeled data, PINNs incorporate PDE constraints through automatic differentiation.
	
	%
	Let $\Omega \subset \mathbb{R}^d$ denote a spatial--temporal domain with boundary $\partial \Omega$. We consider a general partial differential equation (PDE) expressed in operator form
	\begin{equation}
		\mathcal{N}(u)(x) =a(x), \quad x \in \Omega,
	\end{equation}
	where $u:\Omega \rightarrow \mathbb{R}$ represents the unknown solution field, $\mathcal{N}$ denotes a differential operator involving spatial and/or temporal derivatives of $u$, and $a(x)$ is a prescribed source or forcing term defined on $\Omega$. The governing equation is complemented by boundary and/or initial conditions given by
	\begin{equation}
		\mathcal{B}(u)(x) = b(x), \quad x \in \partial \Omega,
	\end{equation}
	where $\mathcal{B}$ is a boundary or initial operator specifying constraints on the solution, such as Dirichlet, Neumann, or Robin conditions, and $b(x)$ denotes the corresponding prescribed data on the boundary or initial manifold. In physics-informed neural networks (PINNs), the unknown solution $u(x)$ is approximated by a neural network parameterized by $\theta$, namely $u_\theta(x) \approx u(x)$, allowing the differential operators $\mathcal{N}$ and $\mathcal{B}$ to be evaluated through automatic differentiation during the training process.
	
	\paragraph{Physics-informed loss construction.}
	Let $\mu_\Omega$ denote a sampling measure over the interior domain $\Omega$, and $\mu_{\partial \Omega}$ denote a sampling measure over the boundary $\partial \Omega$. The PINNs objective can be written as
	
	\begin{equation}
		\mathcal{L}_{PINN}(\theta)
		=
		\lambda_r \, \mathbb{E}_{x \sim \mu_\Omega} \left[   \mathcal{N}[u_\theta](x) - a(x) \right]
		+
		\lambda_b \, \mathbb{E}_{x \sim \mu_{\partial \Omega}} 
		\left[  \mathcal{B}[u_\theta](x) - b(x)  \right].
	\end{equation}
	The expectations are approximated in practice via Monte Carlo sampling over collocation and boundary points.
	
	\paragraph{Hard constraint enforcement.}
	Besides penalty-based enforcement, physical constraints can also be incorporated directly into the network architecture through hard-constrained parameterizations. Instead of penalizing boundary or initial condition violations, the solution ansatz is parameterized as
	\begin{equation}
		u_\theta(x) = \mathcal{G}(x) + \mathcal{H}(x)\,\tilde{u}_\theta(x),
	\end{equation}
	where $\mathcal{G}(x)$ satisfies the prescribed boundary or initial conditions, and $\mathcal{H}(x)$ is a function that vanishes on $\partial \Omega$. The neural network $\tilde{u}_\theta(x)$ represents the unconstrained component of the solution.
	
	By construction, the resulting approximation $u_\theta(x)$ satisfies the constraints exactly for all parameter values $\theta$. Consequently, the training objective reduces to minimizing the interior PDE residual
	\begin{equation}\label{hard}
		\mathcal{L}_{PINN}(\theta)
		=
		\mathbb{E}_{x \sim \mu_\Omega}
		\left[\mathcal{N}[u_\theta](x) -a(x)
		\right].
	\end{equation}
	
	Hard constraint formulations often improve training stability and reduce sensitivity to loss weighting parameters, although they may introduce additional complexity in constructing suitable constraint functions. For clarity and conciseness in the theoretical derivation and exposition, we focus primarily on the hard-constrained formulation of PINNs \eqref{hard} in the subsequent analysis.
	
	%

	\subsection{Solving Partial Differential Equations via Adversarially Trained PINNs}\label{app:adversarial_pinn}
	To mitigate imbalance issues in physics-informed learning, we introduce an adversarial training framework for PINNs based on the vanilla GAN formulation.
	
	Let the PDE residual evaluated at collocation point $\{x_i\}_{i=1}^{N_r}$ be defined as $ \mathcal{R}(x_i;\theta) := \mathcal{N}[u_\theta](x_i)-a(x_i)$ and $\mathcal{T}(x_i) := 0.$
	Here, $\mathcal{R}(\cdot;\theta)$ denotes the PDE residual function induced by the generator parameterized by $\theta$, and is interpreted as the source of generated data in the adversarial framework. In contrast, \(\mathcal T\) denotes the target residual map corresponding to the ideal zero-residual state. The collocation points $\{x_i\}_{i=1}^{N_r}\subset\Omega$ are sampled from the interior sampling measure $\mu_\Omega$, where $N_r$ is the number of residual collocation points and $\Omega$ is the interior domain of the PDE. For notational convenience, define
	\begin{equation}
		r_i := \mathcal{R}(x_i;\theta), \qquad i=1,\dots,N_r,
	\end{equation}
	and collect these residual values into the residual vector
	\begin{equation}
		\mathbf{r} := (r_1,r_2,\dots,r_{N_r})^\top \in \mathbb{R}^{N_r}.
	\end{equation}
	Let
	\begin{equation}
		\hat{\mu}_\Omega := \frac{1}{N_r}\sum_{i=1}^{N_r}\delta_{x_i}
	\end{equation}
	denote the empirical measure associated with the collocation points, where $\delta_{x_i}$ is the Dirac measure centered at $x_i$. Then the generated residual distribution is given by the pushforward empirical measure
	\begin{equation}
		\hat{\mu}_\theta^{\,r}
		:=
		(\mathcal{R}_\theta)_\# \hat{\mu}_\Omega
		=
		\frac{1}{N_r}\sum_{i=1}^{N_r}\delta_{r_i},
	\end{equation}
	where $\mathcal{R}_\theta : x \mapsto \mathcal{R}(x;\theta)$ denotes the residual mapping, and $(\mathcal{R}_\theta)_\# \hat{\mu}_\Omega$ denotes the pushforward of the empirical collocation measure under this map. Therefore, the generated data distribution can be viewed as a finite Dirac mixture supported on the residual samples $\{r_i\}_{i=1}^{N_r}$, which serves as a discrete approximation of the pushforward measure $(\mathcal{R}_\theta)_\#\mu_\Omega$. In contrast, the real data distribution associated with the ideal PDE-satisfying state is concentrated at zero and is thus represented by the Dirac measure
	\begin{equation}
		\mu_{\mathrm{real}} = \delta_0.
	\end{equation}
	
	\paragraph{Adversarial objective.}
	We formulate the adversarial training between the solution network $u_\theta$ and the discriminator network $f(\cdot;\phi)$ within a general adversarial learning framework. Let $\mathcal{R}(x;\theta)$ denote the generated PDE residual induced by the solution network and $\mathcal{T}(x)$ represent the target residual distribution. The generator (solution network) is trained to minimize
	\begin{equation}
		\mathcal{L}(\theta) 
		=\frac{1}{N_r} \sum_{i=1}^{N_r} 
		R\Bigg(f \Big( \mathcal{R}\left(x_i;\theta\right) ;\phi\Big) \Bigg) ,
		\label{g}
	\end{equation}
	while the discriminator aims to distinguish target samples from generated residuals by maximizing
	\begin{equation}
		\mathcal{L}(\phi) 
		=\frac{1}{N_r} \sum_{i=1}^{N_r} \Bigg[
		P\Bigg(f \Big( \mathcal{T}\left(x_i\right) ;\phi\Big) \Bigg) 
		+ 
		Q\Bigg(f \Big( \mathcal{R}\left(x_i;\theta\right) ;\phi\Big) \Bigg) 
		\Bigg].
		\label{d}
	\end{equation}
	where $P(\cdot)$, $Q(\cdot)$, and $R(\cdot)$ are scalar loss functions defined in \eqref{gan}, \eqref{lsgan} and \eqref{wgan}. Different choices of these functions recover various GAN formulations.
	
	In the general setting, the real data correspond to the zero-residual target. For the original GAN and LSGAN formulations, the discriminator is trained to output 1 when the input equals the real data (i.e., $\mathcal{T}(x_i)=0$) and 0 when the input corresponds to generated residual samples ($\mathcal{R}(x_i;\theta)$).
	In contrast, under the WGAN formulation, the discriminator (critic) is trained to assign larger values to the real data (zero residual) and smaller values to generated residual samples, without restricting its output to the interval $[0,1]$.
	Meanwhile, the generator seeks to reduce the PDE residual magnitude so that the generated samples become indistinguishable from the zero-residual target, thereby enforcing the governing equation.
	
	\subsection{Background on the neural tangent kernel regime}
	\label{app:ntk_background}
	
	This appendix recalls the standard assumptions and conclusions of the neural tangent kernel (NTK) regime used in our analysis. The goal is not to rederive NTK theory from scratch, but to state clearly the conditions under which later kernel-based generator and discriminator dynamics are justified. The presentation follows the standard NTK viewpoint established for wide neural networks, where the network output becomes approximately linearized around initialization and the corresponding kernel remains nearly constant during training \cite{jacot2018neural}.
	
	Let \(g(x;\theta)\) denote a neural network with input \(x\in\mathcal X\) and parameters \(\theta\in\mathbb R^P\). We are interested in the regime where gradient-based training of \(g\) can be described directly at the level of its output function, rather than only through the nonlinear parameter trajectory.
	\begin{assumption}[Network architecture]
		\label{ass1}
		The network is a standard architecture, such as a fully connected, convolutional, or residual network, with standard hidden layers and affine transformations.
	\end{assumption}
	\begin{assumption}[Activation regularity]
		\label{ass2}
		The activation function is standard and sufficiently smooth. The activation can be any standard function: tanh, softplus, sigmoid, Gaussian, etc. 
	\end{assumption}
	\begin{assumption}[Nondegenerate bias]
		\label{ass3}
		Linear layers contain nonzero bias terms, which is standard in NTK formulations and helps avoid degenerate kernels in certain settings.
	\end{assumption}
	\begin{assumption}[Initialization]
		\label{ass4}
		The network parameters are randomly initialized with a standard width-dependent scaling. In the usual NTK parameterization, the preactivations remain of order one as the hidden widths grow, which ensures a nontrivial infinite-width limit. Under such initialization, the network output at initialization converges in law to a Gaussian process.
	\end{assumption}
	\begin{assumption}[Infinite-width limit]
		\label{ass5}
		The hidden widths \(n_1,\dots,n_{L-1}\) tend to infinity. In this regime, fluctuations due to finite width vanish, and both the network output at initialization and the NTK admit deterministic limiting descriptions.
	\end{assumption}
	
	Under these assumptions, the network output admits the first-order approximation
	\begin{equation}
		g(x;\theta(t))
		\approx
		g(x;\theta_0)
		+
		\nabla_\theta g(x;\theta_0)^\top\big(\theta(t)-\theta_0\big).
		\label{eq:app_ntk_linearization}
	\end{equation}
	This approximation is the basis of the NTK viewpoint: the nonlinear network is replaced by a linearized model around initialization, while the training dynamics are encoded by a kernel.
	
	The corresponding neural tangent kernel is defined by
	\begin{equation}
		k(x,x')
		:=
		\left\langle
		\nabla_\theta g(x;\theta_0),
		\nabla_\theta g(x';\theta_0)
		\right\rangle.
		\label{eq:app_ntk_def}
	\end{equation}
	For a finite set of points \(\{x_i\}_{i=1}^N\), the associated NTK Gram matrix is
	\begin{equation}
		K
		=
		\big[k(x_i,x_j)\big]_{i,j=1}^N.
		\label{eq:app_ntk_gram}
	\end{equation}
	
	The central NTK conclusions can then be summarized as follows.
	
	\begin{theorem}[Standard NTK regime \cite{jacot2018neural}]
		\label{thm:standard_ntk}
		Under Assumptions \ref{ass1}--\ref{ass5}, the following hold:
		\begin{enumerate}
			\item \textbf{Gaussian-process initialization.}  
			The random function \(g(\cdot;\theta_0)\) converges in law to a Gaussian process.
			
			\item \textbf{Deterministic limiting NTK.}  
			The random NTK \(k(x,x';\theta_0)\) converges to a deterministic limiting kernel \(k_\infty(x,x')\).
			
			\item \textbf{Kernel stability during training.}  
			Along training, the NTK remains approximately constant:
			\begin{equation}
				k(x,x';\theta(t))
				\approx
				k(x,x';\theta_0)
				\approx
				k_\infty(x,x').
			\end{equation}
			
			\item \textbf{Output dynamics in function space.}  
			If \(\mathbf g(t)=[g(x_1;\theta(t)),\dots,g(x_N;\theta(t))]^\top\) denotes the vector of network outputs on training points and \(\mathcal L(\mathbf g)\) is the loss viewed as a function of the outputs, then under gradient flow one has
			\begin{equation}
				\dot{\mathbf g}(t)
				\approx
				-K\,\nabla_{\mathbf g}\mathcal L(\mathbf g(t)),
				\label{eq:app_ntk_output_dyn}
			\end{equation}
			where \(K\) is the limiting NTK Gram matrix.
		\end{enumerate}
	\end{theorem}
	
	\paragraph{Interpretation.}
	The theorem states that, in the wide-network regime, training can be analyzed directly in function space. The nonlinear parameter dynamics are replaced by a kernel-preconditioned gradient flow on the outputs. This is precisely why the NTK framework is useful for studying residual dynamics, modewise convergence, and kernel-induced weighting mechanisms.
	
	Hence the convergence rate is governed by the spectrum of the kernel matrix \(K\): components associated with large eigenvalues decay faster, while components associated with small eigenvalues decay more slowly. This spectral viewpoint is one of the key motivations for using NTK analysis in training-dynamics problems.
	
	\paragraph{Scope of the approximation.}
	The NTK regime is an asymptotic and mechanism-level approximation. Its justification relies on wide hidden layers, standard smooth parameterization, and small parameter drift. Therefore, it is not a complete description of all finite-width training phenomena. Nevertheless, it provides a mathematically tractable and often accurate description of the early and intermediate training dynamics of overparameterized neural networks, which is why it serves as the theoretical basis for the kernel-based analysis in this work.
	
	\paragraph{Relevance for the present paper.}
	In the sequel, this NTK viewpoint will be applied separately to the generator and the discriminator. On the generator side, it yields a kernel description of residual evolution. On the discriminator side, it motivates a function-space flow under alternating training. The resulting interaction between these two kernel-induced dynamics is the basis of our later residual-space analysis.
	
	\section{Detailed derivations of discriminator and generator dynamics}
	\label{app:disc_gen_dynamics}
	
	In this appendix, we provide the detailed derivations of the discriminator and generator dynamics used in the main text. Throughout, we follow the notation introduced in Section~\ref{sec:adversarial_pinn}. Recall that the adversarial objective is written in the unified form
	\begin{equation}
		\mathcal L_D(\phi)
		=
		\frac{1}{N_r}\sum_{i=1}^{N_r}
		\Big[
		P\!\big(f(\mathcal T(x_i);\phi)\big)
		+
		Q\!\big(f(\mathcal R(x_i;\theta);\phi)\big)
		\Big],
	\end{equation}
	for the discriminator, and
	\begin{equation}
		\mathcal L_G(\theta)
		=
		\frac{1}{N_r}\sum_{i=1}^{N_r}
		R\!\big(f(\mathcal R(x_i;\theta);\phi)\big)
	\end{equation}
	for the generator.
	
	Under alternating optimization, gradient ascent is applied to $\phi$ and gradient descent is applied to $\theta$. The corresponding gradients are
	\begin{align}
		\frac{\partial \mathcal{L}_G(\theta)}{\partial \theta}
		&=
		\frac{1}{N_r}\sum_{i=1}^{N_r}
		R'\!\Big(f(\mathcal R(x_i;\theta);\phi)\Big)\,
		\frac{\partial f(\mathcal R(x_i;\theta);\phi)}{\partial \mathcal R(x_i;\theta)}
		\frac{\partial \mathcal R(x_i;\theta)}{\partial \theta(t)},
		\label{eq:appendix_grad_G}
	\end{align}
	and
	\begin{align}
		\frac{\partial \mathcal{L}_D(\phi)}{\partial \phi}
		&=
		\frac{1}{N_r}\sum_{i=1}^{N_r}
		\Bigg[
		P'\!\Big(f(\mathcal T(x_i);\phi)\Big)
		\frac{\partial f(\mathcal T(x_i);\phi)}{\partial \phi(t)}
		+
		Q'\!\Big(f(\mathcal R(x_i;\theta);\phi)\Big)
		\frac{\partial f(\mathcal R(x_i;\theta);\phi)}{\partial \phi(t)}
		\Bigg].
		\label{eq:appendix_grad_D}
	\end{align}
	Hence the discrete parameter updates are
	\begin{equation}
		\theta(n+1)=\theta(n)-a(n)\frac{\partial \mathcal L_G(\theta)}{\partial \theta},
	\end{equation}
	\begin{equation}
		\phi(n+1)=\phi(n)+b(n)\frac{\partial \mathcal L_D(\phi)}{\partial \phi},
	\end{equation}
	where $a(n)$ and $b(n)$ denote the generator and discriminator learning rates, respectively. Passing formally to the continuous-time limit yields the gradient-flow system
	\begin{equation}\label{eq:appendix_d_flow}
		\dot{\phi}(t)=\frac{\partial \mathcal L_D(\phi(t))}{\partial \phi(t)},
	\end{equation}
	\begin{equation}\label{eq:appendix_g_flow}
		\dot{\theta}(t)=-\frac{\partial \mathcal L_G(\theta(t))}{\partial \theta(t)}.
	\end{equation}
	
	\subsection{Discriminator dynamics in parameter space}
	\label{app:disc_dynamics}
	
	We first derive the discriminator evolution. For a fixed generator state $\theta$, we have defined the residual samples
	\begin{equation}
		r_i:=\mathcal R(x_i;\theta),
		\qquad i=1,\dots,N_r.
	\end{equation}
	Then \eqref{eq:appendix_d_flow} becomes
	\begin{equation}\label{eq:appendix_disc_param_flow}
		\dot{\phi}(t)
		=
		\frac{1}{N_r}\sum_{i=1}^{N_r}
		\Bigg[
		P'\!\big(f(\mathcal T(x_i);\phi(t))\big)\,
		\frac{\partial f(\mathcal T(x_i);\phi(t))}{\partial \phi(t)}
		+
		Q'\!\big(f(r_i;\phi(t))\big)\,
		\frac{\partial f(r_i;\phi(t))}{\partial \phi(t)}
		\Bigg].
	\end{equation}
	This is the discriminator gradient flow in parameter space.
	
	Let
	\begin{equation}
		k_t^D(x,y)
		:=
		\left\langle
		\frac{\partial f(x;\phi(t))}{\partial \phi(t)},
		\frac{\partial f(y;\phi(t))}{\partial \phi(t)}
		\right\rangle
	\end{equation}
	denote the discriminator NTK. Under the standard Assumptions \ref{ass1}-\ref{ass5}, this kernel remains approximately constant during training:
	\begin{equation}
		k_t^D(x,y)\approx k^D(x,y).
	\end{equation}
	For any input $x$ in the discriminator input space, the time derivative of the discriminator output is
	\begin{align}
		\frac{d}{dt}f(x;\phi(t))
		&=
		\left\langle
		\frac{\partial f(x;\phi(t))}{\partial \phi(t)},
		\dot{\phi}(t)
		\right\rangle \notag\\
		&=
		\left\langle
		\frac{\partial f(x;\phi(t))}{\partial \phi(t)},
		\frac{1}{N_r}\sum_{i=1}^{N_r}
		\Big[
		P'\!\big(f(\mathcal T(x_i);\phi(t))\big)
		\frac{\partial f(\mathcal T(x_i);\phi(t))}{\partial \phi(t)}
		+
		Q'\!\big(f(r_i;\phi(t))\big)
		\frac{\partial f(r_i;\phi(t))}{\partial \phi(t)}
		\Big]
		\right\rangle \notag\\
		&=
		\frac{1}{N_r}\sum_{i=1}^{N_r}
		\Big[
		P'\!\big(f(\mathcal T(x_i);\phi(t))\big)\,
		k_t^D\!\big(x,\mathcal T(x_i)\big)
		+
		Q'\!\big(f(r_i;\phi(t))\big)\,
		k_t^D(x,r_i)
		\Big].
	\end{align}
	Under the constant-NTK approximation, this reduces to
	\begin{equation}\label{eq:appendix_disc_functional_flow}
		\frac{d}{dt}f(x;\phi(t))
		=
		\frac{1}{N_r}\sum_{i=1}^{N_r}
		\Big[
		P'\!\big(f(\mathcal T(x_i);\phi(t))\big)\,
		k^D\!\big(x,\mathcal T(x_i)\big)
		+
		Q'\!\big(f(r_i;\phi(t))\big)\,
		k^D(x,r_i)
		\Big].
	\end{equation}
	In particular, evaluating at $x=r_j$ yields
	\begin{equation}\label{eq:appendix_disc_functional_flow_rj}
		\frac{d}{dt}f(r_j;\phi(t))
		=
		\frac{1}{N_r}\sum_{i=1}^{N_r}
		\Big[
		P'\!\big(f(\mathcal T(x_i);\phi(t))\big)\,
		k^D\!\big(r_j,\mathcal T(x_i)\big)
		+
		Q'\!\big(f(r_i;\phi(t))\big)\,
		k^D(r_j,r_i)
		\Big].
	\end{equation}
	
	Integrating \eqref{eq:appendix_disc_functional_flow} from $0$ to $t$, we obtain the representation
	\begin{equation}\label{eq:appendix_disc_integral_repr}
		f(x;\phi(t))
		=
		f_0(x)
		+
		\int_0^t
		\frac{1}{N_r}\sum_{i=1}^{N_r}
		\Big[
		P'\!\big(f(\mathcal T(x_i);\phi(s))\big)\,
		k^D\!\big(x,\mathcal T(x_i)\big)
		+
		Q'\!\big(f(r_i;\phi(s))\big)\,
		k^D(x,r_i)
		\Big]\,ds,
	\end{equation}
	where $f_0(x):=f(x;\phi(0))$.

	\subsection{Functional-space notation for discriminator training}
	\label{app:disc_functional_notation}
	
	Before deriving the discriminator dynamics in function space, we briefly introduce the notation that will be used in the following subsection. The purpose of this step is to rewrite the discriminator update not only at the parameter level \(\phi\), but also at the level of the discriminator function \(f(\cdot;\phi)\) acting on the residual domain.
	
	\paragraph{Residual-domain training measure.}
	For a fixed generator state \(\theta\), the discriminator is trained on two types of samples in the residual domain:
	\begin{itemize}
		\item the real target samples \(\mathcal T(x_i)=0\),
		\item the generated residual samples \(\mathcal R(x_i;\theta)=r_i\).
	\end{itemize}
	Accordingly, we introduce the empirical training measure on the residual domain,
	\begin{equation}
		\hat\gamma_\theta^{\,r}
		:=
		\frac{1}{2N_r}\sum_{i=1}^{N_r}\delta_{0}
		+
		\frac{1}{2N_r}\sum_{i=1}^{N_r}\delta_{r_i},
		\qquad
		r_i=\mathcal R(x_i;\theta).
		\label{eq:gamma_hat_residual}
	\end{equation}
	Thus, \(\hat\gamma_\theta^{\,r}\) is the empirical measure supported on the real zero state and the generated residual samples. It plays the role of the discriminator-side training distribution in residual space.
	
	\paragraph{Discriminator as a function on residual space.}
	Instead of only viewing the discriminator through its parameter vector \(\phi\), we may also regard it as a time-dependent function
	\begin{equation}
		f_t(\cdot):=f(\cdot;\phi(t)),
	\end{equation}
	defined on the residual variable \(r\in\mathbb R\). In this viewpoint, the discriminator training flow \eqref{eq:appendix_d_flow} induces an evolution equation for the function \(f_t\) itself.
	
	\paragraph{Functional gradient.}
	Let \(\mathcal L_D(f)\) denote the discriminator objective viewed as a functional of the discriminator function \(f\). Its gradient with respect to the empirical measure \(\hat\gamma_\theta^{\,r}\), denoted by
	\[
	\nabla_{\hat\gamma_\theta^{\,r}}\mathcal L_D(f),
	\]
	is the function on the support of \(\hat\gamma_\theta^{\,r}\) such that, for every perturbation \(\psi\in L^2(\hat\gamma_\theta^{\,r})\),
	\begin{equation}
		\left.\frac{d}{d\varepsilon}\mathcal L_D(f+\varepsilon\psi)\right|_{\varepsilon=0}
		=
		\Big\langle
		\nabla_{\hat\gamma_\theta^{\,r}}\mathcal L_D(f),\,
		\psi
		\Big\rangle_{L^2(\hat\gamma_\theta^{\,r})}.
		\label{eq:functional_gradient_def}
	\end{equation}
	In other words, \(\nabla_{\hat\gamma_\theta^{\,r}}\mathcal L_D(f)\) is the discriminator-side functional gradient evaluated on the empirical residual training measure.
	
	\paragraph{Discriminator NTK and RKHS.}
	Assume now that the discriminator is in the infinite-width or constant-NTK regime, and let \(k^D(r,r')\) denote its Neural Tangent Kernel on the residual domain. Associated with \(k^D\) and the empirical measure \(\hat\gamma_\theta^{\,r}\), one defines the reproducing kernel Hilbert space (RKHS), denoted by
	\[
	\mathcal H_{k^D}^{\,\hat\gamma_\theta^{\,r}},
	\]
	which is the function space generated by the kernel sections \(k^D(r,\cdot)\) over points \(r\) in the support of \(\hat\gamma_\theta^{\,r}\).
	
	\paragraph{Kernel integral operator.}
	The corresponding kernel integral operator
	\[
	T_{k^D,\hat\gamma_\theta^{\,r}}
	:
	L^2(\hat\gamma_\theta^{\,r})
	\to
	\mathcal H_{k^D}^{\,\hat\gamma_\theta^{\,r}}
	\]
	is defined by
	\begin{equation}
		\big(T_{k^D,\hat\gamma_\theta^{\,r}}h\big)(r)
		:=
		\int k^D(r,\tilde r)\,h(\tilde r)\,d\hat\gamma_\theta^{\,r}(\tilde r).
		\label{eq:kernel_integral_operator}
	\end{equation}
	Since \(\hat\gamma_\theta^{\,r}\) is empirical, this becomes
	\begin{equation}
		\big(T_{k^D,\hat\gamma_\theta^{\,r}}h\big)(r)
		=
		\frac{1}{2N_r}\sum_{i=1}^{N_r}k^D(r,0)\,h(0)
		+
		\frac{1}{2N_r}\sum_{i=1}^{N_r}k^D(r,r_i)\,h(r_i).
		\label{eq:kernel_integral_operator_empirical}
	\end{equation}
	Therefore, \(T_{k^D,\hat\gamma_\theta^{\,r}}\) maps a function defined only on the discrete training support into a kernel-smoothed function on the whole residual domain.
	
	\paragraph{Interpretation.}
	The key point is that the discriminator loss is only evaluated on the empirical support \(\{0,r_1,\dots,r_{N_r}\}\), whereas the generator requires the input derivative \(\partial_r f_t(r)\) of the trained discriminator as a genuine function of the residual variable. The kernel operator \(T_{k^D,\hat\gamma_\theta^{\,r}}\) provides precisely this extension: it turns the empirical discriminator training signal into a smooth function on residual space, shaped by the architecture-dependent NTK \(k^D\).
	
	With these notations in place, the next subsection rewrites discriminator training as a functional gradient flow in residual space.
	
	\subsection{Discriminator analysis under alternating training: an NTK interpretation in residual space}
	\label{app:disc_ntk_residual_space}We adapt the discriminator-side NTK viewpoint of alternating GAN training to our residual-space formulation of adversarial PINNs. Concretely, the general NTK function-space flow of the discriminator is adapted from the discriminator-side NTK framework of Franceschi et al.~\cite{franceschi2022neural}, while our contribution here is to rewrite and interpret it in the present residual setting, where the generated distribution is the empirical residual measure \(\hat\mu_\theta^{\,r}\) and the real distribution is the zero state \(\delta_0\). This residual-space reinterpretation is what allows the discriminator flow to be connected directly to the generator-side residual dynamics studied in the main text.
	With the functional-space notation introduced above, the discriminator training dynamics under a fixed generator state can be written as
	\begin{equation}
		\partial_t f_t
		=
		T_{k^D,\hat\gamma_\theta^{\,r}}
		\Big(
		\nabla_{\hat\gamma_\theta^{\,r}} \mathcal L_D(f_t)
		\Big).
		\label{eq:disc_functional_flow_residual}
	\end{equation}
	This equation shows that, under alternating optimization, the discriminator should not be viewed as an arbitrary optimizer on the empirical residual samples, but rather as a finite-time function-space flow shaped jointly by the empirical residual training measure \(\hat\gamma_\theta^{\,r}\) and the discriminator NTK \(k^D\). In particular, the discriminator architecture enters generator training through the kernel operator in \eqref{eq:disc_functional_flow_residual}, and therefore directly affects the gradient field induced on residual space.
	
	A key consequence of \eqref{eq:disc_functional_flow_residual} is that the trained discriminator is uniquely determined by its initial state and admits the representation
	\begin{equation}
		f_t
		=
		f_0
		+
		T_{k^D,\hat\gamma_\theta^{\,r}}
		\left(
		\int_0^t
		\nabla_{\hat\gamma_\theta^{\,r}} \mathcal L_D(f_s)\,ds
		\right).
		\label{eq:disc_general_solution_residual}
	\end{equation}
	Hence the increment \(f_t-f_0\) lies in the RKHS generated by \(k^D\), which means that discriminator training amounts to adding to the initial network a kernel-smoothed correction determined by the empirical residual samples and the real zero state. This observation is important in our setting because the discriminator loss is only evaluated on the discrete support \(\{0,r_1,\dots,r_{N_r}\}\), whereas the generator dynamics require the input derivative \(\partial_r f_t(r)\) as a function defined on the whole residual domain.
	
	Under standard architectural assumptions, the discriminator NTK is differentiable, and therefore so is the trained discriminator \(f_t\). As a result, the derivative field \(\partial_r f_t(r)\) appearing in the generator-side residual dynamics is well-defined. This resolves the gradient-indeterminacy issue that would arise if one treated the empirical discriminator as an unconstrained optimum only defined on a discrete support. In our framework, the discriminator derivative is instead a finite-time, NTK-smoothed quantity determined by both the sample distribution and the discriminator architecture.
	
	This interpretation fits directly into the generator dynamics developed in the main text \eqref{eq:residual_dyn_compact_final}. Once the discriminator is frozen after its substep, the residual evolution of the generator takes the form
	\begin{equation}
		\dot{\mathbf r}
		=
		K_{rr}^G(\theta)\,\Gamma^D\,\mathbf 1,
	\end{equation}
	where the diagonal entries of \(\Gamma^D\) depend on \eqref{eq:gamma_entry_final}:
	\[
	\gamma_i=-\frac{1}{N_r}\,R'\!\big(f_t(r_i)\big)\,\partial_r f_t(r_i).
	\]
	Therefore, the discriminator-induced weighting in adversarial PINNs is not produced by an abstract optimal critic, but by the finite-time residual-space vector field generated by \eqref{eq:disc_functional_flow_residual}. From this viewpoint, the discriminator plays the role of an architecture-dependent smoothing and weighting operator acting on the empirical residual distribution.
	\subsubsection{The IPM case: kernel witness function on residual samples}
	\label{app:disc_ipm_residual}
	
	We now reinterpret the general discriminator-side NTK framework of~\cite{franceschi2022neural} in our residual-space setting. The key point here is that the discriminator behaves as a kernel witness function measuring the discrepancy between the empirical residual distribution and the zero target. This makes the resulting generator-side weighting especially explicit.
	
	\begin{proposition}[IPM discriminator flow as a kernel witness function on residual space]
		\label{prop:ipm_disc_residual}
		Fix the generated residual distribution \(\hat\mu_\theta^{\,r}\) and the target \(\delta_0\). In the IPM case, let \(k^D\) denote the discriminator kernel. Then:
		\begin{enumerate}
			\item The discriminator evolves linearly in function space as
			\begin{equation}
				f_t
				=
				f_0+t\,f_\theta^\ast,
				\label{eq:disc_ipm_solution_residual_prop}
			\end{equation}
			where \(f_\theta^\ast\) is the unnormalized MMD witness function associated with \(k^D\).
			
			\item In residual-space notation, the witness function is given by
			\begin{equation}
				f_\theta^\ast(r)
				=
				-\mathbb E_{\tilde r\sim \hat\mu_\theta^{\,r}}\!\big[k^D(\tilde r,r)\big]
				+
				k^D(0,r),
				\label{eq:mmd_witness_residual_prop}
			\end{equation}
			or equivalently
			\begin{equation}
				f_\theta^\ast(r)
				=
				-\frac{1}{N_r}\sum_{i=1}^{N_r}k^D(r_i,r)+k^D(0,r).
				\label{eq:mmd_witness_empirical_residual_prop}
			\end{equation}
			
			\item The induced generator-side sample weighting is
			\begin{equation}
				\gamma_i^{\mathrm{IPM}}
				=
				\frac{1}{N_r}\,\partial_r f_t(r_i)
				=
				\frac{1}{N_r}\,\partial_r f_0(r_i)
				+
				\frac{t}{N_r}\,\partial_r f_\theta^\ast(r_i).
				\label{eq:wgan_gamma_ntk_witness_prop}
			\end{equation}
		\end{enumerate}
	\end{proposition}
	
	\paragraph{Proof.}
	In the IPM setting, the discriminator-side NTK dynamics admit an explicit linear form in function space. For fixed \(\hat\mu_\theta^{\,r}\) and target \(\delta_0\), the discriminator moves in the direction of the kernel witness function associated with the discrepancy between the generated residual distribution and the zero target. Therefore,
	\begin{equation}
		f_t=f_0+t\,f_\theta^\ast,
	\end{equation}
	which proves \eqref{eq:disc_ipm_solution_residual_prop}.
	
	By definition, the corresponding unnormalized witness function is
	\begin{equation}
		f_\theta^\ast(r)
		=
		-\mathbb E_{\tilde r\sim \hat\mu_\theta^{\,r}}\!\big[k^D(\tilde r,r)\big]
		+
		k^D(0,r),
	\end{equation}
	which proves \eqref{eq:mmd_witness_residual_prop}. Since
	\[
	\hat\mu_\theta^{\,r}
	=
	\frac{1}{N_r}\sum_{i=1}^{N_r}\delta_{r_i},
	\]
	the empirical representation \eqref{eq:mmd_witness_empirical_residual_prop} follows immediately:
	\begin{equation}
		f_\theta^\ast(r)
		=
		-\frac{1}{N_r}\sum_{i=1}^{N_r}k^D(r_i,r)+k^D(0,r).
	\end{equation}
	
	Finally, in the IPM-GAN case, the generator-side weighting is given by
	\begin{equation}
		\gamma_i^{\mathrm{IPM}}
		=
		\frac{1}{N_r}\,\partial_r f_t(r_i).
	\end{equation}
	Substituting \(f_t=f_0+t\,f_\theta^\ast\) yields
	\begin{equation}
		\gamma_i^{\mathrm{IPM}}
		=
		\frac{1}{N_r}\,\partial_r f_0(r_i)
		+
		\frac{t}{N_r}\,\partial_r f_\theta^\ast(r_i),
	\end{equation}
	which proves \eqref{eq:wgan_gamma_ntk_witness_prop}.
	\hfill $\square$
	
	\paragraph{Interpretation.}
	Proposition~\ref{prop:ipm_disc_residual} shows that, in the IPM case, the discriminator is most naturally interpreted as a kernel witness function on residual space. Unlike the LSGAN discriminator, which relaxes toward a regression target, or the vanilla GAN discriminator, which behaves as a nonlinear kernel classifier, the IPM discriminator directly represents the discrepancy between the empirical residual distribution and the zero target through the kernel \(k^D\).
	
	This makes the generator-side effect especially transparent: the discriminator induces a residual-space gradient field through \(\partial_r f_t(r_i)\), and this field is precisely the derivative of the witness function. Therefore, the IPM discriminator contributes not merely a notion of sample realism, but an explicit sample-dependent preconditioning field for residual descent. In the language of this paper, this is the clearest example of how discriminator training reshapes generator dynamics through residual-space weighting.
	
	\subsubsection{The LSGAN case: kernel-smoothed regression target for our choice of labels}
	\label{app:disc_lsgan_residual}
	
	We next specialize the discriminator-side NTK framework to our LSGAN formulation in residual space. The main point here is not to introduce a new NTK tool, but to reinterpret the discriminator dynamics under our labeling convention: generated residual samples are assigned target value \(0\), while the real zero state is assigned target value \(1\). Under this choice, the discriminator becomes a kernel-smoothed regression field on residual space, whose derivative then induces the generator-side residual weighting.
	
	\begin{proposition}[LSGAN discriminator flow as kernel-smoothed regression on residual space]
		\label{prop:lsgan_disc_residual}
		Consider the discriminator objective
		\begin{equation}
			\mathcal L_D^{\mathrm{LS}}(f)
			=
			-\frac{1}{2}\,\mathbb E_{r\sim \hat\mu_\theta^{\,r}}\!\big[f(r)^2\big]
			-\frac{1}{2}\,\mathbb E_{y\sim \delta_0}\!\big[(f(y)-1)^2\big].
			\label{eq:our_lsgan_disc_objective_prop}
		\end{equation}
		Let
		\begin{equation}
			\hat\gamma_\theta^{\,r}
			=
			\frac12 \hat\mu_\theta^{\,r}
			+
			\frac12 \delta_0,
		\end{equation}
		and define the regression target
		\begin{equation}
			\rho_\theta
			:=
			\frac{d\delta_0}{d(\delta_0+\hat\mu_\theta^{\,r})}.
			\label{eq:rho_theta_ours_prop}
		\end{equation}
		Then the following statements hold:
		
		\begin{enumerate}
			\item The \(L^2(\hat\gamma_\theta^{\,r})\) functional gradient of \(\mathcal L_D^{\mathrm{LS}}\) is
			\begin{equation}
				\nabla_{\hat\gamma_\theta^{\,r}} \mathcal L_D^{\mathrm{LS}}(f)
				=
				2(\rho_\theta-f).
				\label{eq:our_lsgan_functional_gradient_prop}
			\end{equation}
			
			\item Consequently, the discriminator NTK flow is
			\begin{equation}
				\partial_t f_t
				=
				2\,T_{k^D,\hat\gamma_\theta^{\,r}}(\rho_\theta-f_t).
				\label{eq:our_lsgan_disc_flow_prop}
			\end{equation}
			
			\item The resulting function-space dynamics are linear, and the solution is
			\begin{equation}
				f_t
				=
				\exp\!\big(-2t\,T_{k^D,\hat\gamma_\theta^{\,r}}\big)(f_0-\rho_\theta)
				+
				\rho_\theta.
				\label{eq:our_lsgan_disc_solution_prop}
			\end{equation}
			Equivalently, if
			\begin{equation}
				\varphi_t(x):=e^{-2tx}-1,
			\end{equation}
			then
			\begin{equation}
				f_t
				=
				f_0
				+
				\varphi_t\!\big(T_{k^D,\hat\gamma_\theta^{\,r}}\big)(f_0-\rho_\theta).
				\label{eq:our_lsgan_disc_solution_phi_prop}
			\end{equation}
			
			\item The induced generator-side sample weighting is
			\begin{equation}
				\gamma_i^{\mathrm{LSGAN}}
				=
				-\frac{1}{N_r}
				\big(f_t(r_i)-1\big)\,
				\partial_r f_t(r_i).
				\label{eq:our_lsgan_gamma_prop}
			\end{equation}
		\end{enumerate}
	\end{proposition}
	
	\paragraph{Proof.}
	Starting from \eqref{eq:our_lsgan_disc_objective_prop}, we rewrite the discriminator loss with respect to the empirical measure \(\hat\gamma_\theta^{\,r}\). Since
	\[
	\hat\gamma_\theta^{\,r}
	=
	\frac12 \hat\mu_\theta^{\,r}
	+
	\frac12 \delta_0,
	\]
	we may express \(\mathcal L_D^{\mathrm{LS}}\) as
	\begin{equation}
		\mathcal L_D^{\mathrm{LS}}(f)
		=
		\int
		\left[
		-\frac12 \frac{d\hat\mu_\theta^{\,r}}{d\hat\gamma_\theta^{\,r}} f^2
		-\frac12 \frac{d\delta_0}{d\hat\gamma_\theta^{\,r}} (f-1)^2
		\right]
		d\hat\gamma_\theta^{\,r}.
	\end{equation}
	Taking the functional derivative with respect to \(f\) gives
	\begin{align}
		\nabla_{\hat\gamma_\theta^{\,r}} \mathcal L_D^{\mathrm{LS}}(f)
		&=
		-\frac{d\hat\mu_\theta^{\,r}}{d\hat\gamma_\theta^{\,r}} f
		-\frac{d\delta_0}{d\hat\gamma_\theta^{\,r}} (f-1) \notag\\
		&=
		-\left(
		\frac{d\hat\mu_\theta^{\,r}}{d\hat\gamma_\theta^{\,r}}
		+
		\frac{d\delta_0}{d\hat\gamma_\theta^{\,r}}
		\right)f
		+
		\frac{d\delta_0}{d\hat\gamma_\theta^{\,r}}.
	\end{align}
	Since
	\[
	\hat\gamma_\theta^{\,r}
	=
	\frac12(\hat\mu_\theta^{\,r}+\delta_0),
	\]
	we have
	\[
	\frac{d\hat\mu_\theta^{\,r}}{d\hat\gamma_\theta^{\,r}}
	+
	\frac{d\delta_0}{d\hat\gamma_\theta^{\,r}}
	=2,
	\qquad
	\frac{d\delta_0}{d\hat\gamma_\theta^{\,r}}
	=2\rho_\theta.
	\]
	Therefore,
	\begin{equation}
		\nabla_{\hat\gamma_\theta^{\,r}} \mathcal L_D^{\mathrm{LS}}(f)
		=
		-2f+2\rho_\theta
		=
		2(\rho_\theta-f),
	\end{equation}
	which proves \eqref{eq:our_lsgan_functional_gradient_prop}.
	
	Substituting this gradient into the general discriminator function-space flow yields
	\begin{equation}
		\partial_t f_t
		=
		T_{k^D,\hat\gamma_\theta^{\,r}}
		\Big(
		\nabla_{\hat\gamma_\theta^{\,r}} \mathcal L_D^{\mathrm{LS}}(f_t)
		\Big)
		=
		2\,T_{k^D,\hat\gamma_\theta^{\,r}}(\rho_\theta-f_t),
	\end{equation}
	which proves \eqref{eq:our_lsgan_disc_flow_prop}.
	
	Since this is a linear differential equation in function space, its solution is
	\begin{equation}
		f_t
		=
		\exp\!\big(-2t\,T_{k^D,\hat\gamma_\theta^{\,r}}\big)(f_0-\rho_\theta)
		+
		\rho_\theta,
	\end{equation}
	which proves \eqref{eq:our_lsgan_disc_solution_prop}. The equivalent representation
	\eqref{eq:our_lsgan_disc_solution_phi_prop} follows immediately from the definition
	\(
	\varphi_t(x)=e^{-2tx}-1.
	\)
	
	Finally, under our generator-side convention, the induced sample-wise weighting is
	\begin{equation}
		\gamma_i^{\mathrm{LSGAN}}
		=
		-\frac{1}{N_r}
		\big(f_t(r_i)-1\big)\,
		\partial_r f_t(r_i),
	\end{equation}
	which proves \eqref{eq:our_lsgan_gamma_prop}.
	\hfill $\square$
	
	
	

    \paragraph{Interpretation.}
Proposition~\ref{prop:lsgan_disc_residual} shows that, under our labeling convention, the LSGAN discriminator is best understood not as an abstract optimal critic, but as a kernel-smoothed regression function on residual space. The target \(\rho_\theta\) encodes the desired regression rule induced by the real zero state and the generated residual samples. When the supports of \(\delta_0\) and \(\hat\mu_\theta^{\,r}\) are disjoint, this rule reduces to the usual binary labeling convention: the real zero state is assigned value \(1\), while generated residual samples are assigned value \(0\). The NTK flow then drives the discriminator toward this regression target in function space, up to the null space of the kernel operator.

This interpretation is useful for adversarial PINNs because the generator does not only need discriminator values on the empirical support, but also requires the derivative field \(\partial_r f_t(r)\). The kernel operator \(T_{k^D,\hat\gamma_\theta^{\,r}}\) extends the target regression signal into a smooth function on the whole residual domain, so that \(\partial_r f_t(r_i)\) is well-defined at finite training time. In this sense, the discriminator acts as a kernel-smoothed regression field whose derivative modulates the generator residual dynamics.

Compared with the IPM case, where the discriminator behaves as a witness function, the LSGAN discriminator instead behaves like a regression map relaxing toward a target field. Therefore, the essential difference is not only the nominal loss form, but the type of residual-space structure induced on the generator: witness-type geometry in the IPM case versus regression-type weighting in the LSGAN case.

\begin{remark}[On the target density and convergence]
	The target \(\rho_\theta=d\delta_0/d(\delta_0+\hat\mu_\theta^{\,r})\) should be interpreted as a Radon--Nikodym density with respect to the combined residual measure, rather than always as a pointwise binary label. In particular, if some generated residual samples satisfy \(r_i=0\), then the supports of \(\delta_0\) and \(\hat\mu_\theta^{\,r}\) overlap, and the value of \(\rho_\theta\) at zero reflects the relative mass of the real zero state and the generated residual samples located at zero. Thus, the simple \(1/0\) labeling interpretation is exact only in the support-disjoint case.

	Moreover, the formula
	\[
	f_t
	=
	\exp\!\big(-2t\,T_{k^D,\hat\gamma_\theta^{\,r}}\big)(f_0-\rho_\theta)
	+
	\rho_\theta
	\]
	should be understood as relaxation toward \(\rho_\theta\) modulo the null space of the kernel operator. If \(T_{k^D,\hat\gamma_\theta^{\,r}}\) is injective on the empirical support, then the discriminator converges exponentially to \(\rho_\theta\) on that support. Otherwise, components of \(f_0-\rho_\theta\) lying in the null space of \(T_{k^D,\hat\gamma_\theta^{\,r}}\) are not dissipated by the flow.
\end{remark}
\subsubsection{The vanilla GAN case: kernel-smoothed nonlinear classifier in residual space}
\label{app:disc_vanilla_residual}

We finally rewrite the vanilla GAN discriminator flow in the same residual-space language, so that it can be compared directly with the IPM and LSGAN cases from the viewpoint of generator-side residual weighting. In contrast to the linear regression structure of LSGAN, the key new feature here is the sigmoid nonlinearity, which turns the discriminator dynamics into a nonlinear kernel classification flow.

\begin{proposition}[Vanilla GAN discriminator flow on residual space]
	\label{prop:vanilla_gan_disc_residual}
	Let the discriminator output be
	\begin{equation}
		D_t(r):=\sigma(f_t(r)),
	\end{equation}
	where \(f_t(r)\) is the pre-sigmoid score. Consider the discriminator objective
	\begin{equation}
		\mathcal L_D^{\mathrm{GAN}}(f)
		=
		\mathbb E_{y\sim \delta_0}\!\big[\log D(y)\big]
		+
		\mathbb E_{r\sim \hat\mu_\theta^{\,r}}\!\big[\log(1-D(r))\big].
		\label{eq:vanilla_gan_disc_objective_residual_prop}
	\end{equation}
	Let
	\begin{equation}
		\hat\gamma_\theta^{\,r}
		=
		\frac12 \hat\mu_\theta^{\,r}
		+
		\frac12 \delta_0,
	\end{equation}
	and define
	\begin{equation}
		\rho_\theta
		:=
		\frac{d\delta_0}{d(\delta_0+\hat\mu_\theta^{\,r})}.
		\label{eq:rho_theta_vanilla_prop}
	\end{equation}
	Then:
	\begin{enumerate}
		\item The \(L^2(\hat\gamma_\theta^{\,r})\) functional gradient of \(\mathcal L_D^{\mathrm{GAN}}\) is
		\begin{equation}
			\nabla_{\hat\gamma_\theta^{\,r}}\mathcal L_D^{\mathrm{GAN}}(f)
			=
			2\big(\rho_\theta-\sigma(f)\big).
			\label{eq:vanilla_gan_functional_gradient_prop}
		\end{equation}
		
		\item Consequently, the discriminator NTK flow is
		\begin{equation}
			\partial_t f_t
			=
			2\,T_{k^D,\hat\gamma_\theta^{\,r}}
			\big(\rho_\theta-\sigma(f_t)\big).
			\label{eq:vanilla_gan_disc_flow_residual_prop}
		\end{equation}
		
		\item If \(f_\infty\) is a stationary point of \eqref{eq:vanilla_gan_disc_flow_residual_prop}, then
		\begin{equation}
			T_{k^D,\hat\gamma_\theta^{\,r}}
			\big(\rho_\theta-\sigma(f_\infty)\big)
			=
			0.
		\end{equation}
		In particular, if the kernel operator is injective on the empirical support, then
		\begin{equation}
			\sigma(f_\infty)=\rho_\theta
			\qquad
			\text{on } \mathrm{supp}(\hat\gamma_\theta^{\,r}).
			\label{eq:vanilla_gan_equilibrium_prop}
		\end{equation}
		
		\item Near a stationary point \(f_\infty\), writing \(f_t=f_\infty+h_t\) with \(\|h_t\|\ll 1\), one has the first-order approximation
		\begin{equation}
			\partial_t h_t
			=
			-2\,T_{k^D,\hat\gamma_\theta^{\,r}}
			\big(\sigma'(f_\infty)h_t\big)
			+
			\mathcal O(\|h_t\|^2).
			\label{eq:vanilla_gan_local_linearization_prop}
		\end{equation}
		
		\item The induced generator-side sample weighting is
		\begin{equation}
			\gamma_i^{\mathrm{GAN}}
			=
			\frac{1}{N_r}\,
			\frac{1}{1-D_t(r_i)}\,\partial_r D_t(r_i)
			=
			\frac{1}{N_r}\,
			D_t(r_i)\,\partial_r f_t(r_i).
			\label{eq:vanilla_gan_gamma_residual_prop}
		\end{equation}
	\end{enumerate}
\end{proposition}

\paragraph{Proof.}
Rewriting \eqref{eq:vanilla_gan_disc_objective_residual_prop} over the empirical measure \(\hat\gamma_\theta^{\,r}\), we have
\begin{equation}
	\mathcal L_D^{\mathrm{GAN}}(f)
	=
	\int
	\left[
	\frac{d\delta_0}{d\hat\gamma_\theta^{\,r}}\log \sigma(f)
	+
	\frac{d\hat\mu_\theta^{\,r}}{d\hat\gamma_\theta^{\,r}}\log(1-\sigma(f))
	\right]
	d\hat\gamma_\theta^{\,r}.
\end{equation}
Differentiating with respect to \(f\) yields
\begin{align}
	\nabla_{\hat\gamma_\theta^{\,r}}\mathcal L_D^{\mathrm{GAN}}(f)
	&=
	\frac{d\delta_0}{d\hat\gamma_\theta^{\,r}}
	\frac{\sigma'(f)}{\sigma(f)}
	-
	\frac{d\hat\mu_\theta^{\,r}}{d\hat\gamma_\theta^{\,r}}
	\frac{\sigma'(f)}{1-\sigma(f)} \notag\\
	&=
	\frac{d\delta_0}{d\hat\gamma_\theta^{\,r}}(1-\sigma(f))
	-
	\frac{d\hat\mu_\theta^{\,r}}{d\hat\gamma_\theta^{\,r}}\sigma(f).
\end{align}
Since
\[
\hat\gamma_\theta^{\,r}
=
\frac12(\hat\mu_\theta^{\,r}+\delta_0),
\qquad
\frac{d\delta_0}{d\hat\gamma_\theta^{\,r}}=2\rho_\theta,
\qquad
\frac{d\hat\mu_\theta^{\,r}}{d\hat\gamma_\theta^{\,r}}=2(1-\rho_\theta),
\]
we obtain
\begin{equation}
	\nabla_{\hat\gamma_\theta^{\,r}}\mathcal L_D^{\mathrm{GAN}}(f)
	=
	2\rho_\theta-2\sigma(f)
	=
	2(\rho_\theta-\sigma(f)),
\end{equation}
which proves \eqref{eq:vanilla_gan_functional_gradient_prop}. Substituting this into the general discriminator NTK flow gives
\eqref{eq:vanilla_gan_disc_flow_residual_prop}.

The equilibrium condition follows immediately by setting \(\partial_t f_t=0\) in
\eqref{eq:vanilla_gan_disc_flow_residual_prop}, which gives
\[
T_{k^D,\hat\gamma_\theta^{\,r}}
\big(\rho_\theta-\sigma(f_\infty)\big)=0.
\]
If the kernel operator is injective on the empirical support, then \eqref{eq:vanilla_gan_equilibrium_prop} follows.

For the local linearization, write \(f_t=f_\infty+h_t\) and use the Taylor expansion
\[
\sigma(f_\infty+h_t)
=
\sigma(f_\infty)+\sigma'(f_\infty)h_t+\mathcal O(\|h_t\|^2).
\]
Substituting this into \eqref{eq:vanilla_gan_disc_flow_residual_prop} and using the equilibrium relation yields
\eqref{eq:vanilla_gan_local_linearization_prop}.

Finally, for the minimax generator objective,
\begin{equation}
	\gamma_i^{\mathrm{GAN}}
	=
	-\frac{1}{N_r}
	R'\!\big(f_t(r_i)\big)\,\partial_r f_t(r_i).
\end{equation}
Equivalently, writing the discriminator output directly as \(D_t(r)=\sigma(f_t(r))\), one obtains
\begin{equation}
	\gamma_i^{\mathrm{GAN}}
	=
	\frac{1}{N_r}\,
	\frac{1}{1-D_t(r_i)}\,\partial_r D_t(r_i).
\end{equation}
Using the chain rule
\[
\partial_r D_t(r_i)=D_t(r_i)(1-D_t(r_i))\,\partial_r f_t(r_i),
\]
this becomes
\begin{equation}
	\gamma_i^{\mathrm{GAN}}
	=
	\frac{1}{N_r}\,
	D_t(r_i)\,\partial_r f_t(r_i),
\end{equation}
which proves \eqref{eq:vanilla_gan_gamma_residual_prop}.
\hfill $\square$

\paragraph{Interpretation.}
Proposition~\ref{prop:vanilla_gan_disc_residual} shows that, in residual space, the vanilla GAN discriminator is best understood as a kernel-smoothed nonlinear classifier rather than as a linear regression field or a witness function. The target \(\rho_\theta\) still separates the real zero residual from the generated residual samples, but the sigmoid nonlinearity changes the discriminator dynamics qualitatively. Formally, if a finite stationary point exists, it satisfies the equilibrium condition in Proposition~\ref{prop:vanilla_gan_disc_residual}. In the separable case, however, the limiting relation \(\sigma(f_\infty)=\rho_\theta\) may correspond to saturation of the logits rather than convergence to a finite discriminator.

This saturation behavior directly affects the generator-side weighting. Although the logit-level discriminator flow is driven by the classification residual \(\rho_\theta-\sigma(f_t)\), the probability-level input derivative satisfies
\[
\partial_r D_t(r)
=
D_t(r)(1-D_t(r))\,\partial_r f_t(r).
\]
Hence the residual-space derivative of the discriminator output is suppressed in saturated regions where \(D_t(r)\approx 0\) or \(D_t(r)\approx 1\). Compared with IPM and LSGAN, vanilla GAN also introduces an additional confidence factor \(D_t(r_i)\) into the sample-wise weighting, so the discriminator acts through both its residual-space score gradient \(\partial_r f_t(r_i)\) and its current classification confidence. In this sense, the essential difference is not only the nominal adversarial loss, but also the way the sigmoid nonlinearity reshapes the discriminator-induced residual weighting.

\begin{remark}
	The purpose of the present appendix is not to claim a new NTK theory of GAN discriminators in itself. Rather, we use the existing discriminator-side NTK framework as an analysis tool and reinterpret it in residual space so that it becomes directly applicable to adversarial PINNs. What is specific to the present setting is that the ``generated samples'' are residual values, the ``real samples'' correspond to the zero residual state, and the resulting discriminator flow can therefore be linked explicitly to the generator-side residual weighting and residual-energy dynamics developed in the main text.
\end{remark}
	\subsection{Generator residual dynamics}
	\label{app:gen_dynamics}
	
	We next derive the generator-side dynamics in residual space. Unlike the discriminator, whose evolution is naturally described in function space, the generator affects training through the evolution of the residual values themselves. It is therefore convenient to rewrite the parameter-space gradient flow of the generator as a closed dynamics for the residual vector. This leads directly to the kernel-preconditioned residual evolution used throughout the main text.
	
	\begin{proposition}[Generator residual dynamics in residual space]
		\label{prop:generator_residual_dynamics}
		Let the generator gradient flow be
		\begin{equation}
			\dot{\theta}(t)
			=
			-\frac{\partial \mathcal L_G(\theta(t))}{\partial \theta(t)}.
		\end{equation}
		For each collocation point \(x_j\), define the residual
		\begin{equation}
			r_j(t):=\mathcal R(x_j;\theta(t)),
			\qquad j=1,\dots,N_r.
		\end{equation}
		Let
		\begin{equation}
			\mathbf r(t):=
			[r_1(t),\dots,r_{N_r}(t)]^\top\in\mathbb R^{N_r}
		\end{equation}
		be the residual vector, and define the generator residual NTK matrix
		\begin{equation}
			(K_{rr}^G)_{ij}(t)
			:=
			\left\langle
			\frac{\partial \mathcal R(x_i;\theta(t))}{\partial \theta(t)},
			\frac{\partial \mathcal R(x_j;\theta(t))}{\partial \theta(t)}
			\right\rangle.
		\end{equation}
		Further define the diagonal matrix
		\begin{equation}
			\Gamma(t):=
			\mathrm{Diag}\!\big(\Gamma_{11}(t),\dots,\Gamma_{N_rN_r}(t)\big),
		\end{equation}
		with entries
		\begin{equation}
			\Gamma_{ii}(t)
			=
			-\frac{1}{N_r}
			R'\!\Big(f(r_i(t);\phi(t))\Big)\,
			\frac{\partial f(r_i(t);\phi(t))}{\partial r_i(t)}.
			\label{eq:appendix_gamma_entry_prop}
		\end{equation}
		Then the generator-induced residual dynamics satisfy
		\begin{equation}
			\dot r_j(t)
			=
			\sum_{i=1}^{N_r}(K_{rr}^G)_{ji}(t)\,\Gamma_{ii}(t),
			\qquad j=1,\dots,N_r,
		\end{equation}
		or equivalently, in vector form,
		\begin{equation}
			\dot{\mathbf r}(t)
			=
			K_{rr}^G(t)\,\Gamma(t)\,\mathbf 1,
			\label{eq:appendix_residual_dyn_gamma_prop}
		\end{equation}
		where \(\mathbf 1\in\mathbb R^{N_r}\) denotes the all-one vector.
		
		Equivalently, if one introduces the weighting vector
		\begin{equation}
			\boldsymbol{\gamma}(t)
			:=
			\Gamma(t)\mathbf 1
			=
			[\gamma_1(t),\dots,\gamma_{N_r}(t)]^\top,
		\end{equation}
		with
		\begin{equation}\label{eq:appendix_gamma_entry}
			\gamma_i(t)
			=
			-\frac{1}{N_r}
			R'\!\Big(f(r_i(t);\phi(t))\Big)\,
			\frac{\partial f(r_i(t);\phi(t))}{\partial r_i(t)},
		\end{equation}
		then the residual dynamics admit the compact form
		\begin{equation}
			\dot{\mathbf r}(t)
			=
			K_{rr}^G(t)\,\boldsymbol{\gamma}(t).
			\label{eq:appendix_residual_dyn_gamma_compact_prop}
		\end{equation}
	\end{proposition}
	
	\paragraph{Proof.}
	Starting from the generator gradient flow,
	\begin{equation}
		\dot{\theta}(t)
		=
		-\frac{\partial \mathcal L_G(\theta(t))}{\partial \theta(t)},
	\end{equation}
	and substituting \eqref{eq:appendix_grad_G}, we obtain
	\begin{equation}
		\dot{\theta}(t)
		=
		-\frac{1}{N_r}\sum_{i=1}^{N_r}
		R'\!\Big(f(\mathcal R(x_i;\theta(t));\phi(t))\Big)\,
		\frac{\partial f(\mathcal R(x_i;\theta(t));\phi(t))}{\partial \mathcal R(x_i;\theta(t))}
		\frac{\partial \mathcal R(x_i;\theta(t))}{\partial \theta(t)}.
		\label{eq:appendix_theta_flow_expanded_prop}
	\end{equation}
	
	For each collocation point \(x_j\), apply the chain rule:
	\begin{align}
		\frac{d}{dt}\mathcal R(x_j;\theta(t))
		&=
		\left\langle
		\frac{\partial \mathcal R(x_j;\theta(t))}{\partial \theta(t)},
		\dot{\theta}(t)
		\right\rangle \notag\\
		&=
		-\left\langle
		\frac{\partial \mathcal R(x_j;\theta(t))}{\partial \theta(t)},
		\frac{1}{N_r}\sum_{i=1}^{N_r}
		R'\!\Big(f(\mathcal R(x_i;\theta(t));\phi(t))\Big)\,
		\frac{\partial f(\mathcal R(x_i;\theta(t));\phi(t))}{\partial \mathcal R(x_i;\theta(t))}
		\frac{\partial \mathcal R(x_i;\theta(t))}{\partial \theta(t)}
		\right\rangle \notag\\
		&=
		-\frac{1}{N_r}\sum_{i=1}^{N_r}
		R'\!\Big(f(\mathcal R(x_i;\theta(t));\phi(t))\Big)\,
		\frac{\partial f(\mathcal R(x_i;\theta(t));\phi(t))}{\partial \mathcal R(x_i;\theta(t))}
		\left\langle
		\frac{\partial \mathcal R(x_j;\theta(t))}{\partial \theta(t)},
		\frac{\partial \mathcal R(x_i;\theta(t))}{\partial \theta(t)}
		\right\rangle.
	\end{align}
	Since \(r_i(t)=\mathcal R(x_i;\theta(t))\), this becomes
	\begin{equation}
		\dot r_j(t)
		=
		-\frac{1}{N_r}\sum_{i=1}^{N_r}
		R'\!\Big(f(r_i(t);\phi(t))\Big)\,
		\frac{\partial f(r_i(t);\phi(t))}{\partial r_i(t)}
		\left\langle
		\frac{\partial \mathcal R(x_j;\theta(t))}{\partial \theta(t)},
		\frac{\partial \mathcal R(x_i;\theta(t))}{\partial \theta(t)}
		\right\rangle.
		\label{eq:appendix_rj_dynamics_prop}
	\end{equation}
	
	Defining the generator residual NTK matrix \(K_{rr}^G(t)\) and the diagonal weighting matrix \(\Gamma(t)\) by
	\eqref{eq:appendix_gamma_entry_prop}, we obtain the componentwise representation
	\begin{equation}
		\dot r_j(t)
		=
		\sum_{i=1}^{N_r}(K_{rr}^G)_{ji}(t)\,\Gamma_{ii}(t),
	\end{equation}
	which yields the vector form
	\begin{equation}
		\dot{\mathbf r}(t)
		=
		K_{rr}^G(t)\,\Gamma(t)\,\mathbf 1.
	\end{equation}
	Finally, introducing \(\boldsymbol{\gamma}(t)=\Gamma(t)\mathbf 1\) gives
	\begin{equation}
		\dot{\mathbf r}(t)
		=
		K_{rr}^G(t)\,\boldsymbol{\gamma}(t).
	\end{equation}
	This proves \eqref{eq:appendix_residual_dyn_gamma_prop} and
	\eqref{eq:appendix_residual_dyn_gamma_compact_prop}.
	\hfill $\square$
	
	\paragraph{Interpretation.}
	Proposition~\ref{prop:generator_residual_dynamics} shows that the generator does not evolve in residual space by an ordinary gradient descent with a scalar loss coefficient. Instead, its residual dynamics are preconditioned by the generator residual NTK \(K_{rr}^G(t)\) and modulated samplewise by the discriminator-induced weighting \(\boldsymbol{\gamma}(t)\).
	
	This decomposition is central to the viewpoint of the main text. The matrix \(K_{rr}^G(t)\) encodes the geometry of residual propagation through the generator, while \(\boldsymbol{\gamma}(t)\) contains the discriminator-side signal through the factor
	\[
	\gamma_i=-\frac{1}{N_r}\,R'\!\Big(f(r_i(t);\phi(t))\Big)\,
	\frac{\partial f(r_i(t);\phi(t))}{\partial r_i(t)}.
	\]
	Hence, different adversarial objectives do not simply change a scalar divergence value; rather, they reshape the residual dynamics through different sample-dependent weighting mechanisms. In this sense, adversarial PINNs training should be understood as a discriminator-modulated, kernel-preconditioned residual descent process.
	
	\subsubsection{Specialization to common adversarial objectives}
	\label{app:gamma_specializations}
	
	We finally record the specialization of $\Gamma(t)$ for several standard adversarial objectives.
	
	\paragraph{Vanilla GAN.}
	If
	\begin{equation}
		R(f)=\log\!\big(1-\sigma(f)\big),
	\end{equation}
	then
	\begin{equation}
		R'(f)=-\sigma(f).
	\end{equation}
	Hence, by \eqref{eq:appendix_gamma_entry},
	\begin{equation}
		[\Gamma(t)]_{ii}
		=
		\frac{1}{N_r}\,
		\sigma\!\big(f(r_i(t);\phi(t))\big)\,
		\frac{\partial f(r_i(t);\phi(t))}{\partial r_i(t)}.
	\end{equation}
	\paragraph{Least-Squares GAN (LSGAN).}
	If
	\begin{equation}
		R(f)=\frac12(f-1)^2,
	\end{equation}
	then
	\begin{equation}
		R'(f)=f-1.
	\end{equation}
	Hence
	\begin{equation}
		[\Gamma(t)]_{ii}
		=
		-\frac{1}{N_r}
		\big(f(r_i(t);\phi(t))-1\big)
		\frac{\partial f(r_i(t);\phi(t))}{\partial r_i(t)}.
	\end{equation}
	
	\paragraph{Integral Probability Metric GAN (IPM-GAN).}
	If
	\begin{equation}
		R(f)=-f,
	\end{equation}
	then
	\begin{equation}
		R'(f)=-1.
	\end{equation}
	Therefore,
	\begin{equation}
		[\Gamma(t)]_{ii}
		=
		\frac{1}{N_r}
		\frac{\partial f(r_i(t);\phi(t))}{\partial r_i(t)}.
	\end{equation}
	
	Equations above show that, once the discriminator state is fixed, different adversarial objectives enter the generator residual dynamics through different sample-wise weighting rules, while the structural form of the residual dynamics remains the same.
	
	\subsubsection{Interpretation in our framework}
	\label{app:disc_ntk_our_interpretation}
	
	The discriminator-side NTK analysis can be summarized in the language of this paper as follows. First, for a fixed generator, discriminator training against \(\hat\mu_\theta^{\,r}\) and \(\delta_0\) produces a finite-time, architecture-dependent residual-space function, rather than an arbitrary empirical optimum. Second, the derivative \(\partial_r f_t(r)\) entering the generator dynamics is therefore well-defined and inherits the smoothing and inductive bias of the discriminator NTK. Third, in the IPM case, this derivative is induced by the NTK-MMD witness function comparing \(\hat\mu_\theta^{\,r}\) and \(\delta_0\); in the LSGAN case, it is induced by a kernel-smoothed regression solution; and in the vanilla GAN case, it is induced by a kernel-smoothed nonlinear classification flow shaped by the sigmoid nonlinearity.
	
	From our perspective, the essential role of the discriminator in adversarial PINNs is therefore not to realize an abstract optimal divergence critic, but to generate, after finite-time NTK-constrained training, a differentiable residual-space weighting field whose derivative directly modulates the generator residual dynamics. This is precisely the mechanism captured by our framework.

    \subsection{The effect of the PDE operator on the NTK spectrum}
Let $u_{\theta}$ denote the neural network in which the prescribed boundary conditions have already been incorporated. Consider the linear PDE $\mathcal{T}u_{\theta}(\mathbf{x})=f(\mathbf{x})$, where $\mathcal{T}$ is an $m$th-order linear differential operator, and define the residual $r(\mathbf{x};\theta)=\mathcal{T}u_{\theta}(\mathbf{x})-f(\mathbf{x})$. The neural-network NTK is
$
K_{uu}^{G}(\mathbf{x},\mathbf{x}')=
\left\langle
\nabla_{\theta}u_{\theta}(\mathbf{x}),
\nabla_{\theta}u_{\theta}(\mathbf{x}')
\right\rangle .
$

Under sufficient smoothness such that the parameter and spatial
derivatives commute, the operator-informed NTK relation gives

\begin{equation}
    K_{rr}^{G}\left(\mathbf{x},\mathbf{x}'\right)
=\mathcal{T}_{\mathbf{x}}\mathcal{T}_{\mathbf{x}'}K_{uu}^{G}\left(\mathbf{x},\mathbf{x}'\right),
\end{equation}

as established in \cite{gan2025neural}. Thus, the PDE operator not
only defines the residual, but also transforms the neural-network
parameter features and reshapes the kernel governing the residual
dynamics.

Let $\{\lambda_j\}$ and $\{\mu_j\}$ denote the eigenvalues of the
integral operators induced by $K_{uu}^{G}$ and $K_{rr}^{G}$,
respectively. For a symmetric operator satisfying
$\|v\|_{L^2}\leq C_{\mathcal{T}} \|\mathcal{T}v\|_{L^2}$, the spectral
comparison in [1] yields

\begin{equation}
     \lambda_j
    \leq
    C_{\mathcal{T}}^{\,2}\mu_j .
\end{equation}

The effect of the differential order is particularly transparent when
$\mathcal{T}$ and the neural-network NTK share an eigenbasis. If
$\mathcal{T}\varphi_j=\tau_j\varphi_j$ and
$\mathcal{I}_{u}\varphi_j=\lambda_j\varphi_j$, then

\begin{equation}
    \mu_j=|\tau_j|^2\lambda_j .
\end{equation}

For an $m$th-order elliptic operator,
$|\tau_{\mathbf{k}}|\asymp|\mathbf{k}|^m$ at high frequencies, giving

\begin{equation}
\mu_{\mathbf{k}}
\asymp
|\mathbf{k}|^{2m}\lambda_{\mathbf{k}}.
\end{equation}

Hence, the operator order appears twice because $\mathcal{T}$ acts on both arguments of the NTK. For example, the second-order Poisson operator produces the multiplier $|\mathbf{k}|^4$. When the operator and the NTK do not share an eigenbasis, the PDE operator may also mix the original neural-network modes. This discussion is limited to known linear operators; for general nonlinear partial differential equations, the residual NTK may also depend on other conditions. The influence of the boundary on the NTK spectrum will be left for future work.

    \begin{remark}[Extension to GAN-Based Regression]
        The essential difference between GAN-based regression problems and our analysis lies in whether the target distribution is available.
        
        For a GAN-based regression problem, after redefining the residual as the prediction error, the resulting dynamics share a similar algebraic form with our framework. However, a critical distinction lies in the target distribution. In regression, the true data distribution $p(\mathbf y)$ is unknown; the training objective is instead defined with respect to the empirical distribution $\hat p(\mathbf y)=\frac{1}{N}\sum_{i=1}^{N}\delta_{\mathbf y_i}$. Consequently, the empirical distribution varies with the number of training samples, and it remains unclear whether the gap between the true and empirical distributions may compromise the validity of the first-order indicator derived from our analysis. In contrast, in our PINN setting, the target distribution is always the fixed Dirac measure $\delta_0$, regardless of the number of collocation points, because this zero-residual state is dictated by the physical laws encoded in the PDE. At present, our work is conducted within the PINNs framework. Although our analysis suggests that the method may be extendable to other models, this direction requires further theoretical and experimental validation, which falls beyond the scope of this paper. We therefore leave it as future work.
    \end{remark}
    
	\section{Analysis of the residual-energy decay}
	\label{app:energy_decay}
	\subsection{Residual-energy law and constant-kernel approximation}
	We now turn from the residual dynamics themselves to the evolution of the residual energy. This provides a more compact scalar description of whether adversarial training is effectively driving the residual toward zero. Starting from the generator-side residual dynamics derived in the previous appendix, we show that the energy dissipation rate is determined by the interaction between the generator residual NTK and the discriminator-induced weighting vector. This identity will serve as the basis for the later decay estimates, failure-mode analysis, and spectral interpretations.
	
	\begin{theorem}[Residual-energy law under adversarial residual dynamics]
		\label{prop:energy_law_residual}
		Let
		\begin{equation}
			E(t):=\frac12\|\mathbf r(t)\|^2
		\end{equation}
		be the residual energy, where the residual vector evolves according to
		\begin{equation}
			\dot{\mathbf r}(t)=K_{rr}^G(t)\boldsymbol{\gamma}(t),
			\label{eq:app_r_dynamics_re_prop}
		\end{equation}
		with generator residual NTK \(K_{rr}^G(t)\in\mathbb R^{N_r\times N_r}\) and discriminator-induced weighting vector
		\begin{equation}
			\boldsymbol{\gamma}(t)
			=
			[\gamma_1(t),\dots,\gamma_{N_r}(t)]^\top.
		\end{equation}
		Then the residual energy satisfies
		\begin{equation}
			\dot E(t)
			=
			\mathbf r(t)^\top K_{rr}^G(t)\boldsymbol{\gamma}(t).
			\label{eq:app_energy_identity_re_prop}
		\end{equation}
		In particular, the sign and magnitude of \(\dot E(t)\) are determined by how the discriminator-induced weighting \(\boldsymbol{\gamma}(t)\) aligns with the residual vector through the kernel geometry imposed by \(K_{rr}^G(t)\).
	\end{theorem}
	
	\paragraph{Proof.}
	By definition,
	\begin{equation}
		E(t)=\frac12\|\mathbf r(t)\|^2=\frac12\,\mathbf r(t)^\top \mathbf r(t).
	\end{equation}
	Differentiating with respect to time yields
	\begin{equation}
		\dot E(t)
		=
		\mathbf r(t)^\top \dot{\mathbf r}(t).
	\end{equation}
	Substituting the residual dynamics
	\eqref{eq:app_r_dynamics_re_prop}, we obtain
	\begin{equation}
		\dot E(t)
		=
		\mathbf r(t)^\top K_{rr}^G(t)\boldsymbol{\gamma}(t),
	\end{equation}
	which proves \eqref{eq:app_energy_identity_re_prop}.
	\hfill $\square$
	
	\paragraph{Interpretation.}
	Theorem~\ref{prop:energy_law_residual} shows that the energy decay in adversarially trained PINNs is not controlled by the generator kernel alone, nor by the discriminator weighting alone, but by their interaction. The matrix \(K_{rr}^G(t)\) determines how a local discriminator signal propagates across residual samples, while \(\boldsymbol{\gamma}(t)\) determines whether that propagated signal is aligned with residual descent. Therefore, the key quantity is not simply the size of the discriminator gradient, but the signed bilinear form
	\[
	\mathbf r(t)^\top K_{rr}^G(t)\boldsymbol{\gamma}(t),
	\]
	which acts as the instantaneous energy-dissipation rate.
	
To obtain explicit convergence rates and more refined decay mechanisms, it is convenient to further work under the constant-NTK approximation
\begin{equation}
	K_{rr}^G(t)\approx K_{rr}^G.
	\label{eq:app_const_ntk_re_prop}
\end{equation}
Under this approximation, the energy law admits a clearer spectral interpretation, which allows us to distinguish exponential decay, slower algebraic decay, and several representative failure modes.

\begin{remark}[On the constant-NTK approximation]
	The approximation \eqref{eq:app_const_ntk_re_prop} is best understood as a mechanism-level simplification rather than a universally exact property. Under Assumptions~\ref{ass1}--\ref{ass5}, for some relatively simple or weakly nonlinear problems---especially linear PDEs and certain regular regimes---the generator NTK is known or empirically observed to remain nearly constant during training\cite{wang2022and}. By contrast, for strongly nonlinear, multiscale, or more complicated equations, the NTK may evolve nontrivially \cite{zhou2024neural, carvalho2025positivity}, so the constant-kernel assumption need not hold exactly. Nevertheless, even in such cases, the approximation remains useful for revealing the dominant spectral mechanisms behind residual-energy decay and failure modes.
\end{remark}

	\subsection{Spectral factorization of the aligned dynamics}

    We first consider the favorable case in which the discriminator-induced weighting is linearly aligned with the negative residual direction. Instead of treating \(K_{rr}^G A\) as a symmetric operator, we use a spectral factorization that converts the effective operator into the symmetric positive semidefinite matrix \(K_{rr}^{G,1/2} A K_{rr}^{G,1/2}\).

Assume throughout this subsection that the generator residual NTK \(K_{rr}^G\) is symmetric positive definite on the active residual subspace, and that the discriminator-induced weighting takes the linear alignment form
\begin{equation}
	\boldsymbol{\gamma}(t)=-A\mathbf r(t),
\end{equation}
where \(A\in\mathbb R^{N_r\times N_r}\) is symmetric positive semidefinite. Then the residual dynamics become
\begin{equation}\label{eq:app_linear_dyn_spectral}
		\dot{\mathbf r}(t)=-K_{rr}^G A\,\mathbf r(t).
	\end{equation}
	
	\begin{theorem}[Spectral factorization and convergence rate under linear alignment]
		\label{thm:spectral_factorization_alignment}
		Let
		\begin{equation}
			K:=K_{rr}^G,
			\qquad
			M:=KA,
			\qquad
			H:=K^{1/2}AK^{1/2}.
		\end{equation}
		Assume that \(K\) is symmetric positive definite and that \(A\) is symmetric positive semidefinite. Then:
		\begin{enumerate}
			\item \(M\) is similar to \(H\), namely
			\begin{equation}
				K^{-1/2}MK^{1/2}
				=
				K^{-1/2}(KA)K^{1/2}
				=
				K^{1/2}AK^{1/2}
				=
				H.
			\end{equation}
			Hence \(M\) and \(H\) have the same eigenvalues.
			
			\item Since \(H\) is symmetric positive semidefinite, all eigenvalues of \(M\) are real and nonnegative. Moreover, if
			\begin{equation}
				\lambda_1(A)\ge \lambda_2(A)\ge \cdots \ge \lambda_{N_r}(A)\ge 0
			\end{equation}
			denote the eigenvalues of \(A\), then for each \(j=1,\dots,N_r\),
			\begin{equation}\label{eq:app_eig_bound_KA}
				\lambda_{\min}(K)\,\lambda_j(A)
				\;\le\;
				\lambda_j(M)
				=
				\lambda_j(H)
				\;\le\;
				\lambda_{\max}(K)\,\lambda_j(A).
			\end{equation}
			
			\item Introducing the transformed variable
			\begin{equation}
				\mathbf z(t):=K^{-1/2}\mathbf r(t),
			\end{equation}
			the dynamics \eqref{eq:app_linear_dyn_spectral} become
			\begin{equation}\label{eq:app_z_dyn}
				\dot{\mathbf z}(t)=-H\mathbf z(t).
			\end{equation}
			Therefore,
			\begin{equation}
				\|\mathbf z(t)\|^2
				\le
				e^{-2\lambda_{\min}(H)t}\|\mathbf z(0)\|^2.
			\end{equation}
			Since
			\begin{equation}
				\lambda_{\min}(K)\|\mathbf z\|^2
				\le
				\|\mathbf r\|^2
				\le
				\lambda_{\max}(K)\|\mathbf z\|^2,
			\end{equation}
			the residual energy satisfies
			\begin{equation}\label{eq:app_energy_decay_spectral}
				E(t)
				\le
				\kappa(K)\,E(0)\,e^{-2\lambda_{\min}(H)t},
				\qquad
				\kappa(K):=\frac{\lambda_{\max}(K)}{\lambda_{\min}(K)}.
			\end{equation}
			In particular, using \eqref{eq:app_eig_bound_KA},
			\begin{equation}\label{eq:app_energy_decay_spectral_lower}
				E(t)
				\le
				\kappa(K)\,E(0)\,
				e^{-2\lambda_{\min}(K)\lambda_{\min}(A)t}.
			\end{equation}
		\end{enumerate}
	\end{theorem}
	
	\paragraph{Proof.}
	The similarity relation follows directly from
	\[
	K^{-1/2}(KA)K^{1/2}=K^{1/2}AK^{1/2}.
	\]
	Since \(H=K^{1/2}AK^{1/2}\) is symmetric positive semidefinite, its spectrum is real and nonnegative, and therefore the same is true for \(M\). The eigenvalue bound \eqref{eq:app_eig_bound_KA} is the standard congruence bound for symmetric positive semidefinite matrices. Now define \(\mathbf z=K^{-1/2}\mathbf r\). Then
	\[
	\dot{\mathbf z}
	=
	K^{-1/2}\dot{\mathbf r}
	=
	-K^{-1/2}KA\mathbf r
	=
	-K^{1/2}AK^{1/2}\mathbf z
	=
	-H\mathbf z,
	\]
	which gives \eqref{eq:app_z_dyn}. Since \(H\succeq 0\),
	\[
	\frac{d}{dt}\frac12\|\mathbf z(t)\|^2
	=
	-\mathbf z(t)^\top H\mathbf z(t)
	\le
	-\lambda_{\min}(H)\|\mathbf z(t)\|^2.
	\]
	Applying Gr\"onwall's inequality yields
	\[
	\|\mathbf z(t)\|^2
	\le
	e^{-2\lambda_{\min}(H)t}\|\mathbf z(0)\|^2.
	\]
	Finally, using the norm equivalence induced by \(K\),
	\[
	\lambda_{\min}(K)\|\mathbf z\|^2
	\le
	\|\mathbf r\|^2
	\le
	\lambda_{\max}(K)\|\mathbf z\|^2,
	\]
	we obtain \eqref{eq:app_energy_decay_spectral}. The lower bound
	\[
	\lambda_{\min}(H)\ge \lambda_{\min}(K)\lambda_{\min}(A)
	\]
	then gives \eqref{eq:app_energy_decay_spectral_lower}.
	\hfill $\square$
	
	\paragraph{Interpretation.}
	Theorem~\ref{thm:spectral_factorization_alignment} shows that, under linear alignment, the effective decay rate is controlled not by the generator residual NTK \(K_{rr}^G\) alone, but by the coupled operator \(K_{rr}^G A\), or equivalently by the symmetric matrix \(K_{rr}^{G,1/2} A K_{rr}^{G,1/2}\). Therefore, adversarially induced weighting may accelerate or slow down convergence through its interaction with the spectrum of the generator kernel. In particular, small eigenvalues of \(A\) may lead to slow-decaying directions even when \(K_{rr}^G\) itself is well-conditioned, while large eigenvalues of \(A\) may selectively amplify useful descent directions.
\begin{remark}[Connection to standard PINN-NTK dynamics]
	The standard non-adversarial PINN-NTK law is recovered as the special case
	\[
	\boldsymbol{\gamma}(t)=-\mathbf r(t),
	\]
	that is, \(A=I\). In this case the aligned dynamics reduce to
	\[
	\dot{\mathbf r}(t)=-K_{rr}^G\mathbf r(t),
	\]
	and the residual energy satisfies
	\[
	E(t)\le E(0)e^{-2\lambda_{\min}(K_{rr}^G)t}.
	\]
	Thus ordinary PINN training corresponds to the reference case in which the residual is fed back without adversarial distortion.
\end{remark}
    
	\subsection{Modal view and slower decay under weaker alignment}
	\label{app:modal_decay}
	
	We now turn to the complementary situation where the residual energy still decreases, but the convergence is no longer fast. In this regime, the discriminator-induced weighting remains descent-oriented, yet its alignment with the residual is weaker than in the strongly aligned case considered above. The resulting decay is therefore slower, and its mechanism is best understood in the eigenbasis of the generator residual NTK.
	
	\begin{theorem}[Modal decay and slower convergence under weaker alignment]
		\label{prop:modal_decay_alignment}
		Assume the constant-NTK approximation
		\begin{equation}
			K_{rr}^G(t)\approx K_{rr}^G,
		\end{equation}
		and let
		\begin{equation}
			K_{rr}^G = U\Lambda U^\top,
			\qquad
			\Lambda=\mathrm{Diag}(\lambda_1,\dots,\lambda_{N_r}),
			\qquad
			\lambda_j\ge 0,
		\end{equation}
		be the eigendecomposition of the generator residual NTK. Define
		\begin{equation}
			\widetilde{\mathbf r}(t):=U^\top \mathbf r(t),
			\qquad
			\widetilde{\boldsymbol{\gamma}}(t):=U^\top \boldsymbol{\gamma}(t).
		\end{equation}
		Then:
		\begin{enumerate}
			\item The residual dynamics in modal coordinates are
			\begin{equation}
				\dot{\widetilde{\mathbf r}}(t)=\Lambda\,\widetilde{\boldsymbol{\gamma}}(t),
				\label{eq:app_modal_dyn_re_prop}
			\end{equation}
			that is,
			\begin{equation}
				\dot{\widetilde r}_j(t)=\lambda_j\,\widetilde\gamma_j(t),
				\qquad j=1,\dots,N_r.
			\end{equation}
			
			\item If the discriminator-induced weighting is modewise aligned in the sense that
			\begin{equation}
				\widetilde\gamma_j(t)=-a_j(t)\widetilde r_j(t),
				\qquad
				a_j(t)\ge a_*>0,
				\label{eq:app_modal_align_re_prop}
			\end{equation}
			then each mode satisfies
			\begin{equation}
				\dot{\widetilde r}_j(t)=-\lambda_j a_j(t)\widetilde r_j(t),
			\end{equation}
			and therefore
			\begin{equation}
				|\widetilde r_j(t)|
				\le
				|\widetilde r_j(0)|e^{-\lambda_j a_* t}.
				\label{eq:app_modal_decay_re_prop}
			\end{equation}
			Consequently, the residual energy satisfies
			\begin{equation}
				E(t)
				\le
				\frac12\sum_{j=1}^{N_r}\widetilde r_j(0)^2e^{-2\lambda_j a_* t}.
				\label{eq:app_energy_modal_sum_re_prop}
			\end{equation}
			
			\item More generally, if the feedback remains descent-oriented but no longer scales linearly with the residual, and one only has
			\begin{equation}
				\mathbf r(t)^\top K_{rr}^G\boldsymbol{\gamma}(t)
				\le
				-c\|\mathbf r(t)\|^{2\alpha},
				\qquad
				\alpha>1,
				\label{eq:app_power_decay_assumption_re_prop}
			\end{equation}
			then the residual energy obeys the algebraic decay estimate
			\begin{equation}
				E(t)\le C(1+t)^{-1/(\alpha-1)}
				\label{eq:app_algebraic_decay_re_prop}
			\end{equation}
			for some constant \(C>0\).
		\end{enumerate}
	\end{theorem}
	
	\paragraph{Proof.}
	Since
	\begin{equation}
		\dot{\mathbf r}(t)=K_{rr}^G\boldsymbol{\gamma}(t),
	\end{equation}
	left-multiplying by \(U^\top\) and using \(K_{rr}^G=U\Lambda U^\top\) gives
	\begin{equation}
		\dot{\widetilde{\mathbf r}}(t)
		=
		U^\top \dot{\mathbf r}(t)
		=
		U^\top K_{rr}^G\boldsymbol{\gamma}(t)
		=
		U^\top U\Lambda U^\top \boldsymbol{\gamma}(t)
		=
		\Lambda\,\widetilde{\boldsymbol{\gamma}}(t),
	\end{equation}
	which proves \eqref{eq:app_modal_dyn_re_prop}.
	
	Now assume the modewise alignment condition
	\eqref{eq:app_modal_align_re_prop}. Then each mode satisfies
	\begin{equation}
		\dot{\widetilde r}_j(t)
		=
		\lambda_j\widetilde\gamma_j(t)
		=
		-\lambda_j a_j(t)\widetilde r_j(t).
	\end{equation}
	Since \(a_j(t)\ge a_*>0\), Gr\"onwall's inequality yields
	\begin{equation}
		|\widetilde r_j(t)|
		\le
		|\widetilde r_j(0)|e^{-\lambda_j a_* t},
	\end{equation}
	which proves \eqref{eq:app_modal_decay_re_prop}. Summing over all modes gives
	\begin{equation}
		E(t)
		=
		\frac12\|\widetilde{\mathbf r}(t)\|^2
		=
		\frac12\sum_{j=1}^{N_r}\widetilde r_j(t)^2
		\le
		\frac12\sum_{j=1}^{N_r}\widetilde r_j(0)^2e^{-2\lambda_j a_* t},
	\end{equation}
	which proves \eqref{eq:app_energy_modal_sum_re_prop}.
	
	For the weaker descent-oriented case, starting from
	\eqref{eq:app_power_decay_assumption_re_prop} and using \(\|\mathbf r(t)\|^2=2E(t)\), we obtain
	\begin{equation}
		\dot E(t)
		=
		\mathbf r(t)^\top K_{rr}^G\boldsymbol{\gamma}(t)
		\le
		-c(2E(t))^\alpha.
	\end{equation}
	Integrating the associated scalar differential inequality yields
	\begin{equation}
		E(t)\le C(1+t)^{-1/(\alpha-1)}
	\end{equation}
	for some constant \(C>0\), which proves \eqref{eq:app_algebraic_decay_re_prop}.
	\hfill $\square$
	
	\paragraph{Interpretation.}
	Theorem~\ref{prop:modal_decay_alignment} makes clear that residual decay is inherently spectral. Under strong modewise alignment, each mode decays at a rate controlled by \(\lambda_j a_j(t)\), so modes associated with larger effective eigenvalues decay first. As a result, long-time training is dominated by the slower unresolved modes. This explains why spectral imbalance can be harmful: once the fast modes have already become small, maintaining a negative overall energy rate requires the discriminator-induced weighting to remain aligned with the slow modes that still dominate the residual energy.
	
	The last part of the proposition shows that if the discriminator feedback remains descent-oriented but is no longer linearly aligned with the residual, then exponential decay is typically lost, and only algebraic decay can be guaranteed. This provides a natural explanation for the long optimization tails often observed in adversarial PINNs training: even when training does not fail, insufficiently strong alignment with the unresolved modes can substantially slow convergence.
	
	\subsection{Failure modes: ascent and stagnation}
	
	The previous three regimes describe successful but quantitatively different decay. The opposite cases are equally important.
	
	\begin{proposition}[Ascent, stagnation, and unstable oscillation regimes]
		\label{prop:failure_modes_energy}
		Under the residual-energy law \eqref{eq:app_energy_identity_re_prop}:
		\begin{enumerate}
			\item If
			\begin{equation}
				\mathbf r(t)^\top K_{rr}^G(t)\boldsymbol{\gamma}(t)\ge 0
			\end{equation}
			on some time interval, then \(E(t)\) fails to decrease monotonically on that interval. In the strictly positive case, the residual energy increases.
			
			\item If
			\begin{equation}
				\|\boldsymbol{\gamma}(t)\|\to 0
			\end{equation}
			while \(\|K_{rr}^G(t)\|\) remains bounded, then
			\begin{equation}
				\|\dot{\mathbf r}(t)\|
				\le
				\|K_{rr}^G(t)\|\,\|\boldsymbol{\gamma}(t)\|
				\to 0.
			\end{equation}
			Thus the first-order residual dynamics stall. If this happens before \(\mathbf r(t)\) becomes small, training enters a plateau regime.
			
			\item Let
			\begin{equation}
				s(t):=\mathbf r(t)^\top K_{rr}^G(t)\boldsymbol{\gamma}(t).
			\end{equation}
			If \(s(t)\) changes sign repeatedly on a time interval, namely if there exist sequences
			\begin{equation}
				t_1<t_2<t_3<\cdots
			\end{equation}
			such that
			\begin{equation}
				s(t_{2m-1})>0,
				\qquad
				s(t_{2m})<0,
			\end{equation}
			for all admissible \(m\), then the residual-energy slope alternates between ascent-oriented and descent-oriented phases. In this case, \(E(t)\) does not exhibit a stable monotone decay trend, but instead undergoes oscillatory evolution. Consequently, training becomes dynamically unstable in the sense that the discriminator-induced update direction repeatedly switches between alignment and misalignment with residual descent.
		\end{enumerate}
	\end{proposition}
	
	\begin{remark}
		Proposition~\ref{prop:failure_modes_energy} identifies three representative failure modes for adversarially trained PINNs. First, if the discriminator-induced weighting is misaligned with the residual after modulation by the generator residual NTK, so that
		\(
		\mathbf r(t)^\top K_{rr}^G(t)\boldsymbol{\gamma}(t)>0,
		\)
		then the residual energy increases rather than decreases. Second, if the discriminator-induced gradient vanishes, namely
		\(
		\|\boldsymbol{\gamma}(t)\|\to 0,
		\)
		then the first-order residual dynamics stall; this is especially likely to occur when the discriminator becomes overly strong and produces near-vanishing generator-side gradients, as can happen in GAN-type or LSGAN-type training. Third, even when the discriminator gradient does not vanish, it may fail to remain consistently descent-oriented, so that the induced weighting alternates between favorable and unfavorable directions. In this case, the first-order indicator changes sign repeatedly, leading to oscillatory and unstable training behavior. These three phenomena are also clearly illustrated in the right panel of Fig.~\ref{fig:poigdloss} in the main text.
	\end{remark}
	
	\subsection{Discrete-time interpretation for the alternating algorithm}
	
	The same logic carries over to the practical alternating algorithm. Recall the first-order discrete expansion
	\begin{equation}
		E^{m+1}-E^m
		=
		\eta_G S^m+\mathcal O(\eta_G^2),
		\qquad
		S^m:=(\mathbf r^m)^\top K_{rr}^{G,m}\boldsymbol{\gamma}^m.
	\end{equation}
	Thus \(S^m\) is the discrete analogue of the continuous energy-dissipation rate.
	
	\paragraph{Contractive regime.}
	If
	\begin{equation}
		S^m\le -cE^m
	\end{equation}
	uniformly and \(\eta_G\) is sufficiently small, then
	\begin{equation}
		E^{m+1}\le (1-c\eta_G)E^m,
	\end{equation}
	which gives exponential decay in discrete time.
	
	\paragraph{Slow-decay regime.}
	More generally, if \(S^m\) scales like a higher power of \(E^m\), then one expects a slower algebraic-type decay, consistent with the continuous-time estimate \eqref{eq:app_algebraic_decay_re_prop}.
	
	\paragraph{Failure regime.}
	If \(S^m\) becomes positive or nearly vanishes, then the discrete dynamics exhibit ascent or stagnation, respectively.
	
	\paragraph{Implicit spectral balancing effect of the acceptance rule.}
	Although the proposed alternating algorithm does not explicitly enforce mode-by-mode spectral balancing, the score-based acceptance rule can still provide an indirect benefit from the spectral viewpoint. Indeed, in the eigenbasis of the generator residual NTK,
	\begin{equation}
		S^m
		=
		(\mathbf r^m)^\top K_{rr}^{G,m}\boldsymbol{\gamma}^m
		=
		\sum_{j=1}^{N_r}\lambda_j^m\,\widetilde r_j^{\,m}\,\widetilde\gamma_j^{\,m}.
	\end{equation}
	Thus, accepting a discriminator update only when \(S^m<0\) tends to favor discriminator states whose induced weighting yields a negative net contribution across the unresolved dominant modes. In particular, when fast modes have already decayed and their amplitudes \(\widetilde r_j^{\,m}\) become small, maintaining \(S^m<0\) increasingly requires useful correction on the slow modes that still dominate the residual energy. Consequently, the acceptance rule may suppress discriminator states that overemphasize already-resolved modes while underweighting unresolved ones.
	
	From this perspective, the proposed method does not explicitly balance the spectrum, but it may exert an \emph{implicit spectral regularization} effect on the alternating dynamics by filtering out discriminator states that produce poor net modal descent.
	
	Overall, the continuous energy analysis provides a direct interpretation of the practical score used in the alternating algorithm. The convergence rate of the residual energy in adversarial PINNs is governed not by the generator residual NTK alone, but by the interaction between the generator-side kernel geometry, the discriminator-induced weighting, and the evolving residual itself.
	
	\section{Different Dynamics Induced by Residual Input and Squared-Residual Input}
	\label{app:x_vs_x2_dynamics}
	
	In most existing adversarial PINNs formulations, the discriminator takes the residual itself as input,
	\[
	r_i(\theta):=\mathcal R(x_i;\theta),\qquad i=1,\dots,N_r.
	\]
	By contrast, standard PINNs training is fundamentally driven by the squared residual, since the canonical physics loss is proportional to
	\[
	\frac{1}{N_r}\sum_{i=1}^{N_r} r_i(\theta)^2.
	\]
	This raises a natural question: what changes if the discriminator input is replaced by \(r_i^2\)? As we show below, this is not a minor reparameterization. It changes the generator dynamics from an \emph{externally forced preconditioned residual descent}
	\[
	\dot{\mathbf r}=K_{rr}^G\Gamma\mathbf 1
	\]
	into a \emph{state-dependent multiplicative feedback law}
	\[
	\dot{\mathbf r}=K_{rr}^G\widetilde{\Gamma}\mathbf r.
	\]
	Consequently, both the residual evolution law and the energy dissipation mechanism become fundamentally different.

	\subsection{Generator dynamics with squared-residual input}
	\label{app:x2_input_dynamics}
	
	We now consider the alternative design in which the discriminator no longer takes the residual \(r_i\) itself as input, but instead takes the squared residual \(r_i^2\). This seemingly simple modification changes the structure of the generator dynamics in an essential way. In the residual-input case, the discriminator induces an additive forcing term on the residual dynamics. By contrast, under squared-residual input, the chain rule introduces an additional factor of \(r_i\), so that the discriminator enters multiplicatively through the current residual state. The following proposition makes this distinction precise.
	
	\begin{proposition}[Generator dynamics with squared-residual discriminator input]
		\label{prop:x2_generator_dynamics}
		Let
		\begin{equation}
			s_i(\theta):=r_i(\theta)^2,
			\qquad i=1,\dots,N_r,
		\end{equation}
		and define
		\begin{equation}
			\mathbf s(\theta):=[s_1(\theta),\dots,s_{N_r}(\theta)]^\top.
		\end{equation}
		Consider the generator loss
		\begin{equation}
			\mathcal L_G^{(X^2)}(\theta,\phi)
			=
			\frac{1}{N_r}\sum_{i=1}^{N_r}
			R\!\big(f(s_i(\theta);\phi)\big).
		\end{equation}
		Then:
		\begin{enumerate}
			\item The generator gradient is
			\begin{equation}
				\nabla_\theta \mathcal L_G^{(X^2)}
				=
				\frac{2}{N_r}\sum_{i=1}^{N_r}
				r_i\,
				R'\!\big(f_i^{(X^2)}\big)\,
				\partial_s f_i^{(X^2)}\,
				\nabla_\theta r_i,
			\end{equation}
			where
			\[
			f_i^{(X^2)}:=f(s_i;\phi),
			\qquad
			\partial_s f_i^{(X^2)}:=\partial_s f(s_i;\phi).
			\]
			
			\item The generator gradient flow becomes
			\begin{equation}
				\dot{\theta}
				=
				-\frac{2}{N_r}\sum_{i=1}^{N_r}
				r_i\,
				R'\!\big(f_i^{(X^2)}\big)\,
				\partial_s f_i^{(X^2)}\,
				\nabla_\theta r_i.
				\label{eq:theta_flow_x2_appendix_prop}
			\end{equation}
			
			\item Let \(J_r\) denote the residual Jacobian and \(K_{rr}^G(\theta)=J_rJ_r^\top\) the generator residual NTK. Then the residual dynamics satisfy
			\begin{equation}
				\dot{\mathbf r}
				=
				-\,K_{rr}^G(\theta)\,\boldsymbol{\gamma}^{(X^2)},
			\end{equation}
			where
			\begin{equation}
				\gamma_i^{(X^2)}
				=
				\frac{2}{N_r}\,
				r_i\,
				R'\!\big(f(r_i^2;\phi)\big)\,
				\partial_s f(r_i^2;\phi).
			\end{equation}
			
			\item Defining
			\begin{equation}
				\widetilde{\gamma}_i
				:=
				-\frac{2}{N_r}
				R'\!\big(f(r_i^2;\phi)\big)\,
				\partial_s f(r_i^2;\phi),
			\end{equation}
			and
			\begin{equation}
				\widetilde{\Gamma}
				=
				\mathrm{diag}(\widetilde{\gamma}_1,\dots,\widetilde{\gamma}_{N_r}),
			\end{equation}
			one has
			\begin{equation}
				\boldsymbol{\gamma}^{(X^2)}=-\widetilde{\Gamma}\mathbf r,
			\end{equation}
			and therefore
			\begin{equation}
				\dot{\mathbf r}
				=
				K_{rr}^G(\theta)\,\widetilde{\Gamma}\,\mathbf r.
				\label{eq:r_dyn_x2_appendix_prop}
			\end{equation}
			
			\item The corresponding residual-energy law is
			\begin{equation}
				\dot E
				=
				\mathbf r^\top K_{rr}^G(\theta)\,\widetilde{\Gamma}\,\mathbf r.
				\label{eq:E_dyn_x2_appendix_prop}
			\end{equation}
		\end{enumerate}
	\end{proposition}
	
	\paragraph{Proof.}
	Since \(s_i=r_i^2\), we have
	\[
	\nabla_\theta s_i = 2r_i\,\nabla_\theta r_i.
	\]
	Therefore,
	\begin{align}
		\nabla_\theta \mathcal L_G^{(X^2)}
		&=
		\frac{1}{N_r}\sum_{i=1}^{N_r}
		R'\!\big(f(s_i;\phi)\big)\,
		\partial_s f(s_i;\phi)\,
		\nabla_\theta s_i \notag\\
		&=
		\frac{2}{N_r}\sum_{i=1}^{N_r}
		r_i\,
		R'\!\big(f_i^{(X^2)}\big)\,
		\partial_s f_i^{(X^2)}\,
		\nabla_\theta r_i,
	\end{align}
	which proves the first statement. The gradient-flow equation
	\eqref{eq:theta_flow_x2_appendix_prop} follows immediately.
	
	Next, multiplying by the residual Jacobian \(J_r\), we obtain
	\begin{equation}
		\dot{\mathbf r}
		=
		J_r\dot{\theta}
		=
		-\,K_{rr}^G(\theta)\,\boldsymbol{\gamma}^{(X^2)},
	\end{equation}
	where
	\begin{equation}
		\gamma_i^{(X^2)}
		=
		\frac{2}{N_r}\,
		r_i\,
		R'\!\big(f(r_i^2;\phi)\big)\,
		\partial_s f(r_i^2;\phi).
	\end{equation}
	Defining
	\[
	\widetilde{\gamma}_i
	=
	-\frac{2}{N_r}
	R'\!\big(f(r_i^2;\phi)\big)\,
	\partial_s f(r_i^2;\phi),
	\]
	we have
	\[
	\gamma_i^{(X^2)}=-\widetilde{\gamma}_i\,r_i,
	\qquad
	\boldsymbol{\gamma}^{(X^2)}=-\widetilde{\Gamma}\mathbf r,
	\]
	and hence
	\begin{equation}
		\dot{\mathbf r}
		=
		K_{rr}^G(\theta)\,\widetilde{\Gamma}\,\mathbf r,
	\end{equation}
	which proves \eqref{eq:r_dyn_x2_appendix_prop}. Finally,
	\begin{equation}
		\dot E
		=
		\mathbf r^\top \dot{\mathbf r}
		=
		\mathbf r^\top K_{rr}^G(\theta)\,\widetilde{\Gamma}\,\mathbf r,
	\end{equation}
	which proves \eqref{eq:E_dyn_x2_appendix_prop}.
	\hfill $\square$
	
	\paragraph{Interpretation.}
	Proposition~\ref{prop:x2_generator_dynamics} shows that changing the discriminator input from \(r\) to \(r^2\) does not merely rescale the adversarial signal: it changes the structure of the residual dynamics. In the residual-input case, the discriminator acts through an additive forcing term of the form
	\[
	\dot{\mathbf r}=K_{rr}^G(\theta)\Gamma\mathbf 1,
	\]
	where the weighting is independent of the current residual amplitudes. By contrast, under squared-residual input, the dynamics take the multiplicative form
	\[
	\dot{\mathbf r}=K_{rr}^G(\theta)\widetilde{\Gamma}\mathbf r,
	\]
	so that the discriminator enters through the current residual state itself.
	
	This difference is fundamental. The squared-residual input removes sign information and makes the discriminator sensitive mainly to residual magnitude, while the additional factor \(\mathbf r\) turns the dynamics into a genuine multiplicative feedback system. As a result, both the residual evolution law and the energy law become quadratic in the residual state. This is why squared-residual adversarial training is structurally different from the standard residual-input formulation, and why its later spectral analysis takes a substantially different form.
	
\subsection{Spectral interpretation}
\label{app:x_vs_x2_spectral}

Under the constant-kernel approximation
\begin{equation}
	K_{rr}^G(\theta)\approx K_{rr}^G=:K,
\end{equation}
we analyze the residual-input and squared-residual-input dynamics in the eigenbasis of the generator residual NTK. The main purpose is to clarify how the two input choices differ at the level of modal evolution, and why the squared-residual-input case leads to a genuinely coupled multiplicative system.

Let
\begin{equation}
	K
	=
	U\Lambda U^\top,
	\qquad
	\Lambda=\mathrm{diag}(\lambda_1,\dots,\lambda_{N_r}),
	\qquad
	\lambda_j\ge 0,
\end{equation}
where \(U=[u_1,\dots,u_{N_r}]\) is orthonormal, and expand
\begin{equation}
	\mathbf r=\sum_{j=1}^{N_r}c_j u_j.
\end{equation}

\begin{proposition}[Modal dynamics under residual and squared-residual inputs]
	\label{prop:x_vs_x2_modal_dynamics}
	Under the above notation:
	\begin{enumerate}
		\item If the discriminator input is the residual itself, so that
		\begin{equation}
			\dot{\mathbf r}=K\Gamma\mathbf 1,
		\end{equation}
		then the modal coefficients satisfy
		\begin{equation}
			\dot c_j
			=
			u_j^\top \dot{\mathbf r}
			=
			\lambda_j\,u_j^\top \Gamma\mathbf 1.
			\label{eq:mode_dyn_x_appendix_prop}
		\end{equation}
		
		\item If the discriminator input is the squared residual, so that
		\begin{equation}
			\dot{\mathbf r}=K\widetilde{\Gamma}\mathbf r,
			\label{eq:x2_linear_dyn_appendix_prop}
		\end{equation}
		then the modal coefficients satisfy
		\begin{equation}
			\dot c_j
			=
			u_j^\top \dot{\mathbf r}
			=
			\lambda_j\sum_{k=1}^{N_r}
			c_k\,u_j^\top \widetilde{\Gamma}u_k.
			\label{eq:mode_dyn_x2_appendix_prop}
		\end{equation}
	\end{enumerate}
\end{proposition}

\paragraph{Proof.}
For the residual-input dynamics,
\[
\dot{\mathbf r}=K\Gamma\mathbf 1,
\]
we compute
\begin{equation}
	\dot c_j
	=
	u_j^\top \dot{\mathbf r}
	=
	u_j^\top K\Gamma\mathbf 1
	=
	\lambda_j\,u_j^\top \Gamma\mathbf 1,
\end{equation}
which proves \eqref{eq:mode_dyn_x_appendix_prop}.

For the squared-residual-input dynamics,
\[
\dot{\mathbf r}=K\widetilde{\Gamma}\mathbf r,
\]
we similarly obtain
\begin{align}
	\dot c_j
	&=
	u_j^\top \dot{\mathbf r}
	=
	u_j^\top K\widetilde{\Gamma}\mathbf r
	=
	\lambda_j\,u_j^\top \widetilde{\Gamma}\mathbf r \notag\\
	&=
	\lambda_j\sum_{k=1}^{N_r}c_k\,u_j^\top \widetilde{\Gamma}u_k,
\end{align}
which proves \eqref{eq:mode_dyn_x2_appendix_prop}.
\hfill $\square$

\paragraph{Interpretation.}
Proposition~\ref{prop:x_vs_x2_modal_dynamics} reveals the structural difference between the two input choices. In the residual-input case, each mode is driven by an external discriminator-induced forcing term \(u_j^\top\Gamma\mathbf 1\). In the squared-residual-input case, by contrast, the evolution of mode \(j\) depends on all modal amplitudes \(\{c_k\}\), so the residual spectrum is fed back into its own dynamics. This is the first indication that squared-residual input leads to a genuinely multiplicative feedback system.

To understand this multiplicative structure more systematically, it is useful to introduce the coupled operator \(K\widetilde{\Gamma}\) and its symmetric counterpart. The same spectral factorization used for the linearly aligned residual-input dynamics applies here, with \(A\) replaced by \(\widetilde{\Gamma}\). The crucial difference is the sign convention: in the residual-input aligned case, decay corresponds to \(A\succeq0\) in \(\dot{\mathbf r}=-KA\mathbf r\), whereas in the squared-residual-input case, decay corresponds to \(\widetilde{\Gamma}\preceq0\) in \(\dot{\mathbf r}=K\widetilde{\Gamma}\mathbf r\).

\begin{proposition}[Spectral factorization of the squared-residual-input dynamics]
	\label{prop:x2_spectral_factorization}
	Define
	\begin{equation}
		M:=K\widetilde{\Gamma},
		\qquad
		H:=K^{1/2}\widetilde{\Gamma}K^{1/2}.
	\end{equation}
	Assume that \(K\) is symmetric positive definite on the active residual subspace, and that \(\widetilde{\Gamma}\in\mathbb R^{N_r\times N_r}\) is symmetric. Then:
	\begin{enumerate}
		\item \(M\) and \(H\) are similar. \(M\) and \(H\) have the same eigenvalues.
		
		\item Since \(H\) is symmetric, all eigenvalues of \(M\) are real.
		
		\item Since \(H\) is congruent to \(\widetilde{\Gamma}\), Sylvester's law of inertia implies that \(H\) and \(\widetilde{\Gamma}\) have the same inertia, and therefore so does \(M\).
		
		\item If \(\widetilde{\Gamma}\preceq 0\), then all eigenvalues of \(M\) are nonpositive; if \(\widetilde{\Gamma}\succeq 0\), then all eigenvalues of \(M\) are nonnegative; if \(\widetilde{\Gamma}\) is indefinite, then \(M\) has both positive and negative eigenvalues.
	\end{enumerate}
\end{proposition}

\paragraph{Proof.}
The similarity relation follows directly from
	\begin{equation}
	K^{-1/2}MK^{1/2}
	=
	K^{-1/2}(K\widetilde{\Gamma})K^{1/2}
	=
	K^{1/2}\widetilde{\Gamma}K^{1/2}
	=
	H.
\end{equation}
Hence \(M\) and \(H\) have the same eigenvalues. Since \(H\) is symmetric whenever \(\widetilde{\Gamma}\) is symmetric, its spectrum is real, and therefore so is the spectrum of \(M\). Moreover, \(H\) is congruent to \(\widetilde{\Gamma}\), so Sylvester's law of inertia implies that they have the same inertia; similarity then transfers the same inertia to \(M\). The sign-definite and indefinite cases follow immediately.
\hfill $\square$

\paragraph{Interpretation.}
Proposition~\ref{prop:x2_spectral_factorization} shows that the squared-residual-input dynamics are governed not by \(K\) or \(\widetilde{\Gamma}\) separately, but by their coupled spectrum. The symmetric matrix \(H=K^{1/2}\widetilde{\Gamma}K^{1/2}\) provides the most transparent representation of this coupling, and its sign structure determines whether the active modes are contractive, expansive, or mixed.

This also gives a direct way to compare the spectrum of the coupled operator with those of \(K\) and \(\widetilde{\Gamma}\).
\begin{proposition}[Spectral bounds and energy behavior]
	\label{prop:x2_energy_spectral}
	Under the assumptions of Proposition~\ref{prop:x2_spectral_factorization}:
	\begin{enumerate}
		\item If \(\widetilde{\Gamma}\succeq 0\), then the extreme eigenvalues of \(M\) satisfy
		\begin{equation}
			\lambda_{\min}(M)
			=
			\lambda_{\min}(H)
			\ge
			\lambda_{\min}(K)\lambda_{\min}(\widetilde{\Gamma}),
			\qquad
			\lambda_{\max}(M)
			=
			\lambda_{\max}(H)
			\le
			\lambda_{\max}(K)\lambda_{\max}(\widetilde{\Gamma}).
			\label{eq:x2_eig_extreme_bounds_prop}
		\end{equation}
		
		\item Introduce
		\begin{equation}
			\mathbf z:=K^{-1/2}\mathbf r.
		\end{equation}
		Then the squared-residual-input dynamics become
		\begin{equation}
			\dot{\mathbf z}=H\mathbf z,
			\label{eq:x2_z_dyn_appendix_prop2}
		\end{equation}
		and the residual energy satisfies
		\begin{equation}
			\lambda_{\min}(K)\|\mathbf z\|^2
			\le
			\|\mathbf r\|^2
			\le
			\lambda_{\max}(K)\|\mathbf z\|^2.
			\label{eq:x2_norm_equiv_prop}
		\end{equation}
		
		\item If \(\widetilde{\Gamma}\preceq 0\), then \(H\preceq 0\). In the strictly negative definite case \(\widetilde{\Gamma}\prec 0\),
		\begin{equation}
			E(t)
			\le
			\kappa(K)\,E(0)\,e^{2\lambda_{\max}(H)t}
			\le
			\kappa(K)\,E(0)\,e^{-2\lambda_{\min}(K)|\lambda_{\max}(\widetilde{\Gamma})|t},
			\label{eq:x2_energy_decay_appendix_prop2}
		\end{equation}
		where \(\lambda_{\max}(H)<0\) and
		\begin{equation}
			\kappa(K):=\frac{\lambda_{\max}(K)}{\lambda_{\min}(K)}.
		\end{equation}
		
		\item If \(\widetilde{\Gamma}\succeq 0\), then the transformed energy
\(\frac12\|\mathbf z(t)\|^2\) is nondecreasing. In the strictly positive definite case \(\widetilde{\Gamma}\succ 0\), the transformed dynamics are expansive, and the original residual energy admits the lower bound
\begin{equation}
	E(t)
	\ge
	\kappa(K)^{-1}E(0)\,e^{2\lambda_{\min}(H)t}
	\ge
	\kappa(K)^{-1}E(0)\,e^{2\lambda_{\min}(K)\lambda_{\min}(\widetilde{\Gamma})t}.
    \label{eq:x2_energy_growth_appendix_prop2}
\end{equation}
		
		\item If \(\widetilde{\Gamma}\) is indefinite, then some modes decay while others grow, and the residual energy is not guaranteed to be monotone.
	\end{enumerate}
\end{proposition}

\paragraph{Proof.}
For the extreme eigenvalue bounds, note that
\[
H=K^{1/2}\widetilde{\Gamma}K^{1/2}
\]
is symmetric. If \(\widetilde{\Gamma}\succeq 0\), then for any \(x\neq 0\),
\begin{equation}
	x^\top Hx
	=
	(K^{1/2}x)^\top \widetilde{\Gamma}(K^{1/2}x)
	\ge
	\lambda_{\min}(\widetilde{\Gamma})\,\|K^{1/2}x\|^2
	\ge
	\lambda_{\min}(K)\lambda_{\min}(\widetilde{\Gamma})\,\|x\|^2.
\end{equation}
Taking the minimum over unit vectors \(x\) yields
\[
\lambda_{\min}(H)\ge \lambda_{\min}(K)\lambda_{\min}(\widetilde{\Gamma}).
\]
Similarly,
\begin{equation}
	x^\top Hx
	\le
	\lambda_{\max}(\widetilde{\Gamma})\,\|K^{1/2}x\|^2
	\le
	\lambda_{\max}(K)\lambda_{\max}(\widetilde{\Gamma})\,\|x\|^2,
\end{equation}
and taking the maximum over unit vectors gives
\[
\lambda_{\max}(H)\le \lambda_{\max}(K)\lambda_{\max}(\widetilde{\Gamma}).
\]
Since \(M\) and \(H\) are similar by Proposition~\ref{prop:x2_spectral_factorization}, they have the same eigenvalues, which proves \eqref{eq:x2_eig_extreme_bounds_prop}.

Next, setting \(\mathbf z=K^{-1/2}\mathbf r\), we obtain
\[
\dot{\mathbf z}
=
K^{-1/2}\dot{\mathbf r}
=
K^{-1/2}K\widetilde{\Gamma}\mathbf r
=
K^{1/2}\widetilde{\Gamma}K^{1/2}\mathbf z
=
H\mathbf z,
\]
which proves \eqref{eq:x2_z_dyn_appendix_prop2}. Since \(\mathbf r=K^{1/2}\mathbf z\), the norm equivalence \eqref{eq:x2_norm_equiv_prop} follows immediately.

If \(\widetilde{\Gamma}\preceq 0\), then \(H\preceq 0\). In the strictly negative definite case, \(\lambda_{\max}(H)<0\), and
\begin{equation}
	\frac{d}{dt}\frac12\|\mathbf z(t)\|^2
	=
	\mathbf z(t)^\top H\mathbf z(t)
	\le
	\lambda_{\max}(H)\|\mathbf z(t)\|^2.
\end{equation}
Gr\"onwall's inequality yields
\[
\|\mathbf z(t)\|^2
\le
e^{2\lambda_{\max}(H)t}\|\mathbf z(0)\|^2.
\]
Combining this with \eqref{eq:x2_norm_equiv_prop} gives
\[
E(t)
\le
\kappa(K)\,E(0)\,e^{2\lambda_{\max}(H)t}.
\]
Moreover, since \(\widetilde{\Gamma}\prec 0\), we have
\[
\lambda_{\max}(H)
\le
\lambda_{\min}(K)\lambda_{\max}(\widetilde{\Gamma})
=
-\lambda_{\min}(K)|\lambda_{\max}(\widetilde{\Gamma})|,
\]
which proves \eqref{eq:x2_energy_decay_appendix_prop2}.

If \(\widetilde{\Gamma}\succeq 0\), then \(H\succeq 0\), and therefore
\begin{equation}
	\frac{d}{dt}\frac12\|\mathbf z(t)\|^2
	=
	\mathbf z(t)^\top H\mathbf z(t)\ge 0,
\end{equation}
so the energy is nondecreasing. In the strictly positive definite case \(\widetilde{\Gamma}\succ 0\), one has \(\lambda_{\min}(H)>0\), and
\begin{equation}
	\frac{d}{dt}\frac12\|\mathbf z(t)\|^2
	\ge
	\lambda_{\min}(H)\|\mathbf z(t)\|^2.
\end{equation}
Applying Gr\"onwall's inequality gives
\[
\|\mathbf z(t)\|^2
\ge
e^{2\lambda_{\min}(H)t}\|\mathbf z(0)\|^2,
\]
which, together with \eqref{eq:x2_norm_equiv_prop}, yields
\[
E(t)
\ge
\kappa(K)^{-1}E(0)e^{2\lambda_{\min}(H)t}.
\]
Using
\[
\lambda_{\min}(H)\ge \lambda_{\min}(K)\lambda_{\min}(\widetilde{\Gamma}),
\]
we obtain \eqref{eq:x2_energy_growth_appendix_prop2}.

Finally, if \(\widetilde{\Gamma}\) is indefinite, then by Proposition~\ref{prop:x2_spectral_factorization}, \(H\) has both positive and negative eigenvalues. Hence the linear system
\[
\dot{\mathbf z}=H\mathbf z
\]
contains both contractive and expansive modes, so some components decay while others grow, and monotonicity of the residual energy is no longer guaranteed.
\hfill $\square$
\paragraph{Interpretation.}
The three propositions above make the spectral distinction between the two input choices precise. With residual input, the discriminator acts as an external forcing projected onto each kernel mode. With squared-residual input, the residual state is fed back into the dynamics through the coupled operator \(K\widetilde{\Gamma}\). As a result, the squared-residual-input case forms a genuine multiplicative feedback system.

The energy behavior is therefore controlled by the sign structure of the coupled spectrum. If \(\widetilde{\Gamma}\preceq 0\), the active modes are non-expansive; if \(\widetilde{\Gamma}\prec 0\), the residual energy decays exponentially; if \(\widetilde{\Gamma}\succeq 0\), the active modes become expansive and the energy grows; and if \(\widetilde{\Gamma}\) is indefinite, the system contains both damping and amplification, which naturally leads to oscillatory or unstable behavior. This is why replacing the discriminator input \(r\) by \(r^2\) should be viewed not as a cosmetic modification, but as a structural change in the adversarial PINNs dynamics.

	\subsection{Interpretation and implications}
	\label{app:x_vs_x2_interpretation_implications}
	
	The analysis above shows that replacing the discriminator input \(r\) by \(r^2\) changes the adversarial PINNs dynamics in a structural rather than merely parametric manner. In the residual-input case,
	\[
	\dot{\mathbf r}=K_{rr}^G\Gamma\mathbf 1,
	\]
	so the discriminator acts as an external sample-wise forcing term. In the squared-residual-input case,
	\[
	\dot{\mathbf r}=K_{rr}^G\widetilde{\Gamma}\mathbf r,
	\]
	so the discriminator enters multiplicatively through the current residual state. At the same time, the map \(r\mapsto r^2\) removes sign information, making the discriminator sensitive only to residual magnitude rather than to its sign.
	
	This modification has two opposite consequences. On the positive side, the extra factor \(r_i\) naturally amplifies the contribution of large-residual samples, which makes the adversarial signal more consistent with the squared-residual structure of standard PINNs training and may help concentrate optimization on poorly fitted regions. On the negative side, the same mechanism automatically weakens the update as \(r_i\to 0\), which can slow late-stage refinement, and it may over-emphasize a small subset of large-residual modes, thereby worsening spectral imbalance.
	
	Correspondingly, the energy law changes from the linear form
	\[
	\dot E=\mathbf r^\top K_{rr}^G\Gamma\mathbf 1
	\]
	to the quadratic form
	\[
	\dot E=\mathbf r^\top K_{rr}^G\widetilde{\Gamma}\mathbf r.
	\]
	Thus, changing the discriminator input from \(r\) to \(r^2\) should not be viewed as a minor implementation choice: it changes the continuous-time training law, the NTK-level spectral coupling, and the mechanism by which the discriminator influences convergence.
	
	\section{Soft-Constrained PINNs under Adversarial Training}
	\label{app:soft_adv_pinn}
	
	In the hard-constrained setting considered in the main text, the solution ansatz is designed so that boundary and initial conditions are satisfied identically, and adversarial training is applied only to the interior PDE residual. In the soft-constrained setting, by contrast, the solution network does not enforce these constraints exactly. As a result, the PDE residual, the boundary-condition mismatch, and the initial-condition mismatch must all be controlled during training. This naturally leads to a multi-discriminator adversarial formulation, in which a single generator \(u_\theta\) is coupled to three discriminators \(f_1,f_2,f_3\), each acting on a different violation channel.
	
	In this appendix, we derive the resulting coupled dynamics in a unified way and show that, unlike the hard-constrained case, soft-constrained adversarial PINNs training is governed by a block-structured kernel dynamics with cross-coupling among the interior, boundary, and initial-condition channels.
	
	\subsection{Problem setup and violation channels}
	\label{app:soft_setup}
	
	Consider a PDE of the form
	\begin{equation}
		\mathcal N[u](x)=a(x), \qquad x\in\Omega,
	\end{equation}
	with boundary condition
	\begin{equation}
		\mathcal G_B[u](x)=g_B(x), \qquad x\in\partial\Omega,
	\end{equation}
	and initial condition
	\begin{equation}
		\mathcal G_0[u](x)=g_0(x), \qquad x\in\Omega_0,
	\end{equation}
	where \(\Omega_0\) denotes the initial manifold (for example, \(t=0\) in time-dependent problems).
	
	Let \(u_\theta\) be the generator (solution network). In the soft-constrained setting, we define three types of violations:
	
	\paragraph{Interior residual.}
	Given interior collocation points \(\{x_i^r\}_{i=1}^{N_r}\subset\Omega\), define
	\begin{equation}\label{residual}
		r_i(\theta)
		:=
		\mathcal R(x_i^r;\theta)
		:=
		\mathcal N[u_\theta](x_i^r)-a(x_i^r),
		\qquad i=1,\dots,N_r.
	\end{equation}
	Collect them into
	\begin{equation}
		\mathbf r(\theta)
		=
		[r_1(\theta),\dots,r_{N_r}(\theta)]^\top
		\in\mathbb R^{N_r}.
	\end{equation}
	
	\paragraph{Boundary-condition mismatch.}
	Given boundary collocation points \(\{x_j^b\}_{j=1}^{N_b}\subset\partial\Omega\), define
	\begin{equation}\label{bound}
		b_j(\theta)
		:=
		\mathcal B(x_j^b;\theta)
		:=
		\mathcal G_B[u_\theta](x_j^b)-g_B(x_j^b),
		\qquad j=1,\dots,N_b.
	\end{equation}
	Collect them into
	\begin{equation}
		\mathbf b(\theta)
		=
		[b_1(\theta),\dots,b_{N_b}(\theta)]^\top
		\in\mathbb R^{N_b}.
	\end{equation}
	
	\paragraph{Initial-condition mismatch.}
	Given initial-condition points \(\{x_k^0\}_{k=1}^{N_0}\subset\Omega_0\), define
	\begin{equation}\label{initial}
		c_k(\theta)
		:=
		\mathcal I(x_k^0;\theta)
		:=
		\mathcal G_0[u_\theta](x_k^0)-g_0(x_k^0),
		\qquad k=1,\dots,N_0.
	\end{equation}
	Collect them into
	\begin{equation}
		\mathbf c(\theta)
		=
		[c_1(\theta),\dots,c_{N_0}(\theta)]^\top
		\in\mathbb R^{N_0}.
	\end{equation}
	
	Thus, instead of a single residual vector, soft-constrained PINNs involve three generator-dependent violation vectors:
	\[
	\mathbf r,\qquad \mathbf b,\qquad \mathbf c.
	\]
	
	\subsection{Adversarial formulation with three discriminators}
	\label{app:soft_objective}
	
	As in the main text, the target for each violation channel is the zero state. Therefore, each channel induces a generated empirical distribution and a real distribution \(\delta_0\).
	
	For the interior residual channel,
	\begin{equation}
		\hat{\mu}^{\,r}_{\theta}
		=
		\frac{1}{N_r}\sum_{i=1}^{N_r}\delta_{r_i(\theta)},
		\qquad
		\mu_{\mathrm{real}}^{\,r}=\delta_0.
	\end{equation}
	For the boundary channel,
	\begin{equation}
		\hat{\mu}^{\,b}_{\theta}
		=
		\frac{1}{N_b}\sum_{j=1}^{N_b}\delta_{b_j(\theta)},
		\qquad
		\mu_{\mathrm{real}}^{\,b}=\delta_0.
	\end{equation}
	For the initial channel,
	\begin{equation}
		\hat{\mu}^{\,0}_{\theta}
		=
		\frac{1}{N_0}\sum_{k=1}^{N_0}\delta_{c_k(\theta)},
		\qquad
		\mu_{\mathrm{real}}^{\,0}=\delta_0.
	\end{equation}
	
	We introduce three discriminators with separate parameters:
	\[
	f_1(\cdot;\phi_1),\qquad f_2(\cdot;\phi_2),\qquad f_3(\cdot;\phi_3),
	\]
	where
	\begin{itemize}
		\item \(f_1\) acts on interior residual samples \(r_i\),
		\item \(f_2\) acts on boundary mismatch samples \(b_j\),
		\item \(f_3\) acts on initial mismatch samples \(c_k\).
	\end{itemize}
	
	Using the unified GAN notation \(P,Q,R\) from the main text, the discriminator objectives are
	\begin{align}
		\max_{\phi_1}\ \mathcal L_{D_1}(\theta,\phi_1)
		&=
		\frac{1}{N_r}\sum_{i=1}^{N_r} P\!\big(f_1(0;\phi_1)\big)
		+
		\frac{1}{N_r}\sum_{i=1}^{N_r} Q\!\big(f_1(r_i(\theta);\phi_1)\big),
		\\
		\max_{\phi_2}\ \mathcal L_{D_2}(\theta,\phi_2)
		&=
		\frac{1}{N_b}\sum_{j=1}^{N_b} P\!\big(f_2(0;\phi_2)\big)
		+
		\frac{1}{N_b}\sum_{j=1}^{N_b} Q\!\big(f_2(b_j(\theta);\phi_2)\big),
		\\
		\max_{\phi_3}\ \mathcal L_{D_3}(\theta,\phi_3)
		&=
		\frac{1}{N_0}\sum_{k=1}^{N_0} P\!\big(f_3(0;\phi_3)\big)
		+
		\frac{1}{N_0}\sum_{k=1}^{N_0} Q\!\big(f_3(c_k(\theta);\phi_3)\big).
	\end{align}
	
	Correspondingly, the generator minimizes the sum of three adversarial objectives:
	\begin{equation}
		\mathcal L_G(\theta,\phi_1,\phi_2,\phi_3)
		=
		\lambda_r \mathcal L_G^{(r)}
		+
		\lambda_b \mathcal L_G^{(b)}
		+
		\lambda_0 \mathcal L_G^{(0)},
		\label{eq:soft_total_generator_loss}
	\end{equation}
	where
	\begin{align}
		\mathcal L_G^{(r)}
		&=
		\frac{1}{N_r}\sum_{i=1}^{N_r} R\!\big(f_1(r_i(\theta);\phi_1)\big),
		\\
		\mathcal L_G^{(b)}
		&=
		\frac{1}{N_b}\sum_{j=1}^{N_b} R\!\big(f_2(b_j(\theta);\phi_2)\big),
		\\
		\mathcal L_G^{(0)}
		&=
		\frac{1}{N_0}\sum_{k=1}^{N_0} R\!\big(f_3(c_k(\theta);\phi_3)\big).
	\end{align}
	Here \(\lambda_r,\lambda_b,\lambda_0>0\) are channel weights balancing the three soft constraints.
	
	\subsection{Generator gradient flow and channel-wise adversarial weights}
	\label{app:soft_generator_flow}
	
	We now derive the continuous-time generator dynamics with the discriminator parameters \(\phi_1,\phi_2,\phi_3\) frozen.
	
	For notational convenience, define
	\begin{align}
		f_{1,i} &:= f_1(r_i;\phi_1),
		&
		\partial_r f_{1,i} &:= \partial_r f_1(r_i;\phi_1),
		\\
		f_{2,j} &:= f_2(b_j;\phi_2),
		&
		\partial_b f_{2,j} &:= \partial_b f_2(b_j;\phi_2),
		\\
		f_{3,k} &:= f_3(c_k;\phi_3),
		&
		\partial_c f_{3,k} &:= \partial_c f_3(c_k;\phi_3).
	\end{align}
	
	Differentiating \eqref{eq:soft_total_generator_loss} with respect to \(\theta\) yields
	\begin{align}
		\nabla_\theta \mathcal L_G
		&=
		\frac{\lambda_r}{N_r}\sum_{i=1}^{N_r}
		R'(f_{1,i})\,\partial_r f_{1,i}\,\nabla_\theta r_i
		+
		\frac{\lambda_b}{N_b}\sum_{j=1}^{N_b}
		R'(f_{2,j})\,\partial_b f_{2,j}\,\nabla_\theta b_j
		\nonumber\\
		&\quad
		+
		\frac{\lambda_0}{N_0}\sum_{k=1}^{N_0}
		R'(f_{3,k})\,\partial_c f_{3,k}\,\nabla_\theta c_k.
		\label{eq:soft_grad_theta}
	\end{align}
	
	Under generator gradient flow,
	\begin{equation}
		\dot{\theta}
		=
		-\nabla_\theta \mathcal L_G,
	\end{equation}
	we define the three discriminator-induced sample-weight vectors
	\begin{align}
		\gamma_i^{(r)}
		&:=
		-\frac{\lambda_r}{N_r}
		R'(f_{1,i})\,\partial_r f_{1,i},
		\qquad i=1,\dots,N_r,
		\\
		\gamma_j^{(b)}
		&:=
		-\frac{\lambda_b}{N_b}
		R'(f_{2,j})\,\partial_b f_{2,j},
		\qquad j=1,\dots,N_b,
		\\
		\gamma_k^{(0)}
		&:=
		-\frac{\lambda_0}{N_0}
		R'(f_{3,k})\,\partial_c f_{3,k},
		\qquad k=1,\dots,N_0.
	\end{align}
	Let
	\[
	\boldsymbol{\gamma}^{(r)}=[\gamma_1^{(r)},\dots,\gamma_{N_r}^{(r)}]^\top,\qquad
	\boldsymbol{\gamma}^{(b)}=[\gamma_1^{(b)},\dots,\gamma_{N_b}^{(b)}]^\top,\qquad
	\boldsymbol{\gamma}^{(0)}=[\gamma_1^{(0)},\dots,\gamma_{N_0}^{(0)}]^\top.
	\]
	Then \eqref{eq:soft_grad_theta} can be rewritten compactly as
	\begin{equation}
		\dot{\theta}
		=
		J_r^\top \boldsymbol{\gamma}^{(r)}
		+
		J_b^\top \boldsymbol{\gamma}^{(b)}
		+
		J_0^\top \boldsymbol{\gamma}^{(0)},
		\label{eq:theta_dot_soft_compact}
	\end{equation}
	where the three Jacobians are
	\begin{align}
		J_r(\theta)
		&:=
		\begin{bmatrix}
			(\nabla_\theta r_1)^\top\\
			\vdots\\
			(\nabla_\theta r_{N_r})^\top
		\end{bmatrix}
		\in\mathbb R^{N_r\times p},
		\\
		J_b(\theta)
		&:=
		\begin{bmatrix}
			(\nabla_\theta b_1)^\top\\
			\vdots\\
			(\nabla_\theta b_{N_b})^\top
		\end{bmatrix}
		\in\mathbb R^{N_b\times p},
		\\
		J_0(\theta)
		&:=
		\begin{bmatrix}
			(\nabla_\theta c_1)^\top\\
			\vdots\\
			(\nabla_\theta c_{N_0})^\top
		\end{bmatrix}
		\in\mathbb R^{N_0\times p},
	\end{align}
	and \(p=\dim(\theta)\).
	
	Equation \eqref{eq:theta_dot_soft_compact} already reveals the key difference from the hard-constrained case: the generator update is now the sum of three adversarially weighted Jacobian-transpose terms, one for each constraint channel.
	\subsection{Coupled violation dynamics and block kernel structure}
	\label{app:soft_block_dynamics}
	
	We now derive the generator-induced dynamics for the three violation channels in the soft-constrained setting. Unlike the hard-constrained case, where the state is a single residual vector, the soft-constrained formulation involves three coupled channels: the interior residual, the boundary violation, and the initial-condition violation. Since all three channels depend on the same generator parameters, their dynamics are not independent. Instead, the resulting evolution is governed by a block kernel operator with both self-kernel and cross-kernel interactions.
	
	\begin{proposition}[Coupled soft-constrained violation dynamics]
		\label{prop:soft_block_dynamics}
		Let \(\mathbf r\), \(\mathbf b\), and \(\mathbf c\) denote the interior, boundary, and initial-condition violation vectors, respectively, and let the generator flow be given by \eqref{eq:theta_dot_soft_compact}. Then the induced channel-wise dynamics satisfy
		\begin{align}
			\dot{\mathbf r}
			&=
			J_r \dot{\theta}
			=
			J_rJ_r^\top \boldsymbol{\gamma}^{(r)}
			+
			J_rJ_b^\top \boldsymbol{\gamma}^{(b)}
			+
			J_rJ_0^\top \boldsymbol{\gamma}^{(0)},
			\label{eq:r_dot_soft_pre_prop}
			\\
			\dot{\mathbf b}
			&=
			J_b \dot{\theta}
			=
			J_bJ_r^\top \boldsymbol{\gamma}^{(r)}
			+
			J_bJ_b^\top \boldsymbol{\gamma}^{(b)}
			+
			J_bJ_0^\top \boldsymbol{\gamma}^{(0)},
			\label{eq:b_dot_soft_pre_prop}
			\\
			\dot{\mathbf c}
			&=
			J_0 \dot{\theta}
			=
			J_0J_r^\top \boldsymbol{\gamma}^{(r)}
			+
			J_0J_b^\top \boldsymbol{\gamma}^{(b)}
			+
			J_0J_0^\top \boldsymbol{\gamma}^{(0)}.
			\label{eq:c_dot_soft_pre_prop}
		\end{align}
		
		Define the self-kernel blocks
		\begin{equation}
			K_{rr}^G:=J_rJ_r^\top,\qquad
			K_{bb}^G:=J_bJ_b^\top,\qquad
			K_{00}^G:=J_0J_0^\top,
		\end{equation}
		and the cross-kernel blocks
		\begin{equation}
			K_{rb}^G:=J_rJ_b^\top,\qquad
			K_{r0}^G:=J_rJ_0^\top,\qquad
			K_{b0}^G:=J_bJ_0^\top,
		\end{equation}
		with
		\[
		K_{br}^G=(K_{rb}^G)^\top,\qquad
		K_{0r}^G=(K_{r0}^G)^\top,\qquad
		K_{0b}^G=(K_{b0}^G)^\top.
		\]
		Then the channel-wise dynamics become
		\begin{align}
			\dot{\mathbf r}
			&=
			K_{rr}^G \boldsymbol{\gamma}^{(r)}
			+
			K_{rb}^G \boldsymbol{\gamma}^{(b)}
			+
			K_{r0}^G \boldsymbol{\gamma}^{(0)},
			\label{eq:r_dot_soft_prop}
			\\
			\dot{\mathbf b}
			&=
			K_{br}^G \boldsymbol{\gamma}^{(r)}
			+
			K_{bb}^G \boldsymbol{\gamma}^{(b)}
			+
			K_{b0}^G \boldsymbol{\gamma}^{(0)},
			\label{eq:b_dot_soft_prop}
			\\
			\dot{\mathbf c}
			&=
			K_{0r}^G \boldsymbol{\gamma}^{(r)}
			+
			K_{0b}^G \boldsymbol{\gamma}^{(b)}
			+
			K_{00}^G \boldsymbol{\gamma}^{(0)}.
			\label{eq:c_dot_soft_prop}
		\end{align}
		
		Further define the stacked violation vector
		\begin{equation}
			\mathbf z
			:=
			\begin{bmatrix}
				\mathbf r\\
				\mathbf b\\
				\mathbf c
			\end{bmatrix}
			\in\mathbb R^{N_r+N_b+N_0},
		\end{equation}
		the stacked adversarial-weight vector
		\begin{equation}
			\boldsymbol{\gamma}
			:=
			\begin{bmatrix}
				\boldsymbol{\gamma}^{(r)}\\
				\boldsymbol{\gamma}^{(b)}\\
				\boldsymbol{\gamma}^{(0)}
			\end{bmatrix},
		\end{equation}
		and the block kernel matrix
		\begin{equation}
			K_{\mathrm{soft}}^G
			:=
			\begin{bmatrix}
				K_{rr}^G & K_{rb}^G & K_{r0}^G\\
				K_{br}^G & K_{bb}^G & K_{b0}^G\\
				K_{0r}^G & K_{0b}^G & K_{00}^G
			\end{bmatrix}.
			\label{eq:block_kernel_soft_prop}
		\end{equation}
		Then the coupled soft-constrained dynamics admit the compact form
		\begin{equation}
			\dot{\mathbf z}
			=
			K_{\mathrm{soft}}^G\,\boldsymbol{\gamma}.
			\label{eq:soft_block_dynamics_compact}
		\end{equation}
	\end{proposition}
	
	\paragraph{Proof.}
	By differentiating the three violation vectors \(\mathbf r\), \(\mathbf b\), and \(\mathbf c\) along the generator flow \eqref{eq:theta_dot_soft_compact}, we obtain
	\begin{align}
		\dot{\mathbf r}
		&=
		J_r \dot{\theta}
		=
		J_rJ_r^\top \boldsymbol{\gamma}^{(r)}
		+
		J_rJ_b^\top \boldsymbol{\gamma}^{(b)}
		+
		J_rJ_0^\top \boldsymbol{\gamma}^{(0)},
		\\
		\dot{\mathbf b}
		&=
		J_b \dot{\theta}
		=
		J_bJ_r^\top \boldsymbol{\gamma}^{(r)}
		+
		J_bJ_b^\top \boldsymbol{\gamma}^{(b)}
		+
		J_bJ_0^\top \boldsymbol{\gamma}^{(0)},
		\\
		\dot{\mathbf c}
		&=
		J_0 \dot{\theta}
		=
		J_0J_r^\top \boldsymbol{\gamma}^{(r)}
		+
		J_0J_b^\top \boldsymbol{\gamma}^{(b)}
		+
		J_0J_0^\top \boldsymbol{\gamma}^{(0)}.
	\end{align}
	This proves \eqref{eq:r_dot_soft_pre_prop}--\eqref{eq:c_dot_soft_pre_prop}.
	
	Introducing the self-kernel and cross-kernel blocks by
	\[
	K_{rr}^G:=J_rJ_r^\top,\quad
	K_{bb}^G:=J_bJ_b^\top,\quad
	K_{00}^G:=J_0J_0^\top,
	\]
	and
	\[
	K_{rb}^G:=J_rJ_b^\top,\quad
	K_{r0}^G:=J_rJ_0^\top,\quad
	K_{b0}^G:=J_bJ_0^\top,
	\]
	with the transpose relations
	\[
	K_{br}^G=(K_{rb}^G)^\top,\qquad
	K_{0r}^G=(K_{r0}^G)^\top,\qquad
	K_{0b}^G=(K_{b0}^G)^\top,
	\]
	we immediately obtain \eqref{eq:r_dot_soft_prop}--\eqref{eq:c_dot_soft_prop}.
	
	Finally, stacking the three channels into the single vector
	\[
	\mathbf z=
	\begin{bmatrix}
		\mathbf r\\
		\mathbf b\\
		\mathbf c
	\end{bmatrix},
	\qquad
	\boldsymbol{\gamma}=
	\begin{bmatrix}
		\boldsymbol{\gamma}^{(r)}\\
		\boldsymbol{\gamma}^{(b)}\\
		\boldsymbol{\gamma}^{(0)}
	\end{bmatrix},
	\]
	and defining the block kernel matrix \(K_{\mathrm{soft}}^G\) as in \eqref{eq:block_kernel_soft_prop}, the three coupled equations are exactly equivalent to the compact block form
	\[
	\dot{\mathbf z}=K_{\mathrm{soft}}^G\,\boldsymbol{\gamma},
	\]
	which proves \eqref{eq:soft_block_dynamics_compact}.
	\hfill $\square$
	
	\paragraph{Interpretation.}
	Proposition~\ref{prop:soft_block_dynamics} is the soft-constrained analogue of the hard-constrained adversarial PINNs dynamics in the main text. The key difference is that the state is no longer a single residual vector, but a stacked multi-channel violation vector. Accordingly, the generator-side kernel is no longer a single NTK matrix, but a block operator containing both self-channel and cross-channel interactions.
	
	The cross-kernel blocks \(K_{rb}^G\), \(K_{r0}^G\), and \(K_{b0}^G\) are especially important. They show that even if a discriminator acts only on one channel, the resulting generator update will generally propagate to the other channels through the shared generator parameters. Thus, soft-constrained adversarial PINNs training is intrinsically coupled: interior, boundary, and initial-condition corrections cannot be analyzed independently, but must be treated as a single block dynamical system.

	\subsection{Diagonal-weight representation}
	\label{app:soft_diag_weight}
	
	To make the connection with the main text more explicit, define diagonal matrices
	\begin{equation}
		\Gamma_r
		=
		\mathrm{diag}(\gamma_1^{(r)},\dots,\gamma_{N_r}^{(r)}),\qquad
		\Gamma_b
		=
		\mathrm{diag}(\gamma_1^{(b)},\dots,\gamma_{N_b}^{(b)}),\qquad
		\Gamma_0
		=
		\mathrm{diag}(\gamma_1^{(0)},\dots,\gamma_{N_0}^{(0)}).
	\end{equation}
	Let \(\mathbf 1_r,\mathbf 1_b,\mathbf 1_0\) denote the all-one vectors of dimensions \(N_r,N_b,N_0\), respectively. Then
	\[
	\boldsymbol{\gamma}^{(r)}=\Gamma_r\mathbf 1_r,\qquad
	\boldsymbol{\gamma}^{(b)}=\Gamma_b\mathbf 1_b,\qquad
	\boldsymbol{\gamma}^{(0)}=\Gamma_0\mathbf 1_0.
	\]
	Define
	\begin{equation}
		\Gamma_{\mathrm{soft}}
		:=
		\mathrm{diag}(\Gamma_r,\Gamma_b,\Gamma_0),
		\qquad
		\mathbf 1_{\mathrm{soft}}
		:=
		\begin{bmatrix}
			\mathbf 1_r\\
			\mathbf 1_b\\
			\mathbf 1_0
		\end{bmatrix}.
	\end{equation}
	Then \eqref{eq:soft_block_dynamics_compact} can be rewritten as
	\begin{equation}
		\dot{\mathbf z}
		=
		K_{\mathrm{soft}}^G\,\Gamma_{\mathrm{soft}}\,\mathbf 1_{\mathrm{soft}}.
		\label{eq:soft_block_dynamics_gamma}
	\end{equation}
	
	This form makes clear that soft-constrained adversarial training is still a discriminator-induced sample-dependent preconditioned dynamics, but now at the level of a coupled three-channel system rather than a single residual channel.
	
	\subsection{Energy dynamics}
	\label{app:soft_energy}
	
	A natural total violation energy is
	\begin{equation}
		E_{\mathrm{soft}}(t)
		:=
		\frac12\|\mathbf r(t)\|^2
		+
		\frac12\|\mathbf b(t)\|^2
		+
		\frac12\|\mathbf c(t)\|^2
		=
		\frac12\|\mathbf z(t)\|^2.
		\label{eq:soft_energy_def}
	\end{equation}
	Differentiating \eqref{eq:soft_energy_def} and using \eqref{eq:soft_block_dynamics_compact} gives
	\begin{equation}
		\dot E_{\mathrm{soft}}
		=
		\mathbf z^\top \dot{\mathbf z}
		=
		\mathbf z^\top K_{\mathrm{soft}}^G\,\boldsymbol{\gamma}
		=
		\mathbf z^\top K_{\mathrm{soft}}^G\,\Gamma_{\mathrm{soft}}\,\mathbf 1_{\mathrm{soft}}.
		\label{eq:soft_energy_dot}
	\end{equation}
	
	Equivalently, in expanded form,
	\begin{align}
		\dot E_{\mathrm{soft}}
		&=
		\mathbf r^\top
		\Big(
		K_{rr}^G \boldsymbol{\gamma}^{(r)}
		+
		K_{rb}^G \boldsymbol{\gamma}^{(b)}
		+
		K_{r0}^G \boldsymbol{\gamma}^{(0)}
		\Big)
		\nonumber\\
		&\quad
		+
		\mathbf b^\top
		\Big(
		K_{br}^G \boldsymbol{\gamma}^{(r)}
		+
		K_{bb}^G \boldsymbol{\gamma}^{(b)}
		+
		K_{b0}^G \boldsymbol{\gamma}^{(0)}
		\Big)
		\nonumber\\
		&\quad
		+
		\mathbf c^\top
		\Big(
		K_{0r}^G \boldsymbol{\gamma}^{(r)}
		+
		K_{0b}^G \boldsymbol{\gamma}^{(b)}
		+
		K_{00}^G \boldsymbol{\gamma}^{(0)}
		\Big).
		\label{eq:soft_energy_dot_expanded}
	\end{align}
	
	Therefore, unlike in the hard-constrained case, the decrease of total violation energy is jointly determined by
	\begin{enumerate}
		\item the self-kernel blocks \(K_{rr}^G,K_{bb}^G,K_{00}^G\),
		\item the cross-kernel blocks \(K_{rb}^G,K_{r0}^G,K_{b0}^G\),
		\item the three discriminator-induced sample-weight vectors
		\(\boldsymbol{\gamma}^{(r)},\boldsymbol{\gamma}^{(b)},\boldsymbol{\gamma}^{(0)}\).
	\end{enumerate}
	This means that improving one channel adversarially may either help or hurt the others, depending on the sign and structure of the corresponding cross-kernel couplings.
	
	\subsection{Constant-kernel regime and NTK interpretation}
	\label{app:soft_ntk}
	
	In the infinite-width or lazy-training regime, it is natural to approximate the block kernel matrix by a constant operator,
	\begin{equation}
		K_{\mathrm{soft}}^G(\theta)
		\approx
		K_{\mathrm{soft}}^G.
	\end{equation}
	The soft-constrained adversarial dynamics then reduce to
	\begin{equation}
		\dot{\mathbf z}
		=
		K_{\mathrm{soft}}^G\,\Gamma_{\mathrm{soft}}\,\mathbf 1_{\mathrm{soft}},
		\label{eq:soft_const_kernel_dyn}
	\end{equation}
	where
	\[
	\mathbf z=
	\begin{bmatrix}
		\mathbf r\\
		\mathbf b\\
		\mathbf c
	\end{bmatrix},
	\qquad
	\Gamma_{\mathrm{soft}}
	=
	\mathrm{diag}(\Gamma_r,\Gamma_b,\Gamma_0).
	\]
	Equation \eqref{eq:soft_const_kernel_dyn} shows that, unlike the hard-constrained case, the generator is no longer governed by a single residual-regression dynamics, but by a coupled block-NTK system in which the three discriminators jointly modulate the evolution of the interior residual, boundary mismatch, and initial-condition mismatch.
	
	The essential new feature is the presence of cross-channel coupling. Although \(f_1,f_2,f_3\) act on the three channels separately and induce channel-wise diagonal weightings \(\Gamma_r,\Gamma_b,\Gamma_0\), these weightings are all propagated through the same shared generator kernel \(K_{\mathrm{soft}}^G\). As a consequence, an update encouraged by one discriminator generally affects not only its own channel, but also the other two through the off-diagonal kernel blocks. Therefore, the success of soft-constrained adversarial training depends not only on whether each discriminator provides a useful weighting within its own channel, but also on whether the three channels remain balanced under the shared generator dynamics.
	
	A convenient coarse diagnostic for this balance can be obtained by examining the diagonal block interactions between the generator kernel and the discriminator-induced weighting matrices. More specifically, define
	\begin{equation}
		T_r:=\operatorname{Tr}\!\big(K_{rr}^G\Gamma_r\big),\qquad
		T_b:=\operatorname{Tr}\!\big(K_{bb}^G\Gamma_b\big),\qquad
		T_0:=\operatorname{Tr}\!\big(K_{00}^G\Gamma_0\big).
		\label{eq:soft_trace_diag}
	\end{equation}
	Each of these quantities summarizes, at an aggregated level, how strongly the generator-side sensitivity and the discriminator-induced sample weighting interact within the corresponding channel. In this sense, \(T_r,T_b,T_0\) can be viewed as coarse channel-wise driving strengths for the residual, boundary, and initial-condition modules, respectively.
	
	To compare the three channels on a common scale, we further introduce the normalized balance indicators
	\begin{equation}
		\rho_r
		:=
		\frac{T_r}{T_r+T_b+T_0},
		\qquad
		\rho_b
		:=
		\frac{T_b}{T_r+T_b+T_0},
		\qquad
		\rho_0
		:=
		\frac{T_0}{T_r+T_b+T_0},
		\label{eq:soft_balance_ratio}
	\end{equation}
	whenever the denominator is nonzero. By construction,
	\begin{equation}
		\rho_r+\rho_b+\rho_0=1.
	\end{equation}
	These ratios provide a simple first-order measure of how the total effective adversarial drive is distributed across the three channels. If
	\[
	\rho_r\approx \rho_b\approx \rho_0\approx \frac13,
	\]
	then the residual, boundary, and initial-condition modules are being driven at roughly comparable levels. If one ratio is significantly larger than the others, then the corresponding channel is dominating the generator update; if one ratio is much smaller, then that channel is comparatively under-driven.
	
	To quantify this deviation from balanced evolution, one may further define a scalar imbalance score
	\begin{equation}
		\mathcal B_{\mathrm{imb}}
		:=
		\left(\rho_r-\frac13\right)^2
		+
		\left(\rho_b-\frac13\right)^2
		+
		\left(\rho_0-\frac13\right)^2.
		\label{eq:soft_imbalance_score}
	\end{equation}
	A smaller value of \(\mathcal B_{\mathrm{imb}}\) indicates that the three channels are being updated in a more balanced manner, whereas a larger value indicates that the training dynamics are biased toward only one or two components. In particular, a persistently large \(\mathcal B_{\mathrm{imb}}\) suggests that the soft-constrained adversarial training may be over-emphasizing one class of violations while neglecting the others.
	
	Of course, these trace-based quantities do not capture the full off-diagonal coupling structure of \(K_{\mathrm{soft}}^G\). Nevertheless, they provide a useful and easily interpretable first-order diagnostic: they isolate the interaction between the generator-side diagonal kernel blocks and the discriminator-side weighting matrices, and thereby allow one to infer whether the residual, boundary, and initial-condition modules are evolving in a balanced way before analyzing the more detailed cross-channel effects.
	
	\subsection{Discriminator-side functional dynamics}
	\label{app:soft_discriminator_ntk}
	
	For completeness, we also write the discriminator-side gradient-flow dynamics. Since the three discriminators are independent given the current generator state, each evolves on its own violation channel.
	
	For \(\ell\in\{1,2,3\}\), let \(f_\ell(\cdot;\phi_\ell)\) denote the corresponding discriminator and let \(k_\ell^D\) be its NTK kernel. Let the fake samples be
	\[
	\xi_i^{(1)}:=r_i,\qquad
	\xi_j^{(2)}:=b_j,\qquad
	\xi_k^{(3)}:=c_k.
	\]
	Then, under discriminator gradient ascent,
	\begin{equation}
		\dot{\phi}_\ell = \nabla_{\phi_\ell}\mathcal L_{D_\ell},
	\end{equation}
	the induced functional dynamics satisfy, for any input \(z\),
	\begin{equation}
		\frac{d}{dt}f_\ell(z;\phi_\ell(t))
		=
		\frac{1}{N_\ell}\sum_{m=1}^{N_\ell}
		k_\ell^D\!\big(z,0\big)\,
		P'\!\big(f_\ell(0;\phi_\ell(t))\big)
		+
		\frac{1}{N_\ell}\sum_{m=1}^{N_\ell}
		k_\ell^D\!\big(z,\xi_m^{(\ell)}\big)\,
		Q'\!\big(f_\ell(\xi_m^{(\ell)};\phi_\ell(t))\big),
		\label{eq:soft_disc_functional_dynamics}
	\end{equation}
	where \(N_1=N_r\), \(N_2=N_b\), and \(N_3=N_0\). Thus each discriminator learns a channel-specific weighting function shaped by its own NTK and its own empirical violation distribution.
	
	\section{Existing Adaptive and Weak-Adversarial PINNs Methods as Constrained-Discriminator Special Cases}
	\label{app:constrained_discriminator_unification}
	
	In this appendix, we show that several existing PINNs variants, including SA-PINN, LA-PINN, and weak-adversarial methods, can be embedded into a unified adversarial PINNs framework by restricting the discriminator to a prescribed functional subspace. Under this viewpoint, these methods do not require a fundamentally different generator-side training law; rather, they differ in the admissible class of discriminator functions and therefore in the induced sample-weighting or mode-weighting structure acting on the generator dynamics. However, in contrast to the label convention adopted in the main text---where the real target $0$ is assigned label $1$ and the generated data is assigned label $0$---the following methods use a different normalization. Specifically, the discriminator is constrained to satisfy $f(0)=y_{\mathrm{real}}:=0$ rather than 1.
	
	\subsection{General constrained-discriminator formulation}
	\label{app:general_constrained_disc_formulation}
	
	Let \(x\in\mathcal X\) denote the generated discrepancy variable. Depending on the method, \(x\) may represent a pointwise residual, a boundary or initial mismatch, or a weak-form discrepancy. The real target is always the zero-violation state, so the real distribution is concentrated at \(x=0\). 
	
	Consider the generator-side adversarial objective
	\begin{equation}
		\mathcal L_G(\theta,\phi)
		=
		\frac{1}{N}\sum_{i=1}^N R\!\big(f(x_i(\theta);\phi)\big),
		\label{eq:general_constrained_generator_loss}
	\end{equation}
	where \(f(\cdot;\phi)\) is the discriminator (or critic), \(R\) is the generator-side adversarial potential, and \(\{x_i(\theta)\}_{i=1}^N\) are the generated discrepancies induced by the solution network \(u_\theta\).
	
	We now restrict \(f\) to a structured family
	\[
	\mathcal F_{\mathrm{constr}}
	\subset \mathcal F,
	\]
	and assume that every admissible discriminator satisfies the anchoring condition
	\begin{equation}
		f(0;\phi)=y_{\mathrm{real}}=0,
		\label{eq:disc_anchor_condition}
	\end{equation}
	where \(y_{\mathrm{real}}\) is the prescribed real label or real-score baseline. In other words, the discriminator is forced to match the real target exactly at the zero-violation state, and only its response on generated discrepancies \(x\neq 0\) is trainable.
	
	Let
	\[
	\mathbf x(\theta)
	=
	[x_1(\theta),\dots,x_N(\theta)]^\top
	\]
	and define
	\[
	J_x(\theta)
	:=
	\begin{bmatrix}
		(\nabla_\theta x_1)^\top\\
		\vdots\\
		(\nabla_\theta x_N)^\top
	\end{bmatrix},
	\qquad
	K_{xx}^G(\theta):=J_xJ_x^\top.
	\]
	If the generator is trained by gradient flow,
	\[
	\dot\theta=-\nabla_\theta\mathcal L_G,
	\]
	then the induced discrepancy dynamics are given by
	\begin{equation}
		\dot{\mathbf x}
		=
		K_{xx}^G(\theta)\,\boldsymbol{\gamma},
		\qquad
		\gamma_i
		=
		-\frac{1}{N}
		R'\!\big(f(x_i;\phi)\big)\,
		\partial_x f(x_i;\phi).
		\label{eq:constrained_disc_general_dynamics}
	\end{equation}
	Therefore, once the discriminator class is restricted, the generator-side dynamics are completely determined by the admissible form of \(f(x;\phi)\), or equivalently by the induced weighting law \(\gamma_i\).

	\subsection{SA-PINN as a monotone masked discriminator}
	\label{app:sapinn_prop}
	
	In SA-PINN, the discriminator acts on pointwise violation quantities. More precisely, the input variable \(x\) denotes a generic pointwise discrepancy, which may represent any one of the three soft-constrained PINNs channels \eqref{residual}, \eqref{bound} and \eqref{initial}:
	\[
	x \in \{\, r_i,\; b_j,\; c_k \,\},
	\]
	where \(r_i\) is the PDE residual at an interior collocation point, \(b_j\) is the boundary-condition mismatch at a boundary point, and \(c_k\) is the initial-condition mismatch at an initial point.
	
	From the viewpoint of the present paper, SA-PINN can be interpreted within the IPM-type \eqref{wgan} adversarial framework, namely the same general class as WGAN-type objectives, but with the discriminator restricted to a highly structured function family. Instead of optimizing over an unconstrained critic class, SA-PINN introduces trainable nonnegative pointwise weights \(m(\lambda_i)\), where the mask \(m\) is nonnegative, differentiable, and strictly increasing, and the mask parameters are updated by ascent while the solution network is updated by descent. This is equivalent to constraining the discriminator to the family
	\begin{equation}
		f_{\mathrm{SA}}(x;\lambda)
		=-\frac{1}{2}
		m(\lambda)x^2,
		\qquad
		m(\lambda)\ge 0,
		\qquad
		m'(\lambda)>0.
		\label{eq:sa_disc_family_app}
	\end{equation}
	Thus the discriminator is restricted to a one-dimensional monotone family acting on squared discrepancies, rather than being free over an arbitrary nonlinear hypothesis class.
	
	Moreover,
	\[
	\partial_x f_{\mathrm{SA}}(x;\lambda)
	=
	-m(\lambda)x,
	\]
	and, for the IPM/WGAN-type generator choice \(Q(t)=R(t)=-t\), the induced generator-side weighting reduces to
	\begin{equation}
		\gamma_i^{\mathrm{SA}}
		=
		-\frac{1}{N}
		m(\lambda_i)\,x_i.
		\label{eq:sa_gamma_app}
	\end{equation}
	Hence SA-PINN can be interpreted as a constrained-discriminator adversarial PINNs under an IPM-type objective, whose critic is limited to a monotone mask family on \(x^2\), consistent with its original trainable-mask construction~\cite{mcclenny2023self}.
	\subsection{LA-PINN as a linear attentional discriminator}
\label{app:lapinn_prop}

In LA-PINN, the discriminator also acts on pointwise violation quantities. As in the SA-PINN case, the input variable \(x\) denotes a generic pointwise discrepancy, which may represent any one of the three soft-constrained PINNs channels:
\[
x \in \{\, r_i,\; b_j,\; c_k \,\}.
\]

From the viewpoint of the present paper, LA-PINN can also be interpreted within an IPM-type adversarial framework, namely the same general class as WGAN-type objectives, but with the discriminator restricted to a structured linear attentional family. LA-PINN feeds pointwise squared errors into independent loss-attentional networks (LANs) and explicitly emphasizes that the attentional transformation is linear, without nonlinear activation in the LAN weighting stage. This leads naturally to the constrained discriminator family
\begin{equation}
	f_{\mathrm{LA}}(x;\xi)
	=
	-\frac{1}{2}\,\mathrm{LAN}_{\xi}(x^2),
	\qquad
	\mathrm{LAN}_{\xi}(0)=0,
	\label{eq:la_disc_family_app}
\end{equation}
where \(\mathrm{LAN}_{\xi}\) is a linear map parameterized by \(\xi\). The condition \(\mathrm{LAN}_{\xi}(0)=0\) ensures that
\[
f_{\mathrm{LA}}(0;\xi)=0,
\]
which is consistent with the zero-discrepancy target in the IPM setting.

If \(\mathrm{LAN}_{\xi}\) acts pointwise, for instance,
\[
\mathrm{LAN}_{\xi}(x_i^2)=a_i(\xi)x_i^2+b_i(\xi),
\]
then the constraint \(\mathrm{LAN}_{\xi}(0)=0\) implies \(b_i(\xi)=0\), and hence
\[
f_{\mathrm{LA}}(x_i;\xi)
=
-\frac{1}{2}a_i(\xi)x_i^2.
\]
In our notation, this form is exactly equivalent to Eq.~(17) in \cite{song2024loss}.
Therefore,
\[
\partial_x f_{\mathrm{LA}}(x_i;\xi)
=
-a_i(\xi)x_i,
\]
and, for the IPM/WGAN-type choice \(Q(t)=R(t)=-t\), the induced weighting becomes
\begin{equation}
	\gamma_i^{\mathrm{LA}}
	=
	-\frac{1}{N}
	a_i(\xi)\,x_i.
	\label{eq:la_gamma_app}
\end{equation}
Thus LA-PINN may be viewed as a constrained-discriminator adversarial PINNs under an IPM-type objective, with a more flexible linear attentional family than SA-PINN, but still far more structured than a generic MLP critic. This interpretation matches the original LAN design, where the pointwise squared errors are the inputs and the weighting transformation is purely linear~\cite{song2024loss}.

	\subsection{Weak-adversarial methods as projection-type discriminators}
	\label{app:wan_prop}
	
	Weak-adversarial methods such as WAN \cite{zang2020weak} do not act on pointwise strong-form residuals directly. Instead, they optimize a saddle-point problem induced by the weak formulation, where the adversarial network parameterizes a test function and seeks the most challenging weak direction for the current solution network. From the viewpoint of the present paper, this can be interpreted within the same IPM-type adversarial framework, namely the same general class as WGAN-type objectives with
	\[
	Q(t)=R(t)=-t,
	\]
	but with the discriminator restricted to a normalized projection-type family in weak space:
	\begin{equation}
		f_{\mathrm{W}}(x;\eta)
		=
		-\frac{1}{2}
		\log\!\left(
		\frac{|\langle x,\varphi_\eta\rangle|^2}{\|\varphi_\eta\|_2^2}
		\right),
		\label{eq:wan_disc_family_app}
	\end{equation}
	where \(\varphi_\eta\) is the adversarial test function. Here \(x\) denotes the discrepancy object in weak space, such as the weak residual functional induced by the current generator. By contrast, the boundary and initial conditions are not modified through the discriminator, but are still enforced through standard \(L^2\)-type errors.
	
	Unlike SA-PINN or LA-PINN, the discriminator is now restricted not to pointwise masks on \(x^2\), but to normalized rank-one projection-type forms in function space. The adversarial variable \(\eta\) searches for the most challenging direction \(\varphi_\eta\), while the generator attempts to suppress that weak projection.
	
	Formally, writing
	\[
	g(x,\eta):=\frac{|\langle x,\varphi_\eta\rangle|^2}{\|\varphi_\eta\|_2^2},
	\]
	we have
	\begin{equation}
		\partial_x f_{\mathrm{W}}(x;\eta)
		=
		-\frac{1}{2}\frac{1}{g(x,\eta)}\,\partial_x g(x,\eta).
	\end{equation}
	Since
	\[
	\partial_x g(x,\eta)
	=
	\frac{2\langle x,\varphi_\eta\rangle}{\|\varphi_\eta\|_2^2}\,\varphi_\eta,
	\]
	it follows that
	\begin{equation}
		\partial_x f_{\mathrm{W}}(x;\eta)
		=
		-\frac{\varphi_\eta}{\langle x,\varphi_\eta\rangle}.
		\label{eq:wan_disc_derivative_app}
	\end{equation}
	Thus, under the IPM/WGAN-type choice \(Q(t)=R(t)=-t\), the induced generator update is still directed by the most adversarial weak test direction, but now with an additional inverse-projection normalization coming from the logarithmic form. The induced weighting becomes
	\begin{equation}
		\gamma_i^{\mathrm{W}}
		=
		-\frac{1}{N}\,
		\frac{\varphi_\eta}{\langle x,\varphi_\eta\rangle}.
		\label{eq:la_weak_app}
	\end{equation}
	Therefore, weak-adversarial PINNs can be interpreted as constrained-discriminator adversarial PINNs under an IPM-type objective, whose critic is limited to normalized weak-space projection forms, consistent with the original WAN saddle-point formulation based on induced operator norms~\cite{zang2020weak}.
    
	\subsection{Unified special-case theorem}
	\label{app:unified_special_case_theorem}
	\paragraph{Theorem 2 (SA-PINN, LA-PINN, and weak-adversarial methods as constrained-discriminator special cases).}
	Each of the following methods can be embedded into the general generator-side dynamics
	\[
	\dot{\mathbf x}
	=
	K_{xx}^G(\theta)\,\boldsymbol{\gamma}
	\]
	by restricting the discriminator to a prescribed structured family:
	\begin{align}
		\textnormal{SA-PINN:}\qquad
		&f(x)=-\frac{1}{2}m(\lambda)x^2,
		\qquad
		m(\lambda)\ge 0,\ \ m'(\lambda)>0,
		\\
		\textnormal{LA-PINN:}\qquad
		&f(x)=-\frac{1}{2}\mathrm{LAN}_{\xi}(x^2),
		\qquad
		\mathrm{LAN}_{\xi}\ \textnormal{linear},
		\\
		\textnormal{Weak-adversarial PINN:}\qquad
		&f(x)=-\frac{1}{2}
		\log\!\left(
		\frac{|\langle x,\varphi_\eta\rangle|^2}{\|\varphi_\eta\|_2^2}
		\right), \qquad \varphi_\eta \ \textnormal{is the test function}.
	\end{align}
	In all three cases, the discriminator is anchored at the zero-violation state,
	\[
	f(0)=0,
	\]
	and is only allowed to vary on generated violations inside a prescribed structured subspace. Consequently, these methods differ primarily in the admissible structure of the discriminator-induced weighting \(\boldsymbol{\gamma}\), rather than in the generator-side kernel form itself.
	
	\paragraph{Proof.}
	For each of the three constructions above, the discriminator family is explicitly restricted to a structured class and satisfies the zero-state anchoring condition \(f(0)=0\). Moreover, in all cases the generator-side weighting can be written in the unified form
	\[
	\gamma_i
	=
	-\frac{1}{N}
	R'\!\big(f(x_i)\big)\,
	\partial_x f(x_i),
	\]
	or its weak-space analogue in the weak-adversarial case. Therefore, by Proposition~1, each method admits the same generator-side representation
	\[
	\dot{\mathbf x}
	=
	K_{xx}^G(\theta)\,\boldsymbol{\gamma}.
	\]
	Hence they are all realizations of a broader adversarial PINNs framework with constrained discriminator classes. \(\square\)
	
	Theorem~2 has an immediate consequence for the NTK-based analysis developed in the main text.
	
	\paragraph{Corollary 3.}
	The unified discriminator-induced weighting framework is not limited to standard GAN or IPM critics, but also applies to SA-PINN, LA-PINN, and weak-adversarial PINNs once their discriminators are interpreted as belonging to constrained subspaces. 
	
	\paragraph{Interpretation.}
	Under this viewpoint,
	\begin{itemize}
		\item SA-PINN corresponds to a monotone mask-induced weighting on pointwise squared violations;
		\item LA-PINN corresponds to a linear attentional weighting on pointwise squared violations;
		\item weak-adversarial methods correspond to a projection-induced weighting in weak space.
	\end{itemize}
	Therefore, these methods can be analyzed within the same generator-side NTK framework, even though their discriminator classes are much more restricted than those of generic adversarial neural networks.
	
	More importantly, this viewpoint clarifies that the essential distinction among these methods is not whether they are formally written as GANs, but rather the functional subspace in which the discriminator is allowed to live. Once this subspace is fixed, the induced weighting law \(\boldsymbol{\gamma}\) is fixed accordingly, and the generator dynamics follow the same kernel-driven mechanism as in the general adversarial PINNs analysis.
	
	\section{A broader design space for adversarial PINNs training / Future work}
	\label{app:design_space_adv_pinn}
	
	The analysis developed in this paper suggests that adversarial PINNs training should be viewed as a broad design space rather than as a small collection of isolated named algorithms. Two largely orthogonal design dimensions emerge naturally: the \emph{discriminator input} and the \emph{discriminator objective or functional class}. Existing methods only occupy a limited subset of this space.
	
	An additional important point is that several methods that appear distinct from standard adversarial PINNs can in fact be understood within the same general framework. More precisely, many such methods do not leave the GAN/IPM paradigm altogether; instead, they can be interpreted as standard GAN-type or IPM-type adversarial schemes in which the discriminator is restricted to a structured function space. From this viewpoint, the essential distinction among methods is often not whether they are ``adversarial'' in a completely different sense, but rather what discriminator input is used and in what functional class the discriminator is allowed to live.
	
	\subsection{Dimension I: residual input versus squared-residual input}
	\label{app:design_space_input_dimension}
	
	The first design choice concerns the quantity fed into the discriminator. Two natural options are the residual itself,
	\[
	x=r,
	\]
	and the squared residual,
	\[
	x=r^2.
	\]
	
	When the discriminator takes \(r\) as input, the sign information of the violation is preserved. The discriminator can therefore distinguish positive and negative residuals and induce a direction-sensitive correction signal. In the NTK residual-dynamics formulation of the main text, this leads to an additive forcing law
	\[
	\dot{\mathbf r}=K_{rr}^G\Gamma\mathbf 1.
	\]
	
	When the discriminator instead takes \(r^2\) as input, the sign information is removed and only the magnitude of the violation is retained. In this case, the discriminator naturally emphasizes large residuals, and the generator dynamics become multiplicatively coupled:
	\[
	\dot{\mathbf r}=K_{rr}^G\widetilde\Gamma\mathbf r.
	\]
	Thus the distinction between \(r\) and \(r^2\) is structural: the former yields a sign-sensitive additive mechanism, while the latter yields a sign-free magnitude-gated multiplicative mechanism.
	
	\subsection{Dimension II: discriminator objective and functional class}
	\label{app:design_space_disc_dimension}
	
	The second design choice concerns the discriminator itself. At a broad level, the methods considered in this work may be grouped into the following categories.
	
	\paragraph{Classification-type discriminators (GAN).}
	These discriminators are trained through a classification-style adversarial objective and produce a signal tied primarily to separation of the zero target from the generated residual samples.
	
	\paragraph{Regression-type discriminators (LSGAN).}
	These discriminators learn a smooth regression from residual samples to target labels. Compared with classification-type objectives, they generate a smoother and more local correction signal, which can be particularly suitable when the training goal is to drive every residual point toward zero.
	
	\paragraph{Witness-function discriminators (IPM/WGAN).}
	These discriminators learn a witness function measuring the discrepancy between the empirical generated distribution and the real target distribution. Under the NTK view, the generator is driven by the input derivative of this witness field, so these methods emphasize global distributional discrepancy rather than direct pointwise fitting.
	
	\paragraph{Regularized witness discriminators (WGAN-GP).}
	These methods retain the witness-function interpretation of IPM/WGAN while additionally regularizing the discriminator, for instance through gradient penalties, in order to improve smoothness and stability.
	
	\paragraph{Monotone constrained discriminators (SA-PINN-type).}
	Here the discriminator is restricted to a trainable monotone mask acting on pointwise discrepancy magnitudes. From the present viewpoint, SA-PINN is naturally interpreted as a standard adversarial framework with the discriminator restricted to a monotone mask family.
	
	\paragraph{Linear constrained discriminators (LA-PINN-type).}
	In this class, the discriminator is restricted to a linear attentional transform of pointwise squared errors. Thus LA-PINN may be understood as an adversarial scheme with a more expressive but still highly structured discriminator class.
	
	\paragraph{Weak-space discriminators (WAN-type).}
	Weak adversarial methods do not act directly on strong-form pointwise residuals. Instead, they parameterize an adversarial test function in weak space and maximize the normalized weak-form discrepancy. From the present viewpoint, this again fits naturally into the same broader adversarial framework, but with the discriminator restricted to a weak-space projection class.
	
	\subsection{Why SA-PINN, LA-PINN, and WAN each have distinctive advantages}
	\label{app:design_space_existing_methods_advantages}
	
	This taxonomy helps explain why SA-PINN, LA-PINN, and WAN can all be effective, but for different reasons.
	
	\paragraph{SA-PINN.}
	The strength of SA-PINN lies in its simplicity and robustness. By assigning a trainable nonnegative weight to each point and updating these weights by ascent, it automatically increases the emphasis on points whose losses remain large. This makes it particularly effective at highlighting stiff or hard-to-fit regions.
	
	\paragraph{LA-PINN.}
	LA-PINN inherits the pointwise weighting idea but replaces the simple monotone mask with a more expressive linear attentional map. This gives the discriminator more flexibility in redistributing emphasis across points while preserving the stability and interpretability of a structured weighting rule.
	
	\paragraph{WAN.}
	The advantage of WAN is of a different nature. Rather than refining pointwise weighting, WAN changes the space in which adversarial training is performed. The discriminator becomes a weak test function, and the adversarial interaction takes place over the weak formulation of the PDE. This is particularly attractive when the weak form is more natural or more stable than the strong form.
	
	\subsection{A two-dimensional taxonomy of adversarial PINNs training strategies}
	\label{app:taxonomy_adv_pinn}
	
	The discussion in Appendix~\ref{app:design_space_input_dimension}--\ref{app:design_space_existing_methods_advantages}
	is summarized here by the two-dimensional taxonomy in Tables~\ref{tab:input_objective_taxonomy} and~\ref{tab:objective_functionclass_taxonomy}. The first table emphasizes the role of the discriminator input, while the second highlights the role of the discriminator objective or structured function class. Here, a check mark indicates that a representative method already exists in the literature or has a clear correspondence, while $\triangle$ highlights combinations that we believe are particularly promising.
	
	\begin{table}[t]
		\centering
		\caption{Combinations of discriminator input and basic adversarial objective in adversarial PINNs. A check mark indicates that a representative method already exists in the literature or has a clear correspondence, while $\triangle$ highlights combinations that we believe are particularly promising.}
		\label{tab:input_objective_taxonomy}
		\resizebox{0.5\linewidth}{!}{
			\begin{tabular}{lccc}
				\toprule
				\textbf{Discriminator input}
				& \textbf{GAN}
				& \textbf{LSGAN}
				& \textbf{IPM-GAN} \\
				\midrule
				Residual \(r\)
				& \checkmark \cite{bullwinkel2022deqgan}
				& \checkmark [This paper]
				& \checkmark \cite{zang2020weak} \\
				Squared residual \(r^2\)
				& $\times$
				& $\triangle$
				& \checkmark \cite{ciftci2024physics,song2024loss,mcclenny2023self} \\
				\bottomrule
			\end{tabular}
		}
	\end{table}
	
	\begin{table*}[t]
		\centering
		\caption{Adversarial objectives and discriminator function classes. The first column corresponds to the unconstrained setting, while the other columns correspond to additional structural restrictions on the discriminator.}
		\label{tab:objective_functionclass_taxonomy}
		\resizebox{\textwidth}{!}{
			\begin{tabular}{lcccccc}
				\toprule
				\textbf{Adversarial framework}
				& \textbf{Unconstrained}
                &\textbf{Spectra Normalization}
				& \textbf{Gradient-Penalty(GP)}
				& \textbf{Monotone-Mask}
				& \textbf{Linear-Transform}
				& \textbf{Weak-Space} \\
				\midrule
				GAN
                & $\times$
				& \checkmark \cite{bullwinkel2022deqgan}
				& $\times$
				& $\times$
				& $\times$
				& $\times$ \\
				LSGAN
                & \checkmark[This paper]
				& $\times$
				& $\times$
				& $\times$
				& $\times$
				& $\triangle$ \\
				IPM-GAN
                & $\times$
				& $\times$ 
				& \checkmark \cite{ciftci2024physics}
				& \checkmark \cite{mcclenny2023self}
				& \checkmark \cite{song2024loss}
				&  \checkmark \cite{zang2020weak}\\
				\bottomrule
			\end{tabular}
		}
	\end{table*}
	
	Several observations follow immediately. First, the current literature only explores a limited subset of the full design space. Second, many methods that appear quite different can still be interpreted within a unified adversarial framework once the discriminator is viewed as living in a structured function space. Third, the missing entries are themselves informative: they indicate combinations that are meaningful from the present theory but have not yet been systematically explored.
	
	In particular, we believe that the two highlighted cases, namely \(r^2\)+LSGAN and LSGAN with weak-space constraints, are especially promising and may potentially yield stronger performance.
    
	\subsection{Summary}
	\label{app:design_space_summary}
	
	In summary, adversarial PINNs training is best understood as a broad design space. The choice between \(r\) and \(r^2\) determines whether the discriminator preserves sign information or focuses on magnitude, and whether the induced generator dynamics are additive or multiplicative. The choice of discriminator objective or function class determines whether the discriminator behaves like a classifier, a regression map, a witness function, or a constrained weighting operator. Existing methods such as SA-PINN, LA-PINN, and WAN should therefore be viewed not as isolated constructions, but as representative points in a much larger structured family of adversarial PINNs training strategies.

	\section{Experiments}\label{experiment}
	\subsection{Experimental setup details}
	\label{app:exp_setup_details}
	
	This appendix provides the detailed experimental setup corresponding to the results reported in Section~\ref{sec:experiments}.
	
	
		\subsubsection{Compared methods}
	\label{app:exp_compared_methods}
	We compare the following adversarial PINNs methods:
	\begin{itemize}
		\item \textbf{DEQGAN},
		\item \textbf{GAN},
		\item \textbf{LSGAN},
		\item \textbf{WGAN-GP},
		\item \textbf{GAN-RB},
		\item \textbf{LSGAN-RB},
		\item \textbf{WGAN-GP-RB}.
	\end{itemize}
	Here the suffix ``RB'' denotes the proposed rollback strategy. For the rollback-enhanced variants, rollback selection is performed within at most 20 generator updates and 20 discriminator updates in each outer iteration.
	
	Although the preceding theoretical discussion shows that a broader family of existing methods can also be interpreted as special cases of adversarial training under constrained discriminator classes, in the experiments we deliberately focus on the most basic adversarial objectives, namely GAN, LSGAN, and WGAN-GP. The reason is that these methods represent the canonical adversarial design choices and therefore provide the cleanest setting for evaluating the proposed framework. By starting from these standard baselines, we can more clearly separate the effect of the adversarial objective itself from the effect of the proposed update-control mechanism, without introducing additional architectural or formulation-specific factors.
	
	DEQGAN is included for a different reason. In fact, our study was initially motivated by the distinctive phenomena observed in DEQGAN under different generator--discriminator update ratios. These observations led us to identify a gap between the empirical behavior of adversarial PINNs training and the classical divergence-based explanation, and further motivated the unified residual-dynamics framework developed in this paper. In this sense, DEQGAN serves not only as a strong existing baseline, but also as the starting point from which we first discovered the phenomenon, constructed the explanatory framework, and then generalized the resulting analysis to a broader class of adversarial PINNs methods.
	
	The purpose of including both the original and rollback-enhanced variants is therefore twofold: first, to evaluate the adversarial objectives themselves under a unified setting; and second, to isolate the contribution of the proposed rollback mechanism from the choice of adversarial objective.
	
	\subsubsection{Benchmark equations}
	\label{app:exp_benchmark_equations}
	
	To evaluate the robustness of the proposed framework across different PDE types, we consider the following benchmark problems:
	\begin{itemize}
		\item \textbf{Poisson equation},
		\item \textbf{Laplace equation},
		\item \textbf{Viscous Burgers equation},
		\item \textbf{Reaction--Diffusion equation},
		\item \textbf{Klein--Gordon equation}.
	\end{itemize}
	These problems cover a diverse range of settings, including linear and nonlinear equations, stationary and time-dependent systems, as well as equations with different propagation, diffusion, and stiffness characteristics. 
	This diversity is intended to test whether the conclusions drawn from the proposed residual-dynamics analysis generalize beyond a single PDE family.
	
	\subsubsection{Network architectures}
	\label{app:exp_network_architectures}
	
	For fairness, the generator architecture is kept fixed across all methods and uses residual links. 
	The discriminator architecture depends on the adversarial objective.
	
	For \textbf{DEQGAN}, \textbf{GAN}, and \textbf{GAN-RB}, we use the standard discriminator configuration of GAN-style adversarial training:
	\begin{itemize}
		\item residual links,
		\item spectral normalization,
		\item a non-regressive output with a final sigmoid activation.
	\end{itemize}
	
	For \textbf{LSGAN}, \textbf{LSGAN-RB}, \textbf{WGAN-GP}, and \textbf{WGAN-GP-RB}, the discriminator is treated as a critic:
	\begin{itemize}
		\item no residual links,
		\item no spectral normalization,
		\item no final sigmoid activation.
	\end{itemize}
	All hidden layers use \(\tanh\) activation.
	
	\subsubsection{Optimization settings}
	\label{app:exp_optimization_settings}
	
	A central practical distinction in our study is that, except for DEQGAN, all methods are trained under the same simple default optimization setting:
	\begin{equation}
		\text{lr}_G=\text{lr}_D=10^{-3},
	\end{equation}
	with Adam optimizer and default hyperparameters
	\begin{equation}
		(\beta_1,\beta_2)=(0.9,0.999).
	\end{equation}
	
	By contrast, \textbf{DEQGAN} follows its original tuning-dependent protocol (Table \ref{tab:deqgan_hyperparameters}), which relies on:
	\begin{itemize}
		\item two-timescale generator/discriminator learning rates,
		\item tuned Adam hyperparameters,
		\item Ray Tune-based hyperparameter search.
	\end{itemize}
	This distinction is particularly important because one of the central goals of our work is to demonstrate that favorable alternating training dynamics can be effectively recovered through the proposed rollback principle, while avoiding the substantial hyperparameter tuning burden and implementation cost typically required by DEQGAN.
	\begin{table}[t]
		\centering
		\caption{Hyperparameter settings for DEQGAN on PDE benchmarks.}
		\label{tab:deqgan_hyperparameters}
		\small
		\begin{tabular}{lccccc}
			\toprule
			\textbf{Hyperparameter} & \textbf{Poisson} & \textbf{Laplace} & \textbf{Viscous--Burgers} & \textbf{Reaction--Diffusion} & \textbf{Klein--Gordon} \\
			\midrule
			Num. iterations & 3000 & 3000 & 3300 & 4500 & 4000 \\
			Num. grid points & $32\times32$ & $32\times32$ & $32\times32$ & $32\times32$ & $32\times32$ \\
			G units/layer & 50 & 50 & 50 & 50 & 50 \\
			G num. layers & 4 & 4 & 3 & 3 & 3 \\
			D units/layer & 30 & 50 & 20 & 20 & 20 \\
			D num. layers & 2 & 2 & 5 & 5 & 5 \\
			Activations & tanh & tanh & tanh & tanh & tanh \\
			G learning rate & 0.019 & 0.012 & 0.012 & 0.007 & 0.012 \\
			D learning rate & 0.021 & 0.088 & 0.005 & 0.009 & 0.005 \\
			G $\beta_1$ (Adam) & 0.139 & 0.295 & 0.185 & 0.185 & 0.185 \\
			G $\beta_2$ (Adam) & 0.369 & 0.358 & 0.594 & 0.594 & 0.594 \\
			D $\beta_1$ (Adam) & 0.745 & 0.575 & 0.093 & 0.093 & 0.093 \\
			D $\beta_2$ (Adam) & 0.759 & 0.133 & 0.184 & 0.184 & 0.184 \\
			Exponential LR decay $(\gamma)$ & 0.957 & 0.953 & 0.954 & 0.954 & 0.954 \\
			Decay step size & 9 & 10 & 25 & 10 & 10 \\
			\bottomrule
		\end{tabular}
	\end{table}
		\begin{table}[t]
		\centering
		\caption{Running time (\textbf{hours}) for different PDE benchmarks and compared methods. All experiments were conducted on a laptop workstation equipped with an NVIDIA GeForce RTX 4070 Laptop GPU and an Intel Core i7-14700HX CPU.}
		\label{tab:appendix_compute_resources}
		\begin{tabular}{lccccccc}
			\toprule
			\textbf{PDE} & \textbf{DEQGAN} & \textbf{GAN} & \textbf{LSGAN} & \textbf{WGAN-GP} & \textbf{GAN-RB} & \textbf{LSGAN-RB} & \textbf{WGAN-GP-RB} \\
			\midrule
			Poisson              & 0.24 & 0.33 & 0.28 & 0.29 & 3.17 & 3.02 & 2.98 \\
			Laplace              & 0.28 & 0.27 & 0.27 & 0.29 & 3.33 & 3.27 & 3.24 \\
			Burgers              & 0.46 & 0.41 & 0.42 & 0.37 & 4.62 & 4.48 & 4.47 \\
			Reaction  & 0.33 & 0.32 & 0.29& 3.86 & 4.11 & 4.09 & 4.07 \\
			Klein        & 0.45 & 0.43 & 0.44 & 0.36 & 4.98 & 4.82 & 4.69 \\
			\bottomrule
		\end{tabular}
	\end{table}
	\subsubsection{Training protocol}
	\label{app:exp_training_protocol}
	For each outer training iteration, the generator and discriminator are updated alternately.
	The rollback-enhanced variants additionally evaluate candidate intermediate states and retain only those updates that improve the corresponding first-order criterion or residual-energy criterion, as described in the main text.
	
	Unless otherwise specified, all methods are trained under the same sampling and optimization pipeline on each benchmark problem.
	This design ensures that performance differences are primarily attributable to the adversarial objective and the rollback mechanism, rather than to unrelated implementation choices.
	All experiments are implemented in PyTorch and conducted on a laptop workstation equipped with an NVIDIA GeForce RTX 4070 Laptop GPU and an Intel Core i7-14700HX CPU.
	The detailed hardware configuration and running-time (hours) statistics are summarized in Table~\ref{tab:appendix_compute_resources}.

    \subsubsection{Sampling Strategy and Training--Validation Relationship}
    \label{sec:sampling_strategy}
    
    Different PDEs adopt slightly different sampling strategies according to the spatial size of their computational domains. For each problem, however, the sampling strategy is predetermined before training and is independent of the final validation error. To quantitatively characterize the spatial relationship between the training and validation points, we first normalize the coordinates of all PDEs to $[0,1]^2.$
    For each validation point $\mathbf{x}^{\mathrm{val}}_i$, we compute its two-dimensional Euclidean distance to the nearest training point:
    \[
    d_i =
    \min_j
    \left\|
    \mathbf{x}^{\mathrm{val}}_i-
    \mathbf{x}^{\mathrm{tr}}_j
    \right\|_2.
    \]
    The mean nearest-neighbor distance is then defined as
    \[
    \bar d =
    \frac{1}{N_{\mathrm{val}}}
    \sum_{i=1}^{N_{\mathrm{val}}} d_i.
    \]
    
    To account for differences in validation-grid resolution, we further normalize this quantity by the diagonal spacing $h_{\mathrm{diag}}$ between neighboring validation-grid points:
    \[
    \rho =
    \frac{\bar d}{h_{\mathrm{diag}}}\times 100\%.
    \]
    Therefore, the second column of Table~\ref{tab:sampling_distance} reports the mean distance from each validation point to its nearest training point, while
    the third column reports this distance relative to one diagonal spacing of the validation grid.
    
    For sampling strategies involving random perturbations, we perform ten independent random realizations and report the averaged distance statistics.
    
    \begin{table}[t]
    \centering
    \caption{
    Spatial relationship between training and validation samples after
    normalizing all PDE coordinates to $[0,1]^2$.
    }
    \label{tab:sampling_distance}
    \begin{tabular}{lcc}
    \toprule
    \textbf{PDE}
    & \textbf{Mean nearest-neighbor distance}
    & \textbf{Relative grid spacing} \\
    \midrule
    Poisson  & 0.00971 & 21.3\%  \\
    Laplace & 0.01129 & 31.9\%  \\
    Burgers  & 0.00483 & 21.9\%  \\
    Reaction & 0.01259 & 32.2\% \\
    Klein    & 0.01206 & 32.1\%  \\
    \bottomrule
    \end{tabular}
    \end{table}
    Overall, the training and validation sets are spatially close but are generally not identical. The mean nearest-neighbor distance is approximately $21\%$--$32\%$ of one diagonal validation-grid spacing, indicating that the training samples provide a dense spatial coverage of the computational domain. At the same time, most validation points do not exactly coincide with the training points, so the reported validation error is evaluated predominantly at distinct spatial locations within the same domain.

	\subsubsection{Evaluation metrics}
	\label{app:exp_metrics}
	
	We report both \textbf{training error} (L2 error) and \textbf{validation error} (L2 error) as the primary accuracy metrics. 
	To connect the experiments with the theory developed in the main text, we additionally track:
	\begin{itemize}
		\item the residual energy
		\begin{equation}
			E=\frac12\|\mathbf r\|^2,
		\end{equation}
		\item the first-order indicator
		\begin{equation}
			S=\mathbf r^\top K_{rr}^{G}\boldsymbol{\gamma},
		\end{equation}
		\item and, for rollback-based methods, the selected discriminator and generator rollback depths.
	\end{itemize}
	
    
    These quantities serve as useful diagnostics and provide a direct empirical bridge between adversarial PINN training and the NTK-based residual-dynamics analysis developed in this paper.
    
	\subsubsection{Relation to the main text}
	\label{app:exp_relation_main_text}
	
	The main text reports the core quantitative comparisons and the principal mechanistic observations. 
	Detailed training curves, per-equation dynamics, rollback-step statistics, and additional ablation results are deferred to the appendix in order to keep the main paper focused on the central theoretical and algorithmic message. 
	
	We note that the mechanistic analysis of DEQGAN under different \(G:D\) ratios has already been presented systematically in Section~\ref{sec:exp_mechanistic} of the main text. There, based on the first-order indicator and the NTK-based failure-mode analysis, we explained the phenomena identified in Section~\ref{sec:unresasonable}, clarified the apparent mismatch between empirical observations and the classical theoretical picture, reconstructed the training-dynamics viewpoint, and supported the resulting interpretation with numerical evidence. Since these components already form a coherent narrative in the main paper, we do not repeat them again in the appendix.
	\subsection{Benchmark equations Comparison}
	\label{app:comparision}
	\subsubsection{Laplace}
	\paragraph{Laplace equation:} 
	
	We first consider a two-dimensional Laplace equation posed on the unit square domain
	\[
	\Omega=(0,1)\times(0,1),
	\]
	with Dirichlet boundary conditions. The governing equation is
	\[
	\left\{
	\begin{aligned}\label{laplace_1}
		\frac{\partial^2 u}{\partial x^2}+\frac{\partial^2 u}{\partial y^2}
		&=0,\qquad \qquad \qquad \qquad \qquad \qquad \qquad (x,y)\in\Omega,\\
		u(0,y)&=0,\qquad \qquad \qquad \qquad \qquad \qquad \qquad  y\in[0,1],\\
		u(1,y)&=0,\qquad \qquad \qquad \qquad \qquad \qquad \qquad  y\in[0,1],\\
		u(x,0)&=0,\qquad \qquad \qquad \qquad \qquad \qquad \qquad x\in[0,1],\\
		u(x,1)&=\frac{1}{2\cosh(\pi)}\sin(\pi x)\bigl(e^{\pi}-e^{-\pi}\bigr)),\qquad x\in[0,1].
	\end{aligned}
	\right.
	\]
	
	The Laplace equation is one of the most fundamental elliptic partial differential equations and commonly arises in steady-state heat conduction, electrostatic potential analysis, incompressible fluid flow, and other equilibrium problems. 
	
	In the present example, the problem is defined on a bounded rectangular domain with mixed homogeneous and non-homogeneous Dirichlet boundary conditions, making it a standard benchmark for evaluating the approximation quality of physics-informed methods.
	
	For this problem, an analytical solution is available:
	\[
	u(x,y)=\frac{1}{2\cosh(\pi)}\sin(\pi x)\bigl(e^{\pi y}-e^{-\pi y}\bigr),
	\]
	Hence, the numerical prediction can be directly compared with the exact solution in both solution error and PDE residual.
	\begin{figure*}[t]
		\centering
		\begin{subfigure}[t]{0.32\textwidth}
			\centering
			\includegraphics[width=\linewidth]{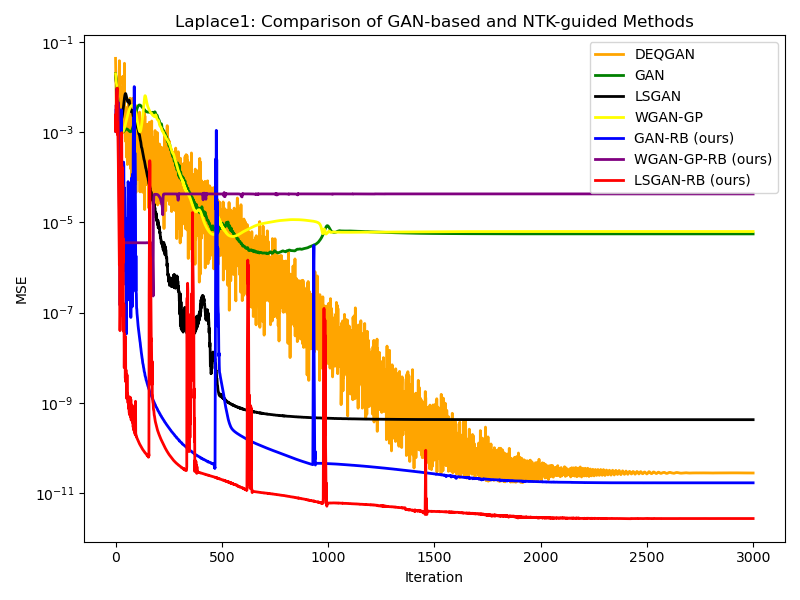}
			\caption{Training MSE}
			\label{fig:lap1_train_mse}
		\end{subfigure}
		\hfill
		\begin{subfigure}[t]{0.32\textwidth}
			\centering
			\includegraphics[width=\linewidth]{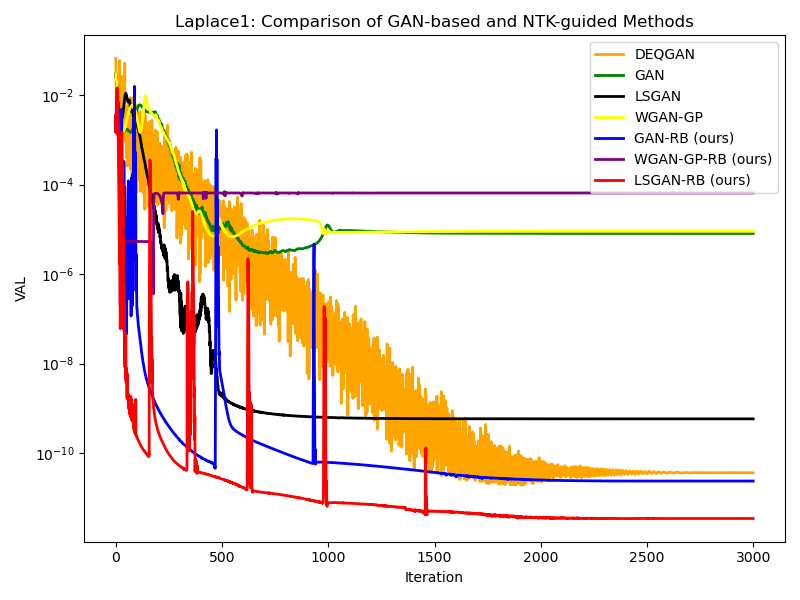}
			\caption{Validation MSE}
			\label{fig:lap1_val_mse}
		\end{subfigure}
		\hfill
		\begin{subfigure}[t]{0.32\textwidth}
			\centering
			\includegraphics[width=\linewidth]{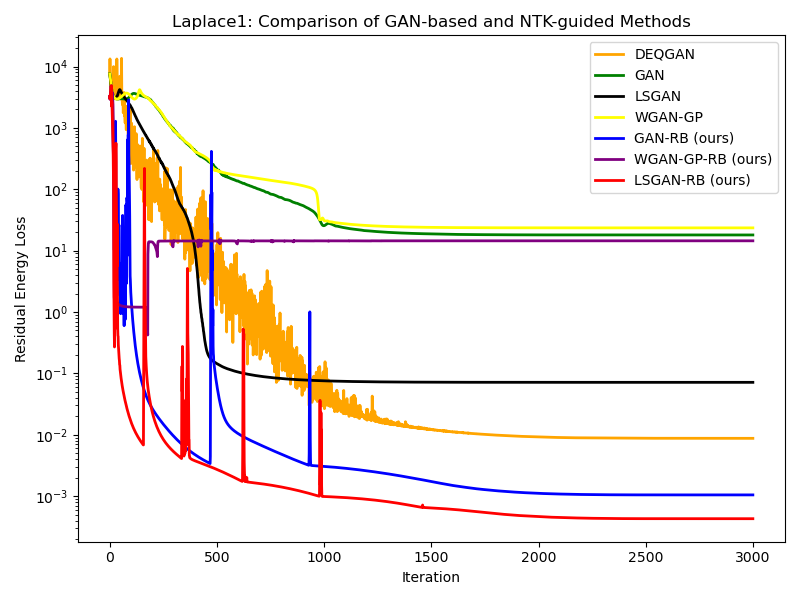}
			\caption{Residual energy}
			\label{fig:lap1_residual_energy}
		\end{subfigure}
		
		\vspace{0.6em}
		
		\begin{subfigure}[t]{0.32\textwidth}
			\centering
			\includegraphics[width=\linewidth]{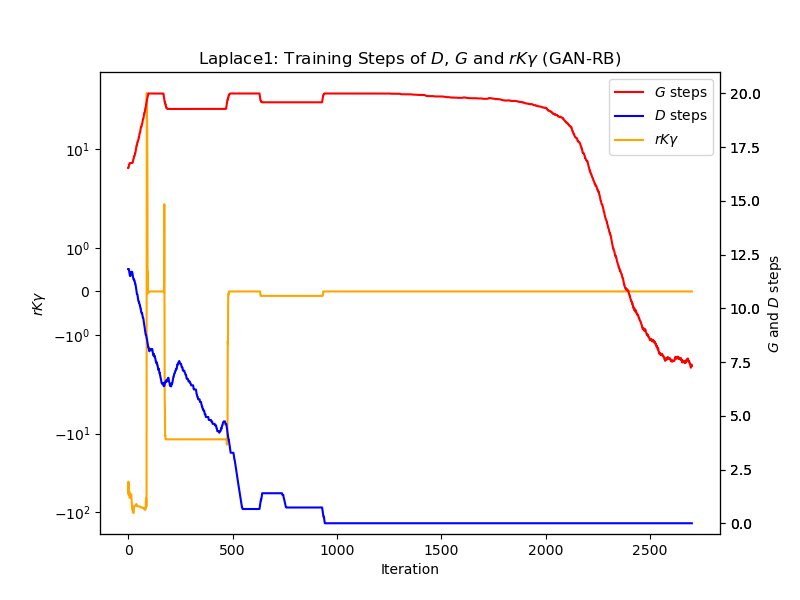}
			\caption{GAN-RB}
			\label{fig:lap1_gan_rb_dynamics}
		\end{subfigure}
		\hfill
		\begin{subfigure}[t]{0.32\textwidth}
			\centering
			\includegraphics[width=\linewidth]{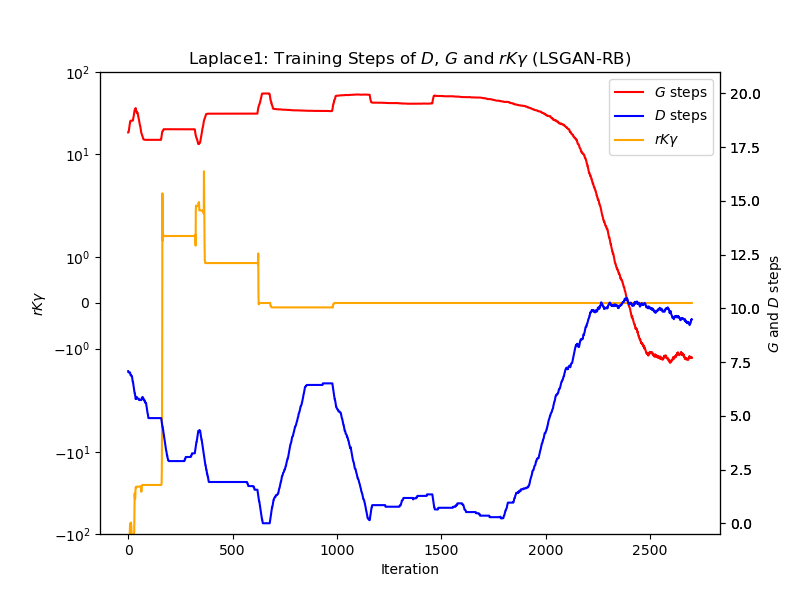}
			\caption{LSGAN-RB}
			\label{fig:lap1_lsgan_rb_dynamics}
		\end{subfigure}
		\hfill
		\begin{subfigure}[t]{0.32\textwidth}
			\centering
			\includegraphics[width=\linewidth]{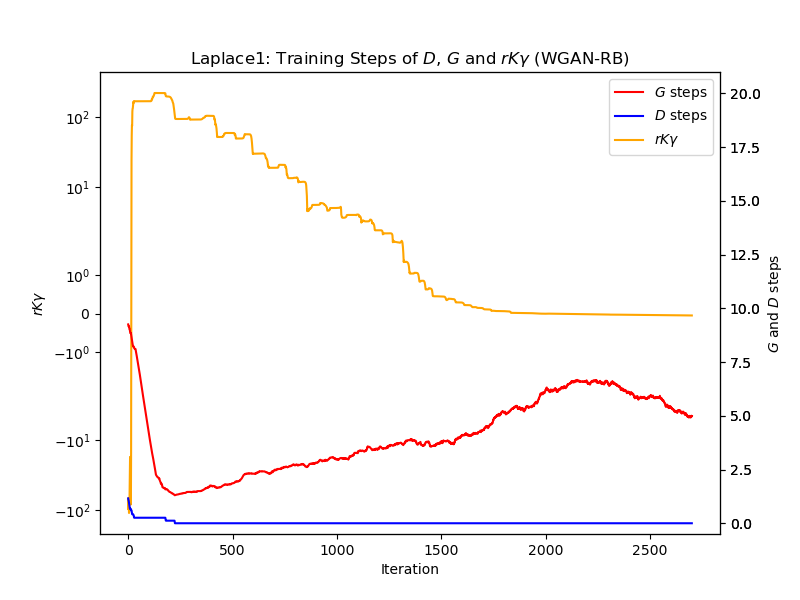}
			\caption{WGAN-GP-RB}
			\label{fig:lap1_wgan_rb_dynamics}
		\end{subfigure}
		
		\caption{Laplace equation: The first row reports the training MSE, validation MSE, and residual energy for all compared methods. The second row shows the rollback-related dynamics for GAN-RB, LSGAN-RB, and WGAN-GP-RB, respectively, including the selected generator/discriminator inner-step behavior and the discriminator-induced quantity \(r^\top K_{rr}\gamma\).}
		\label{fig:lap1_comparison}
	\end{figure*}
	\paragraph{Comparison among different methods: }
	Fig.~\ref{fig:lap1_comparison}(a)--(c) reports the training error, validation error, and residual energy dynamics of different adversarial PINNs variants on the Laplace equation. Overall, the rollback-based methods significantly improve the corresponding original adversarial methods. In particular, LSGAN-RB and GAN-RB achieve the best performance, with both the residual energy and prediction errors reduced by several orders of magnitude compared with their non-rollback counterparts. This demonstrates that the proposed rollback strategy can effectively stabilize and accelerate adversarial PINNs training by selecting more favorable generator--discriminator update states.
	
	An interesting observation is that WGAN-GP does not perform well in this experiment, although it is commonly believed to alleviate the gradient-vanishing issue in standard GAN training. Even after applying the rollback strategy, WGAN-GP-RB still fails to achieve comparable accuracy. This suggests that avoiding discriminator saturation alone is not sufficient for adversarial PINNs training. From the perspective of our framework, the key factor is whether the discriminator-induced weighting $\gamma$ provides an informative and well-aligned residual descent direction under the generator NTK. In this case, the WGAN-type gradient field appears to be less effective for the residual dynamics of the Laplace equation.
	
	We further analyze the three rollback variants in Fig.~\ref{fig:lap1_comparison}(d)--(f), which show the evolution of the first-order indicator, the selected generator update steps, and the selected discriminator update steps during training. For better visualization, all three quantities are smoothed by a moving average over 300 iterations. In the following experiments, we adopt the same analysis strategy. The main quantity of interest is the number of selected generator update steps. According to the rollback criterion, the generator is allowed to continue updating only when the current discriminator-induced direction leads to a decrease in the residual energy. Therefore, a larger selected generator update ratio indicates that the discriminator feedback remains useful over a longer generator trajectory and can continuously drive the residual energy downward.
	
	Fig.~\ref{fig:lap1_comparison}(d) and~(e) correspond to GAN-RB and LSGAN-RB, respectively. We observe that both methods maintain a high generator update ratio for most of the training process, indicating that their discriminator-induced gradient fields provide effective descent directions for the generator. This is consistent with their strong performance in Fig.~\ref{fig:lap1_comparison}(a)--(c). In contrast, Fig.~\ref{fig:lap1_comparison}(f) shows that WGAN-GP-RB keeps the generator update steps at a very low level throughout training. This means that the rollback criterion frequently rejects generator updates, implying that the WGAN-GP discriminator fails to provide a sufficiently useful residual descent direction in this problem.
	
	The behavior of the first-order indicator further supports this interpretation. Under our plotting convention, a negative value corresponds to a favorable energy-decreasing direction. For GAN-RB and LSGAN-RB, the first-order indicator is largely suppressed in the negative region during training, which explains why their generator updates can be repeatedly accepted. By contrast, for WGAN-GP-RB, the indicator rapidly moves toward the positive region at the early stage and then becomes ineffective, indicating poor alignment between the discriminator-induced weighting and the generator NTK descent direction. This provides direct empirical evidence for our theoretical claim that the success of adversarial PINNs training depends not only on the adversarial loss itself, but more fundamentally on the quality of the induced residual-space weighting and its alignment with the generator residual dynamics.

	\subsubsection{Poisson}
	\paragraph{Poisson equation}
	
	We next consider a two-dimensional Poisson equation posed on the unit square domain
	\[
	\Omega=(0,1)\times(0,1),
	\]
	with homogeneous Dirichlet boundary conditions. The governing equation is given by
	\[
	\left\{
	\begin{aligned}
		\frac{\partial^2 u}{\partial x^2}+\frac{\partial^2 u}{\partial y^2}
		&=
		2x (y-1)\bigl(y-2x+xy+2\bigr)e^{x-y},
		\qquad (x,y)\in\Omega,\\
		u(x,y)&=0,\qquad (x,y)\in\partial\Omega.
	\end{aligned}
	\right.
	\]
	
	The Poisson equation is one of the most classical elliptic partial differential equations and arises widely in electrostatics, steady-state heat conduction, gravitational potential modeling, and incompressible flow analysis. In the present example, the problem is defined on a bounded rectangular domain with a prescribed source term and zero boundary values, which makes it a standard benchmark for testing the approximation accuracy and training stability of PINNs-type methods.
	
\begin{figure*}[t]
		\centering
		
		\begin{subfigure}[t]{0.32\textwidth}
			\centering
			\includegraphics[width=\linewidth]{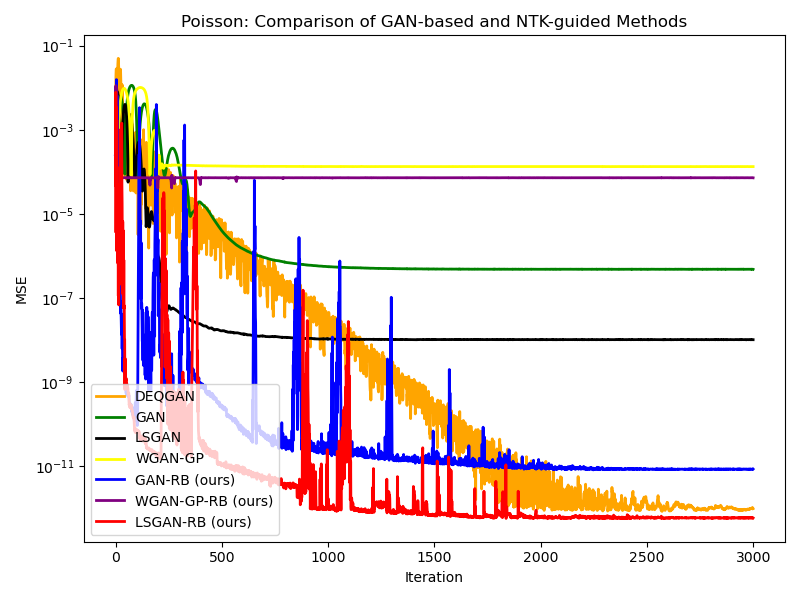}
			\caption{Training MSE}
			\label{fig:poisson_train_mse}
		\end{subfigure}
		\hfill
		\begin{subfigure}[t]{0.32\textwidth}
			\centering
			\includegraphics[width=\linewidth]{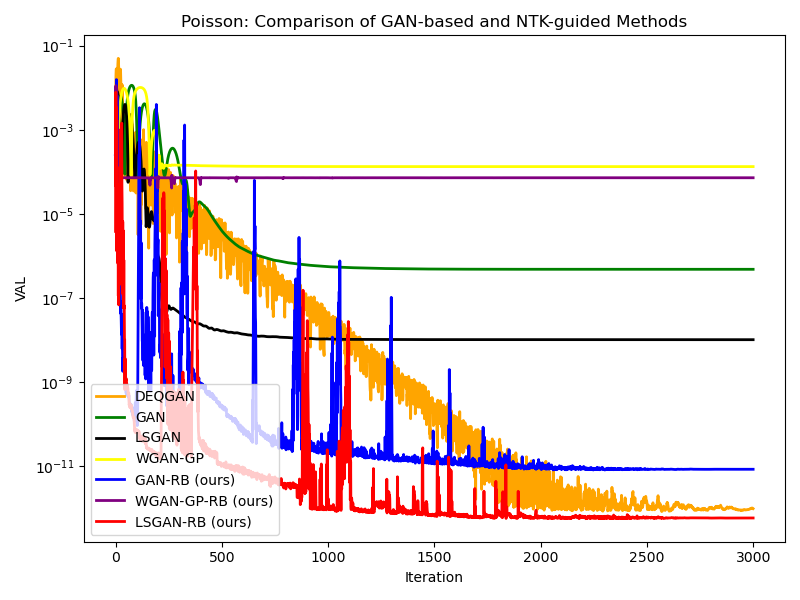}
			\caption{Validation MSE}
			\label{fig:poisson_val_mse}
		\end{subfigure}
		\hfill
		\begin{subfigure}[t]{0.32\textwidth}
			\centering
			\includegraphics[width=\linewidth]{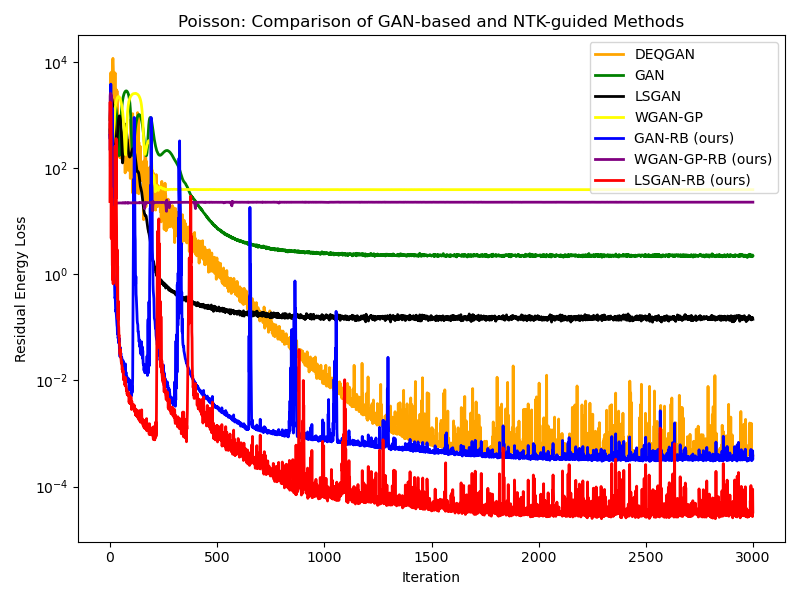}
			\caption{Residual energy}
			\label{fig:poisson_residual_energy}
		\end{subfigure}
		
		\vspace{0.6em}
		
		\begin{subfigure}[t]{0.32\textwidth}
			\centering
			\includegraphics[width=\linewidth]{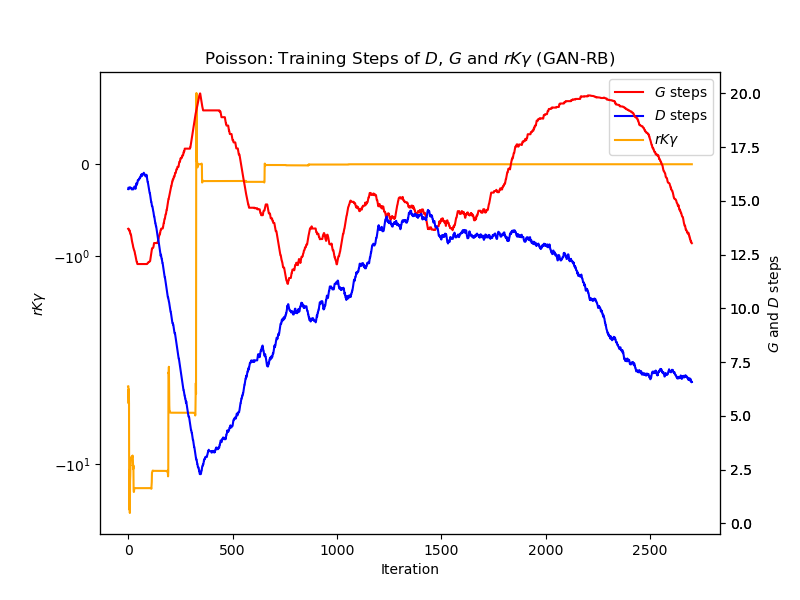}
			\caption{GAN-RB}
			\label{fig:poisson_gan_rb_dynamics}
		\end{subfigure}
		\hfill
		\begin{subfigure}[t]{0.32\textwidth}
			\centering
			\includegraphics[width=\linewidth]{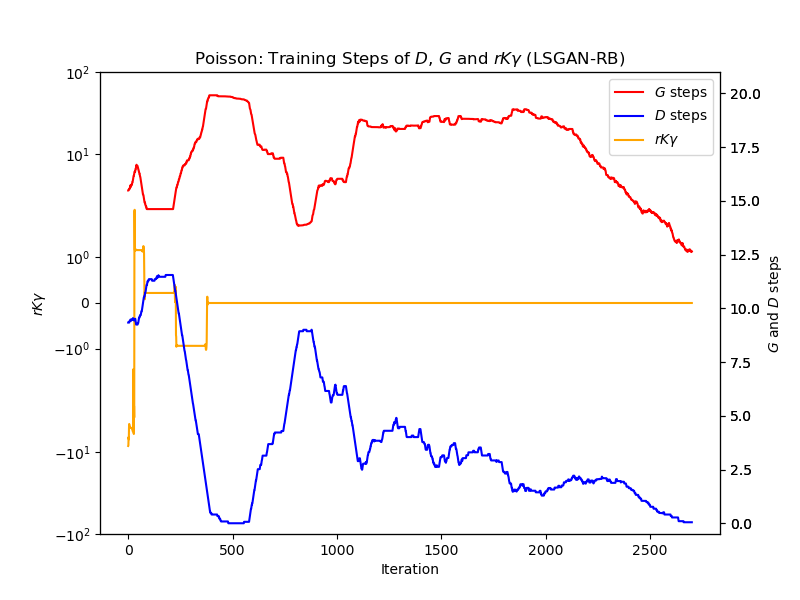}
			\caption{LSGAN-RB}
			\label{fig:poisson_lsgan_rb_dynamics}
		\end{subfigure}
		\hfill
		\begin{subfigure}[t]{0.32\textwidth}
			\centering
			\includegraphics[width=\linewidth]{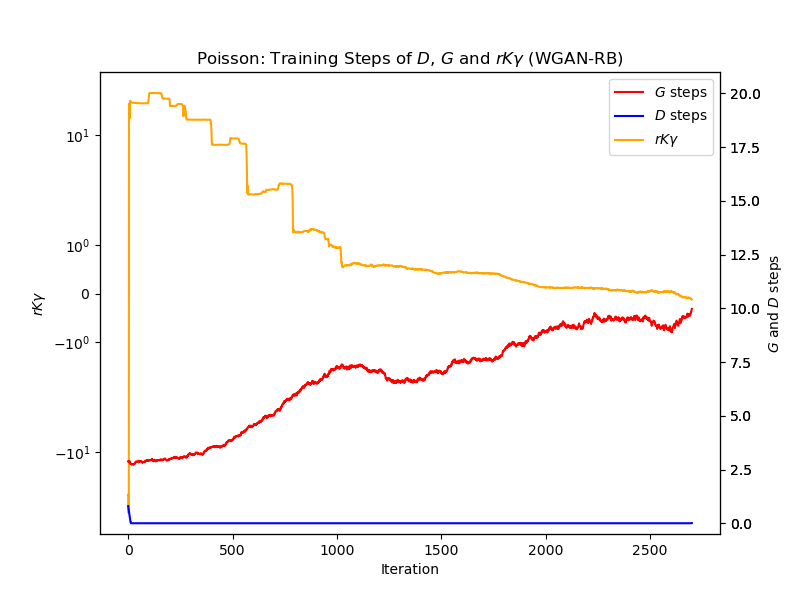}
			\caption{WGAN-GP-RB}
			\label{fig:poisson_wgan_rb_dynamics}
		\end{subfigure}
		
		\caption{Poisson equation: The first row reports the training MSE, validation MSE, and residual energy for all compared methods. The second row shows the rollback-related dynamics for GAN-RB, LSGAN-RB, and WGAN-GP-RB, respectively, including the selected generator/discriminator inner-step behavior and the discriminator-induced quantity \(r^\top K_{rr}\gamma\).}
		\label{fig:poisson_comparison}
	\end{figure*}

	For this problem, an analytical solution is available:
	\[
	u(x,y)=x(1-x)y(1-y)e^{x-y}.
	\]
	Therefore, both the solution error and the PDE residual can be evaluated directly, making this example particularly suitable for quantitative Comparison among different methods: .
	\paragraph{Comparison among different methods: }
	For the Poisson equation, Fig.~\ref{fig:poisson_comparison}(a)--(c) shows that LSGAN-RB achieves the best convergence behavior among all compared methods. It attains lower training error, validation error, and residual energy than the original DEQGAN baseline, indicating that the proposed rollback strategy can further improve adversarial PINNs training beyond the existing GAN-based formulation. Here, WGAN-GP still performs poorly, and the application of rollback does not lead to a clear improvement. This again suggests that the non-saturating property of WGAN-type objectives alone is insufficient to guarantee effective residual descent.
	
	Fig.~\ref{fig:poisson_comparison}(d)--(f) further explain this behavior from the perspective of the rollback dynamics. GAN-RB and LSGAN-RB maintain relatively high selected generator update steps during most of the training process, which means that the discriminator-induced directions can support a longer sequence of energy-decreasing generator updates. In contrast, WGAN-GP-RB keeps the generator update steps at a much lower level, implying that many candidate generator updates are rejected by the rollback criterion.
	
	The first-order indicator shows a consistent trend. For GAN-RB and LSGAN-RB, the indicator remains at a lower level during training, suggesting better alignment between the discriminator-induced residual weighting and the generator NTK dynamics. By contrast, the WGAN-GP-RB indicator becomes large already in the early stage, indicating an unfavorable or poorly aligned descent direction. Since the qualitative interpretation is consistent with the Laplace experiment, we omit repeated details here. Overall, the Poisson results further confirm that the effectiveness of adversarial PINNs training depends on the quality of the induced residual-space weighting rather than merely on the nominal GAN objective.
	
	\subsubsection{Reaction-Difussion}
	\paragraph{Reaction-Difussion equation}
	
	We consider a one-dimensional time-dependent parabolic equation on the space--time domain
	\[
	\Omega=(0,1)\times(0,5),
	\]
	subject to homogeneous Dirichlet boundary conditions and a zero initial condition. The governing equation is
	\[
	\left\{
	\begin{aligned}
		u_t-u_{xx}-10u
		&=
		\frac{2}{5}t-\left(2t-\frac15\right)x(1-x),
		\qquad (x,t)\in\Omega,\\
		u(0,t)&=0,\qquad t\in[0,5],\\
		u(1,t)&=0,\qquad t\in[0,5],\\
		u(x,0)&=0,\qquad x\in[0,1].
	\end{aligned}
	\right.
	\]
	
	This type of parabolic equation is commonly used to describe diffusion and heat-transfer processes, and more generally appears in time-dependent transport-reaction problems. For the present example, an analytical solution is available:
	\[
	u(x,t)=\frac{t}{5}x(1-x).
	\]
	Hence, the predicted solution can be directly compared with the exact solution, which makes this problem convenient for evaluating both approximation accuracy and training behavior.
	
\begin{figure*}[t]
		\centering
		\begin{subfigure}[t]{0.32\textwidth}
			\centering
			\includegraphics[width=\linewidth]{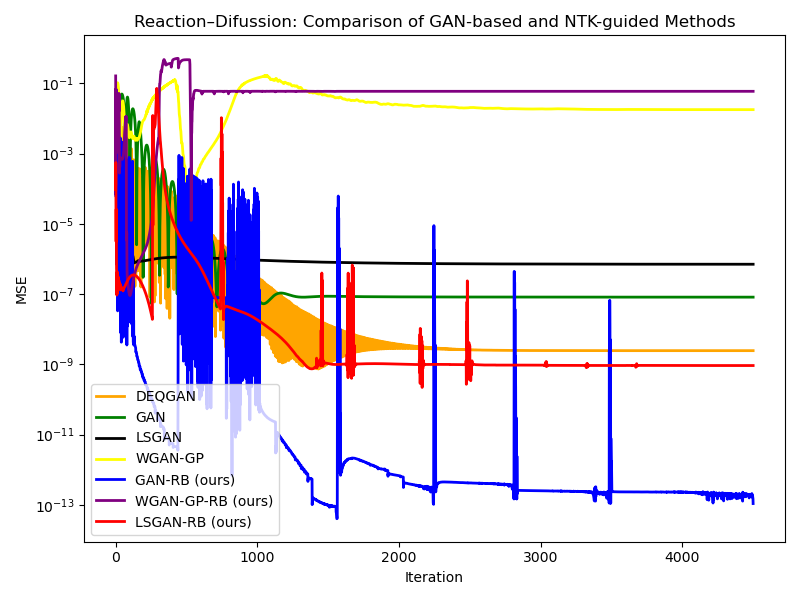}
			\caption{Training MSE}
			\label{fig:heat_train_mse}
		\end{subfigure}
		\hfill
		\begin{subfigure}[t]{0.32\textwidth}
			\centering
			\includegraphics[width=\linewidth]{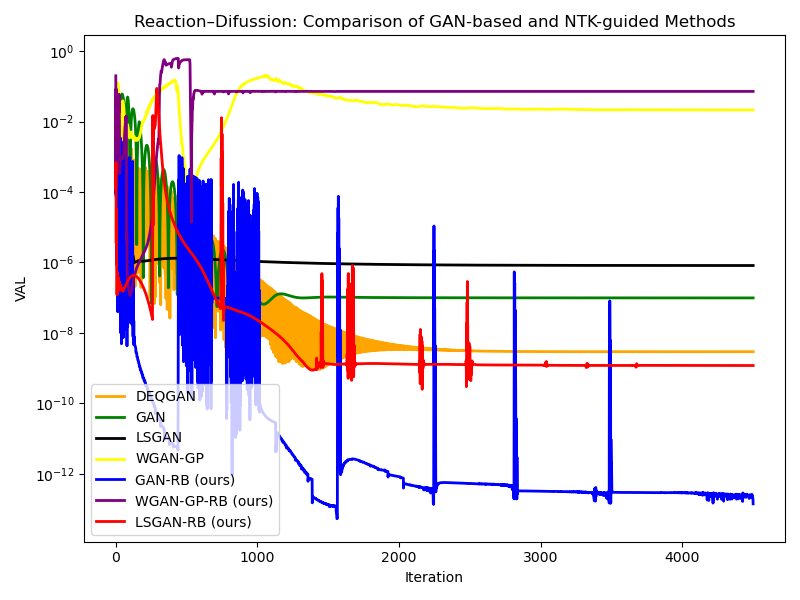}
			\caption{Validation MSE}
			\label{fig:heat_val_mse}
		\end{subfigure}
		\hfill
		\begin{subfigure}[t]{0.32\textwidth}
			\centering
			\includegraphics[width=\linewidth]{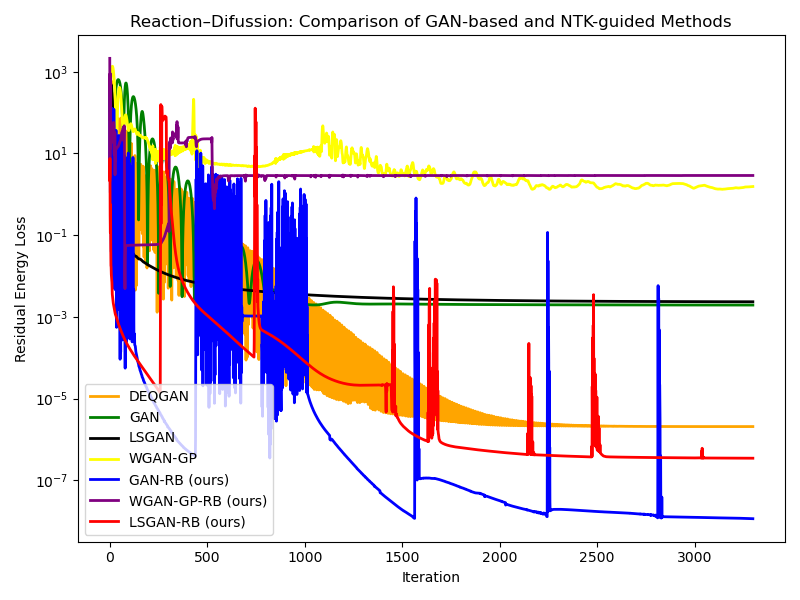}
			\caption{Residual energy}
			\label{fig:heat_residual_energy}
		\end{subfigure}
		
		\vspace{0.6em}
		
		\begin{subfigure}[t]{0.32\textwidth}
			\centering
			\includegraphics[width=\linewidth]{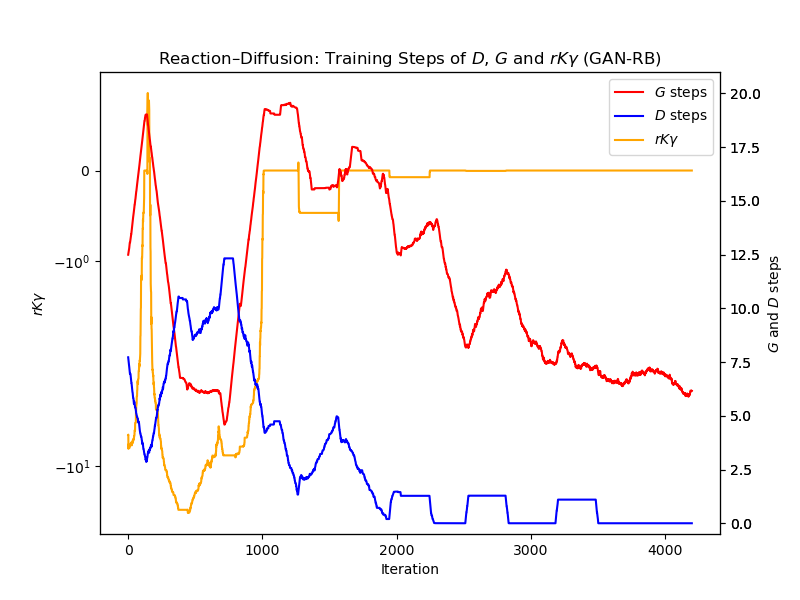}
			\caption{GAN-RB}
			\label{fig:heat_gan_rb_dynamics}
		\end{subfigure}
		\hfill
		\begin{subfigure}[t]{0.32\textwidth}
			\centering
			\includegraphics[width=\linewidth]{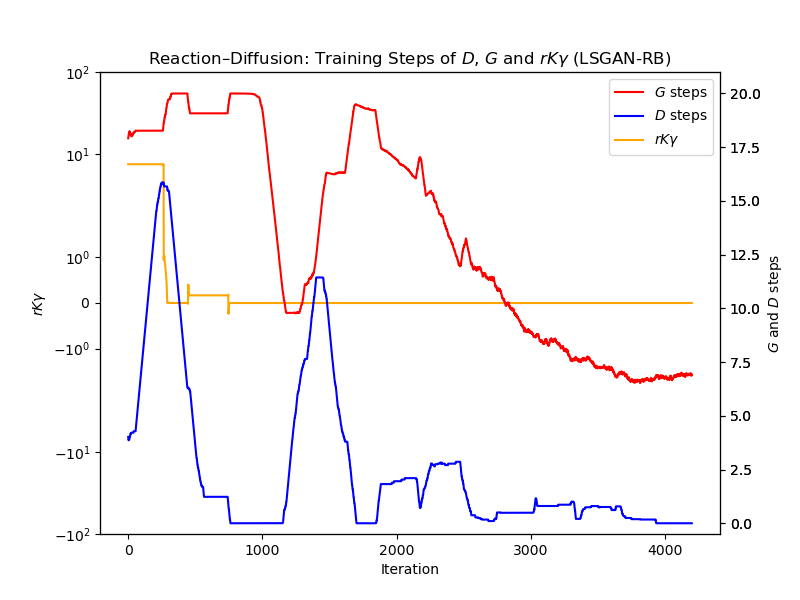}
			\caption{LSGAN-RB}
			\label{fig:heat_lsgan_rb_dynamics}
		\end{subfigure}
		\hfill
		\begin{subfigure}[t]{0.32\textwidth}
			\centering
			\includegraphics[width=\linewidth]{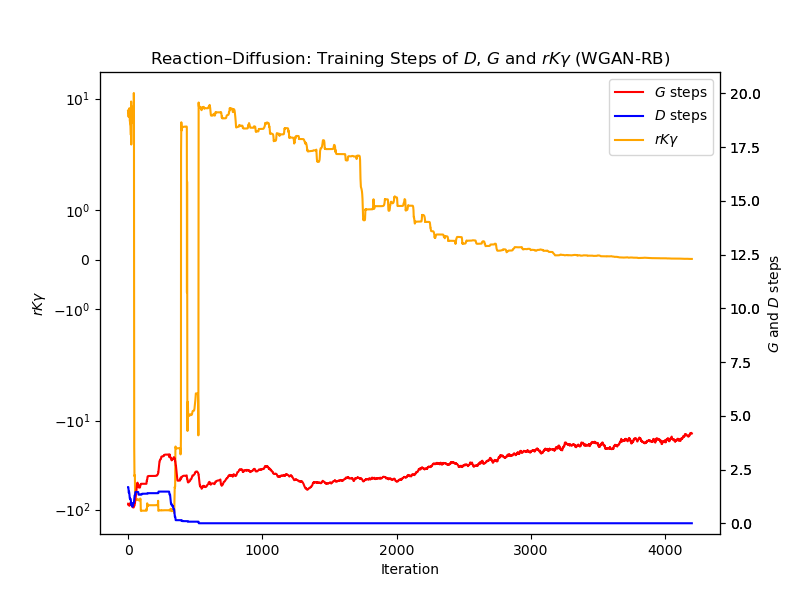}
			\caption{WGAN-GP-RB}
			\label{fig:heat_wgan_rb_dynamics}
		\end{subfigure}
		
		\caption{Reaction-Difussion equation: The first row reports the training MSE, validation MSE, and residual energy for all compared methods. The second row shows the rollback-related dynamics for GAN-RB, LSGAN-RB, and WGAN-GP-RB, respectively, including the selected generator/discriminator inner-step behavior and the discriminator-induced quantity \(r^\top K_{rr}\gamma\).}
		\label{fig:heat_comparison}
	\end{figure*}

	\paragraph{Comparison among different methods: }
	For the reaction--diffusion equation, Fig.~\ref{fig:heat_comparison}(a)--(c) shows that GAN-RB achieves the best accuracy among all compared methods. Compared with the other baselines, GAN-RB improves the training accuracy by approximately four orders of magnitude, demonstrating that the rollback strategy can be particularly effective for nonlinear or reaction-dominated residual dynamics.
	
	Fig.~\ref{fig:heat_comparison}(d)--(f) further illustrate the underlying update behavior. Both GAN-RB and LSGAN-RB maintain relatively high selected generator update steps during training, indicating that their discriminator-induced directions can still support repeated energy-decreasing generator updates. However, compared with GAN-RB, the first-order indicator of LSGAN-RB remains relatively large at the early stage. Since this indicator is an aggregated quantity, this suggests that the discriminator feedback of LSGAN-RB is less well aligned with the global residual-energy descent direction, although it can still produce decreasing residual energy. This explains why LSGAN-RB converges, but not as rapidly or accurately as GAN-RB.
	
	For WGAN-GP-RB, the rollback strategy brings a slight improvement over the original WGAN-GP. As shown in Fig.~\ref{fig:heat_comparison}(f), the WGAN-GP discriminator provides a useful descent direction at the very beginning of training. Nevertheless, this favorable behavior quickly disappears, and the selected generator updates become less effective afterward. As a result, the overall training performance remains unsatisfactory. These results again indicate that the success of the rollback mechanism depends on whether the discriminator-induced weighting can provide a persistently informative residual descent direction.
	
	\subsubsection{Viscous-Burgers}
	\paragraph{Viscous Burgers equation}
	
	We consider a one-dimensional viscous Burgers equation on the space--time domain
	\[
	\Omega=(-5,5)\times(0,2.5),
	\]
	with viscosity coefficient \(\nu=0.001\). The governing equation is
	\[
	\left\{
	\begin{aligned}
		u_t+u\,u_x-\nu u_{xx}&=0,
		\qquad (x,t)\in\Omega,\\
		u(x,0)&=\frac{1}{\cosh(x)},\qquad x\in[-5,5].
	\end{aligned}
	\right.
	\]
	In the implementation, the boundary values at \(x=\pm 5\) are fixed to \(1/\cosh(5)\), which is numerically very close to zero.
	
	The viscous Burgers equation is a classical nonlinear convection--diffusion model and is widely used as a benchmark for studying nonlinear transport, diffusion, and sharp-gradient dynamics. Owing to the competition between nonlinear advection and viscosity, it provides a representative test case for assessing the stability and accuracy of physics-informed methods.
	
	For this problem, the reference solution is not available in closed form. Instead, a high-accuracy numerical solution computed by an FFT-based solver is used for comparison.
	
\begin{figure*}[t]
		\centering
		\begin{subfigure}[t]{0.32\textwidth}
			\centering
			\includegraphics[width=\linewidth]{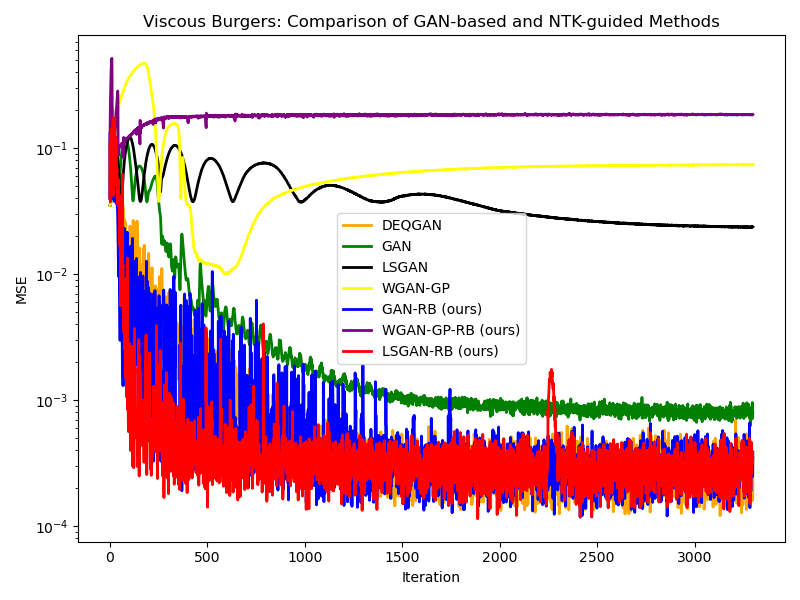}
			\caption{Training MSE}
			\label{fig:burv_train_mse}
		\end{subfigure}
		\hfill
		\begin{subfigure}[t]{0.32\textwidth}
			\centering
			\includegraphics[width=\linewidth]{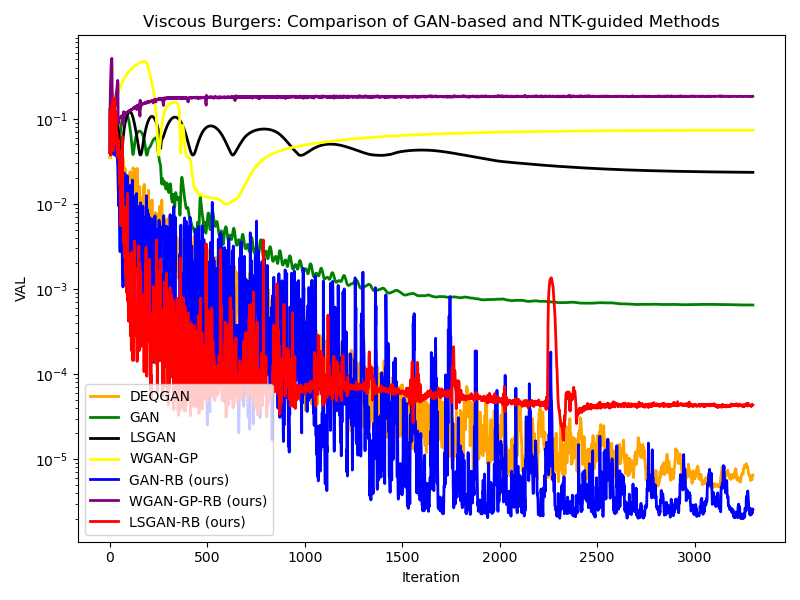}
			\caption{Validation MSE}
			\label{fig:burv_val_mse}
		\end{subfigure}
		\hfill
		\begin{subfigure}[t]{0.32\textwidth}
			\centering
			\includegraphics[width=\linewidth]{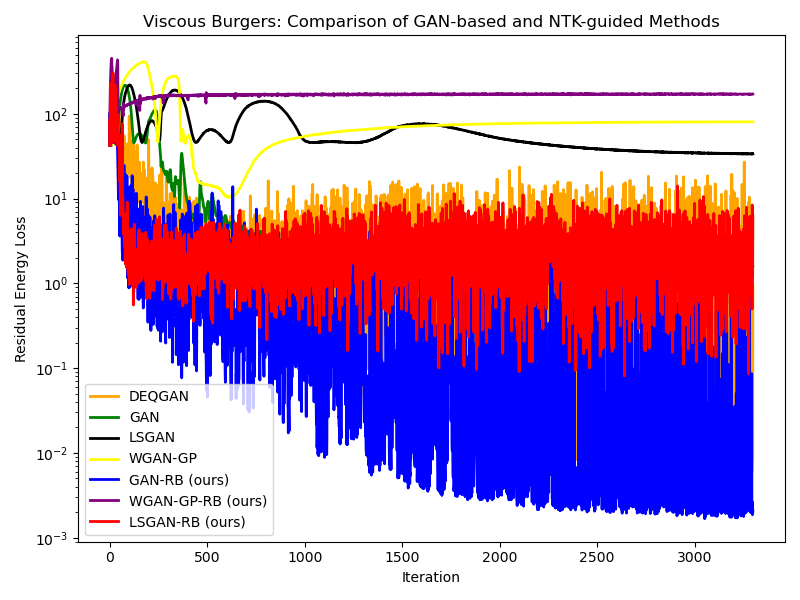}
			\caption{Residual energy}
			\label{fig:burv_residual_energy}
		\end{subfigure}
		
		\vspace{0.6em}
		
		\begin{subfigure}[t]{0.32\textwidth}
			\centering
			\includegraphics[width=\linewidth]{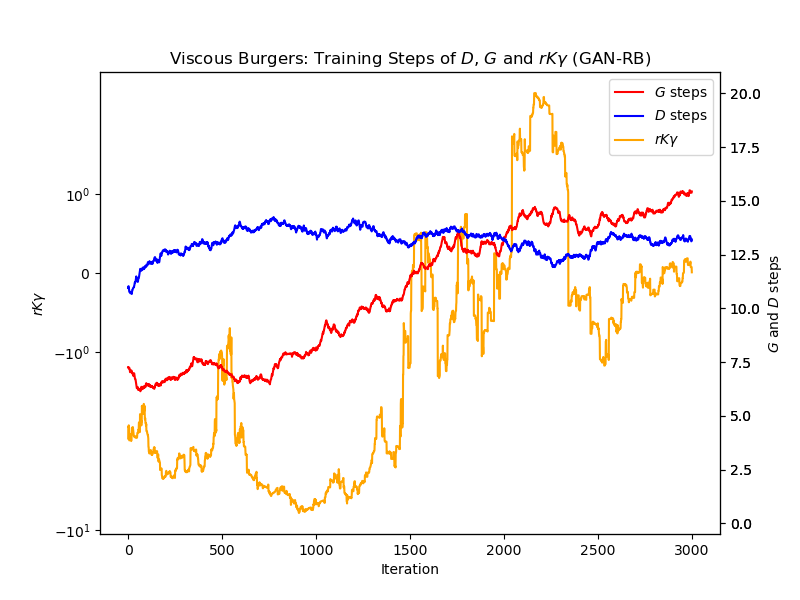}
			\caption{GAN-RB}
			\label{fig:burv_gan_rb_dynamics}
		\end{subfigure}
		\hfill
		\begin{subfigure}[t]{0.32\textwidth}
			\centering
			\includegraphics[width=\linewidth]{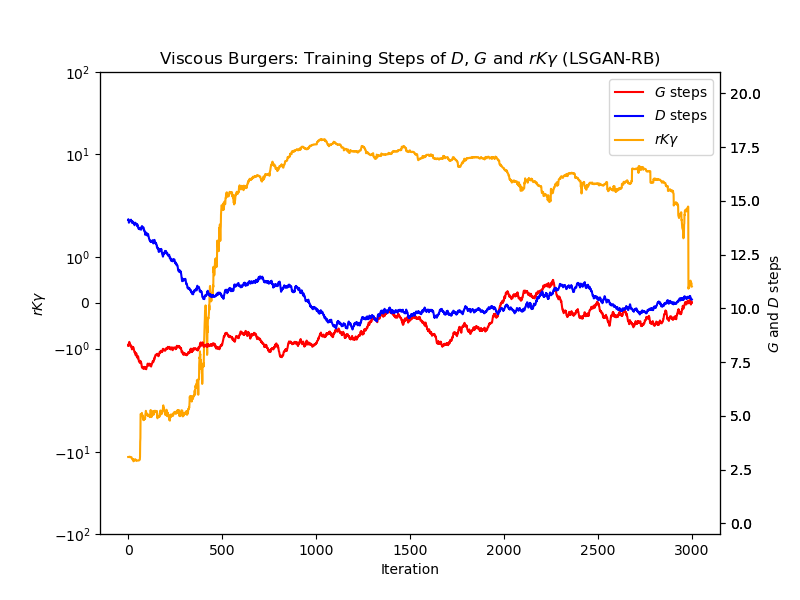}
			\caption{LSGAN-RB}
			\label{fig:burv_lsgan_rb_dynamics}
		\end{subfigure}
		\hfill
		\begin{subfigure}[t]{0.32\textwidth}
			\centering
			\includegraphics[width=\linewidth]{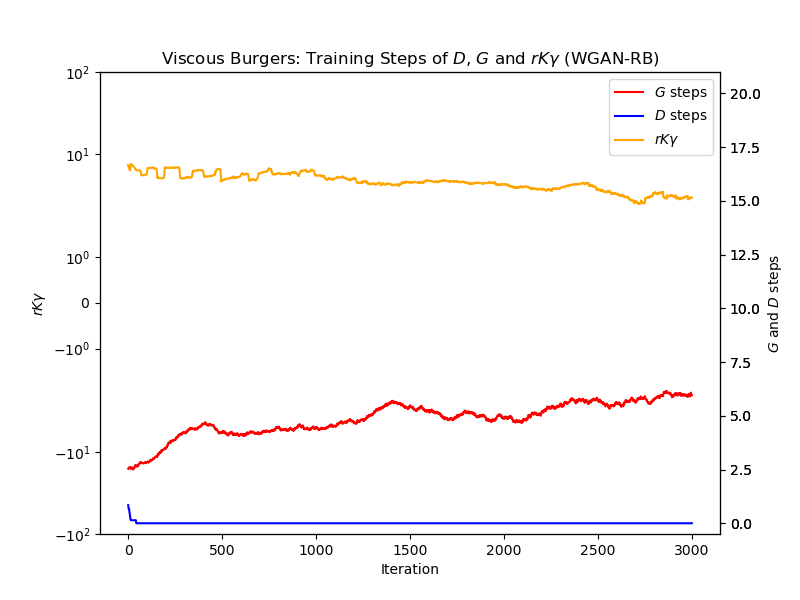}
			\caption{WGAN-GP-RB}
			\label{fig:burv_wgan_rb_dynamics}
		\end{subfigure}
		
		\caption{Viscous Burgers equation: The first row reports the training MSE, validation MSE, and residual energy for all compared methods. The second row shows the rollback-related dynamics for GAN-RB, LSGAN-RB, and WGAN-GP-RB, respectively, including the selected generator/discriminator inner-step behavior and the discriminator-induced quantity \(r^\top K_{rr}\gamma\).}
		\label{fig:burv_comparison}
	\end{figure*}

	\paragraph{Comparison among different methods: }
	For the Burgers equation, Fig.~\ref{fig:burv_comparison}(a)--(c) shows that all methods exhibit strongly oscillatory training dynamics. This behavior is reasonable because the Burgers equation contains nonlinear convection, and its residual landscape is more sensitive to local sharp structures and dynamically evolving error modes. As a result, the adversarial feedback may change rapidly during training, leading to more pronounced fluctuations than in the linear elliptic problems.
	
	Despite these oscillations, GAN-RB achieves the best overall performance. This indicates that, for nonlinear convection-dominated residual dynamics, the sharper discriminator-induced weighting of GAN, together with the rollback selection mechanism, can provide more effective residual descent directions than the smoother LSGAN-type weighting.
	
	Fig.~\ref{fig:burv_comparison}(d)--(f) further support this interpretation. Among the three rollback variants, GAN-RB maintains the largest selected generator update steps, followed by LSGAN-RB, while WGAN-GP-RB keeps the lowest generator update level. This ordering is consistent with their final accuracy. The first-order indicator of GAN-RB is also the most favorable and remains better suppressed during training, suggesting that its discriminator-induced weighting is better aligned with the generator residual dynamics.
	
	For LSGAN-RB, the first-order indicator is negative at the beginning, indicating an initially useful descent direction. However, it gradually moves toward the positive region and remains there, which weakens the effectiveness of subsequent generator updates and leads to an inferior final result compared with GAN-RB. WGAN-GP-RB performs the worst: its first-order indicator is poorly suppressed and the selected generator update steps remain low, implying that the WGAN-GP-RB discriminator fails to provide a persistently useful residual descent direction for this nonlinear problem.
	
	\subsubsection{Klein-Gordon}
	\paragraph{Klein-Gordon equation}
	\begin{figure*}[t]
		\centering
		\begin{subfigure}[t]{0.32\textwidth}
			\centering
			\includegraphics[width=\linewidth]{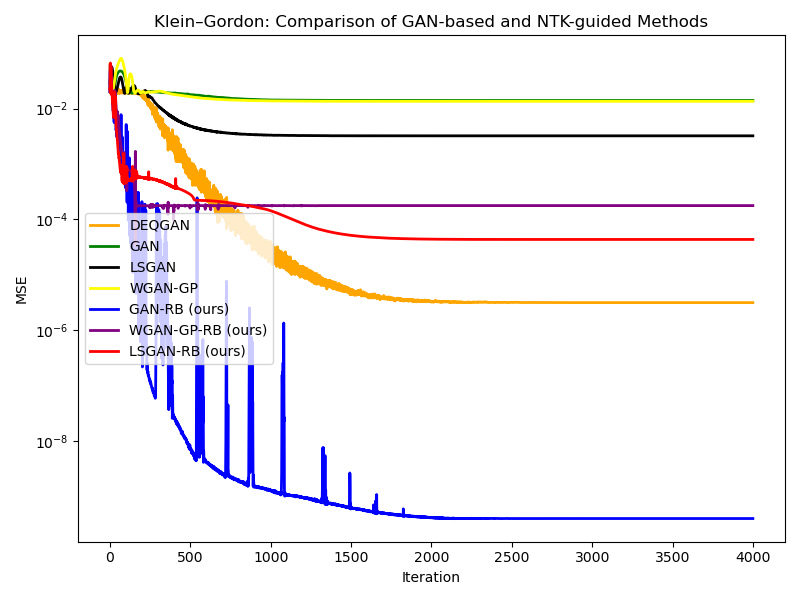}
			\caption{Training MSE}
			\label{fig:wave_train_mse}
		\end{subfigure}
		\hfill
		\begin{subfigure}[t]{0.32\textwidth}
			\centering
			\includegraphics[width=\linewidth]{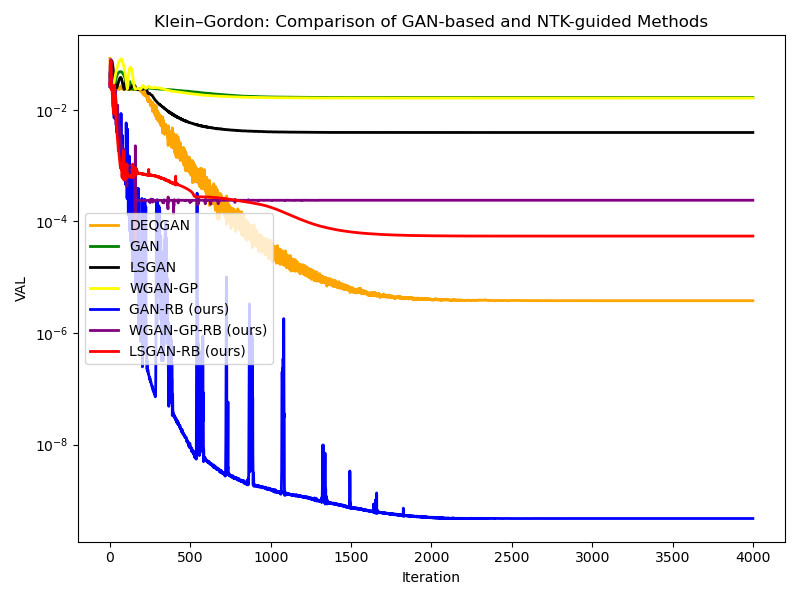}
			\caption{Validation MSE}
			\label{fig:wave_val_mse}
		\end{subfigure}
		\hfill
		\begin{subfigure}[t]{0.32\textwidth}
			\centering
			\includegraphics[width=\linewidth]{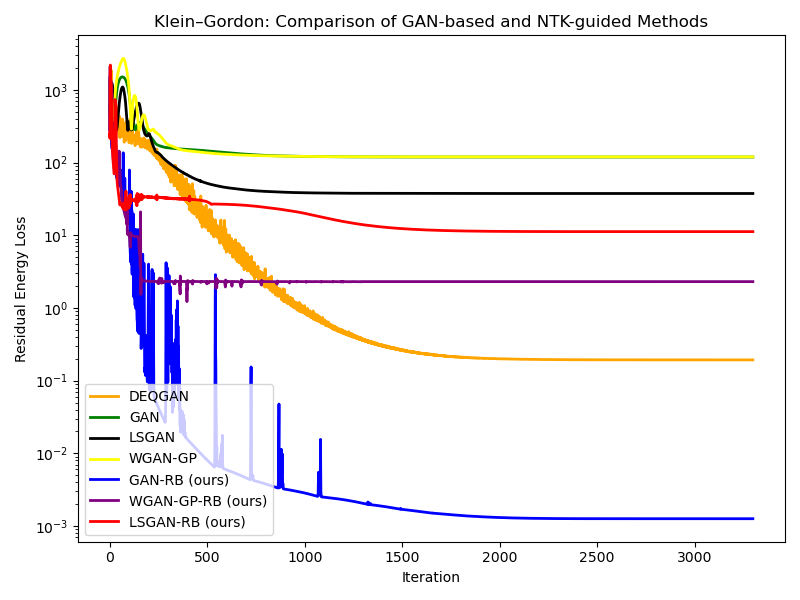}
			\caption{Residual energy}
			\label{fig:wave_residual_energy}
		\end{subfigure}
		
		\vspace{0.6em}
		
		\begin{subfigure}[t]{0.32\textwidth}
			\centering
			\includegraphics[width=\linewidth]{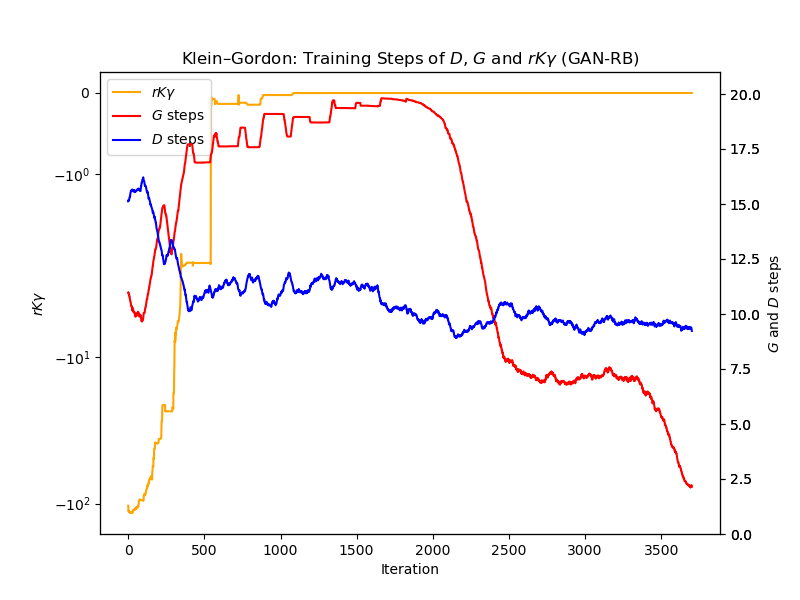}
			\caption{GAN-RB}
			\label{fig:wave_gan_rb_dynamics}
		\end{subfigure}
		\hfill
		\begin{subfigure}[t]{0.32\textwidth}
			\centering
			\includegraphics[width=\linewidth]{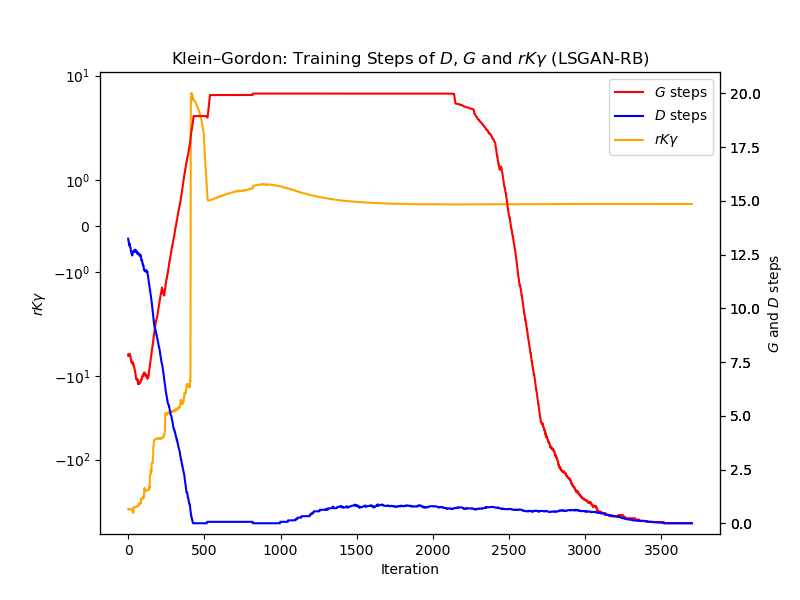}
			\caption{LSGAN-RB}
			\label{fig:wave_lsgan_rb_dynamics}
		\end{subfigure}
		\hfill
		\begin{subfigure}[t]{0.32\textwidth}
			\centering
			\includegraphics[width=\linewidth]{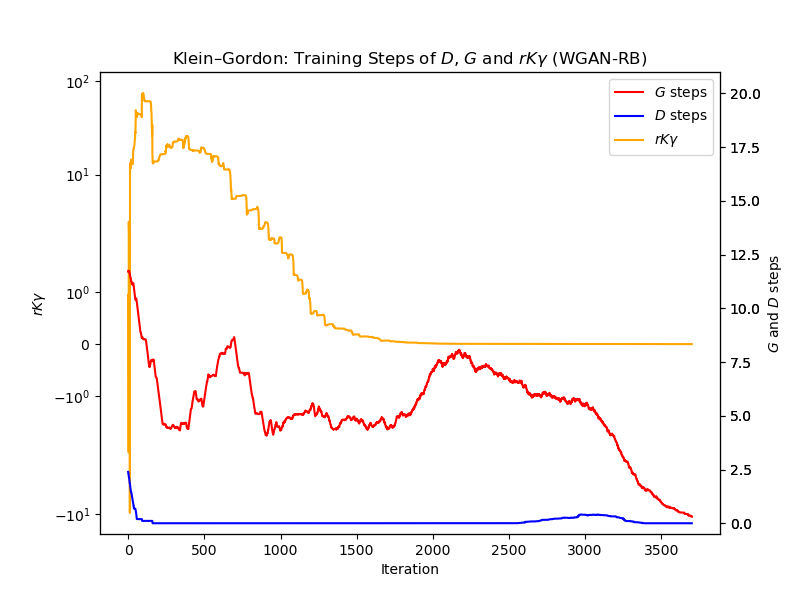}
			\caption{WGAN-GP-RB}
			\label{fig:wave_wgan_rb_dynamics}
		\end{subfigure}
		
		\caption{Klein-Gordon equation: The first row reports the training MSE, validation MSE, and residual energy for all compared methods. The second row shows the rollback-related dynamics for GAN-RB, LSGAN-RB, and WGAN-GP-RB, respectively, including the selected generator/discriminator inner-step behavior and the discriminator-induced quantity \(r^\top K_{rr}\gamma\).}
		\label{fig:wave_comparison} 
	\end{figure*}
	We further consider a one-dimensional Klein--Gordon-type equation on the space--time domain
	\[
	\Omega=(0,1)\times(0,2),
	\]
	with homogeneous Dirichlet boundary conditions. The governing equation is
	\[
	\left\{
	\begin{aligned}\label{wave}
		u_{tt}-c^2u_{xx}-4u
		&=
		0.8\sin(\pi x)\cos(\pi t),
		\qquad (x,t)\in\Omega,\\
		u(0,t)&=0,\qquad t\in[0,2],\\
		u(1,t)&=0,\qquad t\in[0,2],\\
		u(x,0)&=-0.2\sin(\pi x),\qquad x\in[0,1],\\
		u_t(x,0)&=0,\qquad x\in[0,1].
	\end{aligned}
	\right.
	\]
	
	The Klein--Gordon equation is a classical hyperbolic partial differential equation and is widely used to describe wave propagation with restoring or reaction effects. It also serves as a representative benchmark for testing whether learning-based solvers can accurately capture second-order temporal dynamics together with spatial oscillations.
	
	For the default choice \(c=1\), the problem admits the analytical solution
	\[
	u(x,t)=-0.2\sin(\pi x)\cos(\pi t).
	\]
	Hence, the predicted solution can be directly compared with the exact solution in both solution error and PDE residual.
	\paragraph{Comparison among different methods: }
	For the Klein--Gordon equation, Fig.~\ref{fig:wave_comparison}(a)--(c) shows that GAN-RB achieves the best accuracy among all methods. Its final error is approximately four orders of magnitude lower than the other baselines, demonstrating that GAN-RB is particularly effective for this oscillatory and wave-like problem.
	
	From Fig.~\ref{fig:wave_comparison}(d)--(f), we observe that both GAN-RB and LSGAN-RB maintain relatively high selected generator update steps during training. However, GAN-RB obtains a much larger improvement after applying the rollback strategy. This indicates that, in the Klein--Gordon problem, the GAN-induced residual weighting is better aligned with the generator residual dynamics than the smoother LSGAN-type weighting.
	
	The first-order indicator provides further evidence. For GAN-RB, the indicator is globally well suppressed in the favorable region throughout training, which is a strong indication that the discriminator feedback persistently provides useful descent directions. In contrast, LSGAN-RB also produces a favorable indicator at the beginning, but it gradually increases and eventually stays near a small positive value around zero. This means that the discriminator-induced direction becomes less effective in later training stages, explaining why LSGAN-RB is inferior to GAN-RB.
	
	For WGAN-GP-RB, the rollback strategy also brings a clear improvement, reducing the error by about two orders of magnitude compared with the original WGAN-GP. Different from the previous examples, Fig.~\ref{fig:wave_comparison}(f) shows that the selected generator update steps of WGAN-GP-RB are significantly higher in this experiment. This suggests that the WGAN-GP discriminator can provide more useful descent directions for the Klein--Gordon residual dynamics than in the Laplace, Poisson, or reaction--diffusion cases. Nevertheless, its first-order indicator is still less stable and less favorable than that of GAN-RB, so its final performance remains worse.

	\subsection{Ablation experiment}
	\label{app:ablation}
	\subsubsection{Laplace equation (Under different boundary conditions)}
	Here, we conduct the first ablation experiment to examine the robustness of the proposed method under changes in boundary conditions. The motivation for this experiment is that, even when the governing PDE remains unchanged, the solution can be highly sensitive to the prescribed boundary conditions. Therefore, it is important to verify whether the proposed algorithm can still maintain stable and accurate performance when the boundary conditions are modified.
	
	Specifically, we keep the differential operator and the computational domain fixed, and only change the boundary condition setting. This allows us to isolate the effect of boundary-induced changes in the solution structure, while avoiding interference from changes in the PDE type itself. If the proposed method still achieves strong performance under this setting, it provides evidence that the improvement is not limited to one specific boundary configuration, but reflects a more robust residual optimization mechanism.
	\paragraph{Laplace with different boundary condition: }
	We also consider a two-dimensional Laplace equation on the unit square domain
	\[
	\Omega=(0,1)\times(0,1),
	\]
	subject to Dirichlet boundary conditions:
	\[
	\left\{
	\begin{aligned}\label{laplace_2}
		u_{xx}+u_{yy}&=0,
		\qquad (x,y)\in\Omega,\\
		u(0,y)&=0,\qquad y\in[0,1],\\
		u(1,y)&=0,\qquad y\in[0,1],\\
		u(x,0)&=\sin(\pi x),\qquad x\in[0,1],\\
		u(x,1)&=\frac{1}{\cosh(\pi)}\sin(\pi x)+\frac{1}{2}\tanh(\pi)\cos\!\left(\pi x+\frac{\pi}{2}\right),
		\qquad x\in[0,1].
	\end{aligned}
	\right.
	\]
	Compared with simpler homogeneous-boundary examples, this test case involves nontrivial non-homogeneous boundary data and is therefore useful for evaluating the capability of a method to handle more complex boundary structures.
	
\begin{figure*}[t]
		\centering
		\begin{subfigure}[t]{0.32\textwidth}
			\centering
			\includegraphics[width=\linewidth]{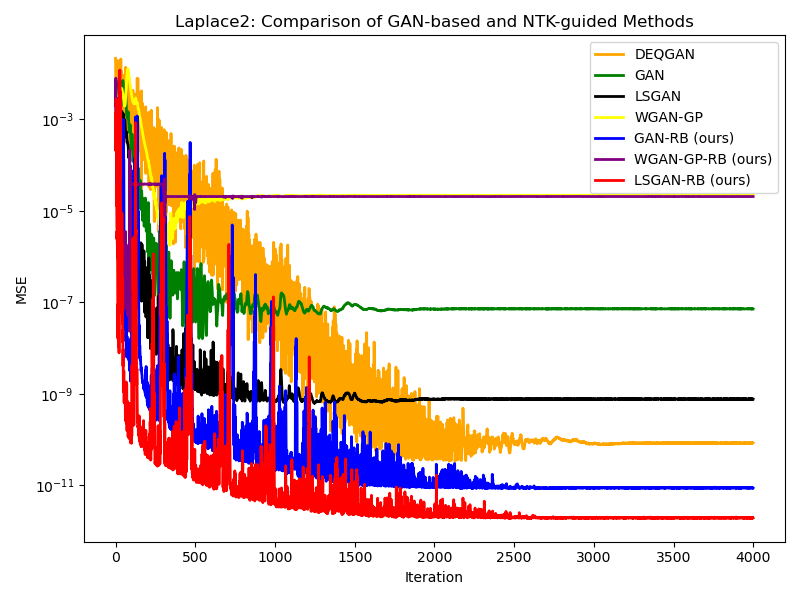}
			\caption{Training MSE}
			\label{fig:lap2_train_mse}
		\end{subfigure}
		\hfill
		\begin{subfigure}[t]{0.32\textwidth}
			\centering
			\includegraphics[width=\linewidth]{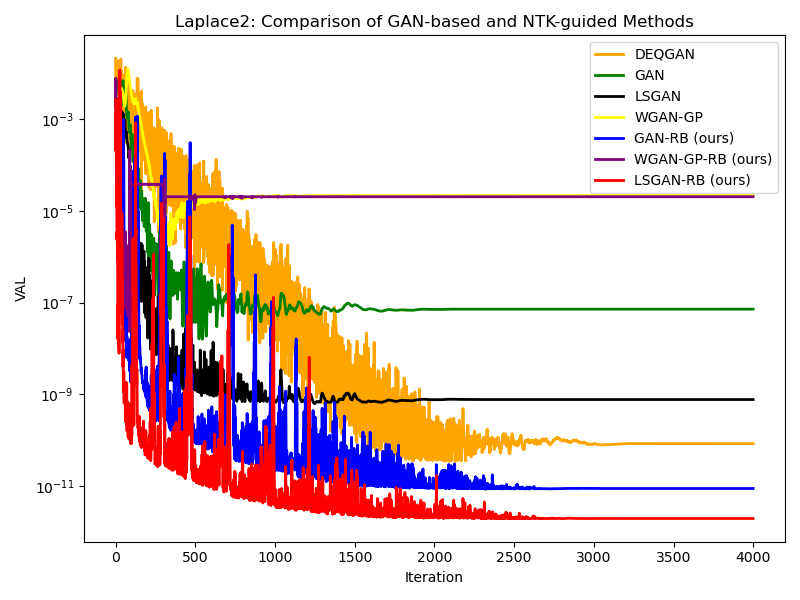}
			\caption{Validation MSE}
			\label{fig:lap2_val_mse}
		\end{subfigure}
		\hfill
		\begin{subfigure}[t]{0.32\textwidth}
			\centering
			\includegraphics[width=\linewidth]{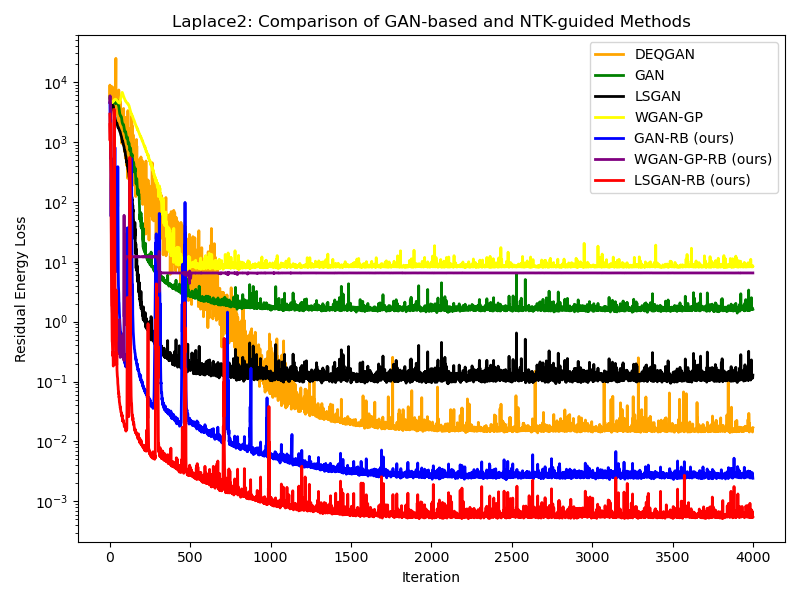}
			\caption{Residual energy}
			\label{fig:lap2_residual_energy}
		\end{subfigure}
		
		\vspace{0.6em}
		
		\begin{subfigure}[t]{0.32\textwidth}
			\centering
			\includegraphics[width=\linewidth]{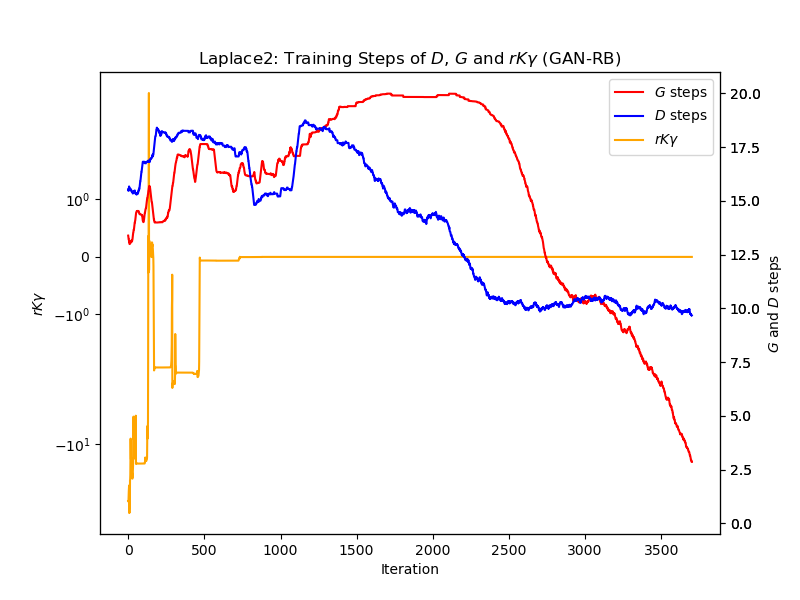}
			\caption{GAN-RB}
			\label{fig:lap2_gan_rb_dynamics}
		\end{subfigure}
		\hfill
		\begin{subfigure}[t]{0.32\textwidth}
			\centering
			\includegraphics[width=\linewidth]{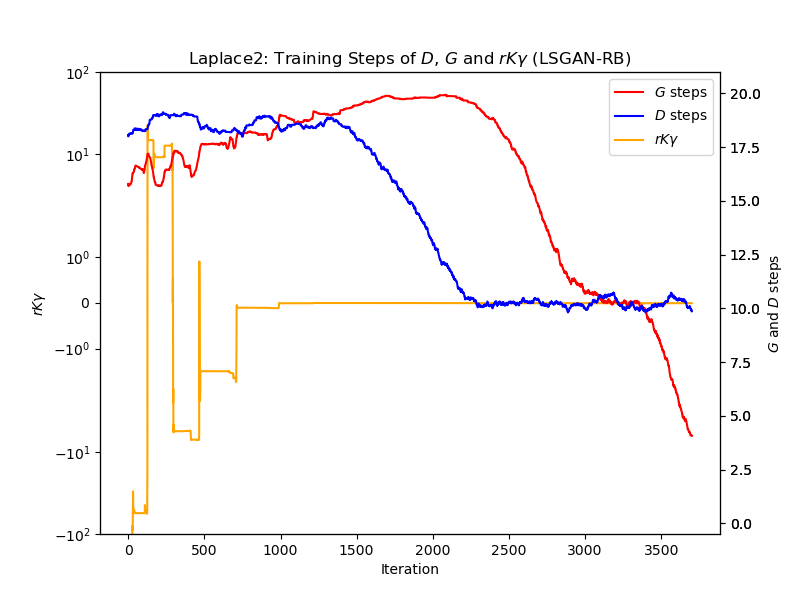}
			\caption{LSGAN-RB}
			\label{fig:lap2_lsgan_rb_dynamics}
		\end{subfigure}
		\hfill
		\begin{subfigure}[t]{0.32\textwidth}
			\centering
			\includegraphics[width=\linewidth]{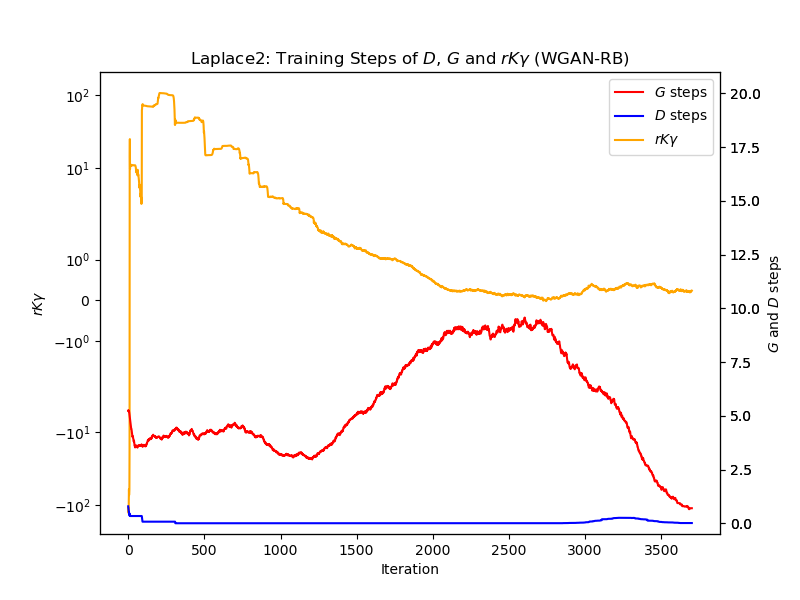}
			\caption{WGAN-GP-RB}
			\label{fig:lap2_wgan_rb_dynamics}
		\end{subfigure}
		
		\caption{Ablation study on the same pde with different boundary condition: The first row reports the training MSE, validation MSE, and residual energy for all compared methods. The second row shows the rollback-related dynamics for GAN-RB, LSGAN-RB, and WGAN-GP-RB, respectively, including the selected generator/discriminator inner-step behavior and the discriminator-induced quantity \(r^\top K_{rr}\gamma\).}
		\label{fig:lap2_comparison}
	\end{figure*}

	For this problem, an analytical solution is available:
	\[
	u(x,y)=\frac{1}{2\cosh(\pi)}\sin(\pi x)\bigl(e^{\pi(y-1)}+e^{\pi(1-y)}\bigr)
	+\frac{1}{4\cosh(\pi)}\cos\!\left(\pi x+\frac{\pi}{2}\right)\bigl(e^{\pi y}-e^{-\pi y}\bigr).
	\]
	Hence, the numerical prediction can be directly compared with the exact solution.
	\paragraph{Comparison among different methods: }
	For the second Laplace setting, Fig.~\ref{fig:lap2_comparison} evaluates the same Laplace equation under different boundary conditions. As shown in Fig.~\ref{fig:lap2_comparison}(a)--(c), both GAN-RB and LSGAN-RB achieve strong convergence performance. Compared with the other methods, they improve the final accuracy by about two orders of magnitude. This result indicates that the proposed rollback strategy remains effective even when the boundary conditions are changed, suggesting good robustness with respect to boundary-induced variations in the solution structure.
	
	Fig.~\ref{fig:lap2_comparison}(d) and~(e) further show that GAN-RB and LSGAN-RB maintain a large number of selected generator update steps during training. At the same time, the discriminator updates help keep the first-order indicator mostly below zero, meaning that the discriminator-induced directions remain favorable for residual-energy reduction. This explains why both methods can achieve stable and accurate convergence in this modified boundary setting.
	
	In contrast, WGAN-GP-RB still performs poorly. As shown in Fig.~\ref{fig:lap2_comparison}(f), its first-order indicator quickly moves to a large positive value at the early training stage, indicating poor alignment between the WGAN-GP-induced weighting and the generator residual dynamics. Consequently, even with rollback, WGAN-GP-RB cannot provide sufficiently effective descent directions for this problem.
	
	\subsubsection{Compatibility with DEQGAN}
	\label{app:ablation_deqgan_rb}
	
	To further evaluate the compatibility of the proposed rollback strategy, we combine it with DEQGAN and test the resulting method on the Laplace equation introduced in Section~\ref{laplace_1}. The purpose of this experiment is not merely to seek another performance gain, but to clarify an important conceptual point: rollback is not a standalone algorithm disconnected from existing adversarial PINNs methods. Rather, it should be understood as a dynamical modification of the alternating optimization process motivated by the NTK-based residual analysis developed in this paper.
	
	From this viewpoint, rollback does not replace the underlying adversarial objective or discriminator design. Instead, it operates at the level of update selection, using the first-order residual-dynamics criterion to filter candidate alternating steps and retain those more consistent with descent-oriented residual evolution. Therefore, if our interpretation is correct, the rollback principle should be compatible not only with GAN and LSGAN, but also with existing adversarial PINNs frameworks such as DEQGAN.
	
	To verify this, we apply rollback on top of DEQGAN under the Laplace benchmark. This setting is particularly informative because DEQGAN already relies on carefully tuned hyperparameters and a specific two-timescale training protocol. If rollback can still improve performance in this case, then its benefit cannot be attributed merely to replacing weak baselines or to using simpler optimization settings. Instead, it would support the stronger claim that rollback captures a more general principle of alternating adversarial PINNs training.
	
	The results confirm this expectation: even when combined with DEQGAN, the rollback-enhanced variant achieves improved convergence behavior and better final accuracy on the Laplace problem. This indicates that the proposed rollback mechanism is compatible with existing adversarial PINNs methods and can serve as an additional optimization layer on top of them. In this sense, the contribution of rollback is not to define a completely separate training paradigm, but to provide a principled update-control mechanism that can be integrated into broader adversarial PINNs frameworks.
	
	\begin{figure*}[t]
		\centering
		
		\begin{subfigure}[t]{0.32\textwidth}
			\centering
			\includegraphics[width=\linewidth]{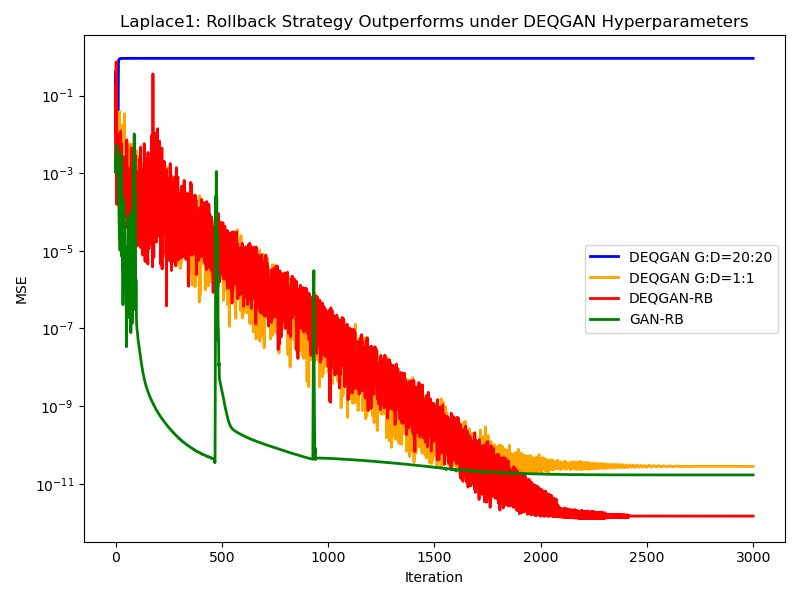}
			\caption{Training MSE}
			\label{fig:app:lap2_train_mse}
		\end{subfigure}
		\hfill
		\begin{subfigure}[t]{0.32\textwidth}
			\centering
			\includegraphics[width=\linewidth]{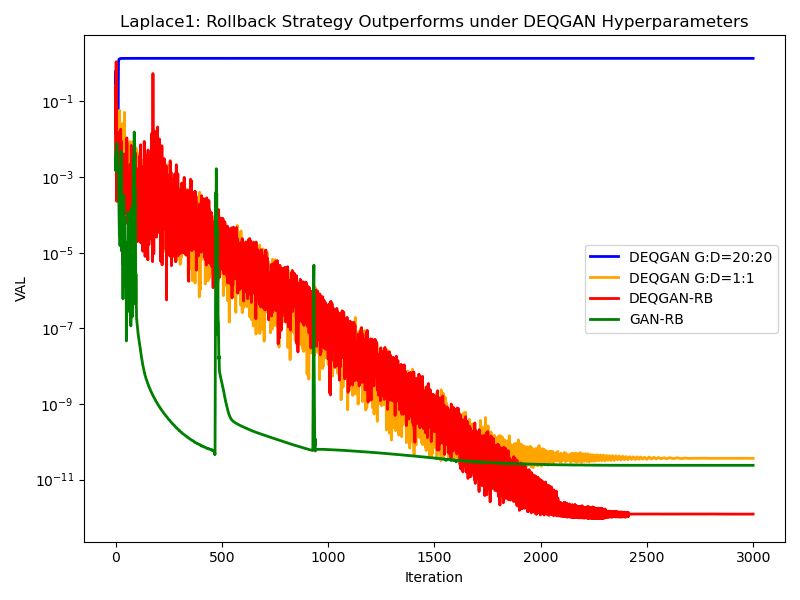}
			\caption{Validation MSE}
			\label{fig:app:lap2_val_mse}
		\end{subfigure}
		\hfill
		\begin{subfigure}[t]{0.32\textwidth}
			\centering
			\includegraphics[width=\linewidth]{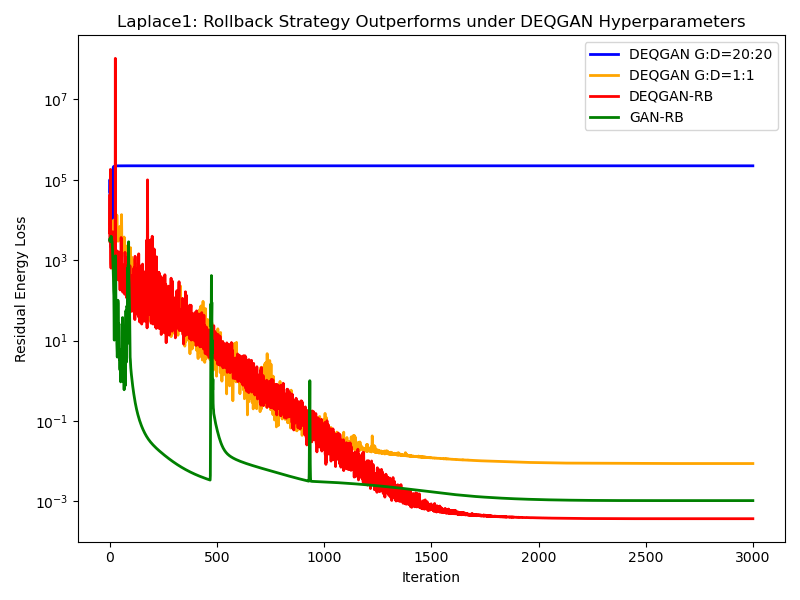}
			\caption{Residual energy}
			\label{fig:app:lap2_residual_energy}
		\end{subfigure}
		
		\vspace{0.7em}
		
		\begin{subfigure}[t]{0.32\textwidth}
			\centering
			\includegraphics[width=\linewidth]{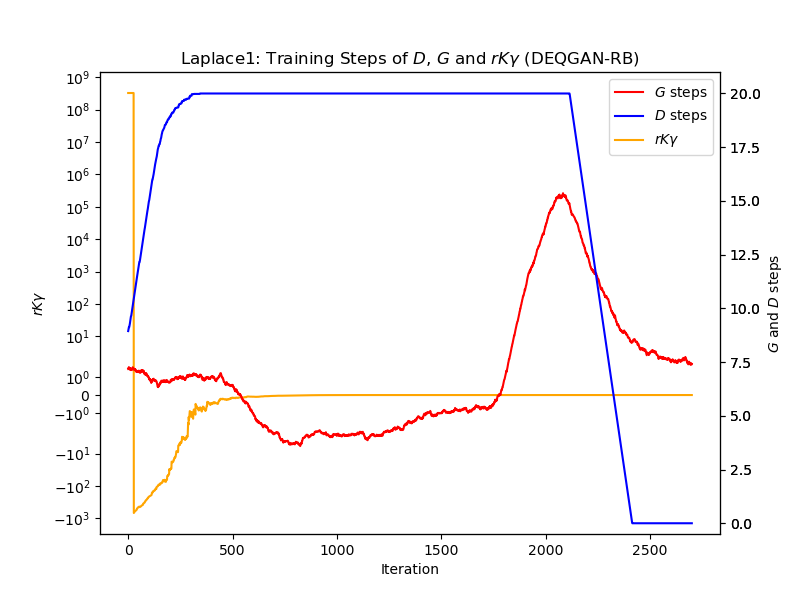}
			\caption{DEQGAN-RB dynamics}
			\label{fig:app:lap2_deqgan_rb_dynamics}
		\end{subfigure}
		\hfill
		\begin{subfigure}[t]{0.32\textwidth}
			\centering
			\includegraphics[width=\linewidth]{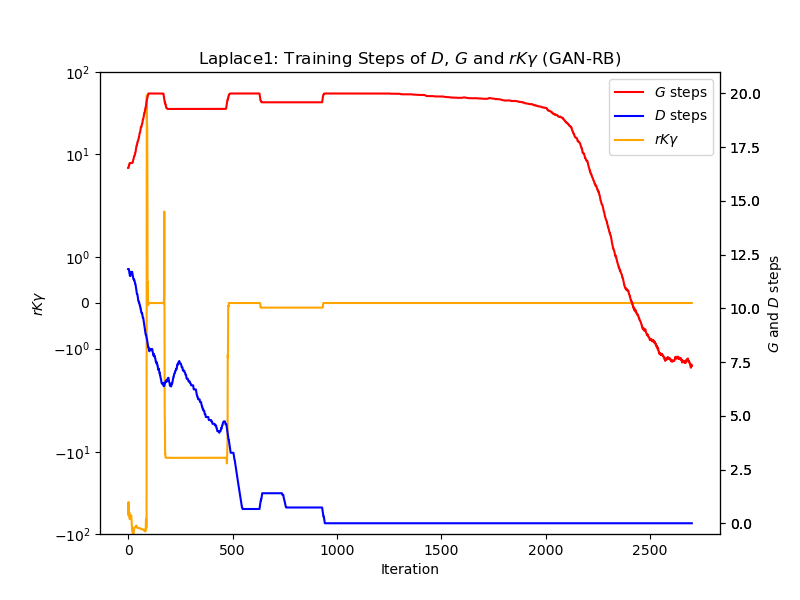}
			\caption{GAN-RB dynamics}
			\label{fig:app:lap2_gan_rb_dynamics}
		\end{subfigure}
		\hfill
		\begin{subfigure}[t]{0.32\textwidth}
			\centering
			\includegraphics[width=\linewidth]{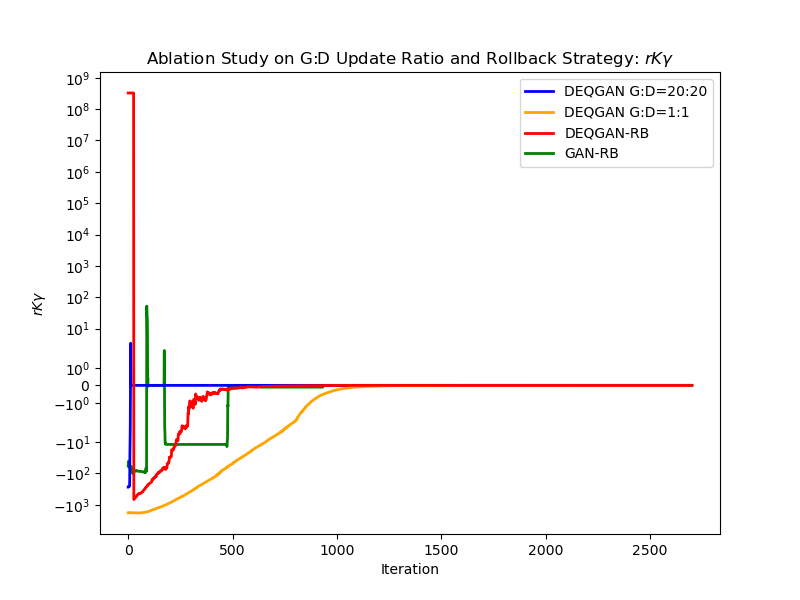}
			\caption{Comparison of \(r^\top K_{rr}\gamma\)}
			\label{fig:appL:lap2_gan_gkr}
		\end{subfigure}
		
		\caption{Ablation study on the DEQGAN and RB. 
			The top row reports the training MSE, validation MSE, and residual energy of all compared methods. 
			The bottom row illustrates the rollback dynamics of adversarial variants, including the selected generator/discriminator inner-step evolution and the discriminator-induced quantity \(r^\top K_{rr}\gamma\).}
		
		\label{fig:app:lap2_comparison}
	\end{figure*}
	
	\paragraph{Comparison among different methods: }
	In Fig.~\ref{fig:app:lap2_comparison}, we further combine the rollback strategy with the existing DEQGAN method to demonstrate the flexibility and broad applicability of the proposed strategy. As shown in Fig.~\ref{fig:app:lap2_comparison}(a)--(c), the resulting DEQGAN-RB method achieves the best accuracy among all compared methods. This indicates that rollback is not restricted to the standard GAN, LSGAN, or WGAN-GP formulations, but can also be incorporated into existing adversarial PINNs frameworks to further improve their stability and accuracy.
	
	An interesting observation is that the original DEQGAN performs poorly when the generator--discriminator update ratio is fixed to \(20:20\). This is reasonable because the original DEQGAN hyperparameters were tuned for a different update setting, such as a \(1:1\) update ratio. In contrast, after incorporating rollback, DEQGAN-RB can adaptively select more favorable update states and therefore achieves much better performance under the same update budget. We also include GAN-RB as a comparison. The results show that, when rollback is applied on top of a properly tuned adversarial training method, it can produce more stable training dynamics and higher final accuracy.
	
	This observation suggests that rollback can serve as a general stabilization module for adversarial PINNs training. Across the previous five numerical experiments, adversarial training methods often exhibit strong fluctuations, especially on nonlinear or oscillatory PDEs. Therefore, combining rollback with existing adversarial PINNs methods is a meaningful direction for improving both robustness and training stability.
	
	Fig.~\ref{fig:app:lap2_comparison}(d) and~(e) compare the rollback dynamics of DEQGAN-RB and GAN-RB. We observe that DEQGAN-RB selects fewer generator update steps than GAN-RB. This does not contradict the previous analysis, because the two methods use different learning rates. DEQGAN-RB adopts a larger learning rate, so each accepted generator update can produce a larger decrease in the residual energy. As a result, fewer accepted generator steps are sufficient to achieve strong convergence. In contrast, GAN-RB uses more generator updates, but its first-order indicator shows much stronger fluctuations. After hyperparameter tuning and integration with the existing DEQGAN framework, the rollback indicator becomes more stable, which explains the superior performance of DEQGAN-RB in this experiment.
	
	In Fig.~\ref{fig:app:lap2_comparison}(f), we compare the first-order indicator across four methods. It can be observed that the \textsc{DEQGAN} method achieves its best performance under a $1{:}1$ generator--discriminator update ratio. This is because, when the number of alternating updates per outer iteration is small, the generator and discriminator—after careful hyperparameter tuning—are less likely to become excessively strong or weak relative to each other. 
	However, since the total number of updates is strictly limited under the $1{:}1$ setting, the overall optimization process remains insufficiently explored, and the final performance is therefore suboptimal. When the update budget is increased, e.g., to a $20{:}20$ ratio, a different phenomenon emerges. We observe that the first-order indicator rapidly approaches zero within a very small number of iterations. This suggests that the discriminator becomes effectively over-trained relative to the generator: the discriminator gradients quickly vanish, leading to $\boldsymbol{\gamma} \approx \mathbf{0}$ and thus a near-degenerate training signal for the generator. In other words, simply increasing the update budget of \textsc{DEQGAN} is not sufficient to guarantee improved optimization performance. For the \textsc{GAN-RB} method, although no extensive hyperparameter tuning is performed, it achieves better results than the $1{:}1$ baseline. However, its first-order indicator is noticeably unstable. As shown in Fig.~F, there exist several abrupt positive spikes (i.e., indicator values exceeding zero), which correspond to regions where the loss exhibits significant oscillations. This instability reflects inconsistent descent behavior in the residual energy. Finally, we emphasize that the proposed \textsc{DEQGAN-RB} method achieves the most stable first-order indicator under the $20{:}20$ update ratio. 
	This stability directly translates into more consistent energy decay and smoother training dynamics, ultimately leading to higher solution accuracy. These observations suggest that combining existing adaptive alternating training strategies with the proposed first-order rollback mechanism could yield new methods with both improved stability and accuracy, which provides an important direction for future research.
	
	\subsubsection{Does rollback help beyond increasing the update budget?}
	\label{app:ablation_budget_control}
	
	A natural concern about the proposed rollback strategy is that it evaluates multiple candidate updates within each outer iteration, and therefore may appear to benefit simply from using more update steps. To rule out this possibility, we perform a controlled ablation and examine whether the observed performance gain truly comes from rollback itself.
	
	We focus on the two rollback-enhanced methods that perform best overall in our experiments, namely \textbf{GAN-RB} and \textbf{LSGAN-RB}. For \textbf{GAN-RB}, we use the Klein--Gordon equation benchmark in Section~\ref{wave}, where GAN-RB performs best. For \textbf{LSGAN-RB}, we use the Laplace equation in Section~\ref{laplace_2}, where this method achieves its strongest performance.
	
	For each method, we compare three training settings: \textbf{(i) no rollback with \(G:D=1\!:\!1\), (ii) no rollback with \(G:D=20\!:\!20\), and (iii) rollback with \(G:D=20\!:\!20\).} This design is chosen to answer two separate questions. The comparison between settings (i) and (ii) tests whether simply increasing the number of generator and discriminator updates can improve the solution accuracy. The comparison between settings (ii) and (iii) then isolates the effect of rollback under the same \(20\!:\!20\) update budget, and asks whether adding the proposed optimization strategy on top of that fixed budget leads to further improvement. Therefore, if the third setting consistently outperforms the second, the gain cannot be attributed merely to more candidate updates, but must come from the rollback mechanism itself.
	\paragraph{Laplace.}
	\begin{figure*}[t]
		\centering
		\begin{subfigure}[t]{0.32\textwidth}
			\centering
			\includegraphics[width=\linewidth]{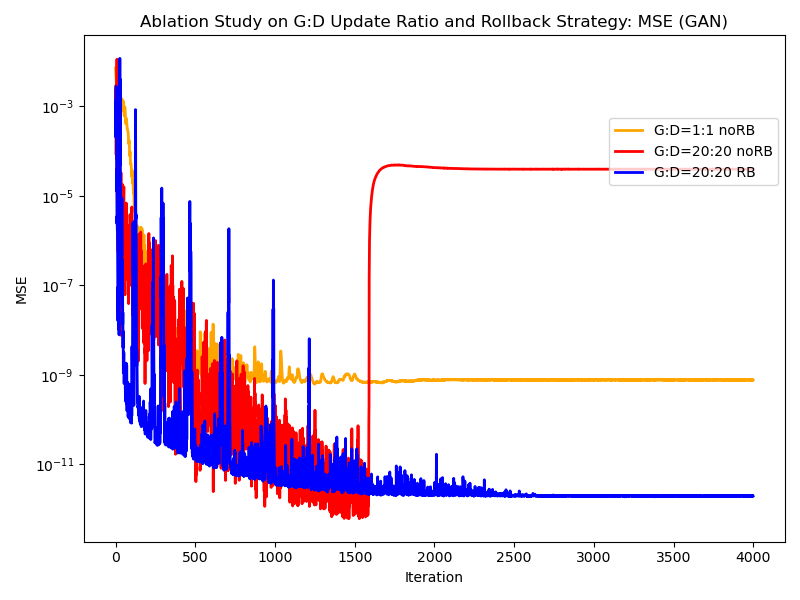}
			\caption{Training MSE}
			\label{fig:ami_lap2_lsgan_mse}
		\end{subfigure}
		\hfill
		\begin{subfigure}[t]{0.32\textwidth}
			\centering
			\includegraphics[width=\linewidth]{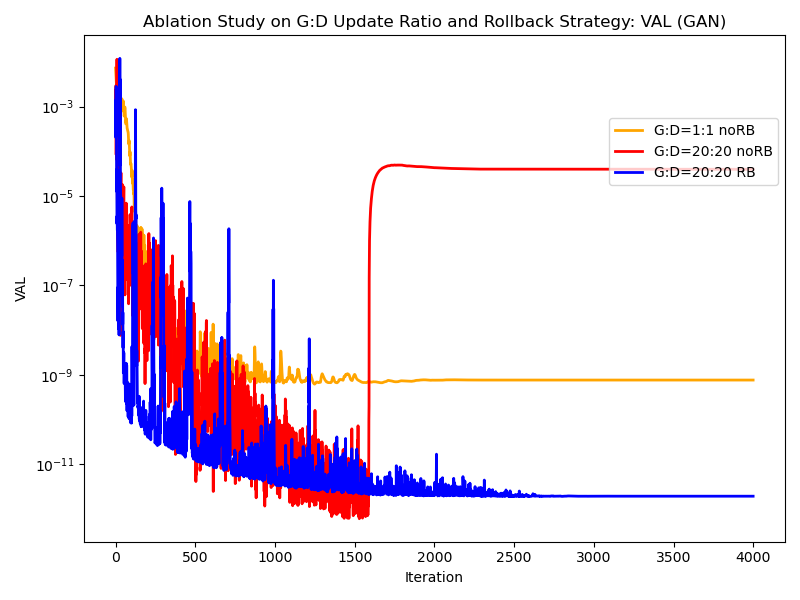}
			\caption{Validation MSE}
			\label{fig:ami_lap2_lsgan_val}
		\end{subfigure}
		\hfill
		\begin{subfigure}[t]{0.32\textwidth}
			\centering
			\includegraphics[width=\linewidth]{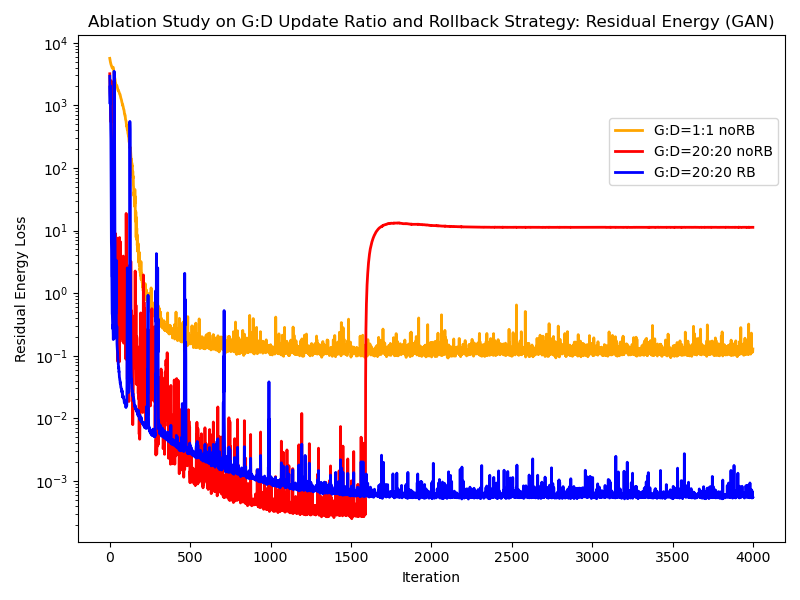}
			\caption{Residual energy}
			\label{fig:ami_lap2_lsgan_energy}
		\end{subfigure}
		
		\vspace{0.6em}
		
		\begin{subfigure}[t]{0.32\textwidth}
			\centering
			\includegraphics[width=\linewidth]{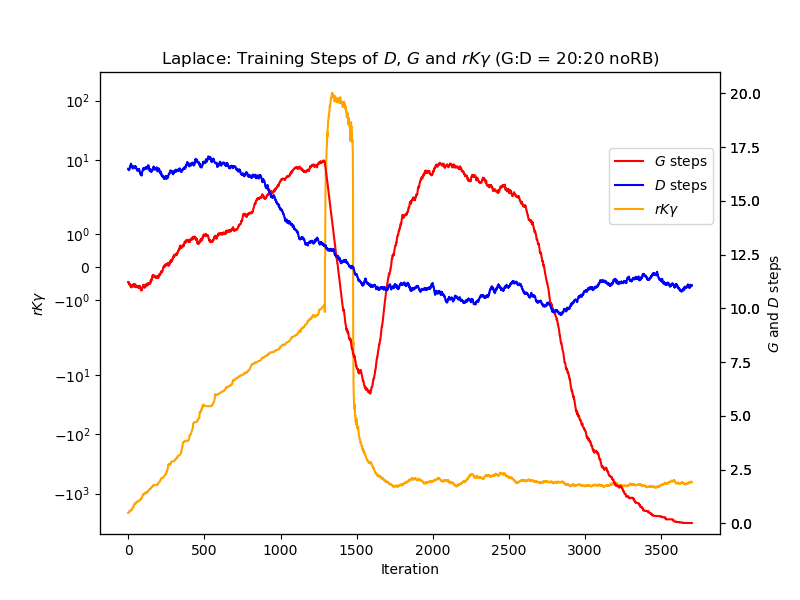}
			\caption{No rollback, \(G:D=20{:}20\)}
			\label{fig:ami_lap2_lsgan_norb_dynamics}
		\end{subfigure}
		\hfill
		\begin{subfigure}[t]{0.32\textwidth}
			\centering
			\includegraphics[width=\linewidth]{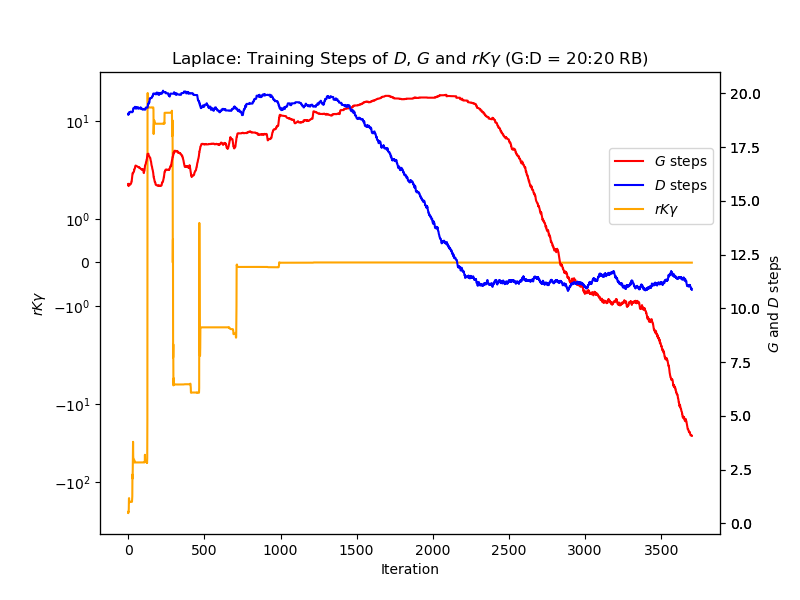}
			\caption{Rollback, \(G:D=20{:}20\)}
			\label{fig:ami_lap2_lsgan_rb_dynamics}
		\end{subfigure}
		\hfill
		\begin{subfigure}[t]{0.32\textwidth}
			\centering
			\includegraphics[width=\linewidth]{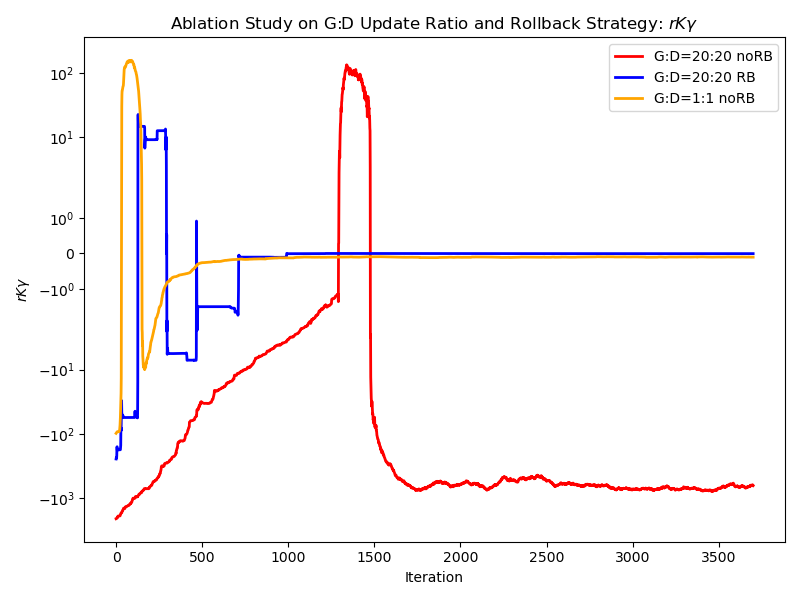}
			\caption{Comparison of \(r^\top K_{rr}\gamma\)}
			\label{fig:ami_lap2_lsgan_gkr}
		\end{subfigure}
		
		\caption{Controlled ablation for LSGAN on the Laplace benchmark. The first row reports the training MSE, validation MSE, and residual energy. The second row compares the convergence behavior without rollback and with rollback under the same update budget \(G:D=20{:}20\), and further shows the corresponding evolution of the first-order quantity \(r^\top K_{rr}\gamma\) across the compared settings.}
		\label{fig:ami_lap2_comparison}
	\end{figure*}
	
	From Fig.~\ref{fig:ami_lap2_comparison}, several clear observations can be made. First, Fig.~\ref{fig:ami_lap2_comparison} (a)--(c) show that among the three compared settings, the fixed \(G:D=20{:}20\) scheme without rollback performs the worst, while the rollback-enhanced \(20{:}20\) scheme achieves the best overall convergence. A particularly interesting phenomenon is that the non-rollback \(20{:}20\) method actually exhibits very good optimization behavior before roughly 1700 iterations, but then suddenly collapses, with both the training and validation errors increasing sharply.
	
	This abrupt failure can be better understood from Fig.~\ref{fig:ami_lap2_comparison} (d), where we plot, for the non-rollback \(20{:}20\) method, the best discriminator/generator inner-step statistics together with the first-order indicator \(r^\top K_{rr}\gamma\). At the beginning of training, this indicator remains stably negative, suggesting that the discriminator-induced signal is well aligned with residual descent. However, around iteration 1500, the indicator undergoes a large jump, and at almost the same time the best generator update depth drops sharply to around 6 steps. Since the generator update budget is fixed to 20 steps, this means that after the useful descent-oriented regime has already been exhausted, the generator continues to update for many additional steps under a deteriorated discriminator signal. As a result, the accumulated error is rapidly amplified, causing the sudden jump in loss and the eventual training failure.
	
	By contrast, Fig.~\ref{fig:ami_lap2_comparison} (e) shows that after introducing rollback, the first-order indicator is maintained within a slightly negative but much more stable range, and the effective generator update depth remains consistently high and stable throughout training. At the same time, the discriminator also keeps a relatively high effective update ratio during the early stage, indicating that it is still able to push the first-order indicator toward a favorable regime. In this sense, rollback acts on both sides of the alternating process: it prevents the generator from taking too many harmful steps once the discriminator signal becomes unreliable, and it also prevents the discriminator from continuing unnecessary updates after a sufficiently good state has already been reached.
	
	This observation also helps explain why the non-rollback \(20{:}20\) scheme can perform well at the beginning of training. In the early stage, allowing the discriminator to take many updates is indeed beneficial, because it helps drive the first-order indicator into a good negative region and thus yields rapid error reduction. However, as training proceeds, keeping the discriminator fixed at 20 updates becomes excessive. At that point, the discriminator no longer needs such a strong update budget, and continuing to optimize it too aggressively starts to destroy the previously favorable dynamical balance. The rollback mechanism effectively corrects this issue by adapting the usable update depth to the current training state, thereby maintaining stability while preserving accuracy.
	
	Finally, Fig.~\ref{fig:ami_lap2_comparison} (f) compares the evolution of \(r^\top K_{rr}\gamma\) across the three settings. One can see that the overall indicator of the \(1{:}1\) baseline is comparable to that of \(20{:}20\) with rollback. However, the important point is that once the generator and discriminator update budgets are both increased, maintaining a favorable first-order indicator becomes much more difficult. This is precisely why the plain \(20{:}20\) scheme fails: without rollback, the enlarged update budget makes the dynamics far more vulnerable to abrupt misalignment, which then causes the catastrophic jump observed in training. In this sense, the comparison shows that the benefit of rollback is not merely to improve performance, but to make larger alternating-update budgets practically usable.
	
	\paragraph{Klein--Gordon.}
	\begin{figure*}[t]
		\centering
		\begin{subfigure}[t]{0.32\textwidth}
			\centering
			\includegraphics[width=\linewidth]{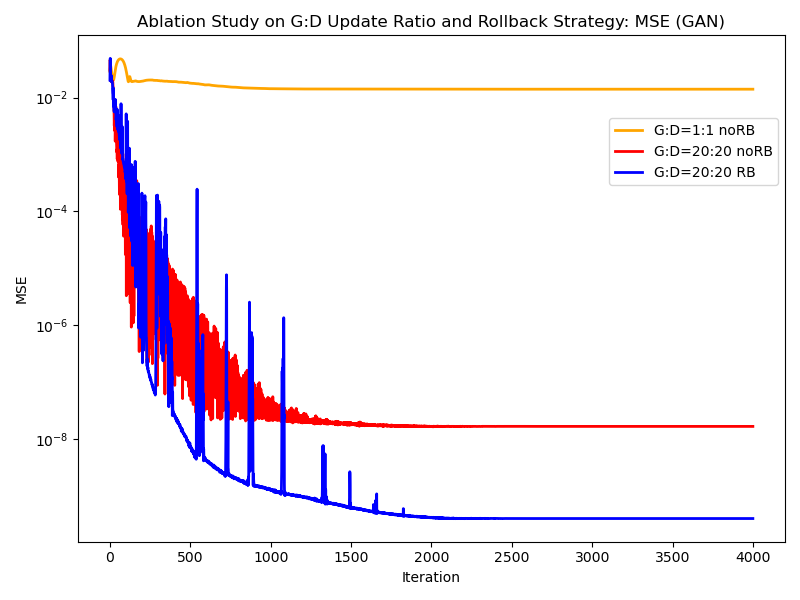}
			\caption{Training MSE}
			\label{fig:ami_wav_train_mse}
		\end{subfigure}
		\hfill
		\begin{subfigure}[t]{0.32\textwidth}
			\centering
			\includegraphics[width=\linewidth]{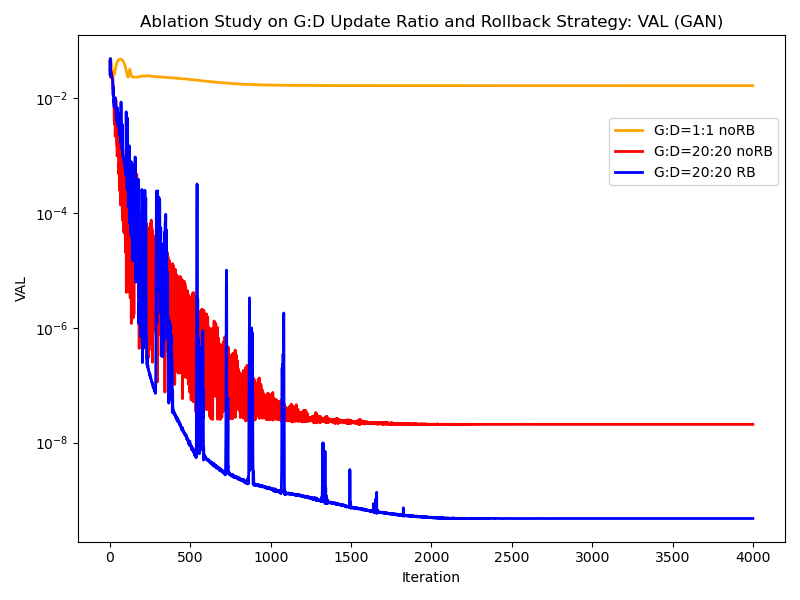}
			\caption{Validation MSE}
			\label{fig:ami_wav_val_mse}
		\end{subfigure}
		\hfill
		\begin{subfigure}[t]{0.32\textwidth}
			\centering
			\includegraphics[width=\linewidth]{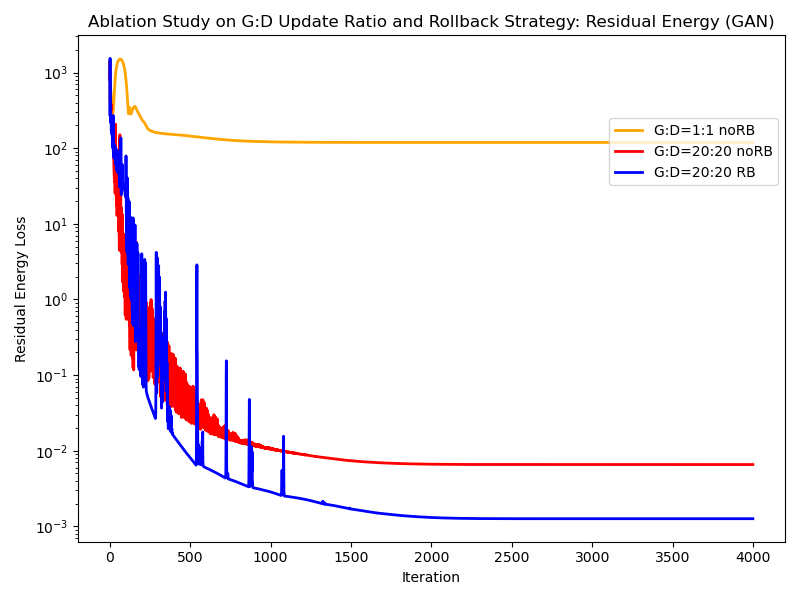}
			\caption{Residual energy}
			\label{fig:ami_wav_residual_energy}
		\end{subfigure}
		
		\vspace{0.6em}
		
		\begin{subfigure}[t]{0.32\textwidth}
			\centering
			\includegraphics[width=\linewidth]{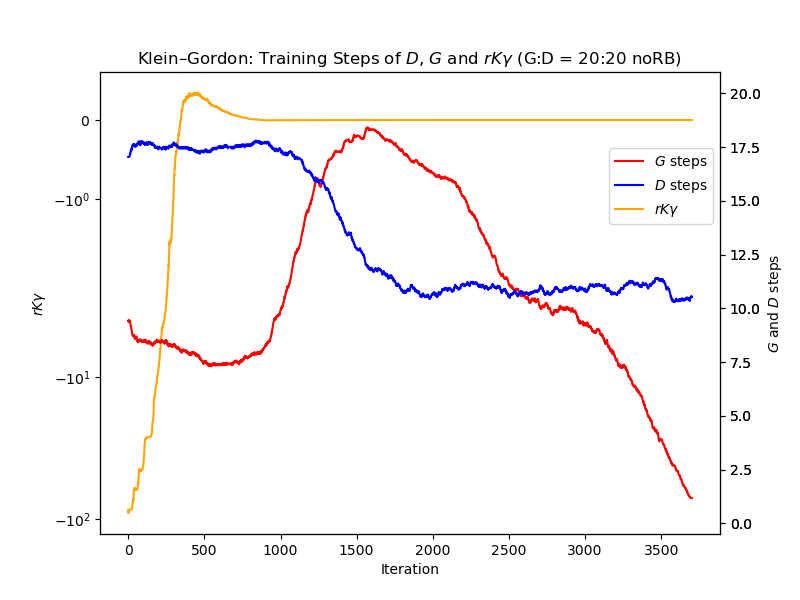}
			\caption{No rollback, \(G:D=20{:}20\)}
			\label{fig:ami_wav_gan_norb_dynamics}
		\end{subfigure}
		\hfill
		\begin{subfigure}[t]{0.32\textwidth}
			\centering
			\includegraphics[width=\linewidth]{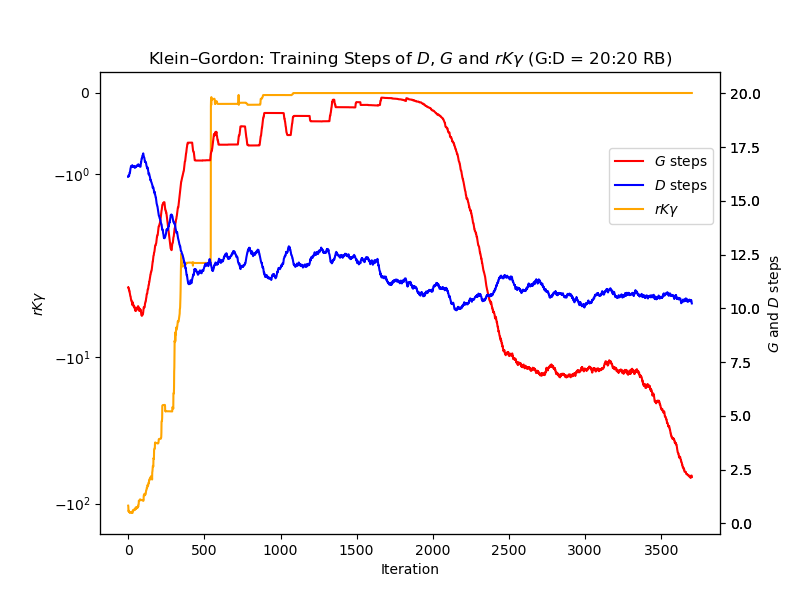}
			\caption{Rollback, \(G:D=20{:}20\)}
			\label{fig:ami_wav_gan_rb_dynamics}
		\end{subfigure}
		\hfill
		\begin{subfigure}[t]{0.32\textwidth}
			\centering
			\includegraphics[width=\linewidth]{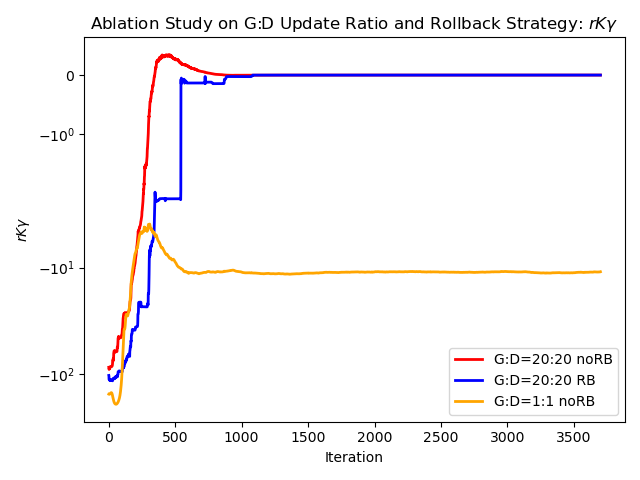}
			\caption{Comparison of \(r^\top K_{rr}\gamma\)}
			\label{fig:ami_wav_gan_gkr}
		\end{subfigure}
		
		\caption{Controlled ablation for GAN on the Klein--Gordon benchmark. The first row reports the training MSE, validation MSE, and residual energy. The second row compares the convergence behavior without rollback and with rollback under the same update budget \(G:D=20{:}20\), and further shows the corresponding evolution of the first-order quantity \(r^\top K_{rr}\gamma\) across the compared settings.}
		\label{fig:ami_wav_comparison}
	\end{figure*}
	
	From Fig.~\ref{fig:ami_wav_comparison}, a different but equally informative pattern can be observed on the Klein--Gordon benchmark. Fig.~\ref{fig:ami_wav_comparison} (a)--(c) show that the \(G:D=1{:}1\) baseline gives the worst final accuracy, the fixed \(20{:}20\) scheme without rollback performs better, and the rollback-enhanced \(20{:}20\) scheme achieves the best overall result. In this example, increasing the generator and discriminator update budgets does indeed improve the solution accuracy, which is different from the Laplace case above.
	
	At the same time, Fig.~\ref{fig:ami_wav_comparison} (d) shows that for the non-rollback \(20{:}20\) scheme, the first-order indicator \(r^\top K_{rr}\gamma\) is kept in a relatively stable negative region throughout most of training. This indicates that the alternating dynamics are more stable here than in the previous example. However, the same Fig.~\ref{fig:ami_wav_comparison} also shows that the best generator update depth is only around 10 steps in the early stage. Therefore, when the generator budget is fixed to 20 updates, a substantial fraction of these updates are already beyond the most effective descent-oriented range. In other words, even though the overall indicator remains negative, the generator is still being updated more than necessary.
	
	Fig.~\ref{fig:ami_wav_comparison} (e) shows the effect of adding rollback under the same \(20{:}20\) budget. In this case, the generator consistently operates within a higher and more stable effective update range, while the first-order indicator is further improved. This suggests that rollback does not merely preserve stability, but also makes the enlarged update budget more efficient by filtering out less useful updates and retaining those that are better aligned with the current descent direction.
	
	Finally, Fig.~\ref{fig:ami_wav_comparison} (f) compares the evolution of \(r^\top K_{rr}\gamma\) across the three settings. One can see that although the \(1{:}1\) baseline has the most favorable first-order indicator overall, its small update budget leads to inferior final accuracy. Once the generator and discriminator update numbers are increased, the first-order indicator deteriorates immediately in the plain \(20{:}20\) setting. After rollback is introduced, however, the indicator is clearly improved again. This explains why the rollback-enhanced method achieves the best performance: it not only allows the larger alternating-update budget to be used, but also restores a better first-order descent quality under that larger budget.
	
	\paragraph{Conclusion:}This ablation supports the interpretation that rollback improves adversarial PINNs training by selecting more descent-consistent alternating updates, rather than by merely enlarging the optimization budget. In other words, the performance improvement is attributable to better update control, not to more iterations.

	\section{Contributions, Limitations, and Future Work}\label{limitation}
	\paragraph{Contributions:}
	Although the present work still has several limitations, we believe that its main contributions are already clear. First, from the perspective of residual dynamics and NTK analysis, we develop a unified framework for adversarially trained PINNs. In contrast to the classical static picture based on the optimal discriminator and divergence minimization, we emphasize that practical training is governed by finite-step alternating dynamics, and we show how different adversarial objectives induce different residual-weighting mechanisms through the discriminator input derivative. Second, based on this framework, we propose a first-order rollback strategy and demonstrate across multiple PDE benchmarks that it can substantially improve GAN-based and LSGAN-based adversarial PINNs without requiring expensive hyperparameter search. Third, we show that this viewpoint is not limited to standard GAN variants: it also provides a unified interpretation of squared-residual input, soft-constrained settings, and existing methods such as SA-PINN, LA-PINN, and WAN, thereby suggesting a broader design space for adversarial PINNs training.
	\paragraph{Limitations:}
	Against this background, several limitations remain.
	
	First, the rollback strategy, while effective, can still introduce noticeable oscillations during training. In some sense, this is natural, since rollback explicitly compares and reverts among candidate alternating updates, so the resulting trajectory is not expected to be completely smooth. Nevertheless, such oscillations indicate that the current rollback mechanism still has room for improvement in terms of stability. At the same time, our ablation results (\ref{app:ablation_deqgan_rb}) show that when rollback is combined with an already well-tuned model, it can achieve both higher accuracy and more stable dynamics. This suggests that rollback should not be viewed only as a substitute for hyperparameter tuning; it may also serve as an additional control layer on top of a good optimization configuration, further improving training quality.
	
	Second, our interpretation of discriminator behavior across different GAN variants is still built on the NTK regime. In particular, the present theory describes how, in the infinite-width and linearized-training limit, the discriminator influences generator residual descent through its input derivative. Compared with the finite-width and often shallow networks used in practice, this analysis remains an approximation. However, the value of the framework is precisely that it does not aim to reproduce every finite-network phenomenon exactly; rather, it provides a mechanistic picture that explains success and failure modes and, more importantly, guides algorithm design. The numerical results suggest that this infinite-width viewpoint remains informative for finite-width training.
	
	Third, the current theory does not explicitly cover common training techniques such as spectral normalization or gradient penalties. This limitation is not specific to our analysis alone, but reflects a broader boundary of the standard NTK framework itself. In particular, classical NTK theory is typically developed under assumptions such as infinite network width, smooth activations, and a near-linearized training regime. By contrast, mechanisms such as spectral normalization actively modify the network parameterization and training geometry, and therefore are not naturally captured by the usual fixed-kernel NTK approximation. For this reason, our present theory does not yet characterize how such techniques alter the discriminator NTK, the induced gradient field, or the resulting residual dynamics in a strict sense. At the same time, this does not prevent us from analyzing the resulting training behavior empirically through the network output and residual-based indicators. One practical strength of the present framework is that these indicators remain informative even when such regularization mechanisms are used in training. A fuller theoretical integration of spectral normalization, gradient penalties, and related techniques would make the framework both more complete and more realistic.
	
	Fourth, the generator-side analysis relies on the assumption that the generator NTK remains approximately constant. For relatively regular problems such as Poisson and Laplace equations, this approximation appears reasonable and is consistent with both existing studies and our numerical observations \cite{wang2022and}. However, for more complex strongly nonlinear, multiscale, stiff, or higher-order PDEs, the generator NTK may evolve more substantially during training, and the constant-kernel approximation may become less accurate \cite{zhou2024neural, carvalho2025positivity}. In this sense, the present theory is better suited to describing early-stage or weakly nonlinear training dynamics than fully global dynamics in strongly nonlinear regimes. Even so, one of the main messages of this paper is that a linearized infinite-width analysis can still provide practically useful guidance for finite-network training control.
	
	Fifth, under finite-width neural network training, the first-order quantity \(r^\top K_{rr}^{G}\gamma\) should be understood as an aggregated and low-cost first-order indicator rather than as a strict per-step descent certificate. It effectively separates successful regimes from representative failure modes and provides a practical basis for rollback, but it remains a low-dimensional global scalar and cannot fully resolve finer interactions among spectral modes, spatial regions, or multiple loss channels. In difficult PDEs, the dominant error may come from a small subset of slow modes, boundary channels, or localized regions, while a single aggregated score may not fully reflect such structural imbalance. At the same time, this indicator is not spectrally blind. Through the decomposition of \(r^\top K_{rr}^{G}\gamma\), it implicitly constrains how discriminator-induced gradients are distributed across modes. In particular, fast modes associated with larger kernel eigenvalues typically decay first, so their residual amplitudes become small early in training. If one still wishes to maintain \(r^\top K_{rr}^{G}\gamma<0\) afterwards, then the discriminator-induced weighting must become more consistent with the slower unresolved modes that continue to dominate the residual energy. In this sense, the first-order indicator does not explicitly optimize spectral modes one by one, but it implicitly favors a training process that first reduces the fast modes and then increasingly emphasizes the slow ones. It is precisely because this quantity is global, low-cost, and available online that it provides a workable bridge between theory and algorithm.
	
	Sixth, while our experiments already cover a diverse set of PDEs, including Poisson, Laplace, viscous Burgers, Reaction--Diffusion, and Klein--Gordon equations, the overall benchmark range is still limited. We have not yet systematically studied higher-dimensional PDEs, complex geometries, noisy inverse problems, parameter identification tasks, or more extreme stiffness and boundary-condition regimes. Therefore, although the current conclusions already show a certain degree of generality, their broader scope still requires further validation. On the other hand, the existing experiments are sufficient to show that the proposed framework is not tied to a single equation or isolated phenomenon, but has explanatory and algorithmic value across multiple PDE types.
	\paragraph{Future Work:}
	These limitations also point naturally to several directions for future work.
	
	First, the rollback strategy itself can be made more stable. For example, one may incorporate rollback as an additional control layer on top of an already well-tuned optimization setting, or design smoother rollback criteria with momentum, confidence thresholds, or multi-step prediction mechanisms to reduce oscillations.
	
	Second, it would be valuable to move beyond the constant-NTK or weakly varying NTK approximation and study more general time-varying kernel dynamics, especially for strongly nonlinear, multiscale, and higher-order PDEs. Such an analysis could lead to a deeper understanding of how generator kernels and discriminator feedback co-evolve across residual modes.
	
	Third, common practical techniques such as spectral normalization, gradient penalties, and residual links should be incorporated into a unified theoretical framework. Understanding how these mechanisms reshape the discriminator NTK, its input-gradient field, and the induced preconditioned residual descent may lead not only to a more faithful theory, but also to better training strategies.
	
	Fourth, the broader design space suggested by this work deserves more systematic exploration. In particular, many combinations among residual input versus squared-residual input, GAN/LSGAN/IPM-type objectives, and constrained discriminator function classes remain unexplored. Such combinations may exhibit better stability, spectral balance, or stiffness-handling properties.
	
	Fifth, future work may develop more refined control criteria beyond the scalar quantity \(r^\top K_{rr}^{G}\gamma\), for instance by using spectral information, channel-wise loss decomposition, or block-structured residual diagnostics for interior, boundary, and initial-condition terms. This could yield a more fine-grained and PDE-aware update-control mechanism. Balancing the computational cost of solving the spectrum and the resulting accuracy during training remains an open problem.
	
	Sixth, the present framework could be extended to broader scientific machine learning settings, including soft-constrained PINNs, inverse problems, optimal control, weak-form methods, and function-space generative models. Since the key idea of this paper is that the discriminator modulates generator dynamics by inducing a gradient field on residual space, this mechanism is not restricted to standard PINNs and may be relevant in a wider class of problems.
	
 \section{Two-Dimensional Heatmap Comparison of GAN-Based Solutions}
    \label{subsec:pde-gan-comparison}
    
    This section presents a comparative visual assessment of the hard-constrained solutions obtained using GAN-RB, LSGAN-RB, and WGAN-RB for the six PDE problems. In each figure, the three rows correspond, from top to bottom, to GAN-RB, LSGAN-RB, and WGAN-RB, respectively. Within each row, the panels from left to right display the hard-constrained prediction, the reference solution, and the pointwise absolute error, defined as \(\lvert \hat{u}(\mathbf{x})-u_{\mathrm{ref}}(\mathbf{x}) \rvert\).

\begin{figure}[H]
        \centering
        \begin{subfigure}{0.96\textwidth}
            \centering
            \includegraphics[width=\linewidth,height=0.24\textheight,keepaspectratio]{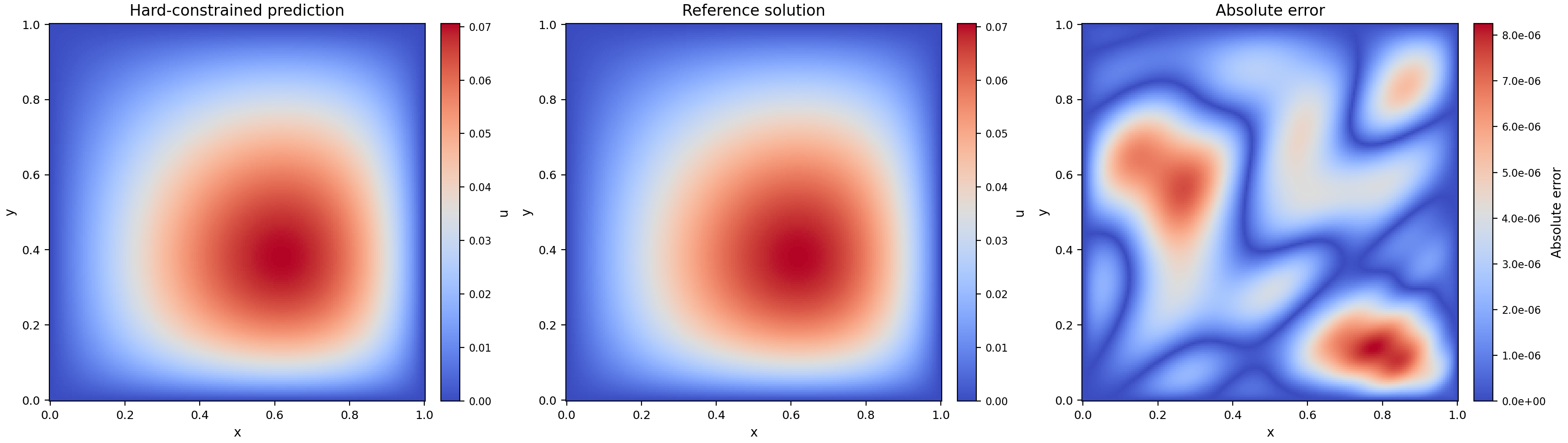}
            \caption{GAN-RB}
        \end{subfigure}
    
        \vspace{0.5em}
        \begin{subfigure}{0.96\textwidth}
            \centering
            \includegraphics[width=\linewidth,height=0.24\textheight,keepaspectratio]{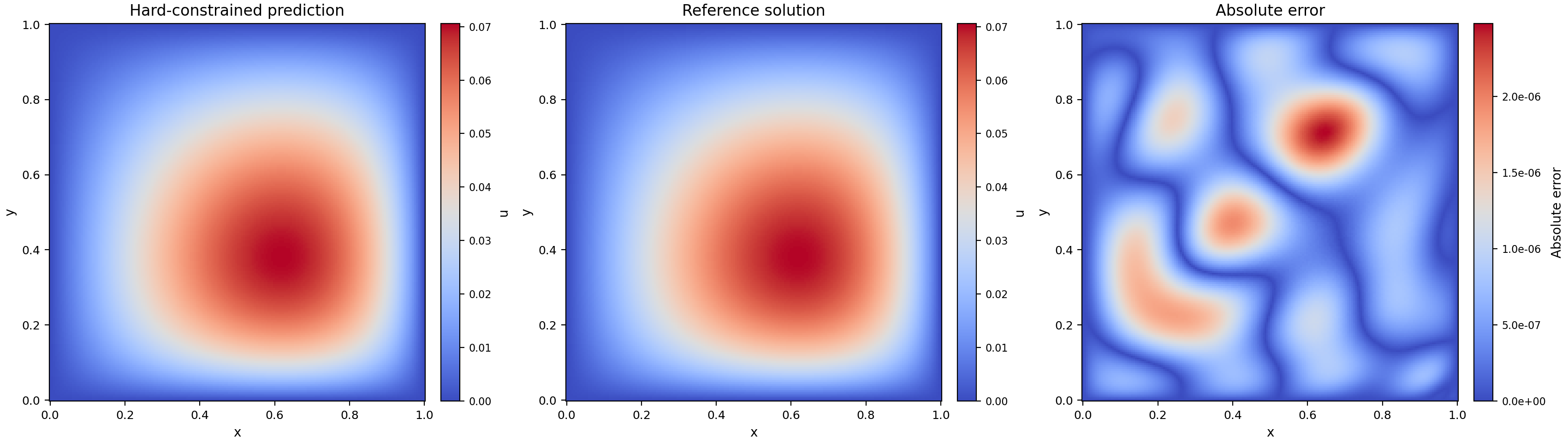}
            \caption{LSGAN-RB}
        \end{subfigure}
    
        \vspace{0.5em}
        \begin{subfigure}{0.96\textwidth}
            \centering
            \includegraphics[width=\linewidth,height=0.24\textheight,keepaspectratio]{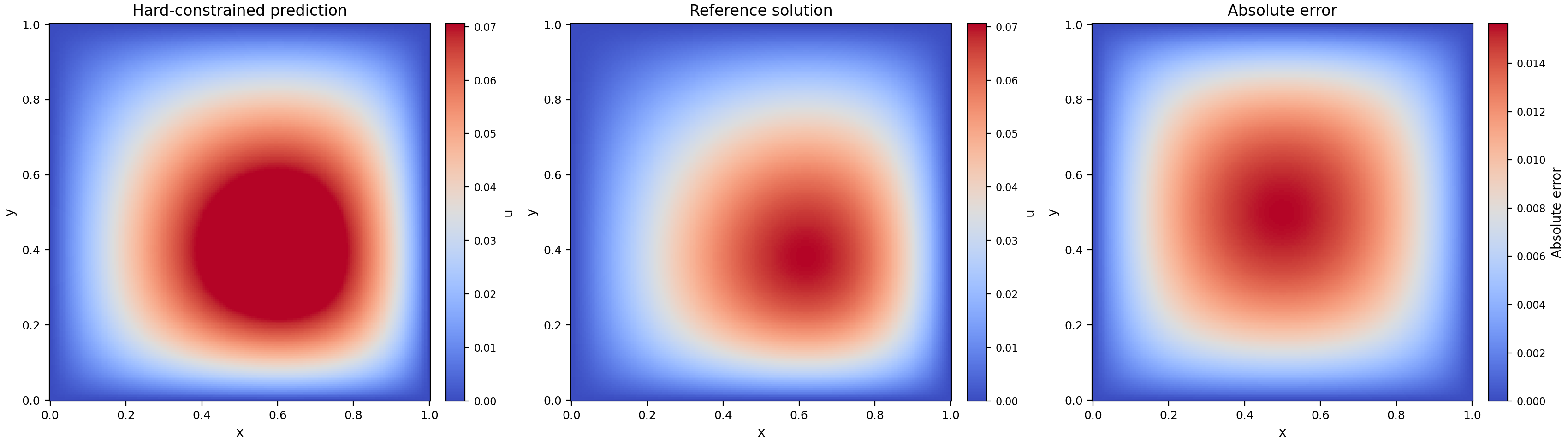}
            \caption{WGAN-RB}
        \end{subfigure}
        \caption{Poisson equation: From top to bottom, the rows present the results obtained using GAN-RB, LSGAN-RB, and WGAN-RB, respectively. Within each row, the panels from left to right show the hard-constrained prediction, the reference solution, and the pointwise absolute error.}
        \label{fig:poisson-gan-comparison}
    \end{figure}

\begin{figure}[H]
        \centering
        \begin{subfigure}{0.96\textwidth}
            \centering
            \includegraphics[width=\linewidth,height=0.24\textheight,keepaspectratio]{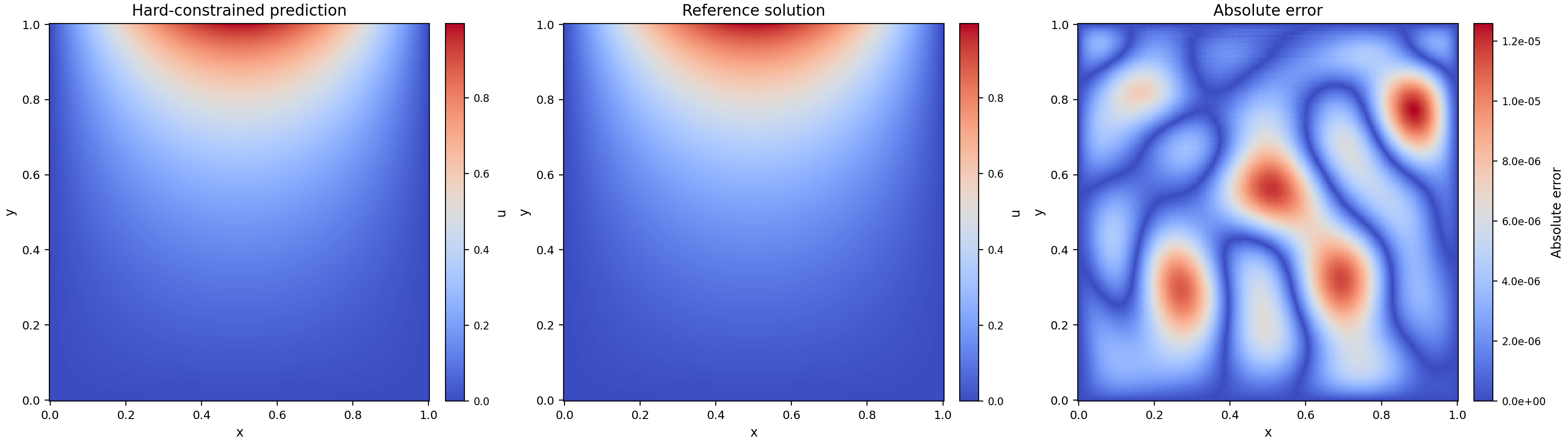}
            \caption{GAN-RB}
        \end{subfigure}
    
        \vspace{0.5em}
        \begin{subfigure}{0.96\textwidth}
            \centering
            \includegraphics[width=\linewidth,height=0.24\textheight,keepaspectratio]{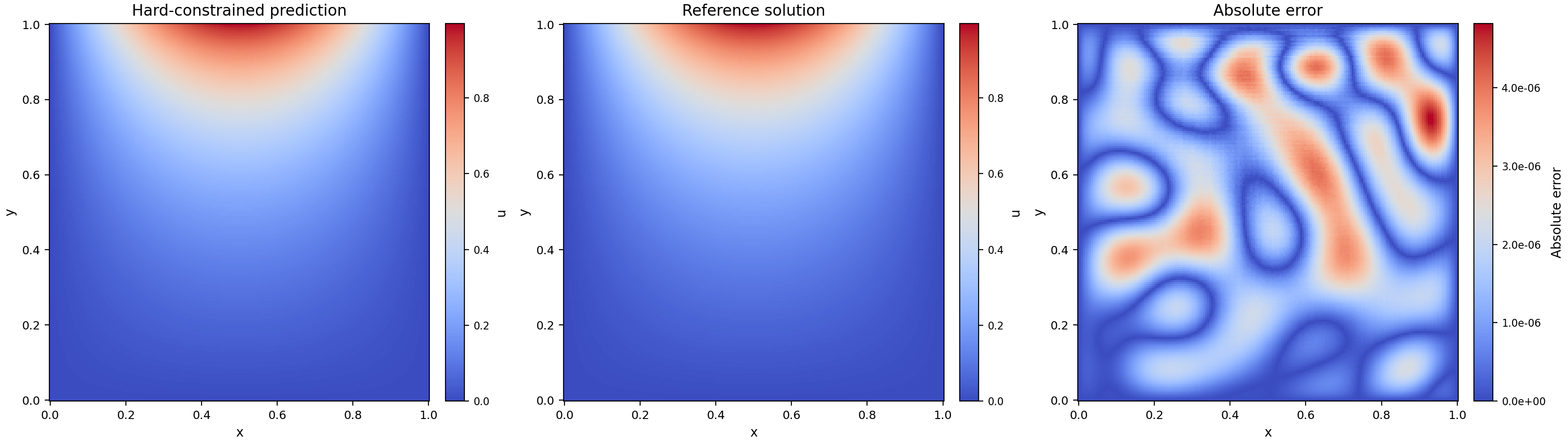}
            \caption{LSGAN-RB}
        \end{subfigure}
    
        \vspace{0.5em}
        \begin{subfigure}{0.96\textwidth}
            \centering
            \includegraphics[width=\linewidth,height=0.24\textheight,keepaspectratio]{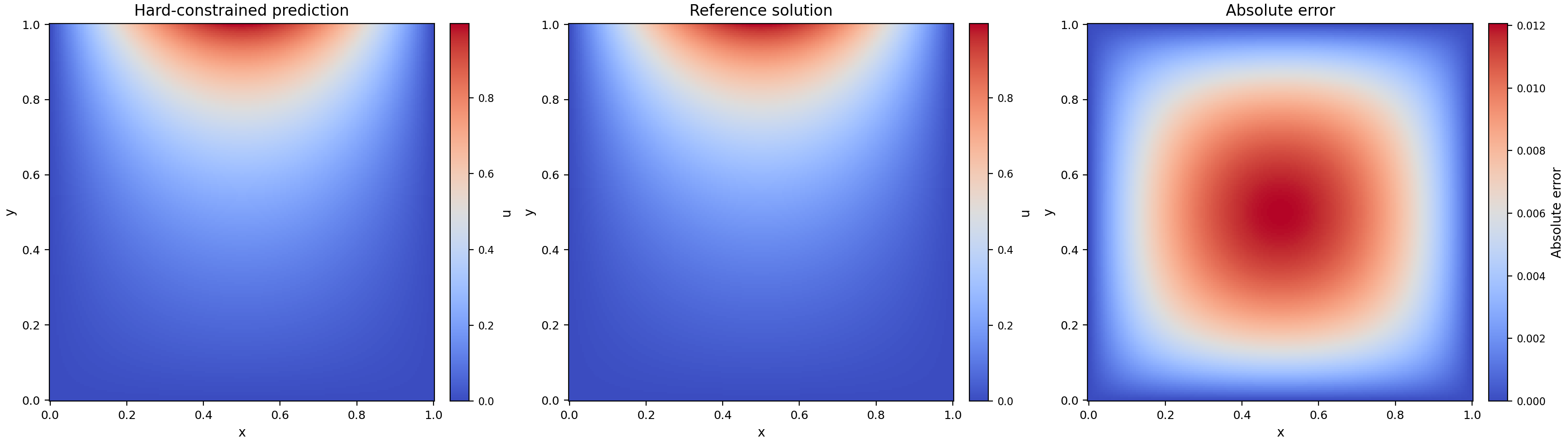}
            \caption{WGAN-RB}
        \end{subfigure}
        \caption{Laplace equation 1: From top to bottom, the rows present the results obtained using GAN-RB, LSGAN-RB, and WGAN-RB, respectively. Within each row, the panels from left to right show the hard-constrained prediction, the reference solution, and the pointwise absolute error.}
        \label{fig:laplace1-gan-comparison}
    \end{figure}

\begin{figure}[H]
        \centering
        \begin{subfigure}{0.96\textwidth}
            \centering
            \includegraphics[width=\linewidth,height=0.24\textheight,keepaspectratio]{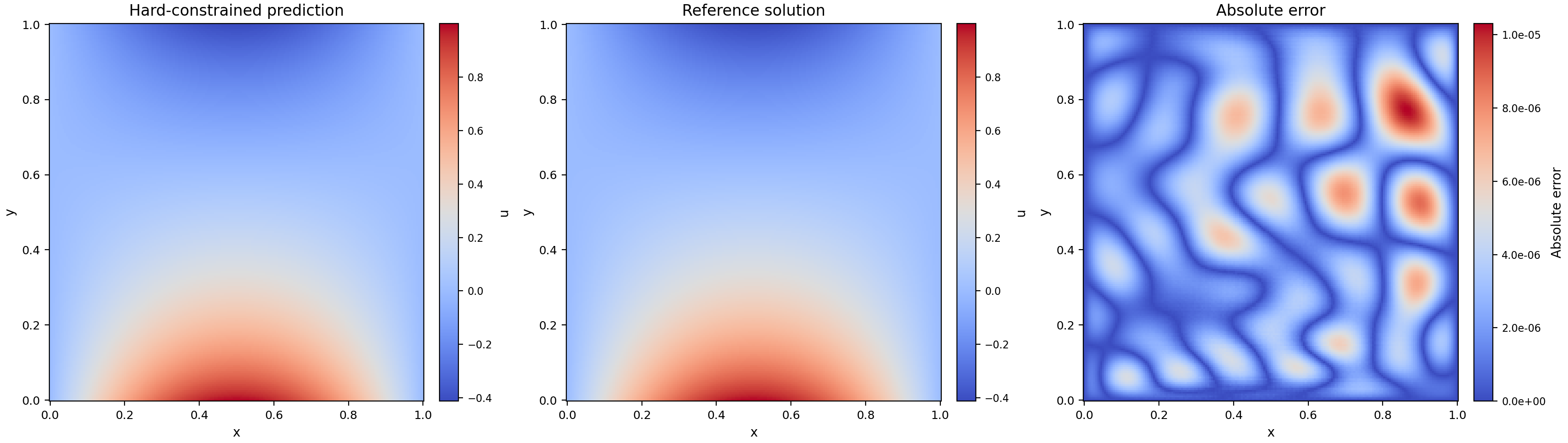}
            \caption{GAN-RB}
        \end{subfigure}
    
        \vspace{0.5em}
        \begin{subfigure}{0.96\textwidth}
            \centering
            \includegraphics[width=\linewidth,height=0.24\textheight,keepaspectratio]{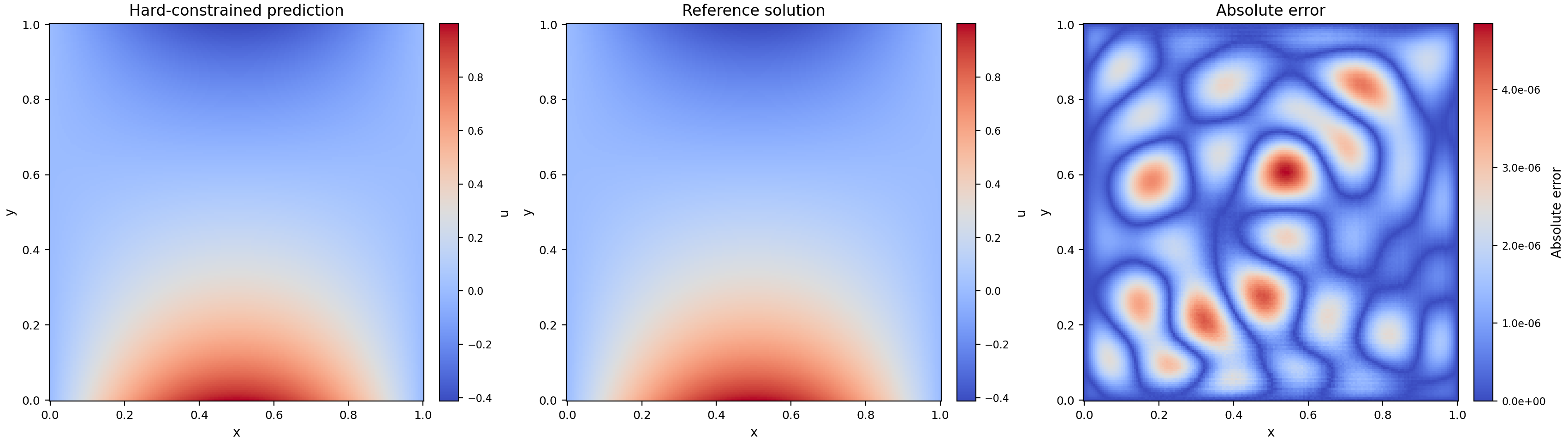}
            \caption{LSGAN-RB}
        \end{subfigure}
    
        \vspace{0.5em}
        \begin{subfigure}{0.96\textwidth}
            \centering
            \includegraphics[width=\linewidth,height=0.24\textheight,keepaspectratio]{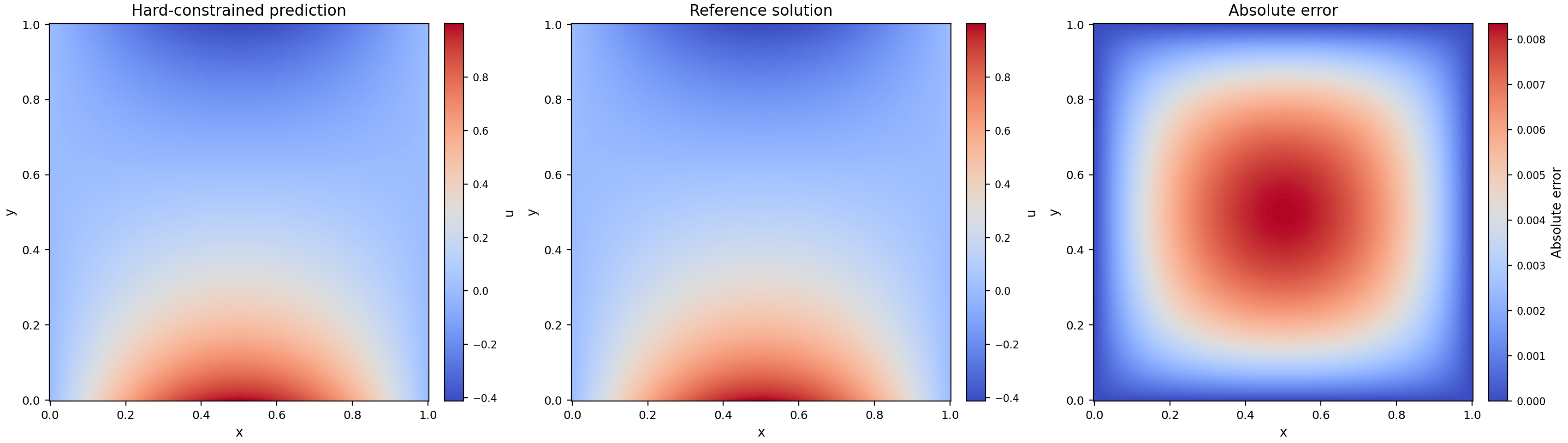}
            \caption{WGAN-RB}
        \end{subfigure}
        \caption{Laplace equation 2: From top to bottom, the rows present the results obtained using GAN-RB, LSGAN-RB, and WGAN-RB, respectively. Within each row, the panels from left to right show the hard-constrained prediction, the reference solution, and the pointwise absolute error.}
        \label{fig:laplace2-gan-comparison}
    \end{figure}

\begin{figure}[H]
        \centering
        \begin{subfigure}{0.96\textwidth}
            \centering
            \includegraphics[width=\linewidth,height=0.24\textheight,keepaspectratio]{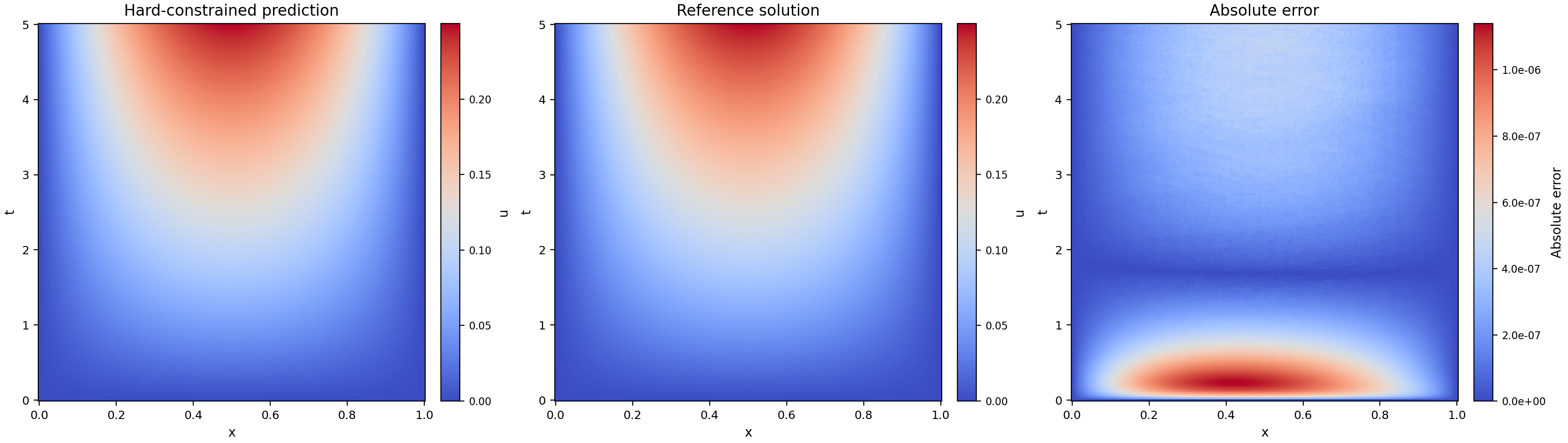}
            \caption{GAN-RB}
        \end{subfigure}
    
        \vspace{0.5em}
        \begin{subfigure}{0.96\textwidth}
            \centering
            \includegraphics[width=\linewidth,height=0.24\textheight,keepaspectratio]{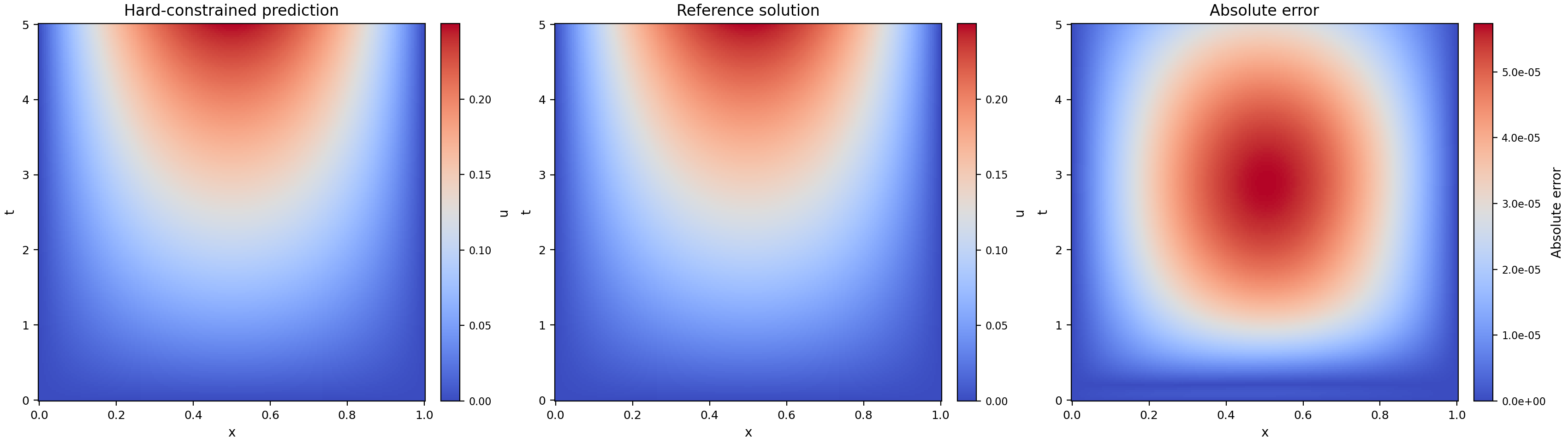}
            \caption{LSGAN-RB}
        \end{subfigure}
    
        \vspace{0.5em}
        \begin{subfigure}{0.96\textwidth}
            \centering
            \includegraphics[width=\linewidth,height=0.24\textheight,keepaspectratio]{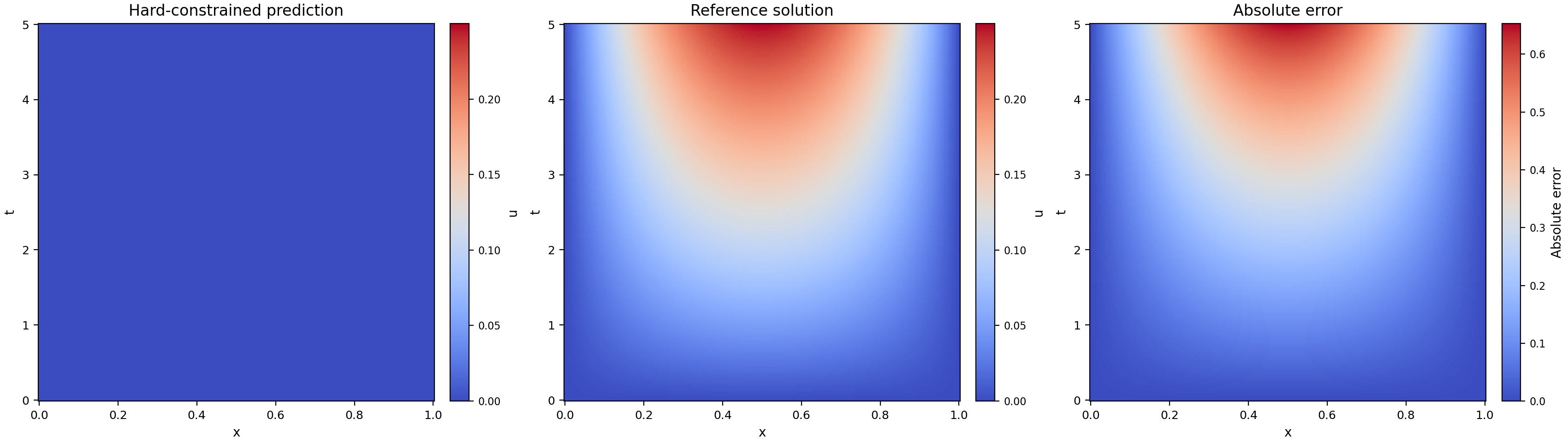}
            \caption{WGAN-RB}
        \end{subfigure}
        \caption{Reaction-Difussion equation: From top to bottom, the rows present the results obtained using GAN-RB, LSGAN-RB, and WGAN-RB, respectively. Within each row, the panels from left to right show the hard-constrained prediction, the reference solution, and the pointwise absolute error.}
        \label{fig:heat-gan-comparison}
    \end{figure}

\begin{figure}[H]
        \centering
        \begin{subfigure}{0.96\textwidth}
            \centering
            \includegraphics[width=\linewidth,height=0.24\textheight,keepaspectratio]{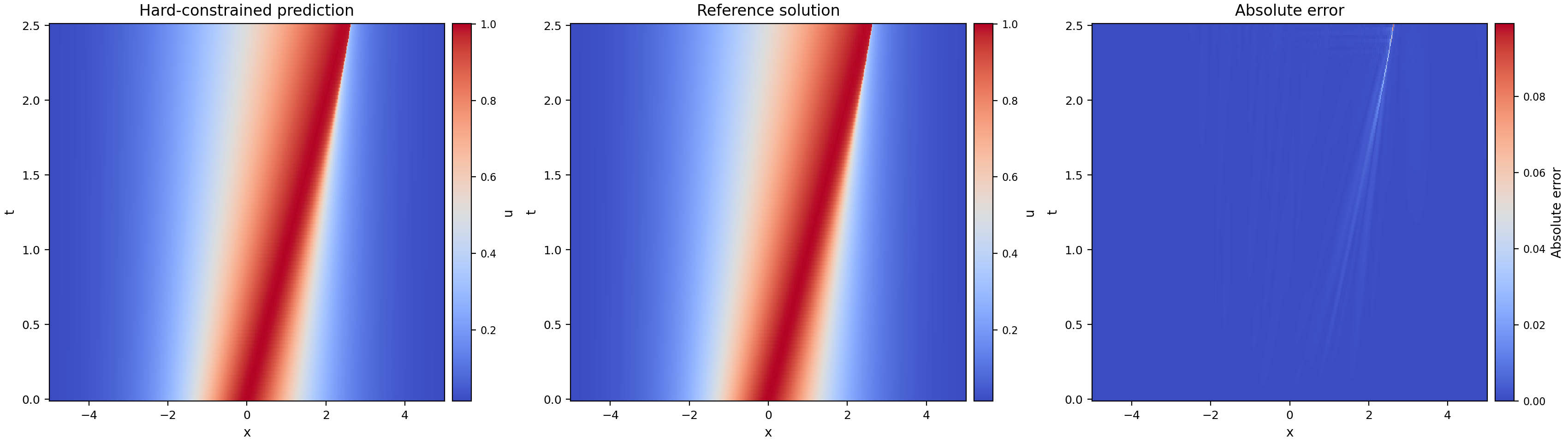}
            \caption{GAN-RB}
        \end{subfigure}
    
        \vspace{0.5em}
        \begin{subfigure}{0.96\textwidth}
            \centering
            \includegraphics[width=\linewidth,height=0.24\textheight,keepaspectratio]{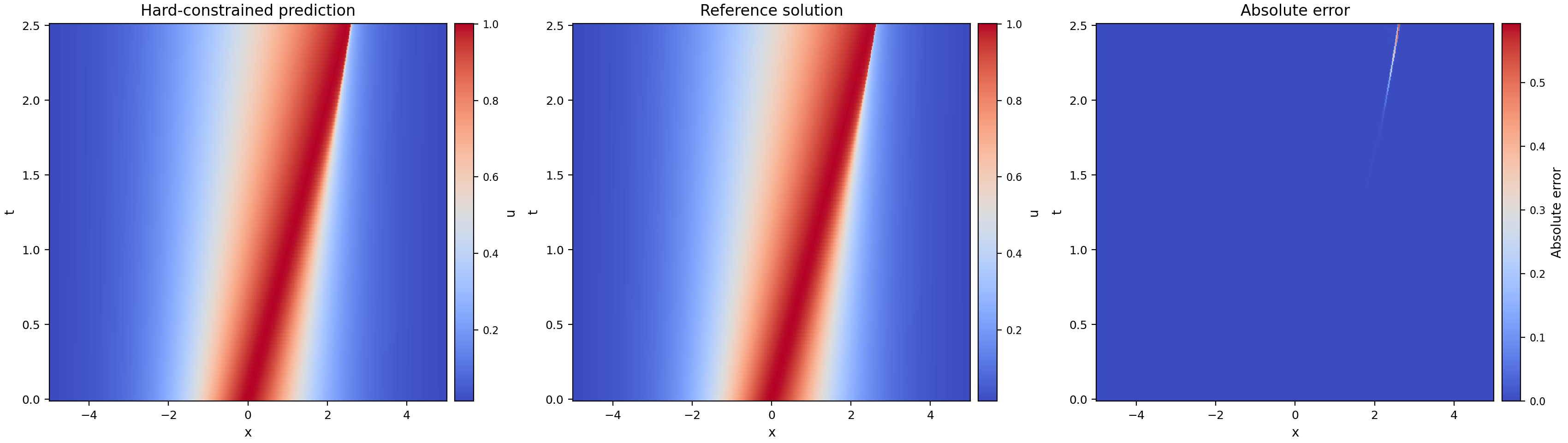}
            \caption{LSGAN-RB}
        \end{subfigure}
    
        \vspace{0.5em}
        \begin{subfigure}{0.96\textwidth}
            \centering
            \includegraphics[width=\linewidth,height=0.24\textheight,keepaspectratio]{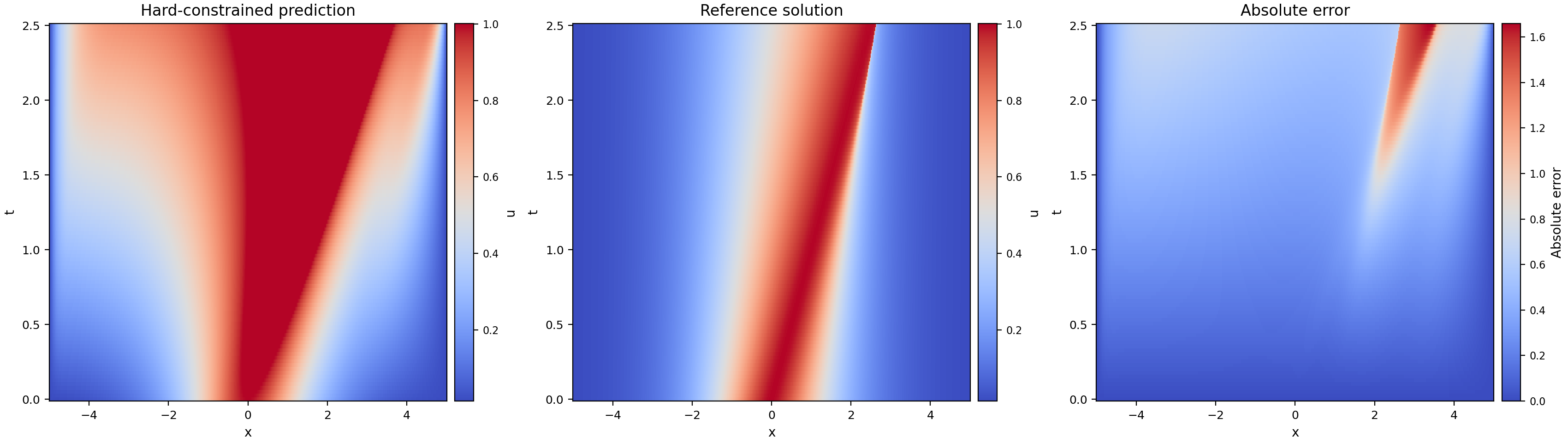}
            \caption{WGAN-RB}
        \end{subfigure}
        \caption{Viscous Burgers equation: From top to bottom, the rows present the results obtained using GAN-RB, LSGAN-RB, and WGAN-RB, respectively. Within each row, the panels from left to right show the hard-constrained prediction, the reference solution, and the pointwise absolute error.}
        \label{fig:burgers-viscous-gan-comparison}
    \end{figure}

\begin{figure}[H]
        \centering
        \begin{subfigure}{0.96\textwidth}
            \centering
            \includegraphics[width=\linewidth,height=0.24\textheight,keepaspectratio]{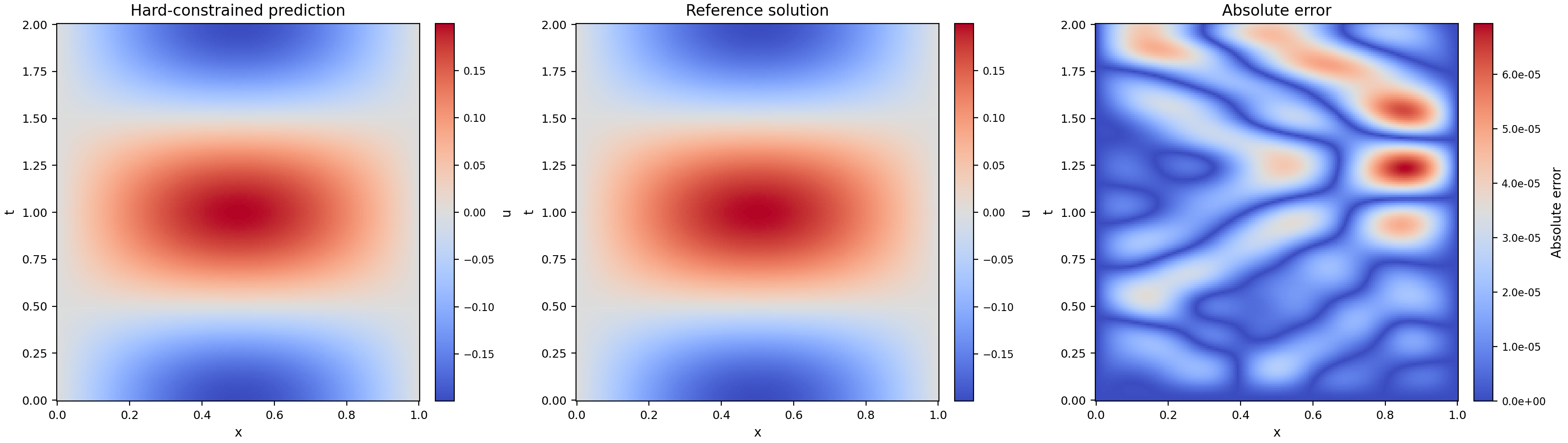}
            \caption{GAN-RB}
        \end{subfigure}
    
        \vspace{0.5em}
        \begin{subfigure}{0.96\textwidth}
            \centering
            \includegraphics[width=\linewidth,height=0.24\textheight,keepaspectratio]{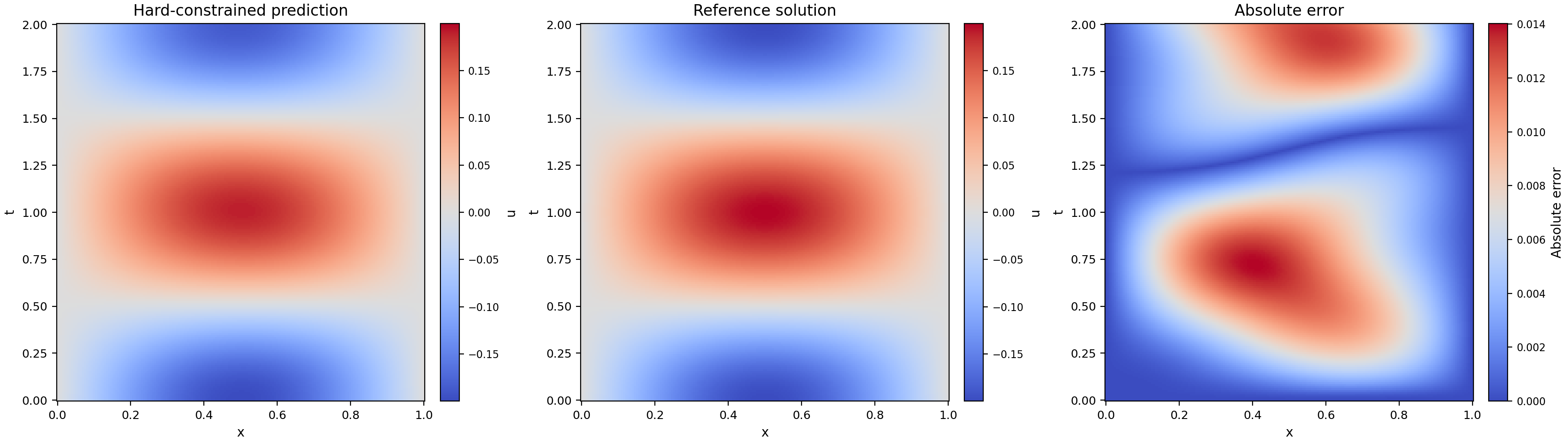}
            \caption{LSGAN-RB}
        \end{subfigure}
    
        \vspace{0.5em}
        \begin{subfigure}{0.96\textwidth}
            \centering
            \includegraphics[width=\linewidth,height=0.24\textheight,keepaspectratio]{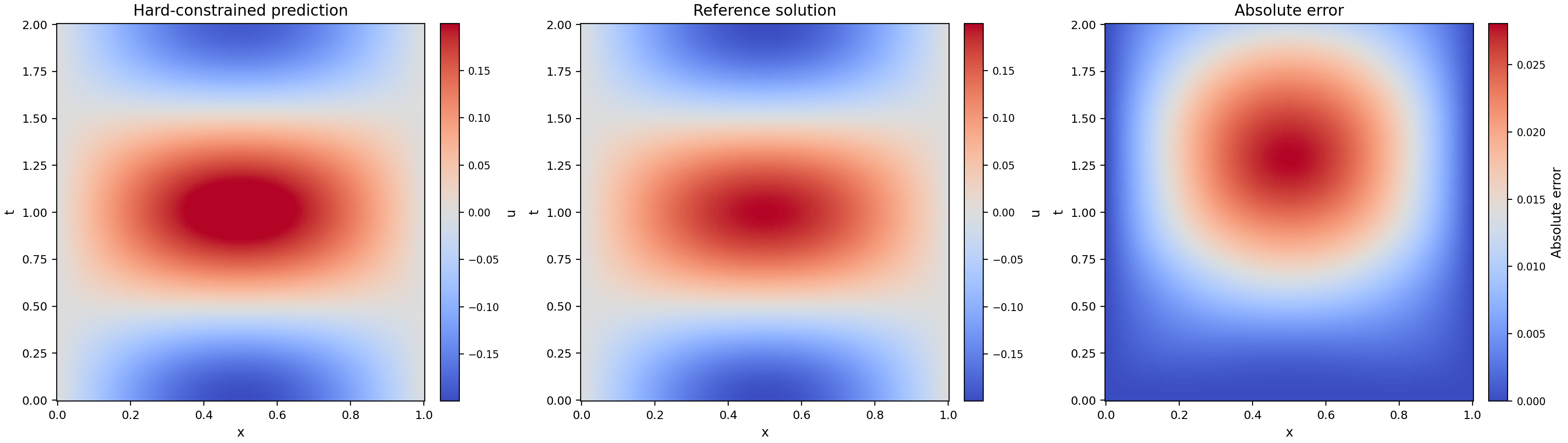}
            \caption{WGAN-RB}
        \end{subfigure}
        \caption{Klein-Gordon equation: From top to bottom, the rows present the results obtained using GAN-RB, LSGAN-RB, and WGAN-RB, respectively. Within each row, the panels from left to right show the hard-constrained prediction, the reference solution, and the pointwise absolute error.}
        \label{fig:wave-gan-comparison}
    \end{figure}

   Figure~\ref{fig:poisson-gan-comparison} compares the predicted and reference solutions and their absolute errors for the Poisson equation. Figures~\ref{fig:laplace1-gan-comparison} and~\ref{fig:laplace2-gan-comparison} present the corresponding results for the two Laplace problems with different boundary conditions. Figures~\ref{fig:heat-gan-comparison}, \ref{fig:burgers-viscous-gan-comparison}, and~\ref{fig:wave-gan-comparison}show the space--time solutions and error distributions for the reaction--diffusion, viscous Burgers, and Klein--Gordon equations, respectively.
   
\end{document}